\documentclass[final]{cvpr}

\usepackage{times}
\usepackage{siunitx}
\usepackage{booktabs}
\usepackage{multicol}
\usepackage{bm}
\usepackage{url}
\usepackage{graphicx} 
\usepackage{mathptmx} 
\usepackage[mathscr]{euscript}  
\usepackage{amsmath} 
\usepackage{amssymb}  
\usepackage{flushend}
\usepackage{siunitx}
\usepackage{adjustbox}
\usepackage{array}
\usepackage{multirow}
\usepackage{enumitem}
\usepackage{microtype}
\usepackage[resetlabels]{multibib}
\usepackage{cuted}

\usepackage{placeins}
\makeatletter
\@namedef{ver@everyshi.sty}{}
\makeatother
\usepackage{pgfplots}
\pgfplotsset{compat=1.7}

\newcommand\blfootnote[1]{%
  \begingroup
  \renewcommand\thefootnote{}\footnote{#1}%
  \addtocounter{footnote}{-1}%
  \endgroup
}

\renewcommand{\eqref}[1]{Eq.~(\ref{#1})}

\newcolumntype{P}[1]{>{\centering\arraybackslash}p{#1}}

\usepackage[pagebackref=true,breaklinks=true,colorlinks,bookmarks=false]{hyperref}

\def\cvprPaperID{5510} 
\def\confYear{CVPR 2021}

\makeatletter
\newcommand{\printfnsymbol}[1]{%
  \textsuperscript{\@fnsymbol{#1}}%
}
\makeatother

\begin{document}

\title{There is More than Meets the Eye:  Self-Supervised Multi-Object Detection and Tracking with Sound by Distilling Multimodal Knowledge}

\author{Francisco Rivera Valverde\thanks{Equal contribution.}
\qquad
Juana Valeria Hurtado\printfnsymbol{1}
\qquad
Abhinav Valada\\
University of Freiburg\\
\tt\small\{riverav, hurtadoj, valada\}@cs.uni-freiburg.de}

\maketitle

\begin{abstract}
Attributes of sound inherent to objects can provide valuable cues to learn rich representations for object detection and tracking. Furthermore, the co-occurrence of audiovisual events in videos can be exploited to localize objects over the image field by solely monitoring the sound in the environment. Thus far, this has only been feasible in scenarios where the camera is static and for single object detection. Moreover, the robustness of these methods has been limited as they primarily rely on RGB images which are highly susceptible to illumination and weather changes. In this work, we present the novel self-supervised MM-DistillNet framework consisting of multiple teachers that leverage diverse modalities including RGB, depth and thermal images, to simultaneously exploit complementary cues and distill knowledge into a single audio student network. We propose the new MTA loss function that facilitates the distillation of information from multimodal teachers in a self-supervised manner. Additionally, we propose a novel self-supervised pretext task for the audio student that enables us to not rely on labor-intensive manual annotations. We introduce a large-scale multimodal dataset with over 113,000 time-synchronized frames of RGB, depth, thermal, and audio modalities. Extensive experiments demonstrate that our approach outperforms state-of-the-art methods while being able to detect multiple objects using only sound during inference and even while moving.
\end{abstract}
\section{Introduction}

\begin{figure}
\centering
\footnotesize
\includegraphics[width=\linewidth]{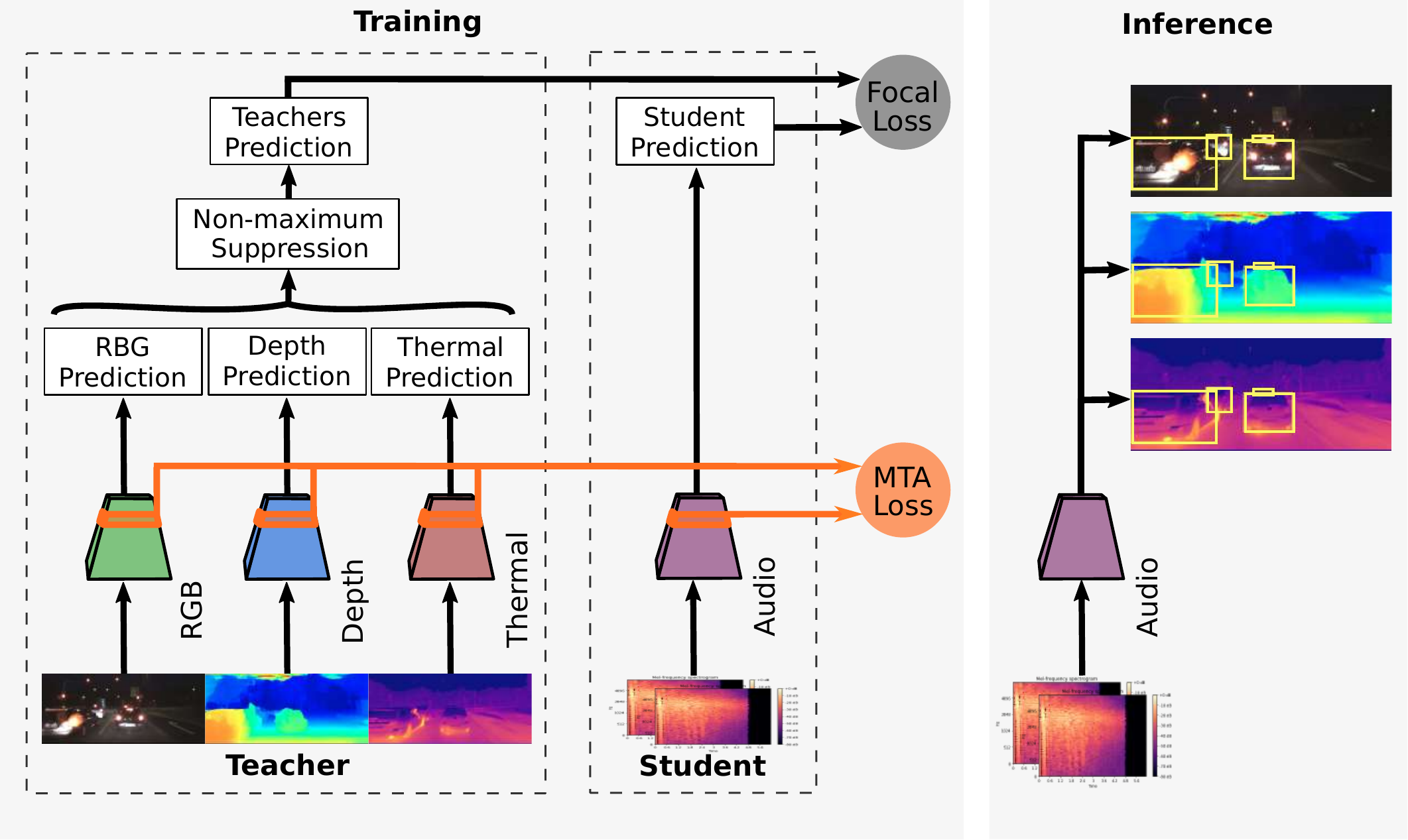}
\caption{Our proposed cross-modal MM-DistillNet distills knowledge exploiting complementary cues from multimodal visual teachers into an audio student. During inference, the model detects and tracks multiple objects in the visual frame using only audio as input.\looseness=-1}
\label{fig:cover}
\end{figure}

Human perception is deceptively effortless, we can sense people behind our backs and in the dark, although we cannot remotely see them. This elucidates that perception is inherently a complex cognitive phenomenon that is facilitated by the integration of various sensory modalities~\cite{calvert1998crossmodal}. ``There is More than Meets the Eye'' aptly summarizes this complexity of our visual perception system. Modeling this ability using learning algorithms is, however, far from being solved. The natural co-occurrence of modalities such as images and audio in videos provides strong cues for supervision that can be exploited to learn more robust perception models in a self-supervised manner. Attributes of sound inherent to objects in the scene also contain rich time and frequency domain information that is valuable for grounding sounds within a visual scene. In this sense, the characteristics of sound are complementary and correlated to the visual information~\cite{senocak2018learning}. Cross-modal learning from images and sound exploits this natural correspondence between audio-visual streams that represent the same event. As a result, the integration of sound with vision enables us to use one modality to supervise the other as well as to use both modalities to supervise each other jointly~\cite{aytar2016soundnet,arandjelovic2017look,zurn2020self}.

Generally, training models to detect objects requires large amounts of groundtruth annotations for supervision. However, we can train models to recognize objects that produce sound without relying on labeled data by jointly leveraging audio-visual learning using the teacher-student strategy~\cite{hinton2015}. With this approach, numerous works~\cite{adanur2019deep, xiao2020audiovisual,aytar2016soundnet} have used the audio-visual correlation to localize sounds sources. Moreover, recent work~\cite{wang2020score} has shown that we can exploit this audio-visual synchronicity to detect and track an object over the visual frame. Thus far, this promising capability has only been shown to be feasible in scenarios where the camera is static and for detecting a single object at a time using stereo sound and metadata containing camera pose information as input. Moreover, it distills knowledge only from models trained with RGB images, which are highly susceptible to perceptual changes such as varying types, scales, and visibility of objects, domain differences in terms of weather, illumination, and seasonality, among many others. Addressing these challenges will enable us to employ the system for detection and tracking in a wide variety of applications.


In this work, we present the novel self-supervised Multi-Modal Distillation Network (MM-DistillNet) that provides effective solutions to the aforementioned problems. Our framework illustrated in Fig.~\ref{fig:cover} consists of multiple teacher networks, each of which takes a specific modality as input, for which we use RGB, depth, and thermal to maximize the complementary cues that we can exploit (appearance, geometry, reflectance). 
The teachers are first individually trained on diverse pre-existing datasets to predict bounding boxes in their respective modalities. We then train the audio student network to learn the mapping of sounds from a microphone array to bounding box coordinates of the combined teachers' prediction, only on unlabeled videos. To do this, we present the novel Multi-Teacher Alignment (MTA) loss to simultaneously exploit complementary cues and distill object detection knowledge from multimodal teachers into the audio student network in a self-supervised manner. During inference, the audio student network detects and tracks objects in the visual frame using only sound as an input. Additionally, we present a self-supervised pretext task for initializing the audio student network in order to not rely on labor-intensive manual annotations and to accelerate training.

To facilitate this work, we collected a large-scale driving dataset with over 113,000 time-synchronized frames of RGB, depth, thermal, and multi-channel audio modalities. We present extensive experimental results comparing the performance of our proposed MM-DistillNet with existing methods as well as baseline approaches, which shows that it substantially outperforms the state-of-the-art. More importantly, for the first time, we demonstrate the capability to detect and track objects in the visual frame, from only using sound as an input, without any meta-data and even while moving in the environment. We also present detailed ablation studies that highlight the novelty of the contributions that we make. Finally, we make our dataset, code and models publicly available at \url{http://rl.uni-freiburg.de/research/multimodal-distill}.

\section{Related Work}

In recent years, several deep learning methods~\cite{arandjelovic2017look, vasudevan2020semantic, hu2020discriminative} have exploited the natural relationship between the co-occurrent vision and sound events found in video sequences. Some of these works rely on groundtruth annotations and propose supervised approaches to learn joint audio-visual embeddings by transferring knowledge between the modality-specific networks. Various tasks such as audio classification~\cite{perez2020audio}, lip reading~\cite{afouras2020asr}, face recognition~\cite{nagrani2018seeing}, and speaker identification~\cite{nagrani2018learnable} have been tackled using these techniques. Another set of approaches exploit self-supervision and they do not rely on any manual annotations. These methods exploit audio-vision synchronicity and learn representations of one modality while using the other counterpart modality. For example, Hershey~\textit{et~al.}~\cite{hershey2017cnn} use audio data as a supervision signal to learn visual representations, and Aytar~\textit{et~al.}~\cite{aytar2016soundnet} propose SoundNet that uses visual imagery as supervision for acoustic scene classification.

More related to our work are methods that use audio-visual correspondence and vision data as a supervisory signal for localizing sound in a given visual input. This task is typically tackled using a pre-trained visual network as supervision~\cite{afouras2020self,arandjelovic2017look}, by generating a common audio-visual representation~\cite{zhao2018sound, owens2018audio, arandjelovic2018objects, qian2020multiple} or by using an attention mechanism~\cite{senocak2018learning, ramaswamy2020see}. StereoSoundNet~\cite{gan2019self} performs object detection and tracking of a single-vehicle using a stereo microphone and camera pose information as the input. While on the other hand, Adanur~\textit{et~al.}~\cite{adanur2019deep} and Ma~\textit{et~al.}~\cite{ma2019phased} demonstrate the advantages of using multiple microphones for spatial detection. Our proposed MM-DistillNet also performs object detection and tracking in the visual frame using only sound from a microphone array, allowing the system to detect multiple vehicles at the same time without using any camera pose information and while moving in the environment.\looseness=-1

Given the low spatial resolution of sound, it is extremely complex and arduous to manually label audio for object localization over the visual domain. Recent techniques~\cite{zhao2018sound, owens2018audio, gan2019self, morgado2018self, vasudevan2020semantic} address this problem by leveraging a vision-teacher's knowledge to supervise and generate the labels to train an audio-student network. Similarly, to reduce the groundtruth label dependency, our approach exploits the co-occurrence of modalities as a self-supervised mechanism to obtain groundtruth annotations. However, all of the aforementioned methods only use RGB images from the visual domain, which are highly susceptible to illumination and weather changes. To address this issue, several approaches have been proposed to leverage multiple modalities such as RGB, depth, and thermal images to exploit complementary cues by fusing them at the input or at the feature level~\cite{hafner2018cross, valada2016convoluted, burgard2020perspectives}. Although these methods have substantially improved the performance of object detection and semantic segmentation in challenging perceptual conditions, they are still constrained by modality limitations such as range-of-vision or occlusions. Moreover, adding new modalities also increases the labeling effort and these fusion techniques typically require all the modalities to be present during inference time, both of which increase the overall system overhead. As opposed to these techniques, we propose a methodology to incorporate the knowledge from multiple pre-trained modality-specific teacher networks into an audio student network that learns from unlabeled videos and only uses audio during inference. Our approach exploits complementary features from the alternate modalities while training, in an effort to improve the robustness of the overall system without increasing the overhead at inference.

Besides for generating pseudo groundtruth labels, we employ the modality-specific teacher networks to guide the training of the audio student network via knowledge distillation. Previous works~\cite{tian2019contrastive, wen2019preparing, cho2019efficacy, zhang2020distilling} use the knowledge from the output logits by softening the labels. Our approach is more related to \cite{romero2014fitnets, zagoruyko2016paying, yim2017gift, aguilar2020knowledge}, which transfers the knowledge from intermediate layers through an alignment loss function. Similar to \cite{park2019feed, zhou2019m2kd}, our approach distills knowledge from multiple teachers. However, our framework does not merely average a dual loss among the teachers, rather it aligns the features of the intermediate teacher-student layers using a probabilistic approach. We show that the conditional knowledge given a synchronized set of modalities can improve the student network's performance. 

In our approach, each modality-specific teacher distills object detection knowledge to the audio student, which can be categorized as cross-modal knowledge distillation. Salem~\textit{et~al.}~\cite{salem2019learning} proposes to distill the knowledge from the logits of a group of different visual modalities. Do~\textit{et~al.}~\cite{do2019compact} and Zhang~\textit{et~al.}~\cite{zhang2018better} employ attention maps to combine the different modalities. These approaches require all the modalities, both during training and inference. Whereas our framework aims to disentangle the need for all the modalities during inference time. Alayrac~\textit{et~al.}~\cite{alayrac2020self} recently propose to address this problem by creating an embedding with a contrastive pairwise loss that facilitates the downstream task. Nevertheless, their approach tackles cross-modal representations that are significantly different. Whereas, our approach distills knowledge from modality-specific teachers that are aligned at the object level, so the information from a common task is distilled into a complementary modality using our proposed MTA loss and the focal loss. We evaluate the performance of existing multi-teacher distillation losses with our proposed strategy in the ablation study.


\section{Technical Approach}
\label{sec:technical}

\begin{figure}
\centering
\footnotesize
\includegraphics[width=\linewidth]{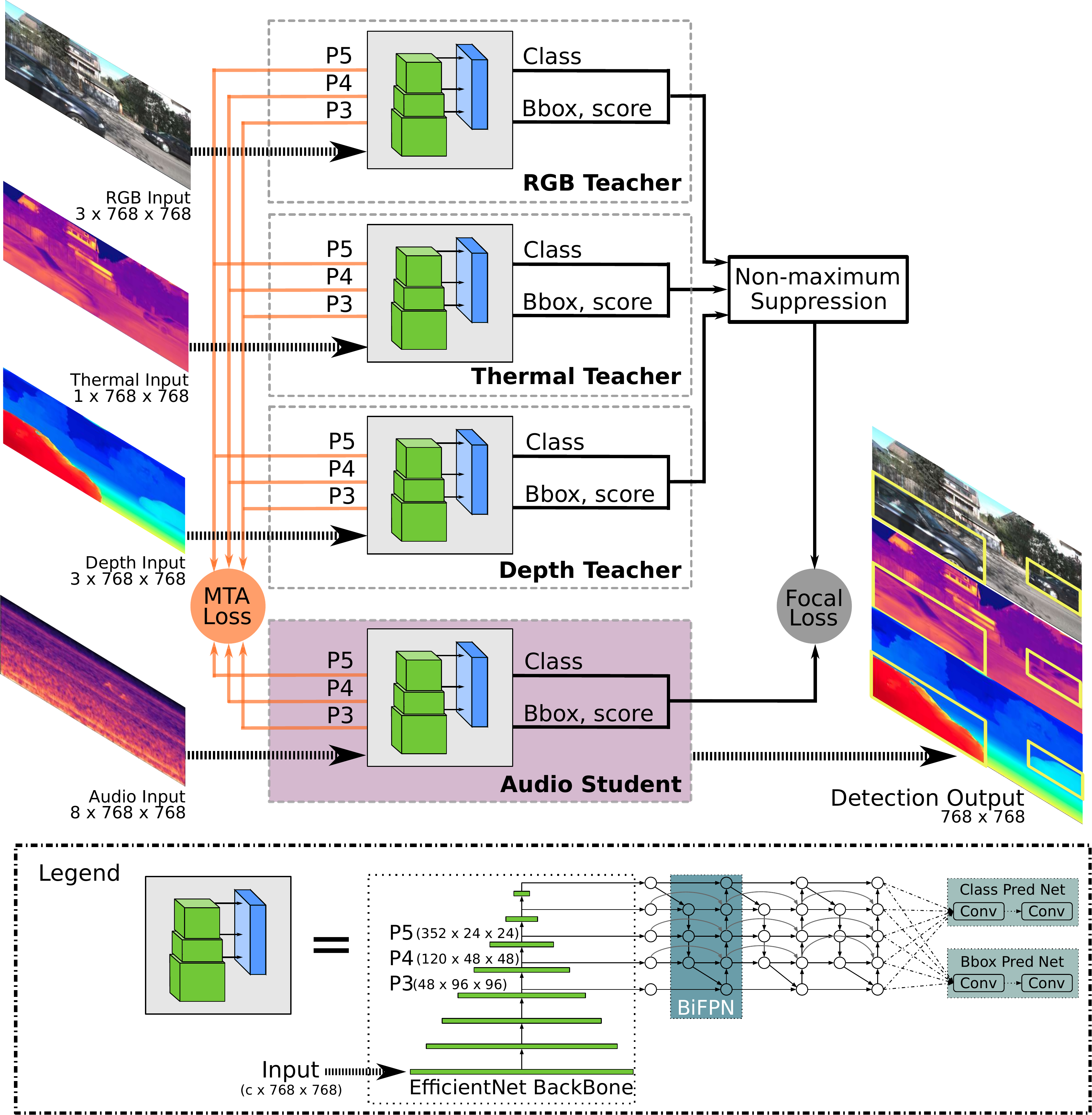}
\caption{Our proposed MM-DistillNet framework consists of three pre-trained modality-specific teachers built upon EfficientDet-D2 that predict bounding boxes in the visual space and an audio student network that takes spectrograms of sound from a microphone array as input. By exploiting the co-occurrence of the modalities, we train the audio student to regress the bounding boxes predicted by the teachers using focal loss and our proposed MTA loss that aligns the intermediate network layers. During inference, the student detects multiple moving vehicles in the visual frame using only sound.}
\label{fig:architecture}
\end{figure}

In this section, we detail our MM-DistillNet framework for distilling the knowledge from a set of pre-trained multimodal teachers into a single student that employs an unlabeled modality as input. We choose RGB, depth, and thermal images as the teacher modalities, and audio from an 8-channel monophonic-microphone array for the student. Specifically, our goal is to learn a mapping from spectrograms of ambient sounds to bounding box coordinates that indicate the vehicle location in the visual space. In our framework illustrated in \figref{fig:architecture}, each pre-trained modality-specific teacher predicts bounding boxes that indicate where the vehicles are located in their respective modality space. These predictions are fused to obtain a single multi-teacher prediction, which is then used as a pseudo label for training the audio student network. To effectively exploit the complementary cues from the modality-specific teachers, we employ our proposed Multi-Teacher Alignment (MTA) loss to align the intermediate representations of the student with that of the teachers. In the rest of this section, we first describe the architecture of the teacher-student networks and the teacher pre-training procedure, followed by the novel pretext task that we propose for better initializing the audio student. We then describe the methodology that we propose for distilling knowledge from multimodal teachers to a single student and finally, the approach we employ to track vehicles over successive frames.\looseness=-1

\subsection{Network Architecture}

We build upon the EfficientDet~\cite{tan2020efficientdet} architecture for the modality-specific teacher networks. EfficientDet has three main components: an EfficientNet~\cite{tan2019efficientnet} backbone, followed by a bidirectional feature pyramid network, and a final regression and classifier branch. The EfficientNet architecture uses multiple stages of mobile inverted bottleneck units~\cite{xie2017aggregated} to extract relevant features from the input data. There are eight variants of this backbone, ranging from B0 to B7 on increasing capacity demand. This allows for trade-off performance and prediction speed, which is achieved through a compound scaling coefficient that uniformly scales the network's depth and width along with the input image resolution. To select from which stages of EfficientNet the features are extracted (and how such features are fused together), EfficientDet introduces a weighted bidirectional feature pyramid through a combination of automatic machine learning and manual tuning. The last stage of the network is a classifier and regressor branch that consist of a sequence of separable convolutions, batch normalization, and a memory-efficient swish~\cite{ramachandran2017searching}.

For this work, we find that EfficientDet-D2 gives us the best speed versus performance trade-off, as demonstrated in the additional experiments in the supplementary material. It is important to note that our framework is not dependent on a specific teacher architecture, as alternate object detection networks can be readily incorporated as drop-in replacements. We use an input image resolution of $768 \times 768$ pixels, with 5 BiFPN cell repetitions, with 112 channels each. We illustrate the EfficientDet architecture in the legend shown in \figref{fig:architecture}. The teacher networks in our framework are comprised of:
\begin{itemize}[noitemsep,topsep=0pt]
  \item \textbf{RGB teacher} that we train on COCO~\cite{lin2014microsoft}, PASCAL~VOC~\cite{everingham2010pascal}, and ImageNet~\cite{deng2009imagenet} for the \emph{car} labels.
  \item \textbf{Depth teacher} that we train on the Argoverse~\cite{chang2019argoverse} dataset using 3D \emph{vehicle} bounding boxes mapped to 2D. Note that Argoverse does not provide direct depth/disparity data. Therefore, we generate it from stereo images using the Guided Aggregation Net~\cite{Zhang2019GANet}.
  \item \textbf{Thermal teacher} that we train on the FLIR~ADAS~\cite{flir} dataset for the \emph{car} and \emph{other vehicle} labels.
\end{itemize}

The audio student network in our MM-DistillNet framework learns to detect vehicles as a regression problem. We adopt the same EfficientDet-D2 topology for the audio student network, which takes eight spectrograms concatenated channel-wise representing the ambient sounds from an 8-channel monophonic-microphone array, as input and predicts bounding boxes localizing vehicles in the visual reference frame. To do so, we first obtain an RGB, depth, and thermal image triplet at a given timestamp, each of which has a resolution of $1920 \times 650$ pixels. Subsequently, we select one second ambient sound clips from the microphone array, centered on the image timestamp and we generate a $80 \times 173$ pixels spectrogram for each of the eight microphones using Short-Time Fourier Transform (STFT). We further detail this procedure in \secref{sec:training}. We then resize the spectrograms to a resolution of $768 \times 768$ pixels to match the input scale of the teachers. Given this 8-channel concatenated spectrograms as input, the audio student yields 4 coordinates ($x_{min}, y_{min}, x_{max}, y_{max}$) for each of the EfficientNet layers at different aspect ratios and scales (EfficientDet uses by default 3 aspect ratios $(1.0, 1.0), (1.4, 0.7), (0.7, 1.4)$ at 3 different scales $[2 ** 0, 2 ** (1.0 / 3.0), 2 ** (2.0 / 3.0)]$.




\subsection{Self-Supervised Pretext Task for Audio Student}

As the input to our audio student network is an 8-channel spectrogram, we cannot leverage pre-trained weights for initializing the EfficientDet architecture, such as from models trained for image detection, which typically take a 3-channel image as the input. It has consistently been shown that models initialized with pre-trained weights from large datasets perform significantly better than models trained from scratch. More recently, self-supervised pretext tasks that learn semantically rich representations by exploiting the supervisory signal from the data itself have shown promising results, even outperforming models initialized with pre-trained weights.

Inspired by this recent progress, we propose a simple pretext task for the audio student that counts the number of cars present in the scene. This task aims at enabling the student to learn audio representations depicting the number of vehicles in the visual field, only using an 8-channel spectrogram as input. To do so, we first use the predictions of multiple pre-trained teachers to identify the number of cars present in the image. Subsequently, we use the corresponding 8-channel spectrogram as the input to EfficientNet with an MLP classifier at its output and we train the network with the cross-entropy loss function to predict the number of cars in the scene. We then use the weights from the model trained on this pretext task to initialize the audio student network in our MM-DistillNet framework while training to detect cars in the visual frame from spectrograms of sound as input.

\subsection{Knowledge Distillation from Multiple Teachers}

To train the audio student network to detect vehicles in the visual frame, given the sound input, we use two different loss functions. First, we employ an object detection loss function at the final prediction of the networks, as shown in \figref{fig:architecture}. Second, we use our Multi-Teacher Alignment (MTA) loss function to align and exploit complementary cues from the intermediate layers of the modality-specific teachers with the audio student. Given that we use multiple teachers, we also obtain multiple sets of bounding box predictions. Each teacher network receives only its input modality and predicts a tuple of bounding boxes, which correspond to their best individual estimation of where the vehicles are located in the visual space. There are often scenes in which each modality-specific teacher predicts a different number of bounding boxes. Therefore, we need to consolidate such predictions. To do so, we obtain three sets of tuples coming from the RGB, depth and thermal teachers, which are consolidated using non-maximum suppression with intersection over union $IoU = 0.5$. This generates a unified prediction from the modality-specific teachers, which is enforced on the student using the Focal loss (FL)~\cite{lin2017focal}. Focal loss is a form of cross-entropy loss with a penalizing parameter that reduces the relative loss for well-classified examples, allowing the network to focus on the training examples that are hard to classify. The Focal loss is given by 
\begin{equation}
\label{eq:focalloss}
L_{focal} = - \alpha (1 - pt)^\gamma * log(pt),
\end{equation}
where $\alpha$ is the weight assigned to hard examples (set to $\alpha=0.25$) and, $\gamma$ is a focusing hyperparameter to balance how much effort to put on hard to classify examples against easy background cases (set to $\gamma=2.0$).

With our proposed MTA loss, we aim to exploit complementary cues contained in the intermediate layers of each modality-specific teacher. In order to achieve this, we train the student network in a manner such that the distribution of activations in specific layers of both the student and the multiple teachers are aligned. Particularly, we enforce the alignment of the $(p3, p4, p5)$ layers of the EfficientNet backbone, as shown in \figref{fig:architecture}. To do so, we compute the distribution of activations using the attention map of each layer normalized to a $[0,1]$ range. We compute the student attention map as $Q_{s}^j = F_{avg}^{r}(A_{s})$, where $F_{avg}$ is a function that collapses the activation tensor $A$ in its channel dimension through the average of the neuron's output at the given layer $j \in \{P3,P4,P5\}$, and $r$ is the exponential over each of the $i-th$ elements of the vector, a hyperparameter that trades-off how much importance to give to high valued activations versus low-valued activations at a given layer.

In the case of the teacher networks, the activation distributions of each modality $P(A_{t_{i}}|m_{i})$ indicates the confidence of each teacher that given an input modality $m_{i}$, the intermediate representations have a high likelihood of detecting a relevant key indicator of a vehicle. With this in mind, we propose to leverage the attention maps of the multiple teachers by means of the product of the modality-specific activation distributions at the selected layers. We assume that the modalities are independent and we use the chain rule of probability so that $P(A_{t_{i}}, ..., A_{t_{N}}|m_{i},...,m_{N}) = P(A_{t_{i}}|m_{i}) * ... * P(A_{t_{N}}|m_{N})$. We believe this assumption is reliable as the teachers have been trained on disjoint datasets, with modalities extracted using different sensor hardware. The intuition behind this idea is to incorporate the knowledge of each modality-specific teacher in an incremental approach. We can consider each pre-trained teacher as a vehicle-sensor engine, and our loss, a mechanism of integrating new measurements in the Bayesian's context. If multiple modalities agree on a bounding box, the probability of this proposal is encouraged. Nevertheless, a modality can also propose a disjoint bounding box with a small probability, allowing the student to learn bounding boxes exclusive to a particular modality. We effectively estimate the probability of detecting a car in a scene, given the privileged knowledge of each modality. This allows for the flexibility to also incorporate other knowledge as confidence scores for each bounding box, so that we reduce the occurrence of false predictions. Therefore, we compute the multi-teacher attention map as $Q_{t}^j = \prod_{i}^{N} F_{avg}^{r}(A_{t_{i}})$, where $i$ denotes each of the $N$ considered modalities. Formally, we define our Multi-Teacher Alignment (MTA) loss as
\begin{equation}
L_{MTA} = \beta * \sum_{j} KL_{div}\left(\frac{Q_{s}^j}{\left \|  {Q_{s}^j}\right \|_{2}}, \frac{Q_{t}^j}{\left \| {Q_{t}^j}\right \|_{2}}\right),
\end{equation}
where the summation iterates over each of the selected EfficientNet layers from the inverted pyramid (e.g., p3, p4, and p5 layers), $s$ and $t$ stand for student and teacher, and $\beta=0.5$ is used for loss balancing. We denote this loss function as Multi-Teacher Alignment loss, as it integrates different modalities privileging the agreement of different inputs while still considering vehicle predictions proposed by one modality. Finally, we optimize our MM-DistillNet framework with the weighted summation of the focal loss and our proposed MTA loss as
\begin{equation}
\label{eq:totalloss}
L_{total} = \delta * L_{focal} + \omega * L_{MTA},
\end{equation}
where $L_{total}$ enforces knowledge transfer from the teachers at the output, as well as at the intermediate network layers.

\subsection{Tracking}

We adopt an approach similar to that of Gan~\textit{et~al.}~\cite{gan2019self} for object tracking. Specifically, we leverage the detected bounding boxes, and we use the IoU values between boxes of consecutive frames to relate the objects to the same tracklet. We set the IoU threshold to 0.5 to assign two bounding boxes from different timesteps to the same object. We initialize a tracklet each time an object is detected with a confidence score higher than $0.8$. The next bounding box related to that tracklet is selected by comparing it to the current frame's detection. The association process between the tracklet and a bounding box is made so that it maximizes the IoU. The tracklet is set as inactive if there are no bounding boxes with $IoU > 0.5$ in the subsequent frames. Given that our contribution with multiple modality-specific teachers improves the quality of the bounding boxes and the number of detected objects, we also expect to enhance the tracking accuracy with this method that primarily relies on IoU object matching.

\section{Experimental Evaluation}
\label{sec:experiments}

In this section, we first describe the data collection methodology that we employ, followed by the protocol that we use for training the MM-DistillNet framework. We then present quantitative results comparing our approach to several strong baselines as well as the state-of-the-art. Subsequently, we present detailed ablation studies and qualitative evaluations to demonstrate the novelty of our contributions.

\subsection{Multimodal Audio-Visual Detection Dataset}
\label{sec:dataset}

As there are no publicly available datasets that consist of synchronized audio, RGB, depth, and thermal images, we collected a large-scale Multimodal Audio-Visual Detection (MAVD) dataset in autonomous driving scenarios. The dataset was gathered from 24 car drives during 3 months and at 20 different locations. Each drive has an average of half hour duration. We recorded data on diverse scenarios ranging from highways to densely populated urban areas and small towns. The recordings consist of high traffic density, freeway driving, and multiple traffic lights (involving transition from static to driving conditions). To capture diverse noise conditions, we recorded sounds not only during conventional city driving but also near trams and while going through tunnels.\looseness=-1

\begin{figure}
\footnotesize
\centering
\setlength{\tabcolsep}{0.03cm}
{\renewcommand{\arraystretch}{0.5}
\begin{tabular}{p{2.7cm}p{2.7cm}p{2.7cm}}
      \includegraphics[width=\linewidth]{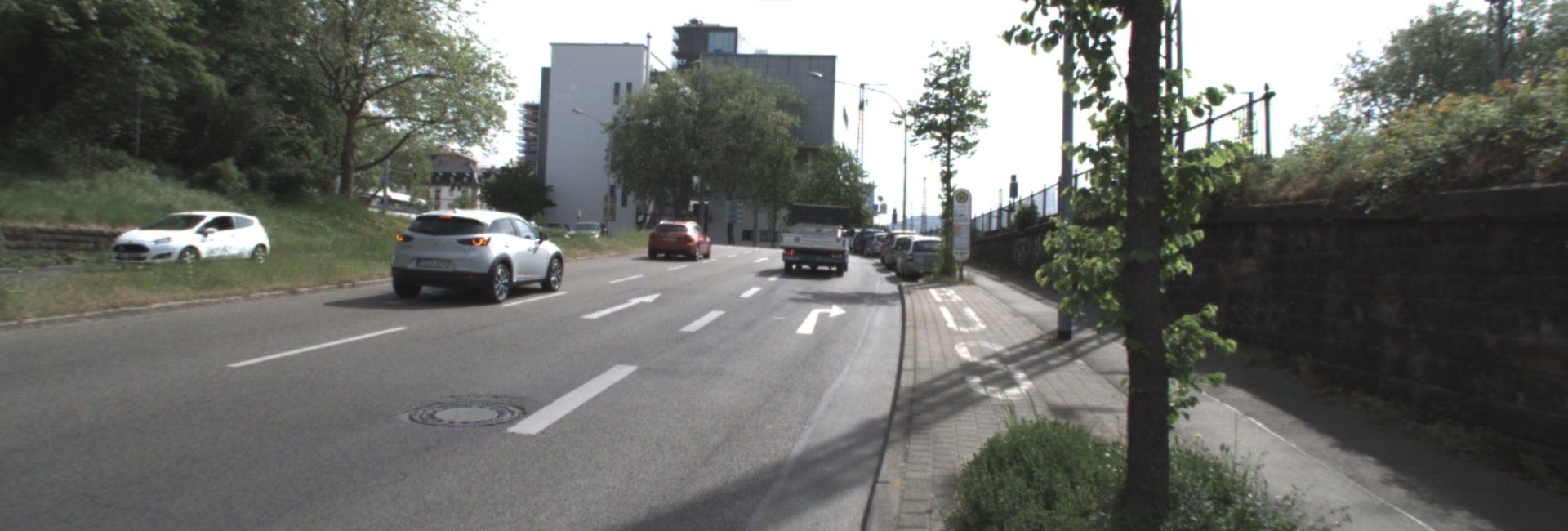} &
      \includegraphics[width=\linewidth]{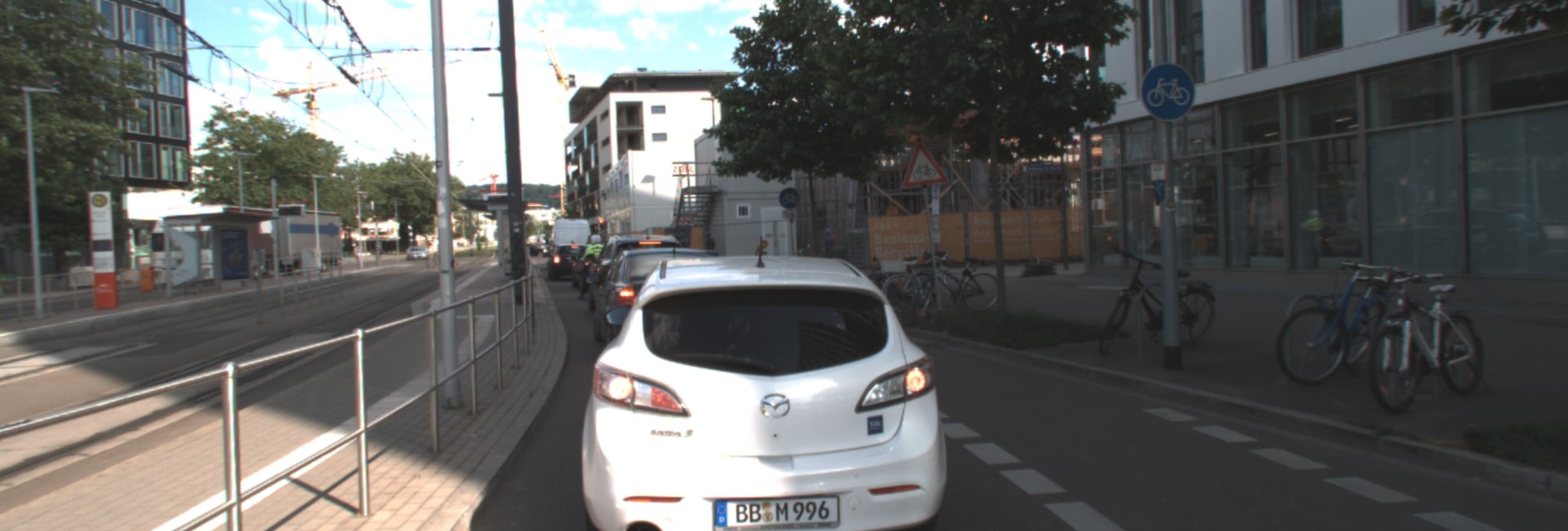} &
      \includegraphics[width=\linewidth]{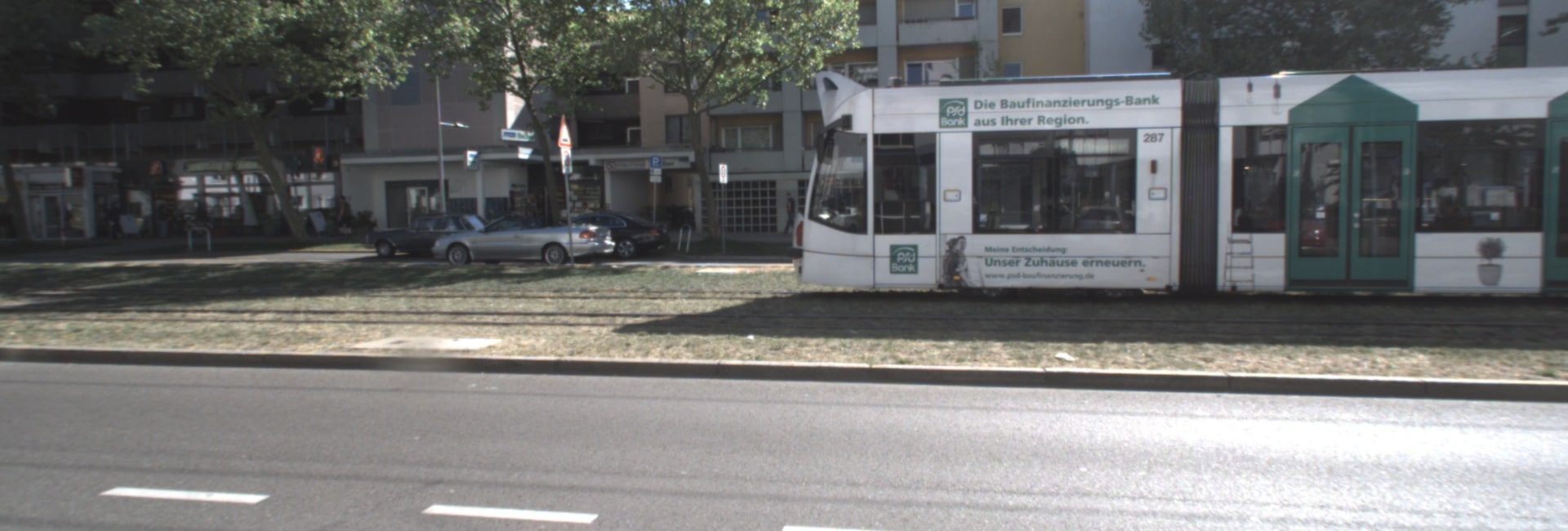} \\ 
      \includegraphics[width=\linewidth]{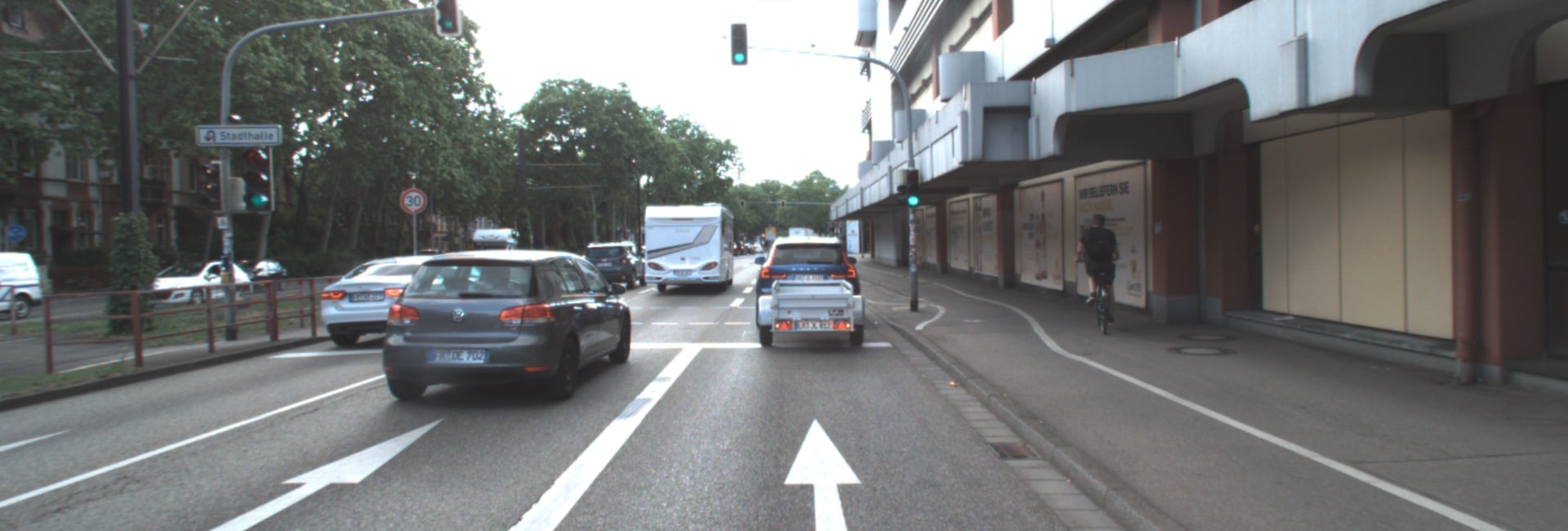} &
      \includegraphics[width=\linewidth]{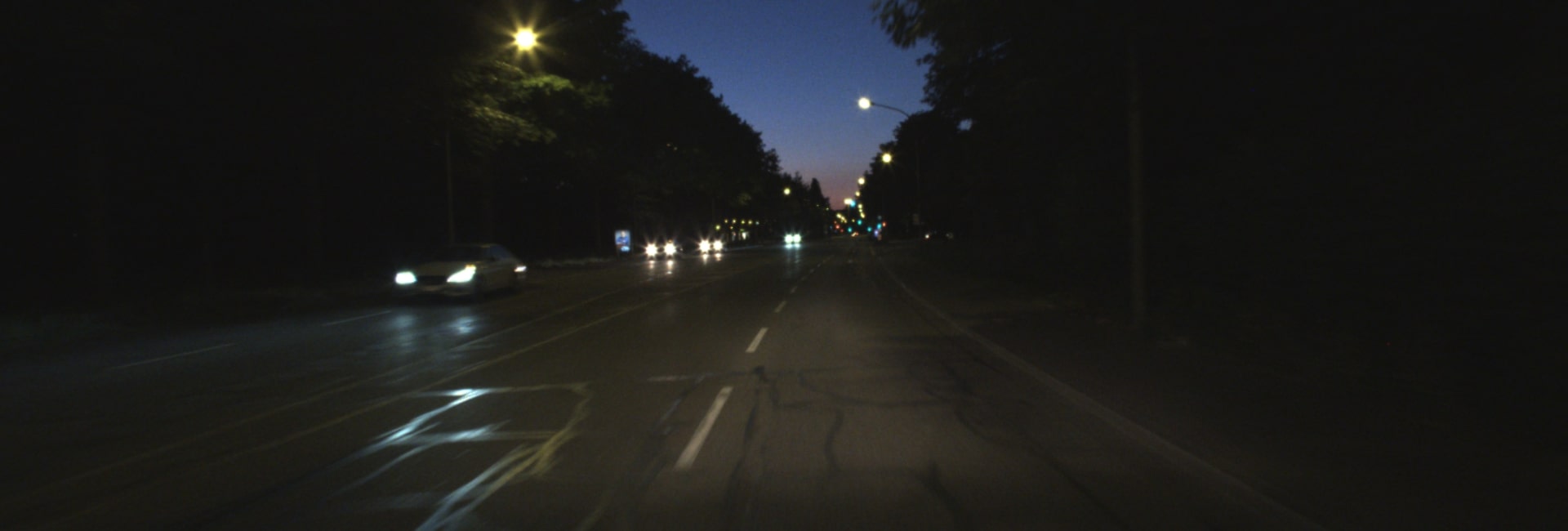} &
      \includegraphics[width=\linewidth]{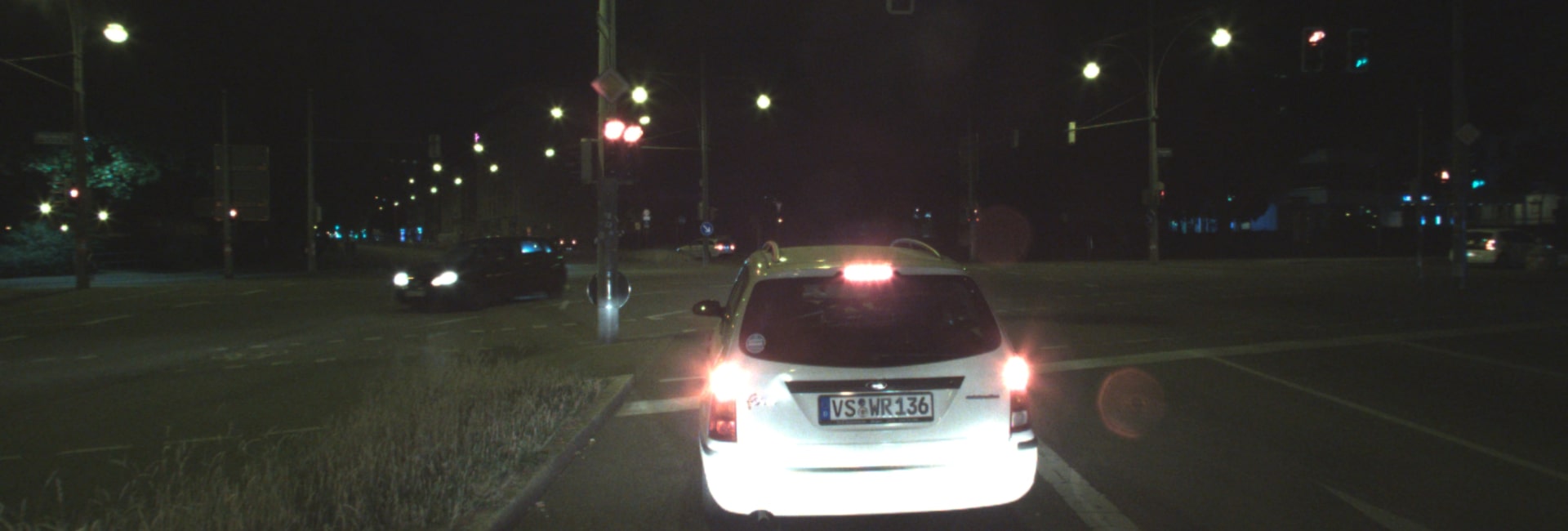} \\
      \includegraphics[width=\linewidth]{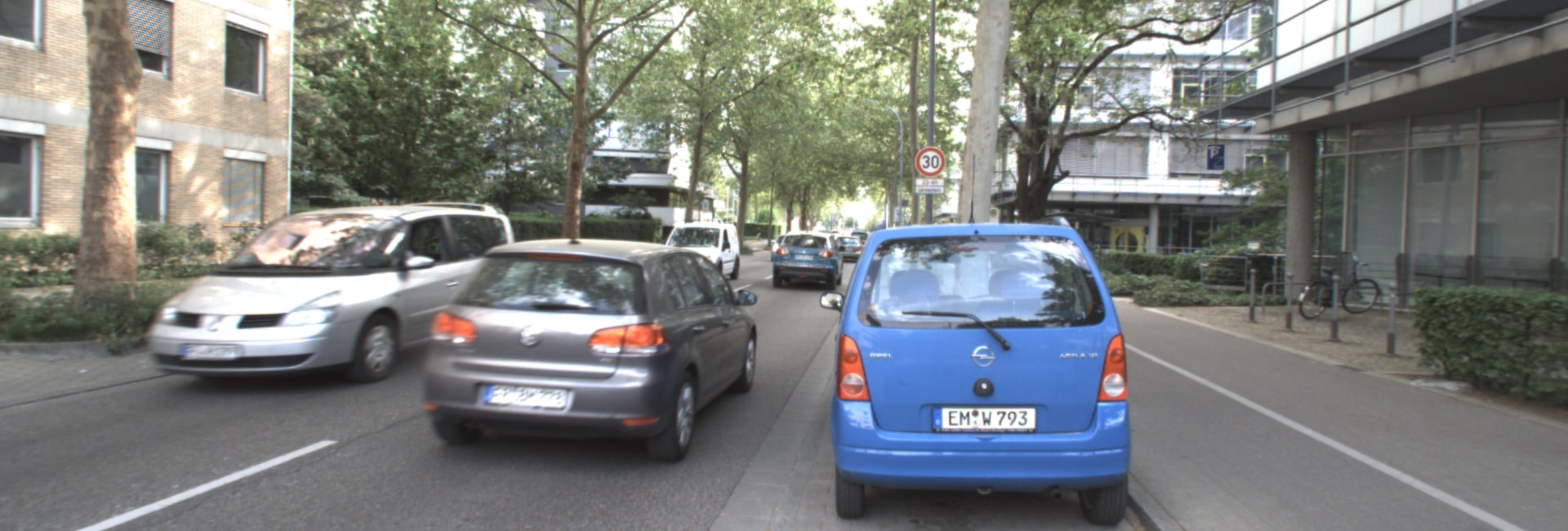} &
      \includegraphics[width=\linewidth]{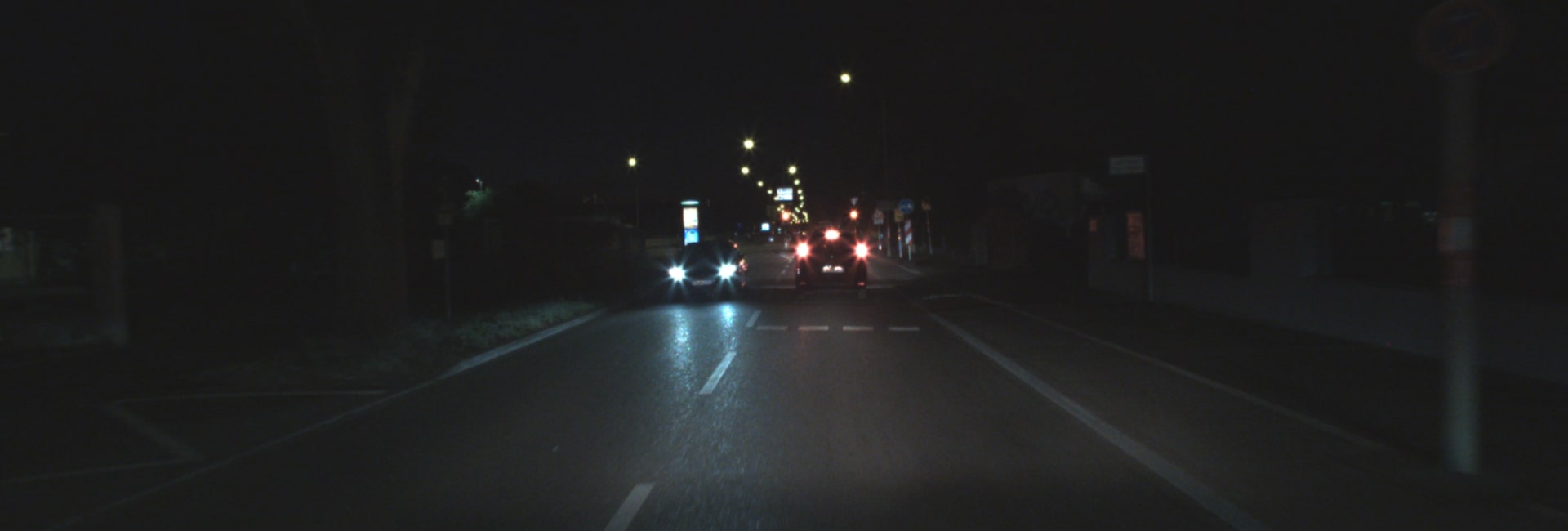} &
      \includegraphics[width=\linewidth]{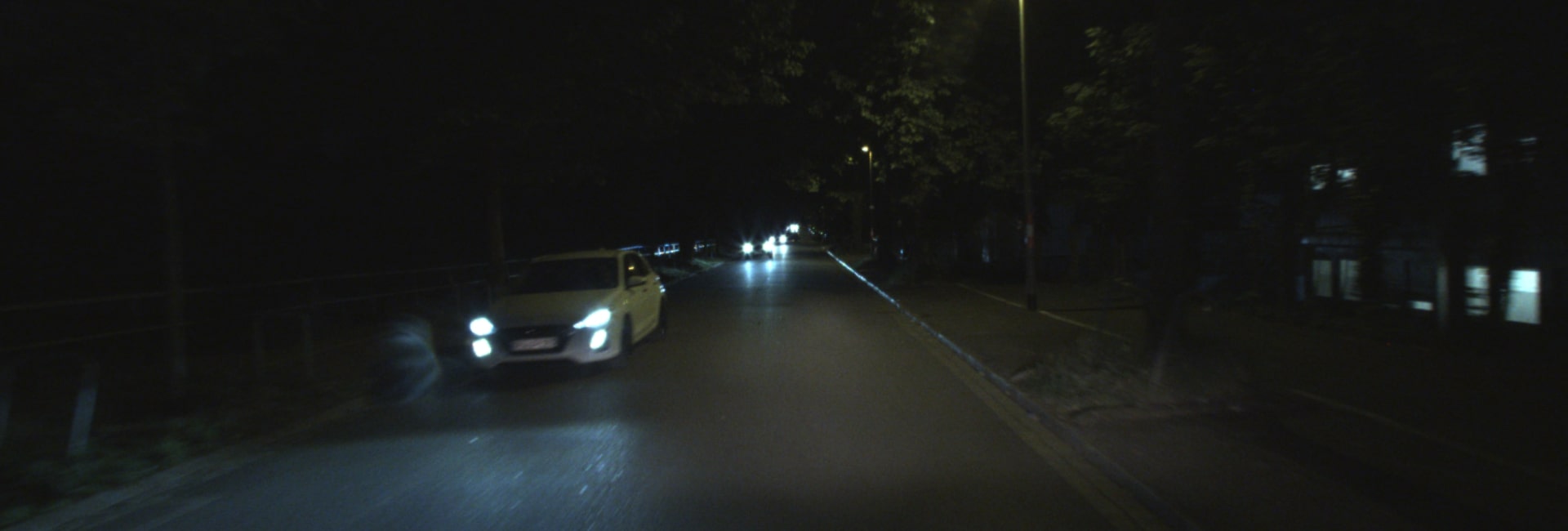} \\
\end{tabular}}
\caption{Example images from our MAVD dataset showing diverse scenes with multiple moving vehicles and low-illumination conditions captured with a camera mounted on a moving car.\looseness=-1}
\label{fig:dataset}
\end{figure}

We provide two types of scenarios, static condition in which the car is motionless and nearly 300 km of driving data. Our dataset contains three cars on average for every image (ranging from 1 to a maximum of 13 cars per scene). We only retained the images with at least one car in the scene. The subset that we use for training the detection stage contains 24589 static day images, 26901 static night images, 26357 day driving images, and 35436 night driving images, amounting to a total of 113283 synchronized multi-channel audio, RGB, depth, and thermal modalities. Additionally, the dataset also contains GPS/IMU data and LiDAR point clouds. An image showing the data collection vehicle and the sensor setup is shown in the supplementary material. The sensors that we used include an RGB stereo camera rig (FLIR Blackfly 23S3C), a thermal stereo camera rig (FLIR ADK), and eight monophonic microphones in an octagon array. The audio was recorded and stored in the 1-channel Microsoft WAVE format with a sampling rate of 44100 Hz. All the sensor data, including the microphone recordings were synchronized to each other via the GPS clock. Example scenes from the dataset are shown in \figref{fig:dataset}.

\subsection{Training Protocol}
\label{sec:training}

\textbf{Data Split}: We use a $60/20/20\%$ split for training, validation, and testing. The validation split was used to perform hyperparameter optimization with Hyperband~\cite{falkner2018bohb}.

\textbf{Evaluation Metric}: We use the standard mean average precision metric for evaluating object detection performance and the center distance proposed by Gan~\textit{~et~al.}~\cite{gan2019self}. Mean average precision is the mean over classes of the interpolated area under each class's precision and recall curve. The Center distance $CDx$ and $CDy$ metrics indicate the prediction accuracy because the spatial information is not directly available for the audio (possible error between the predicted bounding box center and the groundtruth).

\textbf{Training Setup}: We train for 50 epochs with ReduceLRonPlateau learning rate scheduler and an initial learning rate of $1e-5$ , weight decay of $5e-4$, $betas=(0.9, 0.999)$, and Adam optimizer. For the MTA loss, we use $r=2.0$ and $temperature=9.0$ as this selection of hyperparameters provided the best results for individual modalities. We provide additional details in the Supplementary Material. For our loss calculation, we set $\delta=1.0 $ and $\omega = 0.05$ as these settings provided the best performances (more details can be found in the Supplementary Material). The original resolution of all RGB/depth/thermal images is 1920$\times$650. We resize them to be 768$\times$768 as per \cite{tan2020efficientdet} D2 variant. For the audio, we extract 0.5 seconds before and 0.5 seconds after the registered timestamp, an RGB image was taken. We normalize this 1-second raw waveform and further resample it on a Mel-frequency scale with 80 bins resulting in 8 $(80, 173)$ arrays. This is further normalized to [0-1] and re-scaled to 768$\times$768$\times$8 dimensionality.

\subsection{Quantitative Results}

In order to evaluate the performance of multi-teacher distillation from different modalities, we compare the performance of our MM-DistillNet with \textit{StereoSoundNet}~\cite{gan2019self} which uses a single RGB teacher with the Ranking loss to distill the information into an audio student network. We also compare with several strong baselines: \textit{2M-DistillNet Audio} employs a single RGB teacher with our proposed MTA loss to train an audio student network. The comparison with this baseline enables us to evaluate the performance of our proposed MTA loss over the Ranking loss. In order to evaluate the performance of using other modalities representing an object in the student network, we compare with \textit{2M-DistillNet Depth} and \textit{2M-DistillNet Thermal} that use an RGB teacher to train a depth student or a thermal student, respectively using our MTA loss. The comparison with these two models shows the significance of using the audio modality in the student network. Finally, we compare with \textit{MM-DistillNet Avg} that uses a straightforward approach to combine the predictions from RGB, depth, and thermal teachers by averaging the individual modality-specific network activations. Here, we assume that all bounding boxes predicted by any of the modalities are valid (after applying non-maximum suppression with IoU=0.5). The comparison with this baseline demonstrates the utility of our MTA loss function to effectively distill multimodal knowledge from the teaches. All the aforementioned baselines use the 8-channel spectrogram from the microphone array as input and are trained to perform multi-object detection.

We present quantitative comparisons using the same EfficientDet-D2 topology with pre-trained weights as detailed in \secref{sec:technical} for all the baselines, as well as our MM-DistillNet model. Results from this experiment is shown in \tabref{table:results}. We can observe how the knowledge of different teachers improves the performance over the previous state-of-the-art \textit{StereoSoundNet}~\cite{gan2019self}, using the same input (audio only). Furthermore, our baseline \textit{2M-DistillNet Audio}, which also uses an RGB-teacher to train an audio student, yields superior performance than StereoSoundNet. This demonstrates that our MTA loss function outperforms the Ranking loss. \tabref{table:results} also elucidates that audio is a valuable modality to detect moving vehicles. We also observe that combining the prediction of individual RGB, depth, and thermal teachers using averaging does not improve the performance. Nevertheless, we can see that our proposed MM-DistillNet with our MTA loss function exploits complementary cues from the multimodal teachers and facilitates effective distillation.

\begin{table}
\begin{center}
\footnotesize
\begin{tabular}{p{2.9cm}|p{0.65cm}p{0.65cm}p{0.65cm}p{0.4cm}p{0.4cm}}
\toprule
Network & mAP@ & mAP@ & mAP@ & CDx & CDx \\
& Avg & 0.5 & 0.75 \\
\noalign{\smallskip}\hline\hline\noalign{\smallskip}
StereoSoundNet~\cite{gan2019self} & $44.05$ & $62.38$ & $41.46$ & $3.00$ & $2.24$ \\
2M-DistillNet RGB & $57.25$ & $68.01$ & $59.15$ & $2.67$ & $2.13$ \\
2M-DistillNet Depth & $55.41$ & $66.83$ & $57.30$ & $2.60$ & $2.10$ \\
2M-DistillNet Thermal & $56.70$ & $69.15$ & $58.63$ & $2.43$ & $1.98$ \\
MM-DistillNet Avg & $51.63$ & $66.14$ & $52.24$ & $2.14$ & $1.80$ \\
\midrule
MM-DistillNet (Ours) & $\mathbf{61.62}$ & $\mathbf{84.29}$ & $\mathbf{59.66}$ & $\mathbf{1.27}$ & $\mathbf{0.69}$ \\
\bottomrule
\end{tabular}
\end{center}
\caption{Comparison of cross-modal multi-object detection performance with several baselines. '2M-DistillNet Teacher' refers to 2-modal distillation approach to train the audio student using our MTA loss. 'MM-DistillNet Avg' refers to averaging individual modality-specific teacher activations.}
\label{table:results}
\end{table}

\begin{figure*}
\footnotesize
\centering
\setlength{\tabcolsep}{0.05cm}
{\renewcommand{\arraystretch}{0.8}
\begin{tabular}{P{3.39cm}P{3.39cm}P{3.39cm}P{3.39cm}P{3.39cm}}
RGB Teacher & Depth Teacher & Thermal Teacher & StereoSoundNet~\cite{wang2020score} & MM-DistillNet (Ours) \\
    \includegraphics[width=\linewidth]{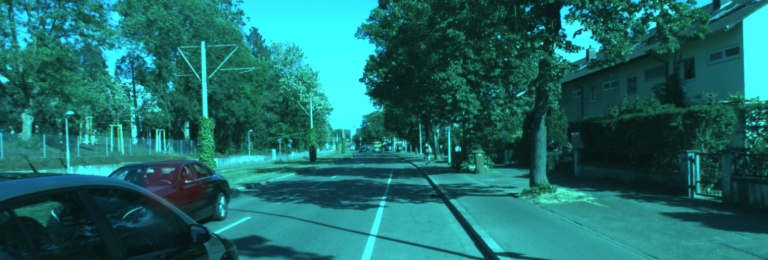} &  \includegraphics[width=\linewidth]{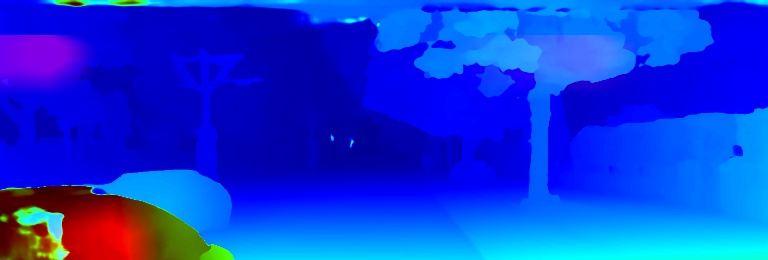} & \includegraphics[width=\linewidth]{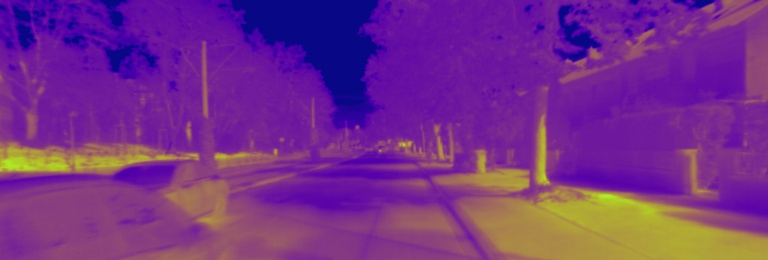} & \includegraphics[width=\linewidth]{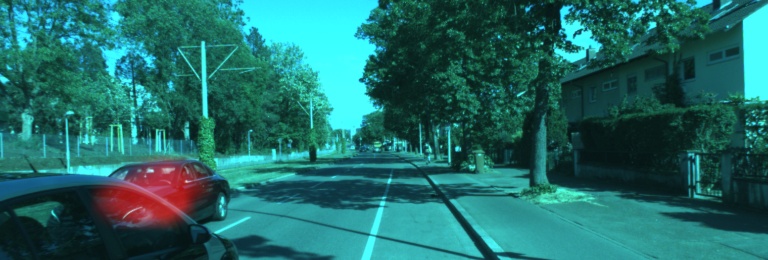} & \includegraphics[width=\linewidth]{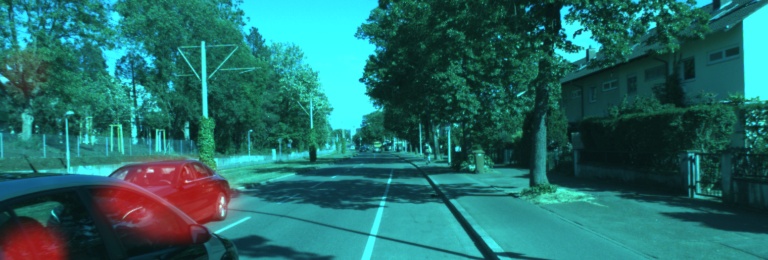} \\
    \includegraphics[width=\linewidth]{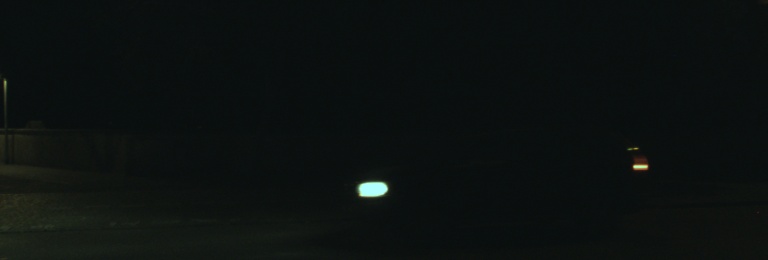} &  \includegraphics[width=\linewidth]{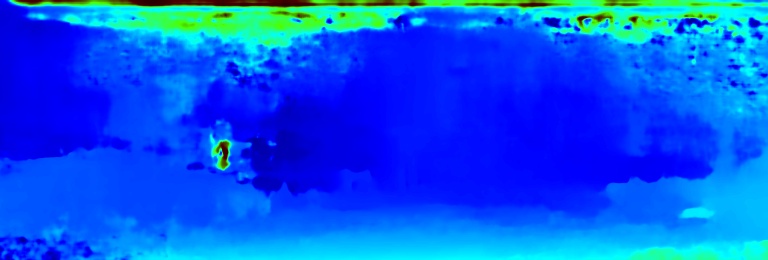} & \includegraphics[width=\linewidth]{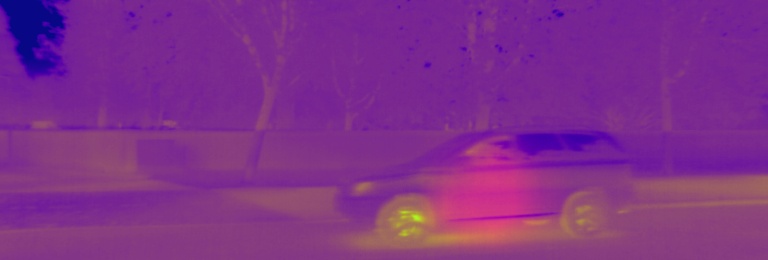} & \includegraphics[width=\linewidth]{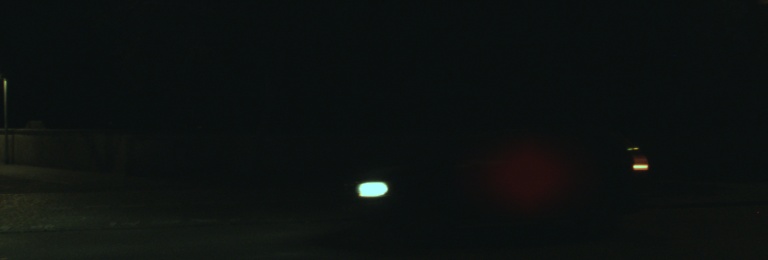} & \includegraphics[width=\linewidth]{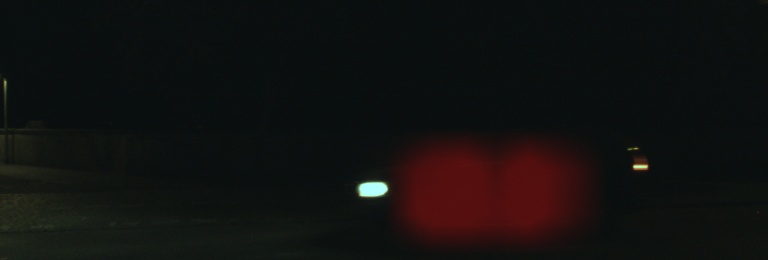} \\
\end{tabular}}
\caption{Comparison of activations visualization from modality-specific teachers with the previous state-of-the-art StereoSoundNet~\cite{wang2020score}, and our proposed MM-DistillNet. High activations (in red) indicate regions where a vehicle is likely to be detected.}
\label{fig:activationpic}
\end{figure*}

We also evaluate the performance of our MTA loss against other knowledge distillation techniques. \tabref{table:results_loss} compares the baseline loss used by \cite{gan2019self}, as well as the pairwise loss conventionally used for similar embedding task \cite{alayrac2020self} and an attention based loss with learnable parameters~\cite{wang2019pay}. We use the audio student and a single RGB teacher. Our loss is intended to distill knowledge from multiple teachers into a single student, yet, in the individual teacher case, it provides appealing results in the object detection task. In the supplementary material, we provide further comparisons against different modalities. Additionally, for multiple teachers, previous methods such as \cite{zhou2019m2kd} proposed to average the predictions of multiple teachers. \tabref{table:results_loss} shows the effect of integrating the prediction of the teachers rather than computing an average. Furthermore, \figref{fig:activationpic} provides an intuition on how the predictions of multiple teachers are integrated into the student by visualizing the activations. We obtain the activation using Score-CAM~\cite{wang2020score} from the P5 layer of EfficientNet. In particular, we show that the baseline's activations are linked only to the RGB teacher, whereas our method's activation is the product of the activation of the RGB, depth and thermal teachers. Furthermore, the thermal teacher in this night setting is the privileged modality which is able to predict cars under poor light conditions. In \tabref{table:tracking}, we report the multiple object tracking accuracy (MOTA), identity switches (ID Sw.), fragment (Frag.), false positive (FP) and false negative (FN) as evaluation metrics \cite{bernardin2008evaluating, li2009learning} on a subset of our test dataset. We excluded scenes that involved tracking of multiple cars for a fair comparison with StereoSoundNet.\looseness=-1

\begin{table}
\begin{center}
\footnotesize
\begin{tabular}{p{2.2cm}|p{0.5cm}p{0.6cm}p{0.6cm}p{0.6cm}p{0.4cm}p{0.4cm}}
\toprule
Loss Function & KD & mAP@ Avg & mAP@ 0.5 & mAP@ 0.75 & CDx & CDx \\
\noalign{\smallskip}\hline\hline\noalign{\smallskip}
Ranking loss~\cite{gan2019self} & RGB & $44.05$ & $62.38$ & $41.46$ & $3.00$ & $2.24$ \\
Pairwise loss~\cite{liu2020structured} & RGB & $40.45$  & $59.72$ & $36.73$ & $2.98$ & $2.20$ \\
AFD loss~\cite{wang2019pay} & RGB & $44.27$ & $62.00$ & $41.90$ & $3.19$ & $2.28$ \\
\midrule
Avg. Ranking loss & R,D,T & $56.16$ & $80.03$ & $52.96$ & $1.46$ & $0.80$ \\
Avg. AFD loss & R,D,T & $58.50$  & $82.18$ & $55.48$ & $1.30$ & $0.70$ \\
Avg. MTA loss & R,D,T & $59.46$ & $82.29$ & $56.94$ & $1.35$ & $0.73$ \\
\midrule
MTA loss (Ours) & RGB & $44.58$ & $62.66$ & $42.39$ & $2.94$ & $2.17$ \\
MTA loss (Ours) & R,D,T &  $\mathbf{61.62}$ & $\mathbf{84.29}$ & $\mathbf{59.66}$ & $\mathbf{1.27}$ & $\mathbf{0.69}$ \\
\bottomrule
\end{tabular}
\end{center}
\caption{Comparison of various loss functions for Knowledge Distilation (KD). All the models were trained with the same MM-DistillNet architecture but with different loss functions. `R,D,T' refers to RGB, Depth, and Thermal teachers. Avg. Loss averages the individual modality-specific teacher activations.} 
\label{table:results_loss}
\end{table}


\begin{table}
\begin{center}
\footnotesize
\begin{tabular}{p{2.5cm}|p{0.8cm}p{0.9cm}p{0.6cm}p{0.35cm}p{0.45cm}}
\toprule
Approach & MOTA$\uparrow$ & ID Sw.$\downarrow$ & Frag.$\downarrow$ & FP$\downarrow$ & FN$\downarrow$ \\
\noalign{\smallskip}\hline\hline\noalign{\smallskip}
StereoSoundNet~\cite{gan2019self} & $16.94\%$ & $1327$ & $1077$ & $3696$ & $3349$ \\
MM-DistillNet (Ours) & $\mathbf{26.96\%}$ & $\mathbf{1078}$ & $\mathbf{1076}$ & $\mathbf{2758}$ & $\mathbf{3524}$ \\
\bottomrule
\end{tabular}
\end{center}
\caption{Comparison of tracking performance.}
\label{table:tracking}
\end{table}

\subsection{Ablation Studies}
\label{sec:ablation}

\begin{table}
\begin{center}
\footnotesize
\begin{tabular}{p{0.4cm}p{2.45cm}P{0.8cm}|p{0.65cm}p{0.65cm}p{0.6cm}}
\toprule
Model & Teacher & Student & mAP@ & AP@ & AP@ \\
& Modalities & Pretext & Avg & 0.5 & 0.75 \\
\noalign{\smallskip}\hline\hline\noalign{\smallskip}
M1 & RGB & - & $44.58$ & $62.66$ & $42.38$  \\
M2 & RGB, Depth & - & $42.89$ &  $62.07$ &  $39.67$  \\
M3 & RGB, Thermal & - & $55.81$ &   $79.84$ &  $54.67$   \\
M4 & Depth, Thermal & - & $44.79$ &   $65.14$ &  $41.82$   \\
M5 & RGB, Depth, Thermal & - & $61.10$ & $83.81$ & $59.07$  \\
M6 & RGB, Depth, Thermal & \checkmark & $\mathbf{61.62}$ & $\mathbf{84.29}$ & $\mathbf{59.66}$ \\ \bottomrule
\end{tabular}
\end{center}
\caption{Ablation study on influence of various modality-specific teachers and self-supervised pretext task of audio student.}
\label{table:ablation}
\end{table}

\begin{figure*}
\footnotesize
\centering
\setlength{\tabcolsep}{0.05cm}
{\renewcommand{\arraystretch}{0.8}
\begin{tabular}{P{3.39cm}P{3.39cm}P{3.39cm}P{3.39cm}P{3.39cm}}
     RGB Teacher & Depth Teacher & Thermal Teacher & StereoSoundNet~\cite{wang2020score} & MM-DistillNet (Ours) \\
     \includegraphics[width=\linewidth]{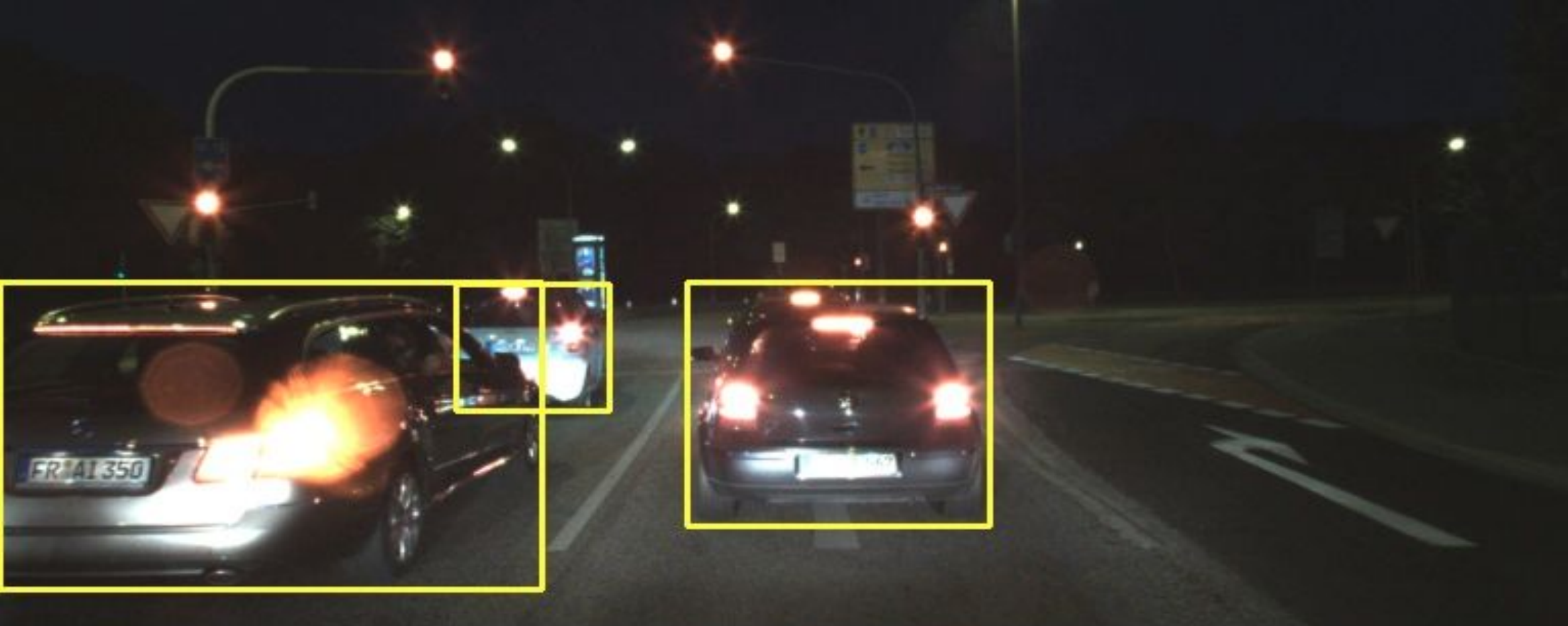} &  \includegraphics[width=\linewidth]{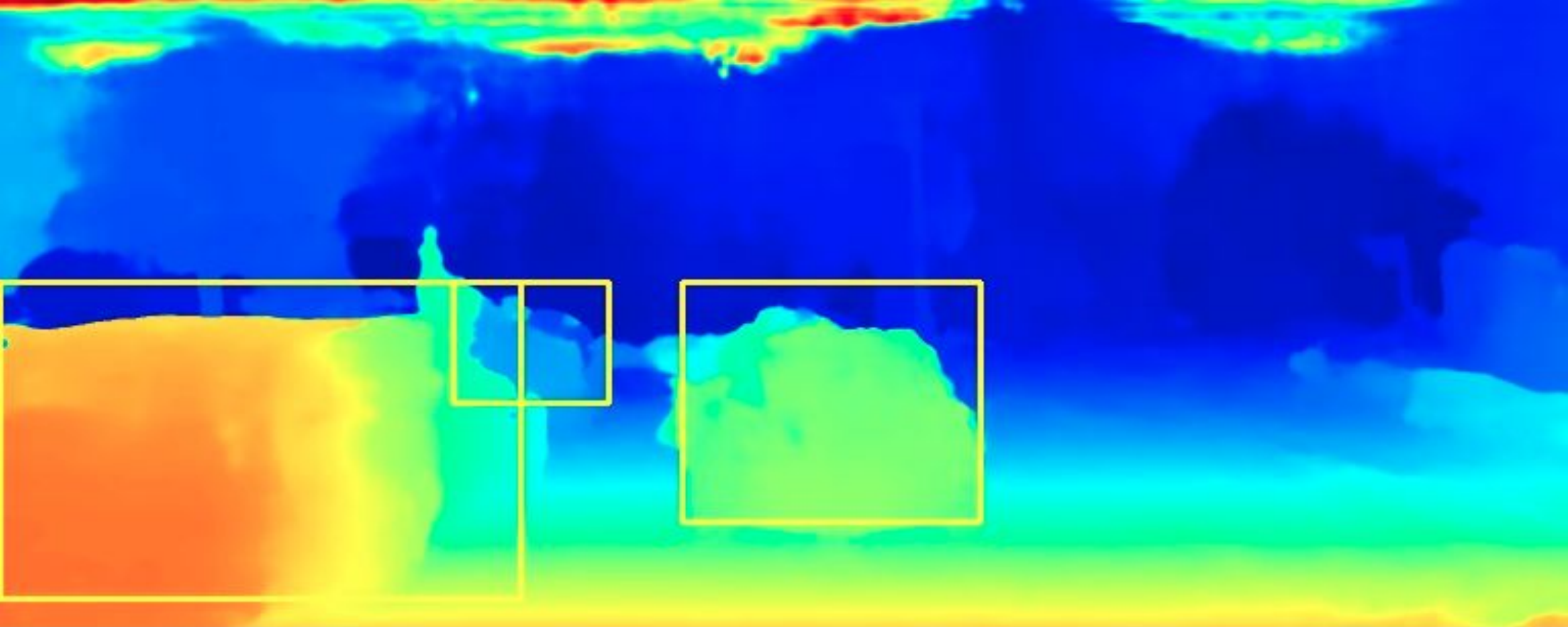} & \includegraphics[width=\linewidth]{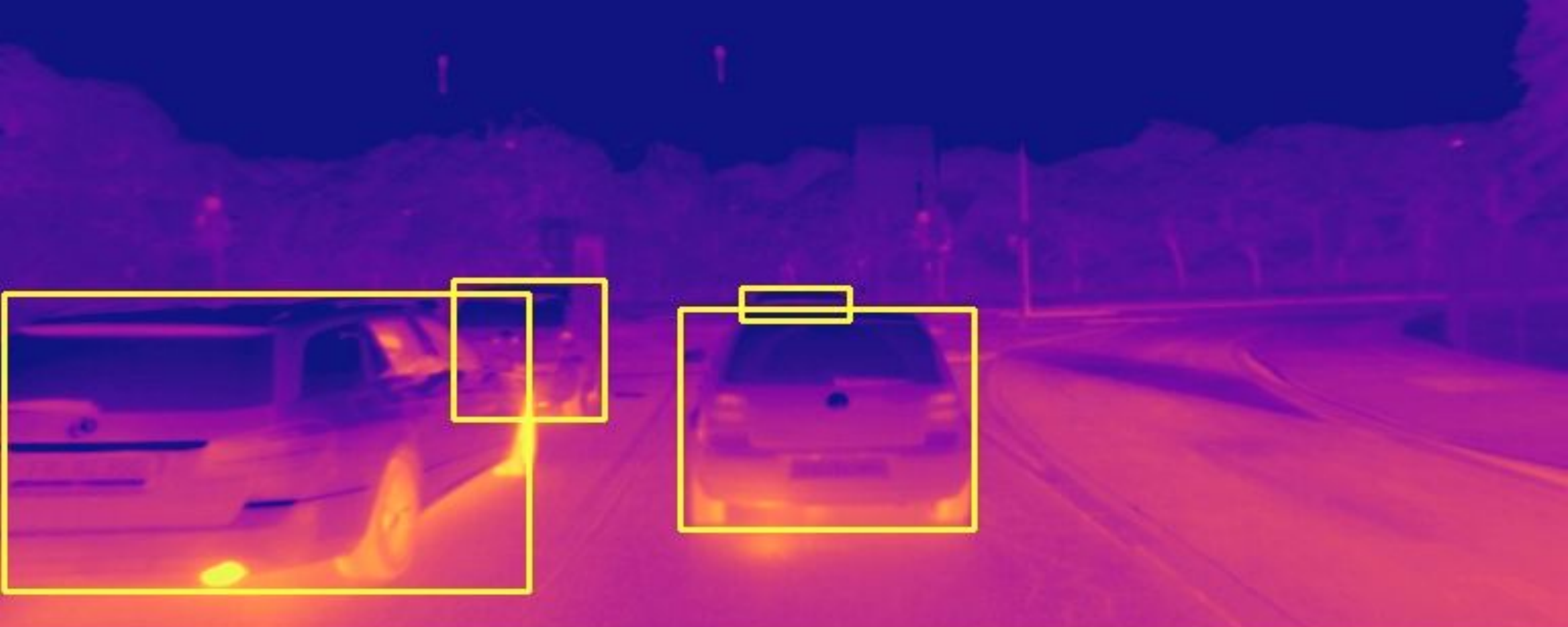} & \includegraphics[width=\linewidth]{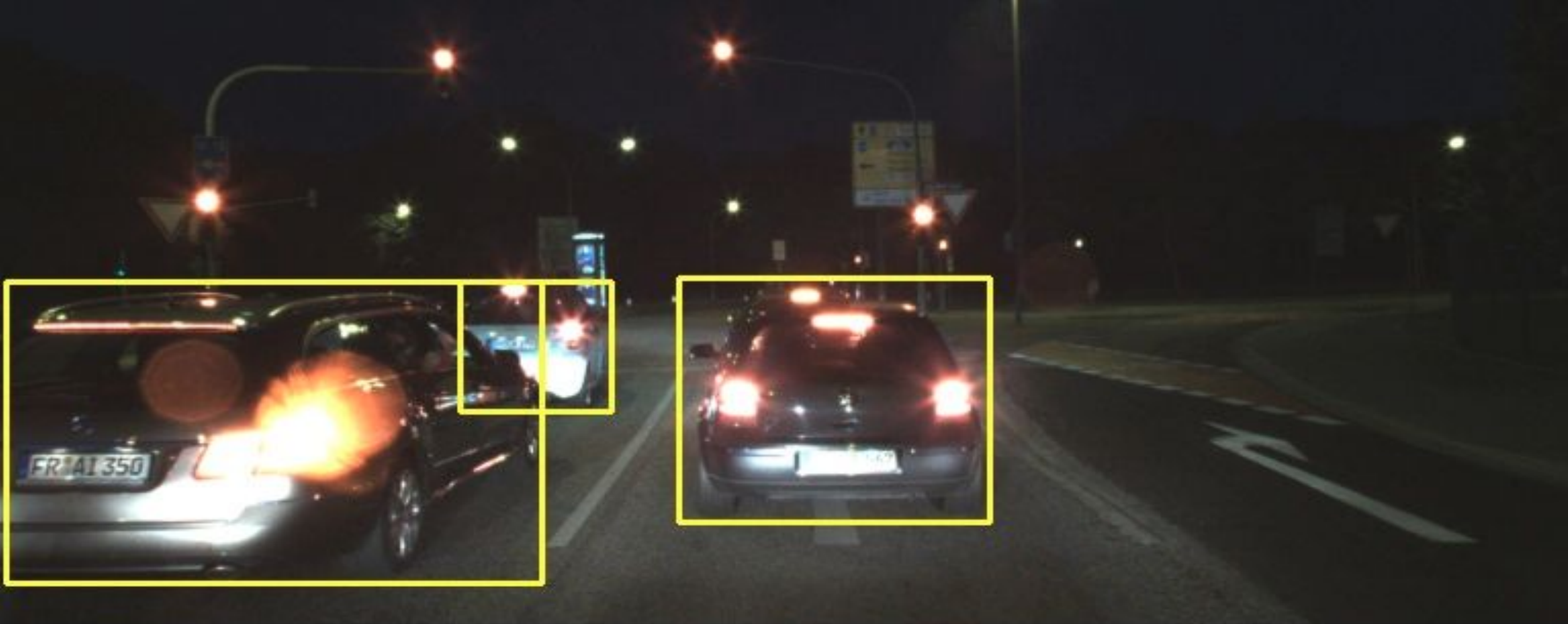} & \includegraphics[width=\linewidth]{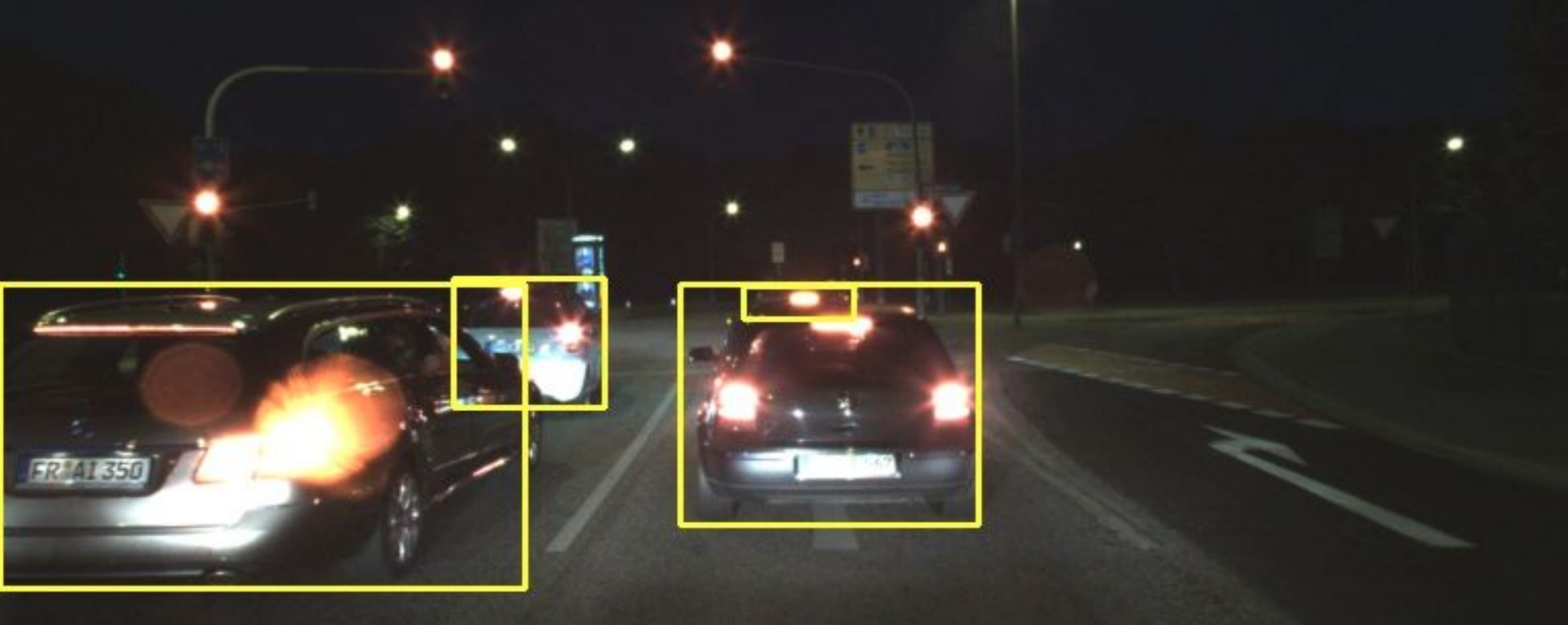} \\
     \includegraphics[width=\linewidth]{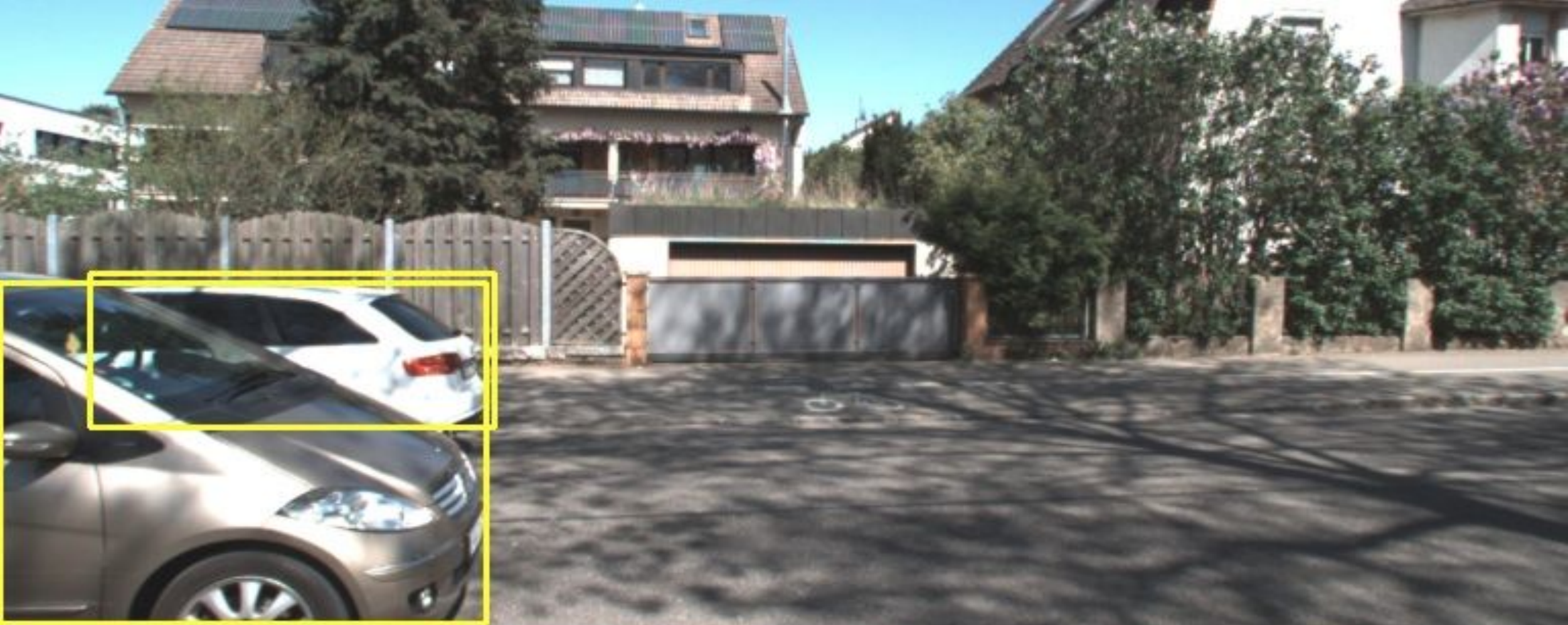} & \includegraphics[width=\linewidth]{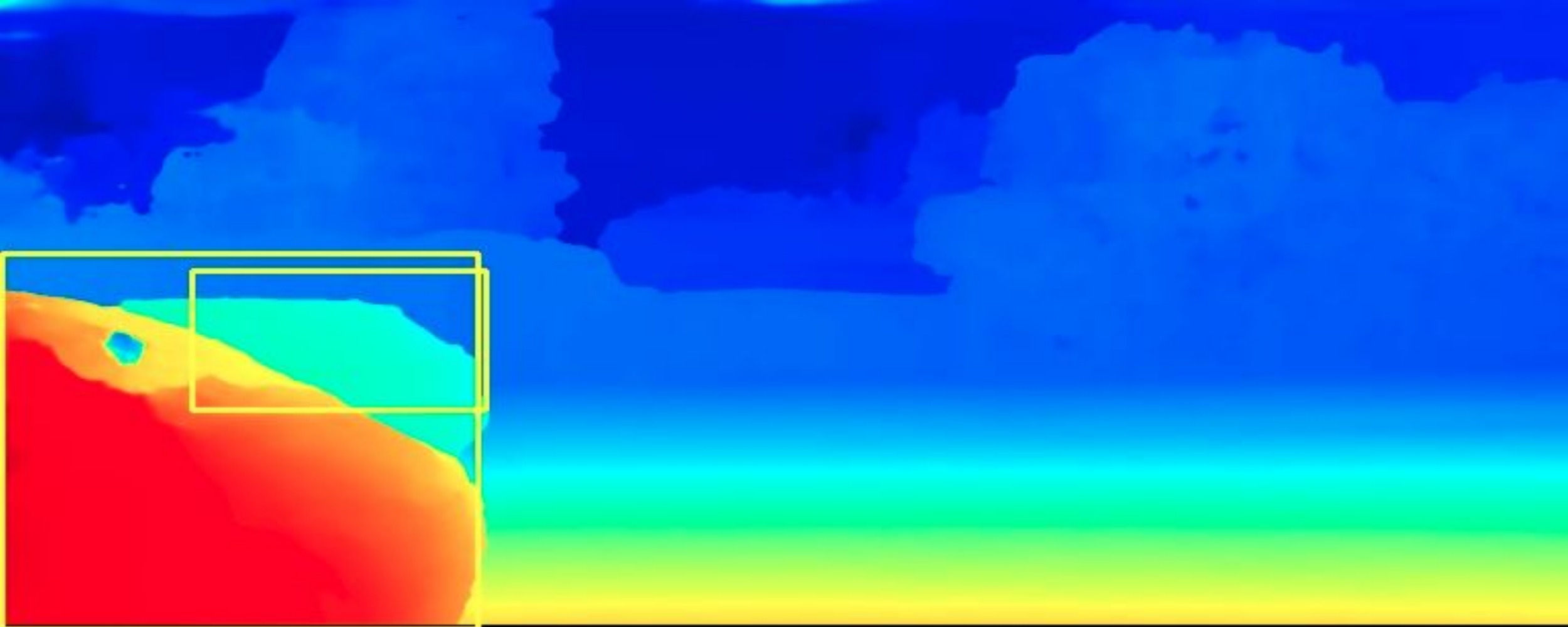} & \includegraphics[width=\linewidth]{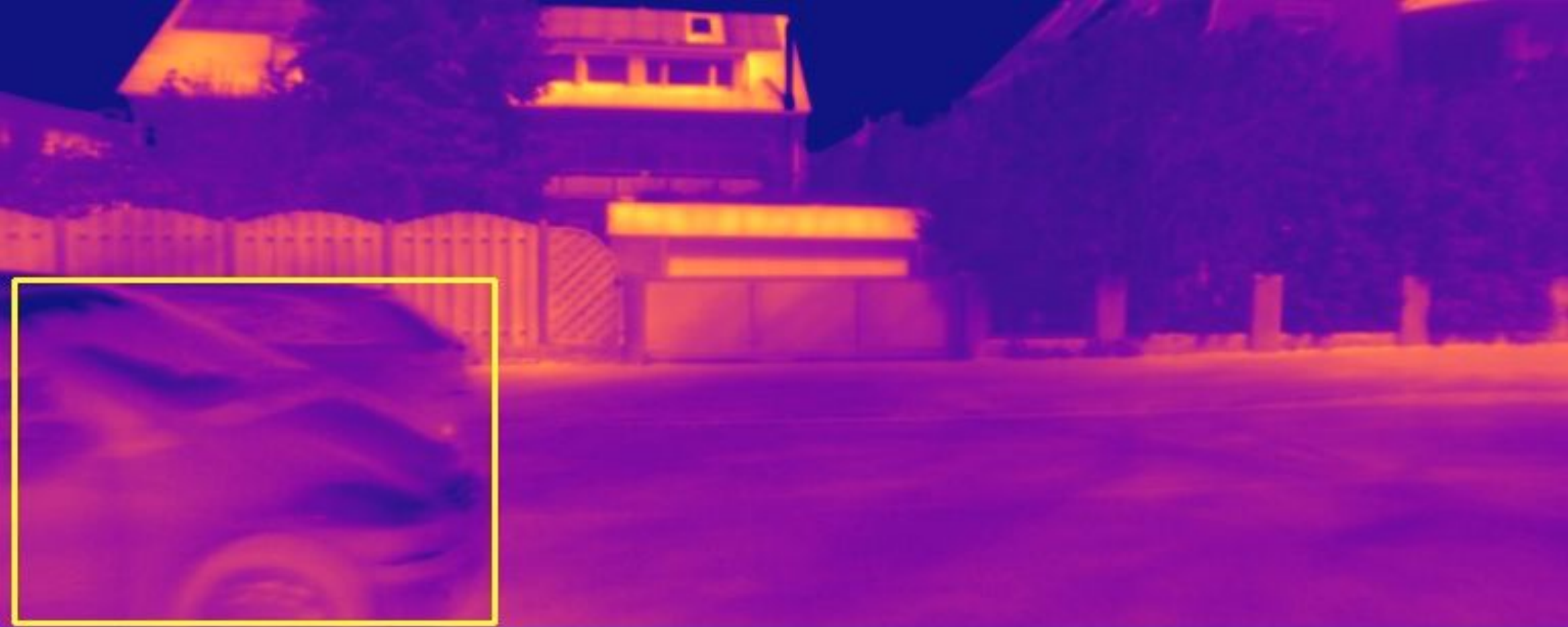} & \includegraphics[width=\linewidth]{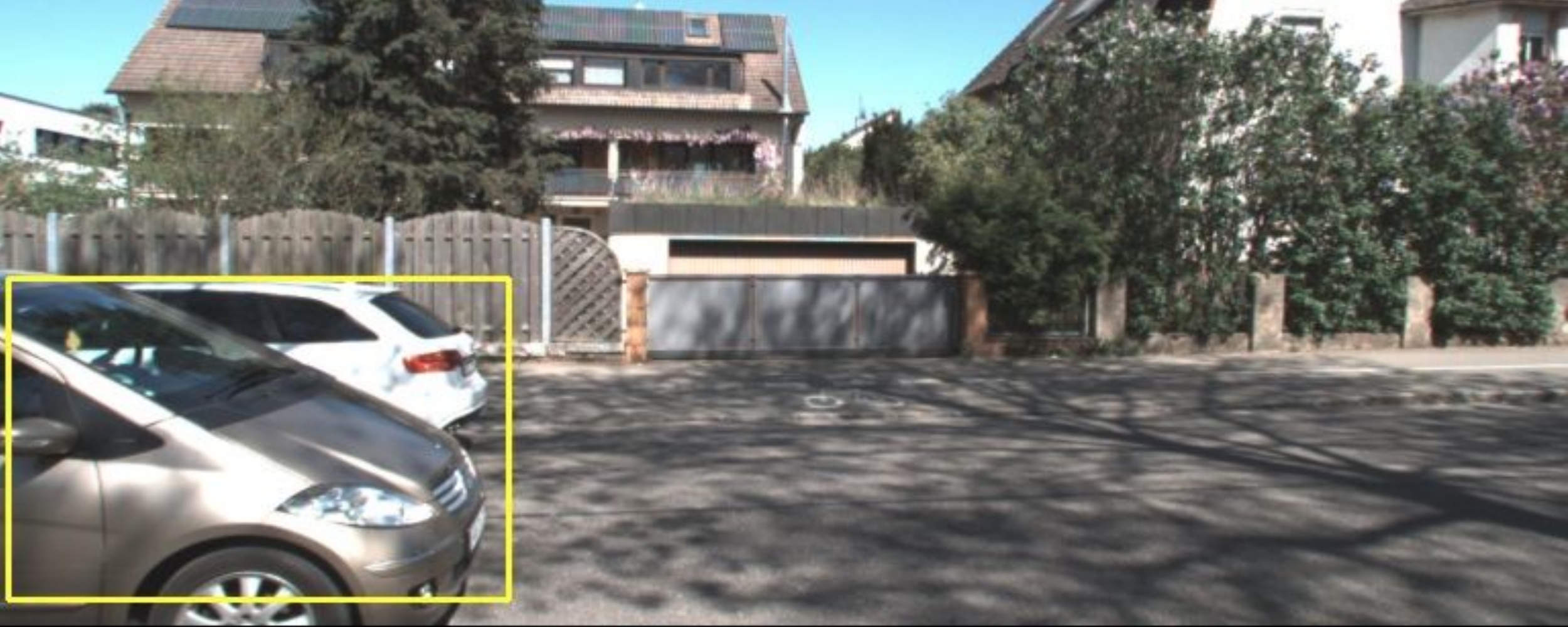} & \includegraphics[width=\linewidth]{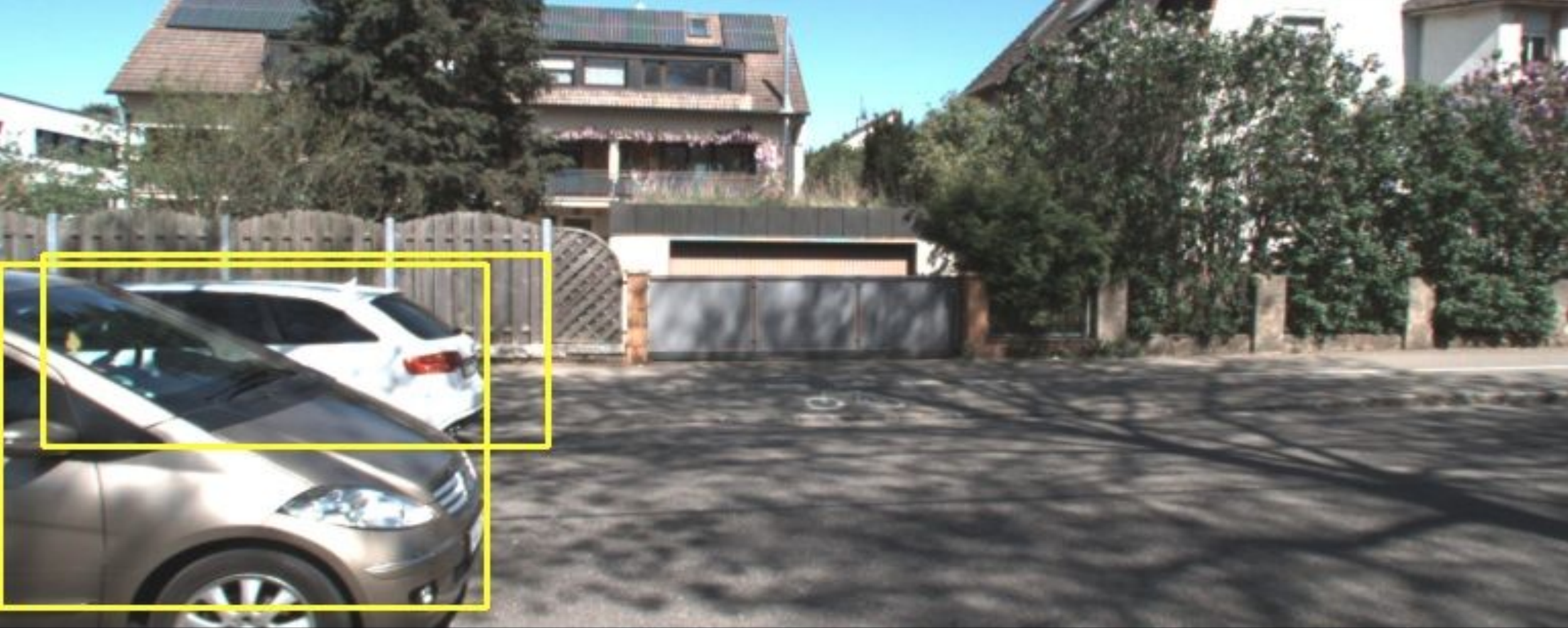} \\
     \includegraphics[width=\linewidth]{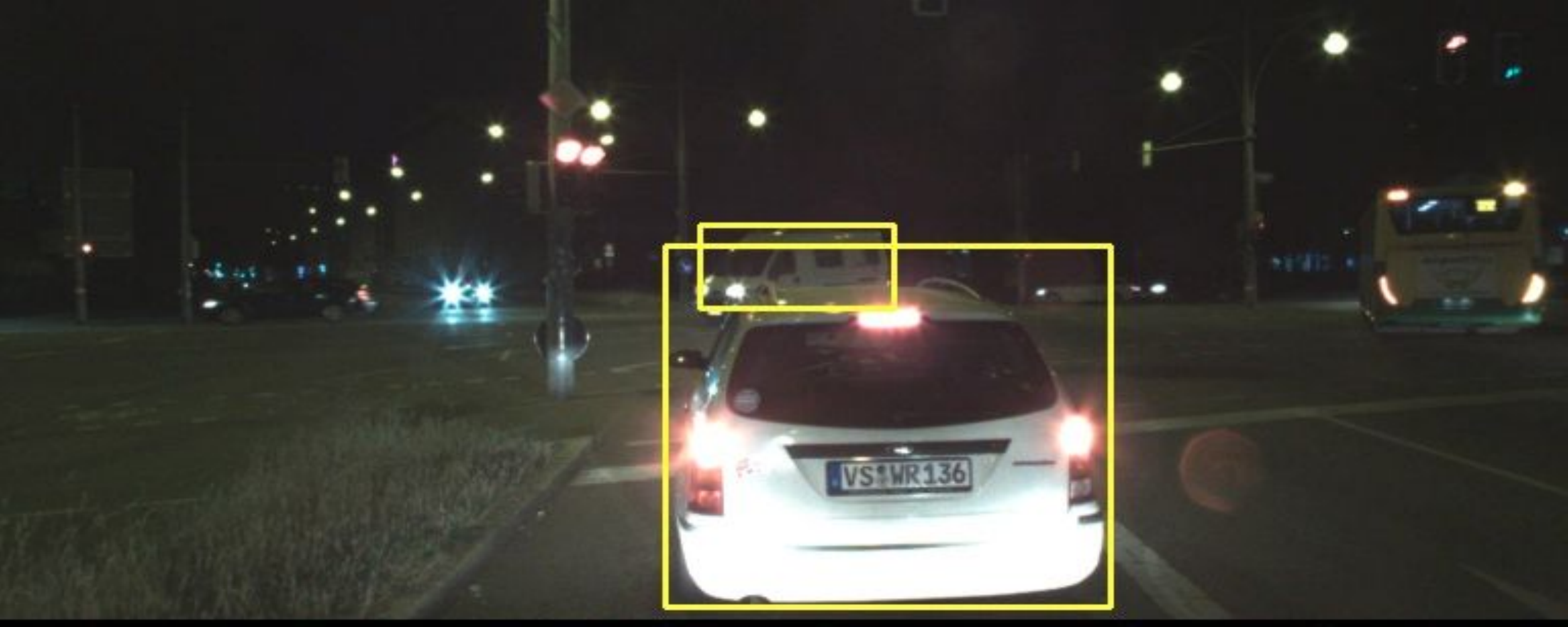} & \includegraphics[width=\linewidth]{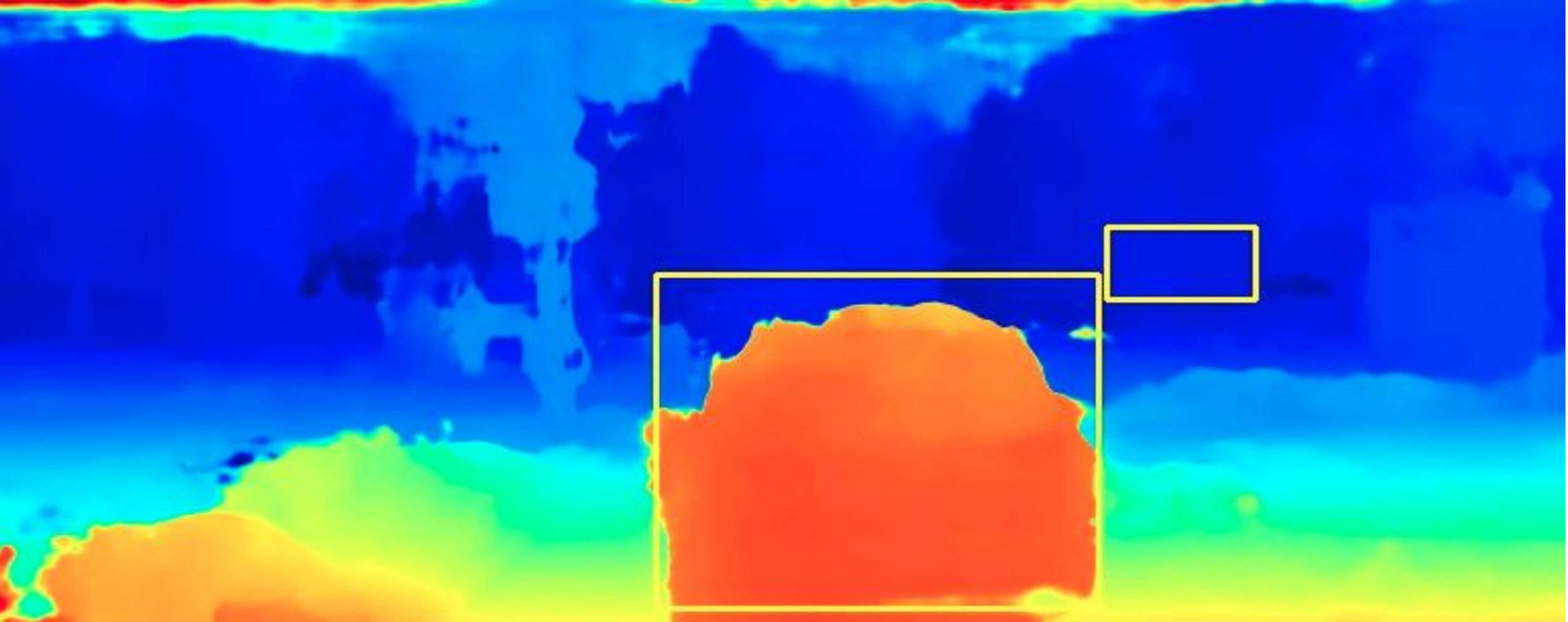} & \includegraphics[width=\linewidth]{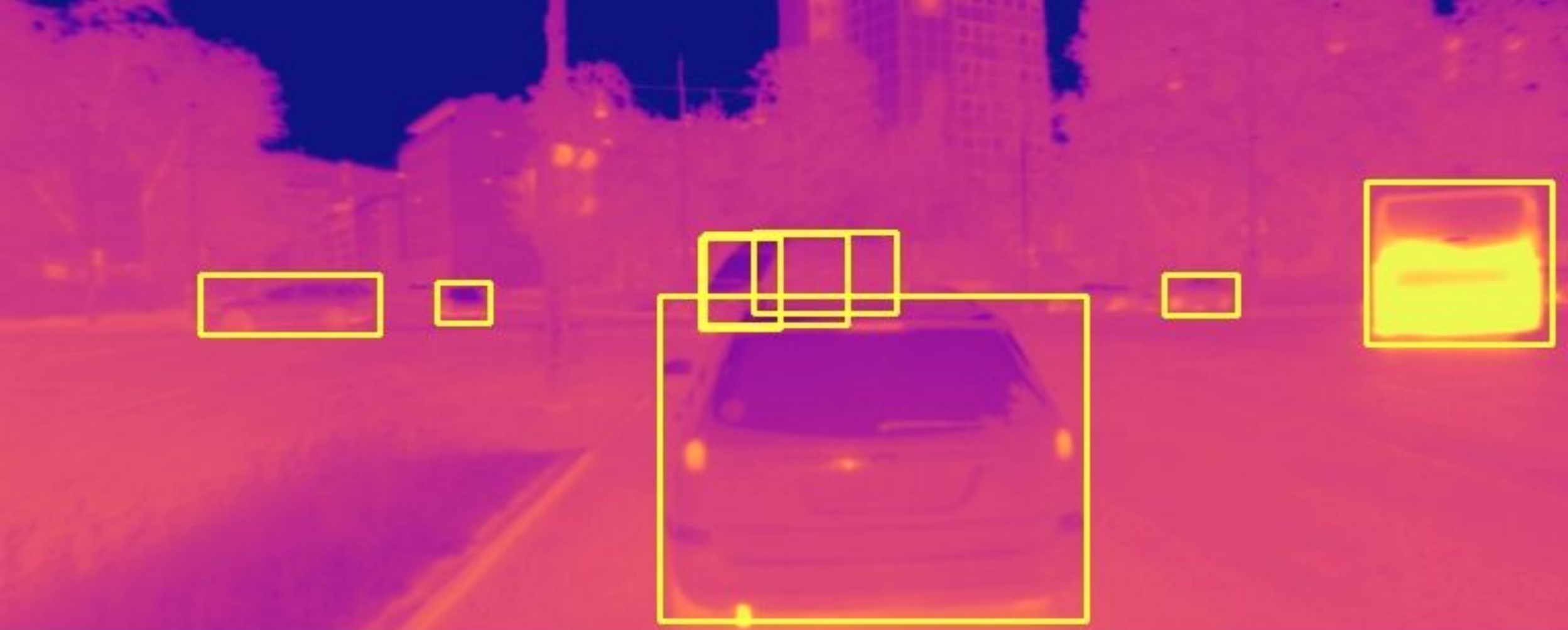} & \includegraphics[width=\linewidth]{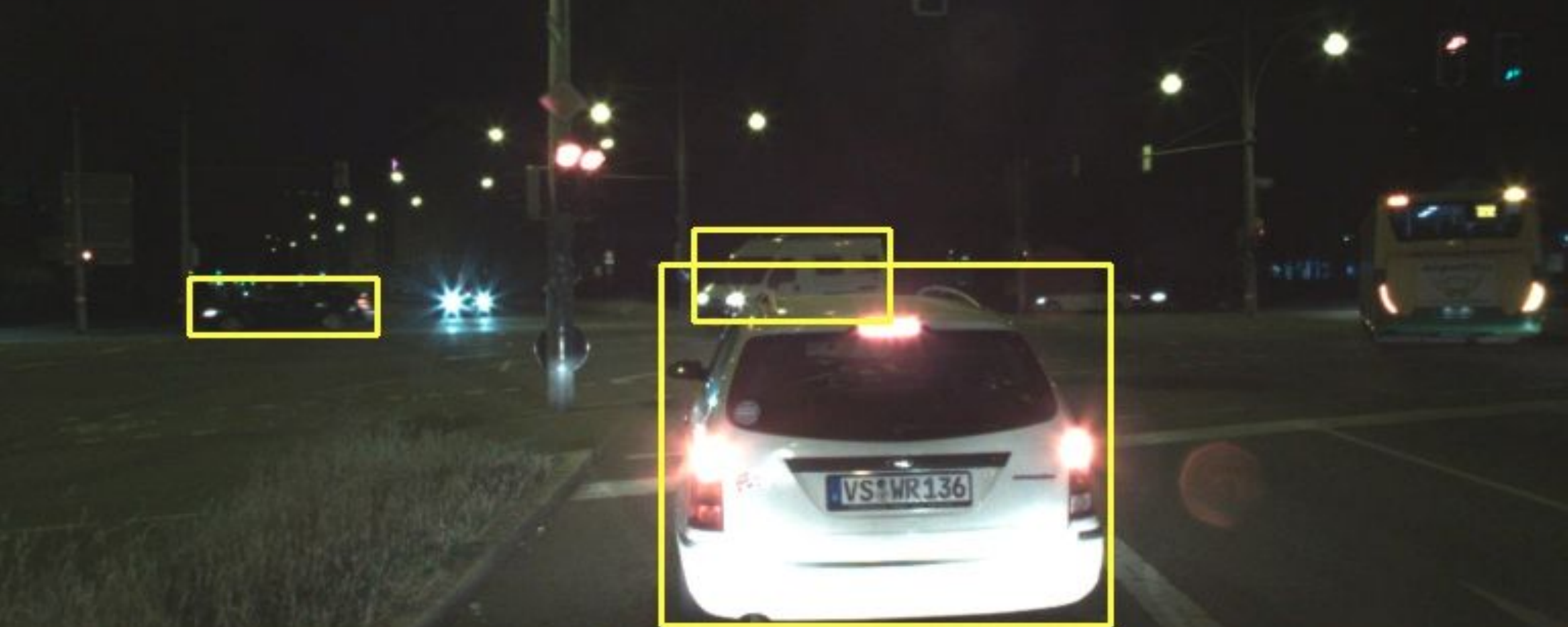} &
     \includegraphics[width=\linewidth]{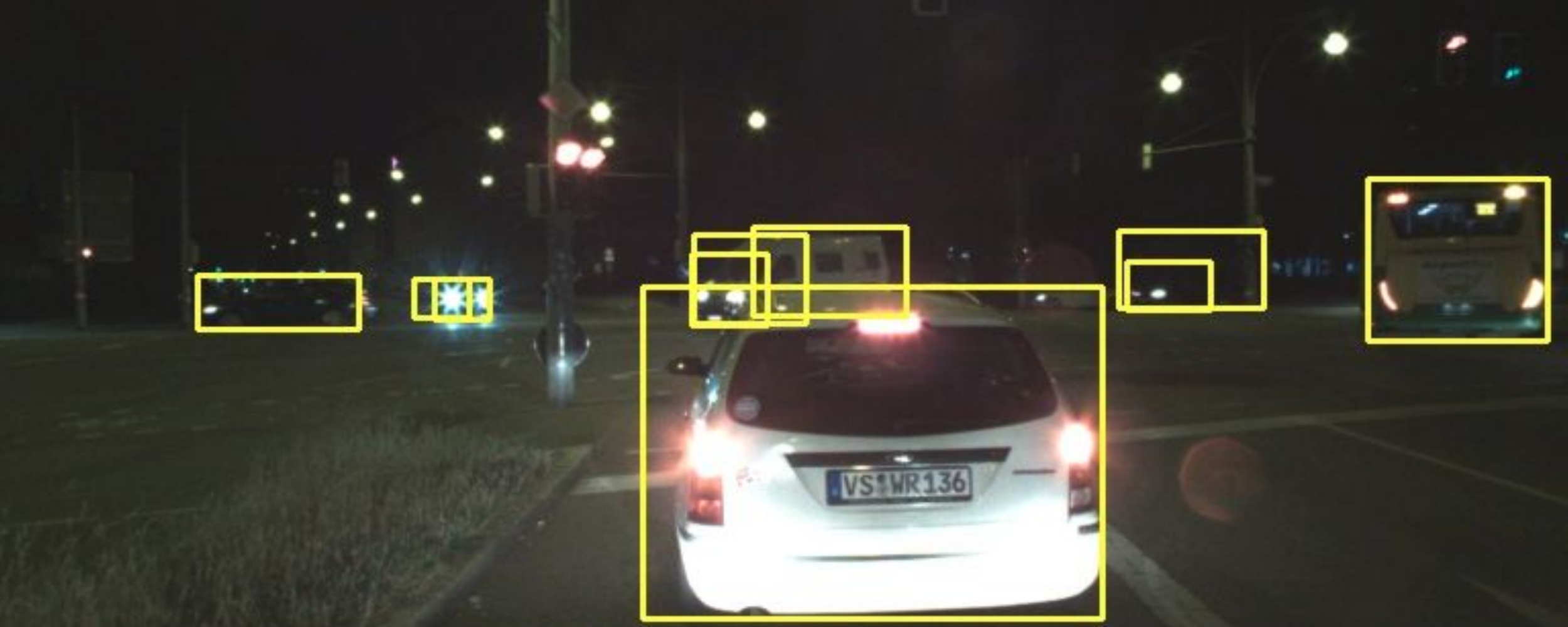} \\
     \includegraphics[width=\linewidth]{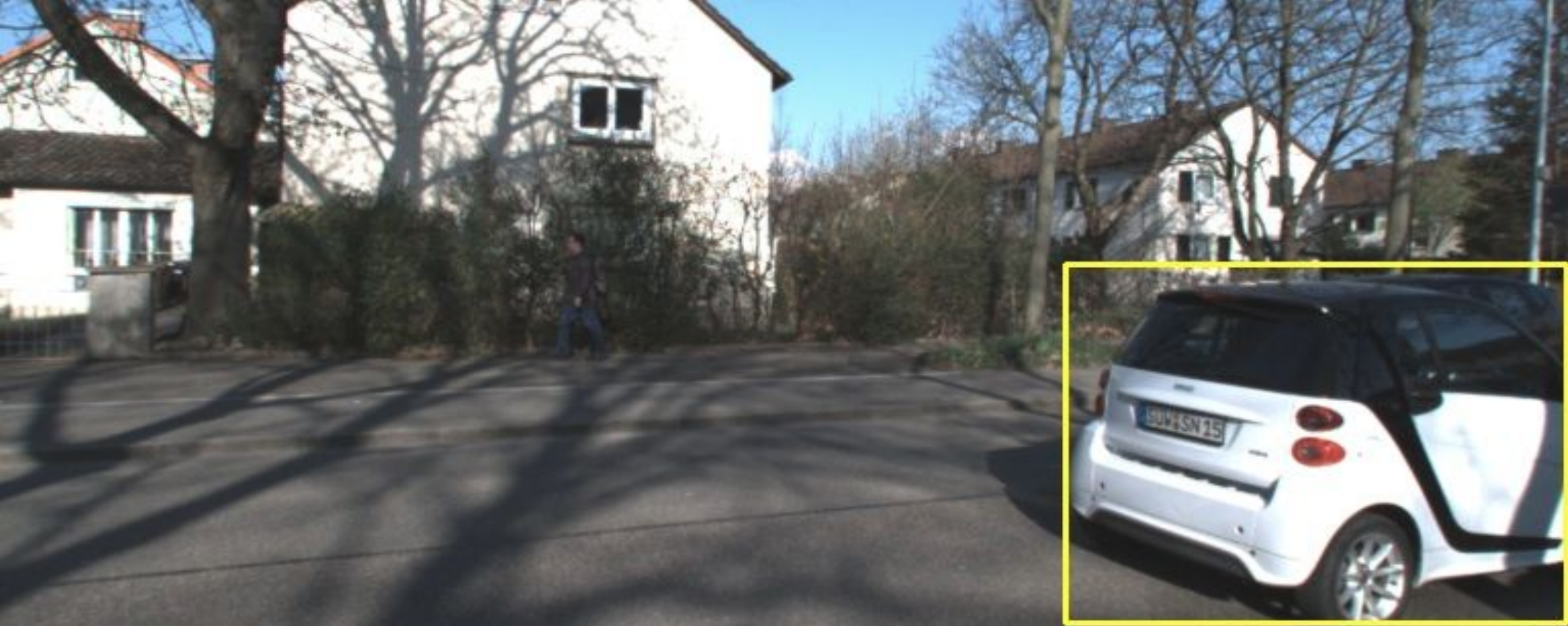} & \includegraphics[width=\linewidth]{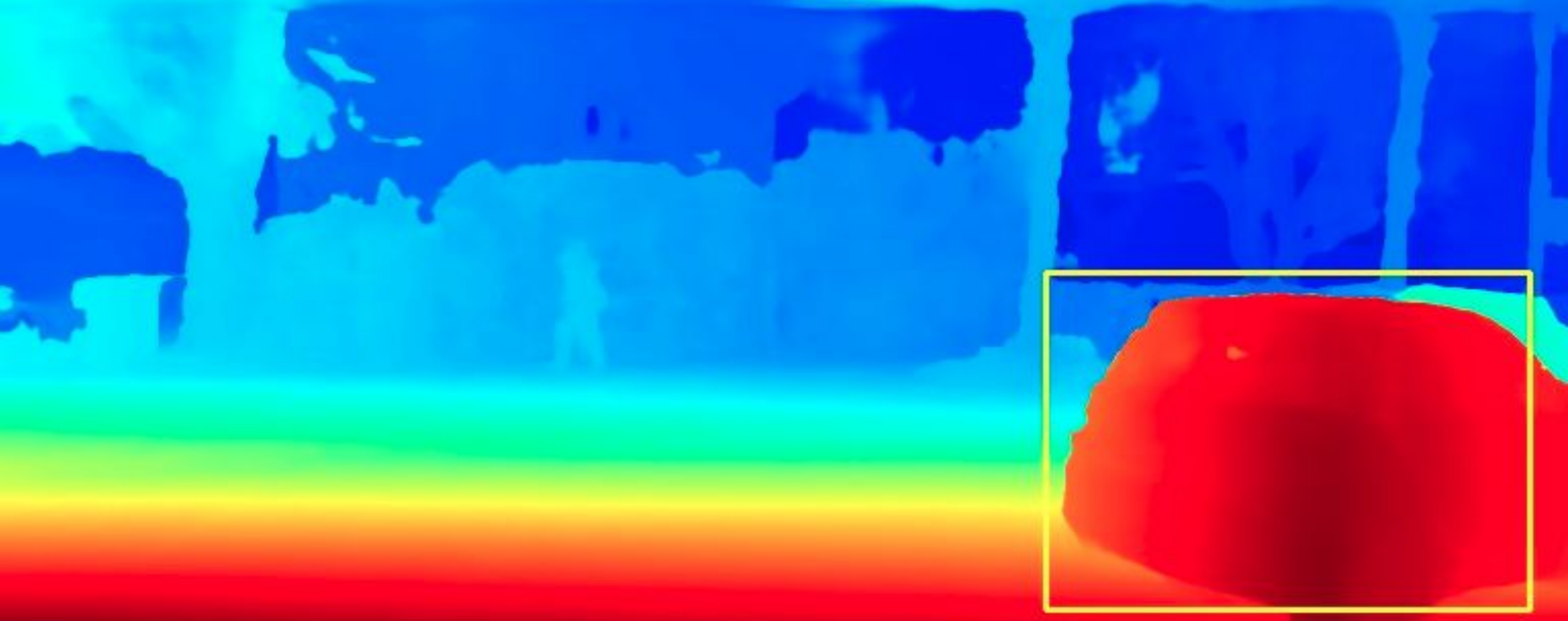} & \includegraphics[width=\linewidth]{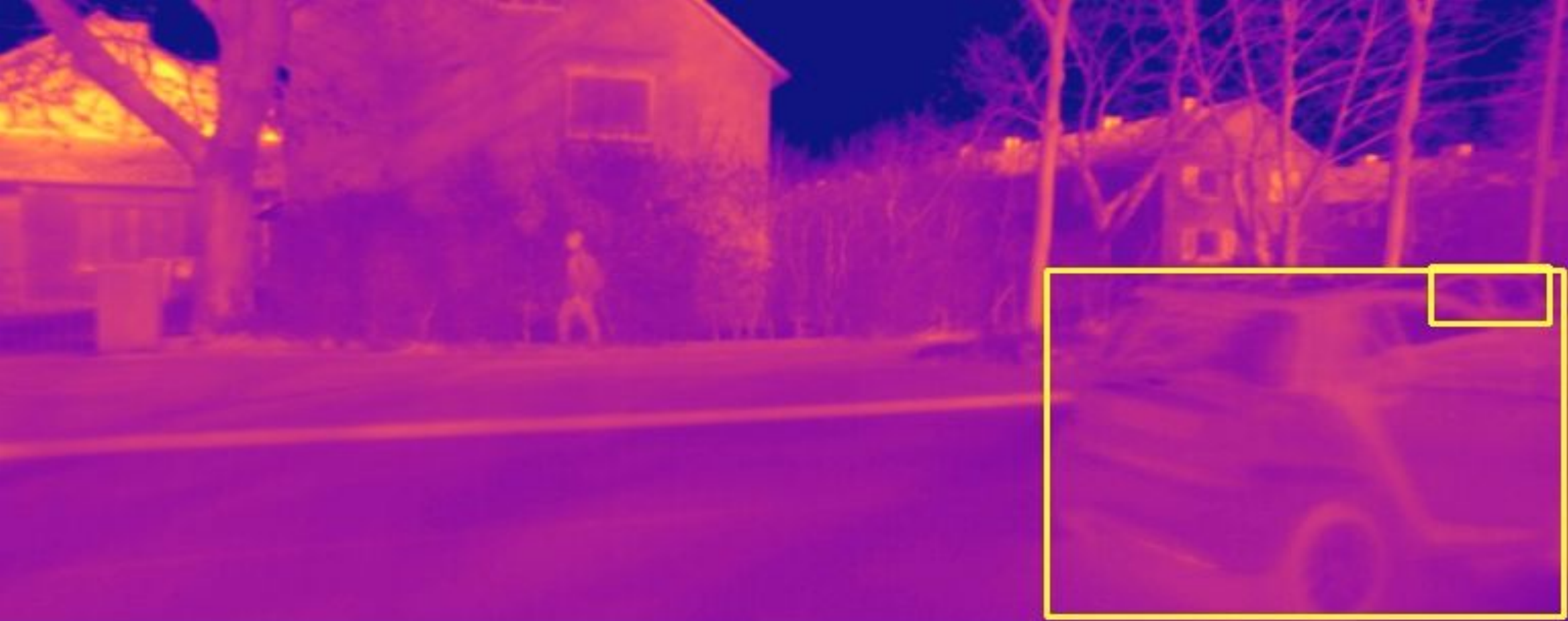} & \includegraphics[width=\linewidth]{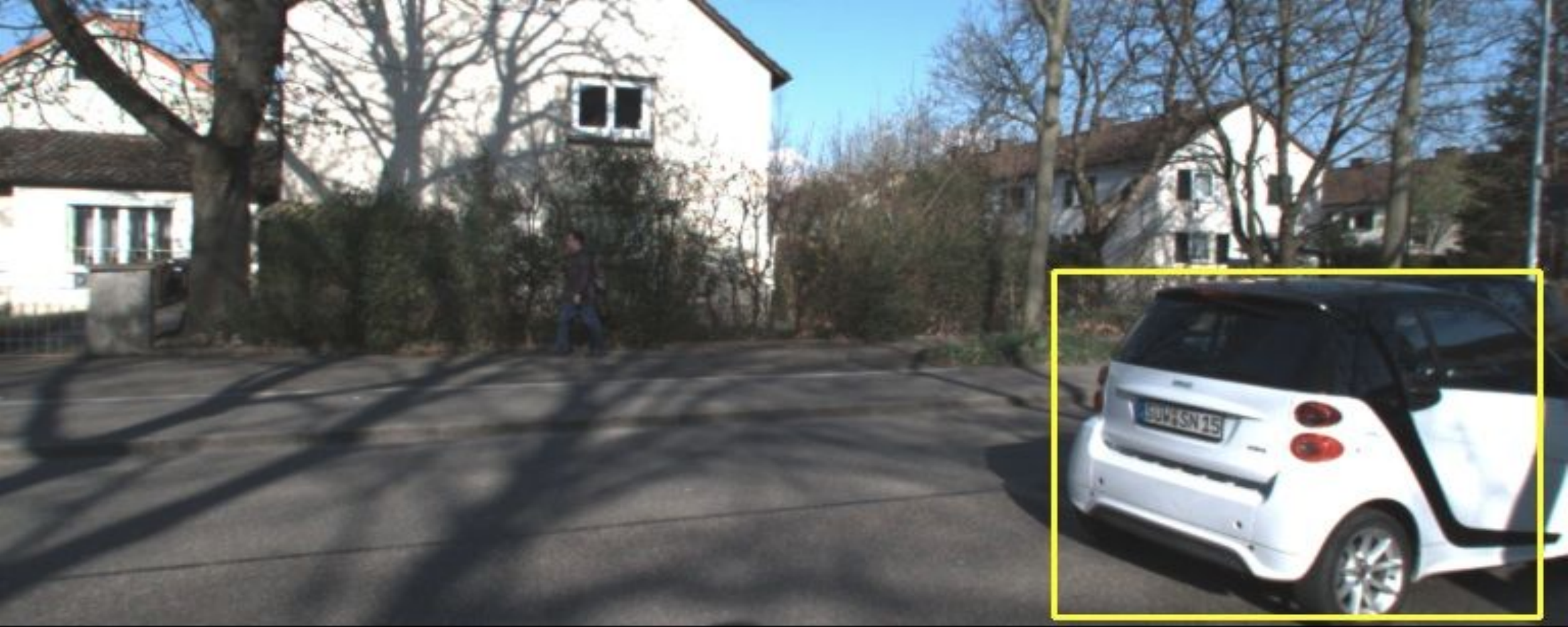} & \includegraphics[width=\linewidth]{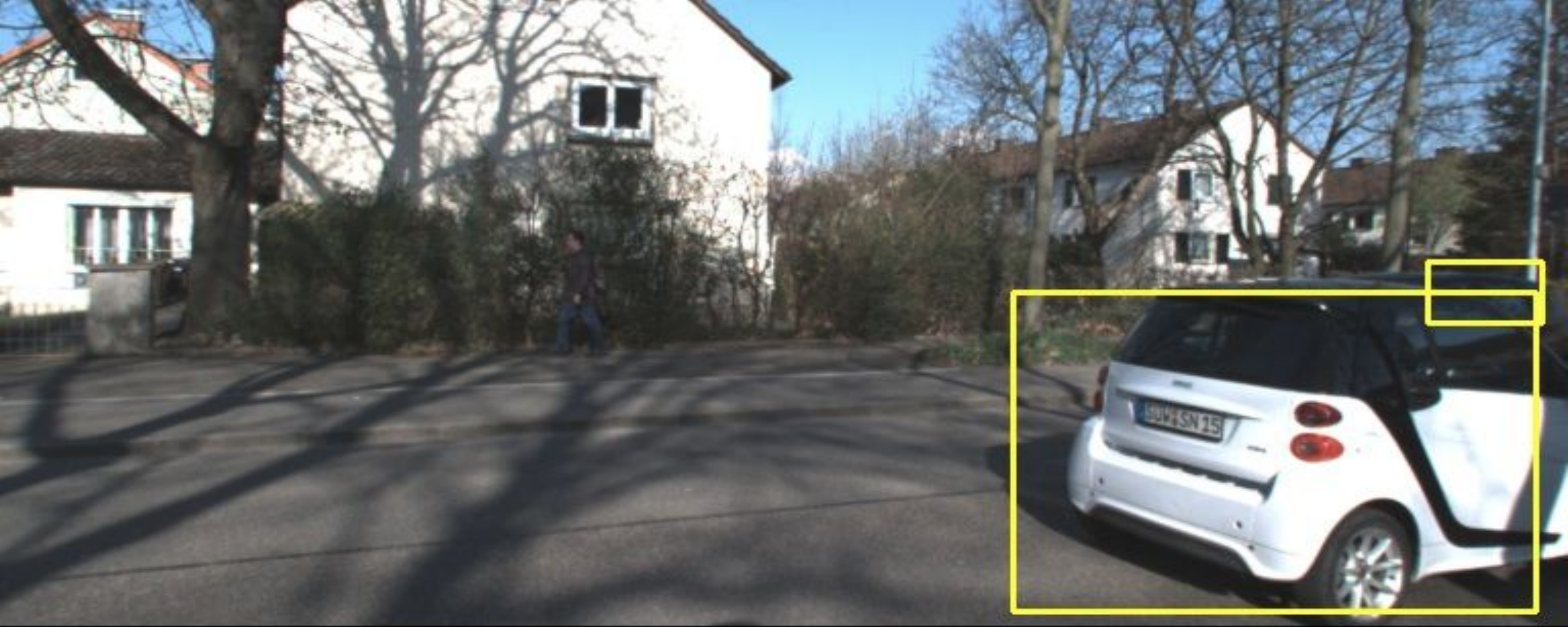} \\
\end{tabular}}
\caption{Qualitative comparisons of the predictions from individual modality-specific teachers with the previous state-of-the-art StereoSoundNet~\cite{wang2020score}, and our MM-DistillNet. Our network consistently detects moving vehicles even in scenes where the baselines fail.}
\label{fig:qualitative}
\end{figure*}

In \tabref{table:ablation}, we ablate the effect of incrementally adding modalities and the impact of our pretext task. It can be seen that RGB and thermal are the main contributors to performance improvement. This can be attributed to the performance of modalities in the day and night, respectively. Nevertheless, integrating depth improves performance, therefore we still employ a depth teacher. Moreover, our pretext task shows an improvement of $0.52$. Additionally, there is an average reduction of $27.55\%$ in the loss value, indicating that the proposed weight initialization accelerates training. We refer to the supplementary material for more details.

\subsection{Qualitative Evaluations}
\label{sec:qualitative}

In this section, we qualitatively evaluate the performance of our proposed MM-DistillNet framework. The audio modality is able to overcome certain limitations of visual sensors, as demonstrated in \figref{fig:qualitative}. The first row highlights how our approach enables us to use the knowledge of the pre-trained teachers to improve the audio student's predictions. The baseline fails to predict a car that the RGB only teacher cannot see. As our model distills knowledge from all the teachers, our MM-DistillNet proactively detects the cars that are not visible to the RGB camera, in this case, coming from the thermal teacher. Our framework also facilitates better student learning, which is highlighted in the second row of \figref{fig:qualitative}. Even though the RGB teacher detects two cars in the image, the baseline does not learn enough cues to predict two cars. Our model uses the RGB and depth teacher to re-enforce the fact that there are two cars in the scene. Our work is not limited to two cars in the scene as in \cite{gan2019self}. We attribute this capability to the incorporation of audio from the microphone array. Finally, the last row of \figref{fig:qualitative} shows how our model can predict cars that are not visible in any of the modalities, such as occluded cars entering the scene.


\section{Conclusions}
\label{sec:conclusions}

This paper proposed a self-supervised framework to distill the knowledge from different expensive sensor modalities into a more accessible one. We do so by leveraging the co-occurrence of modalities and the fact that there exist pre-trained networks for object detection in the visual domain. We use a self-supervised scheme to label audio spectrograms for object detection. During training, we use RGB, depth, and thermal teachers to improve the training of a student network; this enables us to require only audio during inference time. Our results demonstrates how audio is a robust alternative to traditional sensor modalities used in autonomous driving, particularly in overcoming visual limitations. We also publicly released our large-scale MAVD dataset. We compared our approach to different baselines, including different numbers and combinations of modalities, losses, and configurations. We presented qualitative results that highlight the ability of our models to overcome visual limitations such as occlusions and thereby facilitate new applications.


\section{Acknowledgments}

This work was partly funded by the Eva Mayr-Stihl Stiftung and the Graduate School of Robotics in Freiburg.

{\small
\bibliographystyle{ieee_fullname}
\bibliography{egbib}
}

\pagebreak

\begin{strip}
\begin{center}
\vspace{-5ex}
\textbf{\Large \bf
There is More than Meets the Eye:  Self-Supervised Multi-Object Detection and Tracking with Sound by Distilling Multimodal Knowledge} \\
\vspace*{12pt}

\Large{\bf Supplementary Material}\\
\vspace*{12pt}
\large{Francisco Rivera Valverde$^*$ \qquad Juana Valeria Hurtado$^*$ \qquad Abhinav Valada}\\
\large{University of Freiburg}\\
\vspace*{2pt}
\tt\small{\{riverav, hurtadoj, valada\}@cs.uni-freiburg.de}

\end{center}
\end{strip}

\setcounter{section}{0}
\setcounter{equation}{0}
\setcounter{figure}{0}
\setcounter{table}{0}
\makeatletter

%


\normalsize

In this supplementary material, we provide additional experimental results that support the novelty of the contributions made and evaluate our architectural design choices. Particularly, we first provide further details regarding the data collection methodology of our Multimodal Audio-Visual Detection (MAVD) dataset in \secref{sec:SM_sensor}. We then compare the performance of different EfficientDet variants in our proposed MM-DistillNet framework in \secref{sec:SM_EfficientDet}. In \secref{sec:SM_microphones}, we evaluate the performance of our framework with sound from a varying number of microphones as input. Subsequently, we demonstrate the capabilities of our multi-teacher single-student framework to distill knowledge from RGB images into other modalities that are complementary to sound in \secref{sec:SM_modalities}. We then present ablation studies on the influence of various hyperparameters in our proposed MTA loss function in \secref{sec:SM_hyperP} and we compare the performance of our MTA loss with other widely employed knowledge distillation losses in \secref{sec:SM_loses}. Subsequently,  we present additional results on our proposed self-supervised pretext task for the audio student in \secref{sec:SM_pretext}. Then, we present results of our framework in low illumination conditions such as nighttime and dusk in \secref{sec:SM_scenarios}. Finally, we extend our qualitative results with numerous examples in \secref{sec:qualitativeS}. We made the code and models of our approach publicly available at \url{http://rl.uni-freiburg.de/research/multimodal-distill}.\blfootnote{*Equal contribution}\blfootnote{The authors would like to thank Johan Vertens for assistance in the data collection.}


\section{MAVD Dataset}
\label{sec:SM_sensor}

\begin{figure}
\footnotesize
\centering
\setlength{\tabcolsep}{0.05cm}
\includegraphics[width=\linewidth]{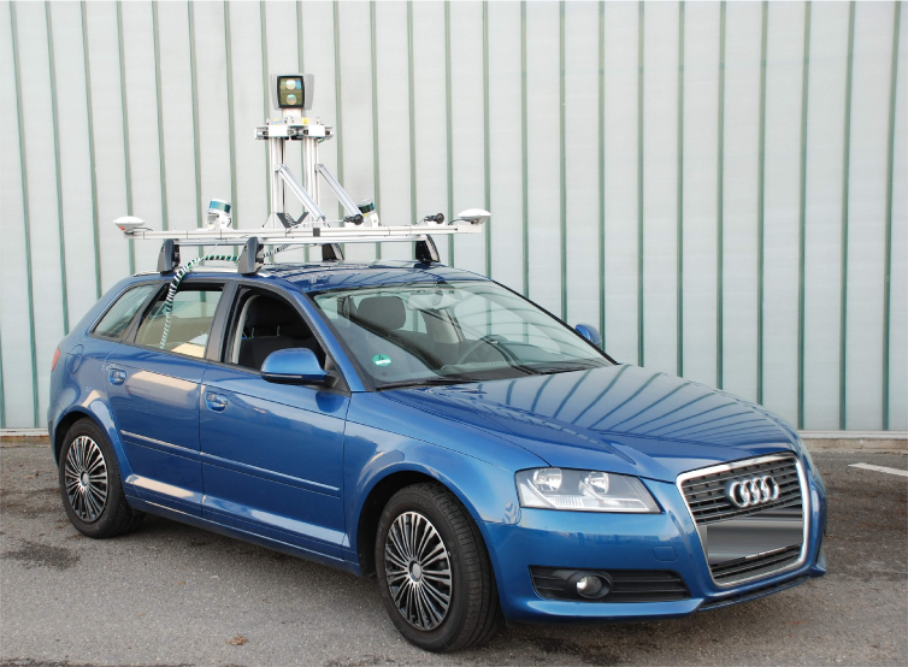}
\includegraphics[width=\linewidth]{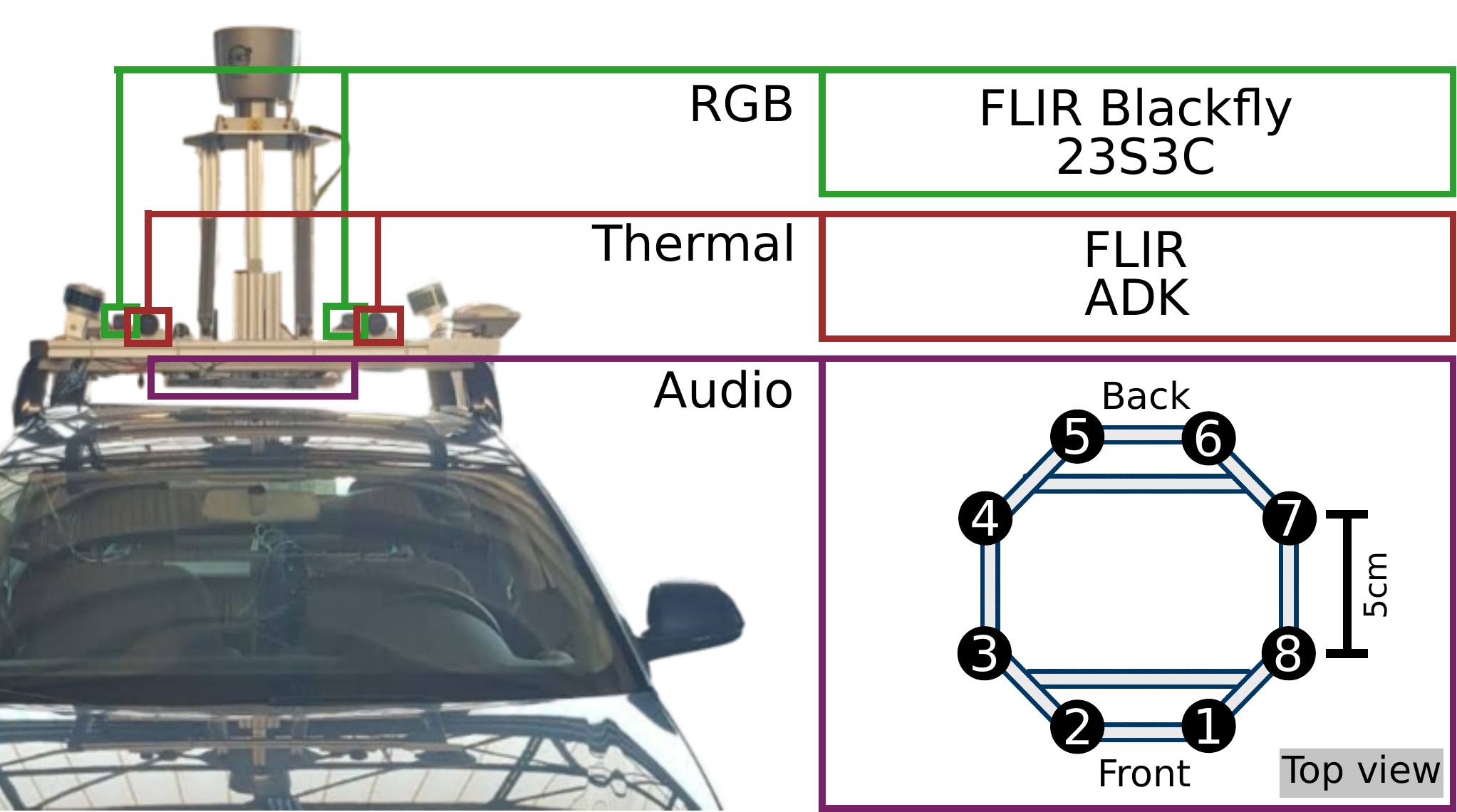}
\caption{Top: The data collection vehicle that we use for our MAVD dataset. Bottom: Closeup view showing the 8-microphone array, stereo cameras and thermal cameras mounted on the roof of the vehicle. The vehicle also contains LiDAR, IMU and GPS, which we also collected for our dataset.}
\label{fig:sensors}
\end{figure}

Our approach employs four different synchronized modalities, including three visual: depth, thermal and RGB, and additional sound. We collected our MAVD dataset using a car with a rack of sensors mounted on the roof as shown in \figref{fig:sensors}. Stereo images were captured using a pair of FLIR~Blackfly~23S3C configured to a resolution of $1920\times650$ pixels and the thermal images were captured at the same resolution using a pair of FLIR ADK cameras. We employ a target-less calibration method~\cite{vertens2020heatnet} for aligning the camera images by formulating a misalignment minimization problem. The miss-alignment is computed as the difference between the gradients of the calibrated RGB image and the transformed thermal image, in the RGB coordinate frame. However, due to the high dimensional nature of this problem, there are ambiguities that cannot be easily resolved without prior information. To this end, the method from \cite{vertens2020heatnet} resolves these ambiguities through a pre-calibration of the camera intrinsics by a pre-processing that align predominant edges of objects that are common to the RGB and thermal cameras.

We employ a stereo rectification and undistortion method with radtan distortion coefficients of (-0.20077378832448342, 0.06858744821624758, -8.318933053823812e-05, 0.0006149164090634826) and camera intrinsics of (1010.5596834500378, 1010.1723409131672, 975.7863331505446, 297.2804298854754). Every RGB and thermal image pair in our dataset is GPS clock synchronized with nano-second precision. The same clock timestamp is used to identify the central audio frame corresponding to the thermal and RGB images. We then identify the frame number in the audio clip using a sampling rate of \SI{44100}{\hertz}, and sample 1 second around the RGB-thermal timestamp. In order to obtain the depth image, we use the network proposed by Zhang~\textit{et~al.}~\cite{Zhang2019GANet} with the left and right images from the stereo rig. We set the maximum disparity to 192 and apply a jet color map to leverage the ImageNet pre-trained weights for initializing the EfficientDet backbone. We made our dataset publicly available at \url{http://rl.uni-freiburg.de/research/multimodal-distill}.

\section{Extended Ablation Study}

\subsection{EfficientDet Compound Coefficient Selection}
\label{sec:SM_EfficientDet}

EfficientDet is a family of object detection models proposed by 
Tan~\textit{et~al.}~\cite{tan2020efficientdet} which contains eight different architectural configurations that trade-off performance and runtime. We evaluated the performance of the variants to identify their suitability for our framework. To this end, we created a large dataset by combining Microsoft~COCO~\cite{lin2014microsoft}, PASCAL~VOC~\cite{everingham2010pascal}, and ImageNet~\cite{deng2009imagenet}. Subsequently, we removed all the scenes that do not contain at least one vehicle and retained the bounding boxes of those that contain cars or moving vehicles. We then trained EfficientDet~D0-D7 to detect the objects in this combined dataset. In \tabref{table:efficientdetruntimes}, we present the performance in terms of the Average Precision (AP) at $IoU=0.5$, as well as the inference time and the Floating Point Operations Per Second (FLOPS). We chose EfficientDet~D2 as the backbone of our framework, given that it presents a higher improvement in the average performance with a lesser increase in the inference time. Nevertheless, our framework provides the flexibility to adopt any of the other variants as a direct drop in replacement.

\begin{table}
\begin{center}
\footnotesize
\begin{tabular}{p{1.8cm}|p{1cm}p{1cm}p{1.2cm}}
\toprule
EfficientDet Variant & AP@ 0.5 & FLOPS & Inference Time (\si{\milli\second}) \\
\noalign{\smallskip}\hline\hline\noalign{\smallskip}
D0 & 0.5165 & 2.5B & 30.04 \\
D1 & 0.6870 & 6.1B& 34.76 \\
\textbf{D2} & \textbf{0.7974} & \textbf{11B} & \textbf{39.99} \\
D3 & 0.8134 & 25B& 59.45 \\
D4 & 0.8680 & 55B& 89.93 \\
D7 & 0.9200 & 325B& 388.90\\
\bottomrule
\end{tabular}
\end{center}
\caption{Performance comparison of different EfficientDet variants for predicting bounding boxes of vehicles in Microsoft COCO~\cite{lin2014microsoft}, PASCAL~VOC~\cite{everingham2010pascal}, and ImageNet~\cite{deng2009imagenet}, with the associated AP at 0.5 IoU.}
\label{table:efficientdetruntimes}
\end{table}

\subsection{Influence of Number of Microphones}
\label{sec:SM_microphones}

\pgfplotsset{width=7cm,compat=1.3}
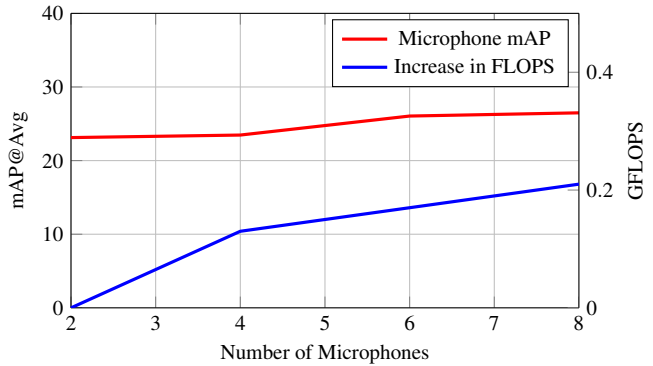
\begin{figure}
    \footnotesize
    \centering
    \begin{tikzpicture}
    \pgfplotsset{
        xmin=2, xmax=8,
    }

    \begin{axis}[
        axis y line*=left,
        ymin=0, ymax=40,
        xlabel = Number of Microphones,
        ylabel = mAP@Avg,
        grid=both,
        height = 5.5cm,
        width =\linewidth,
        every axis plot/.append style={very thick}
    ]
        \addplot[red]
        coordinates {
          (2,  23.1255)
          (4,  23.47)
          (6,  26.0511)
          (8, 26.487)
        };\label{microphoneAP}
    \end{axis}
    
    \begin{axis}[
        axis y line*=right,
        axis x line=none,
        ymin=0, ymax=0.50,
        ylabel=GFLOPS,
        height = 5.5cm,
        width =\linewidth,
        every axis plot/.append style={very thick}
    ]
    \addlegendimage{/pgfplots/refstyle=microphoneAP}\addlegendentry{Microphone mAP}
        \addplot[blue]
        coordinates {
          (2,  0.00)
          (4,  0.13)
          (6,  0.17)
          (8, 0.21)
        };\addlegendentry{Increase in FLOPS}
    \end{axis}
    
    \end{tikzpicture} 
    \caption{The left axis (blue line) shows the performance of the network vs the number of microphones. The right axis (red) shows the GLOPS increase caused by using N microphones. It can be seen, that more channels improve the performance in the given task with negligible impact in FLOPS.}
    \label{fig:mics}
\end{figure}

Our proposed MM-DistillNet framework exploits complementary cues from different modalities such as RGB, depth, and thermal images during training to incorporate multimodal information into a single audio network. The input to the audio network is from a microphone array that captures ambient sounds. We employ multiple monophonic microphones due to the promising results that it has demonstrated for sound source localization~\cite{ma2019phased}. \figref{fig:sensors} shows the microphone array that we employ mounted on our data collection vehicle. In order to estimate the number of microphones that are essential to reliably localize objects, we analyze the performance of our MM-DistillNet for varying number of microphones in a balanced subset of the dataset.

Namely, we hypothesize that there is a relationship between the number of microphones and the complexity of the scene, as measured by the number of vehicles in the environment. Our MAVD dataset contains varying number of vehicles in each scene, with a maximum of 13 vehicles in a single scene. We performed experiments to analyze the improvement that we can achieve by using more number of microphones in the array. To do so, we need to have balanced number of vehicles in the dataset. Therefore, we apply an under-sampling approach to ensure that the number of examples with varying number of vehicles are balanced. We compute the performance of the audio student using multiples of two number of microphones, in order to always consider the microphones that are further away from each other in the hexagonal array.

\figref{fig:mics} shows that increasing the number of microphones consistently improves the performance of the model. Therefore, we utilize sound from all the eight monophonic microphones that are available in our octagonal setup for our experiments. Nevertheless, given that each microphone adds an additional $768\times768\times1$ input to the network, we computed the additional overhead in terms of the increase in number FLOPS in the network. \figref{fig:mics} shows that the increase in FLOPS in the network due to the addition of a microphone is negligible compared to the overall FLOPS of EfficientDet as shown in \tabref{table:efficientdetruntimes}.
    
\subsection{Distillation to Different Student Modalities}
\label{sec:SM_modalities}

\begin{table}
\begin{center}
\footnotesize
\begin{tabular}{p{2.5cm}p{1.2cm}|p{1.35cm}}
\toprule
Teachers & Student & mAP@ Avg  \\
\noalign{\smallskip}\hline\hline\noalign{\smallskip}
RGB & Sound & $57.25$ \\
RGB & Thermal & $56.70$ \\
\midrule
RGB, Depth, Thermal & Sound & $61.62$ \\
RGB, Depth, Thermal & RGB & $81.12$ \\
RGB, Depth, Thermal & Thermal & $\mathbf{81.98}$ \\
\bottomrule
\end{tabular}
\end{center}
\caption{Our framework distills the knowledge from multimodal teachers trained on dataset where supervision is available, to improve the learning of a single student network. Thermal and RGB modalities show significant improvement over their single teacher counterparts.}
\label{table:nonaudio}
\end{table}

By taking advantage of the co-ocurrence of all the modalities present in our dataset (audio, RGB, thermal and depth), it is straightforward to interchange the input modality of the student network from audio to any of the other modalities as described in Sec.~3 of the main paper. By doing so, we can exemplify not only how our method is modality independent, but also how it can further improve the performance of existing object detection frameworks.

In the other experiments, we selected RGB, depth, and thermal images as the teacher modalities and sound as the student modality for our framework. This enables us to tackle limitations of visual modalities such as occlusions and sensor sensitivity (poor performance of RGB cameras during nighttime as well as the limited sensitivity of thermal cameras during the day). Nevertheless, our approach to transfer the knowledge from multiple pre-trained modality-specific networks to a student network, is input agnostic. Under this perspective, instead of employing sound as input to the student network, we can also use conventional RGB, depth, or thermal modalities as input to our framework. \tabref{table:nonaudio} presents results with different modalities as input to the student network. It can be seen that the performance against the single teacher is substantially improved. Particularly, using the thermal modality as input to the student network provides the best performance, as it provides substantial cues for vehicle detection in both day as well as night recordings. Whereas, using RGB as input to the student during night suffers due to low illumination conditions. Nevertheless, using a thermal modality requires expensive hardware and is subject to visual limitations like occlusion. For this reason, we employ audio in our MM-DistillNet as an alternative to the traditional visual inputs used in autonomous driving.

\subsection{Influence of Hyperparameters in MTA Loss}
\label{sec:SM_hyperP}

We define our proposed Multi-Teacher Alignment (MTA) loss function as
\begin{equation}
\label{eq:MTAloss}
L_{MTA} = \beta * \sum_{j} KL_{div}\left(\frac{Q_{s}^j}{\left \|  {Q_{s}^j}\right \|_{2}}, \frac{Q_{t}^j}{\left \| {Q_{t}^j}\right \|_{2}}\right),
\end{equation}
where $Q_{s}^j = F_{avg}^{r}(A_{s})$ and $Q_{t}^j = \prod_{i}^{N} F_{avg}^{r}(A_{t_{i}})$ are the student and multi-teacher attention maps respectively. Additionally, while computing the $KL_{div}$ as a measure of the difference between probability distributions between the teachers and the student, we can apply a temperature $t$ to the softmax as proposed by Hinton~\textit{et~al.}~\cite{hinton2015}. This temperature in the softmax computation is added to adapt the confidence on each individual probability distribution. To this end, we can expand the above notation using the student normalized activation $S_{x}=\frac{Q_{s}^j}{\left \|  {Q_{s}^j}\right \|_{2}}$ and the integrated normalized teacher attention $T_{x}=\frac{Q_{t}^j}{\left \| {Q_{t}^j}\right \|_{2}}$ as
\begin{equation}
\label{eq:klexpanded}
KL_{div}\left(S_{x}||T_{x}\right)= \sum_{x \in \mathcal{X}}\frac{e^{\frac{S_{x, i}}{t}}}{\sum_{k=1}^{n} e^{\frac{S_{x,k}}{t}}} \log \left(\frac{\frac{e^{\frac{S_{x, i}}{t}}}{\sum_{k=1}^{n} e^{\frac{S_{x,k}}{t}}}}{\frac{e^{\frac{T_{x, i}}{t}}}{\sum_{k=1}^{n} e^{\frac{T_{x,k}}{t}}}}\right)
\end{equation}

We employ this loss in addition to the focal loss $L_{focal}$ to optimize our MM-DistillNet framework as
\begin{equation}
\label{eq:totallossS}
L_{total} = \delta * L_{focal} + \omega * L_{MTA}.
\end{equation}

With this formulation, we have two hyperparameters in the MTA loss function that can be selected according to the specific task: the exponential value $r$, which controls the relevance of small valued activations in contrast to large valued activations, and the softmax temperature $t$. We performed experiments to study the influence of these two hyperparameters on the performance of the audio student network. Results from this experiment is shown in \tabref{table:losshyper}. We present the mean average precision of our best recipe using a single RGB teacher and the audio student network. 

\begin{table}
\begin{center}
\footnotesize
\begin{tabular}{p{1.5cm}p{1.5cm}|p{1.4cm}}
\toprule
r & t & mAP@ Avg \\
\noalign{\smallskip}\hline\hline\noalign{\smallskip}
4          & 9          & 60.13       \\
4          & 6          & 61.40       \\
3          & 4          & 64.08       \\
3          & 2          & 64.30       \\
2          & 6          & 64.30       \\
2          & 4          & 66.05       \\
2          & 7          & 66.43       \\
1          & 3          & 67.68       \\
1          & 4          & 67.74       \\
1          & 9          & 69.67       \\
2          & 9          & $\mathbf{69.72}$ \\
\bottomrule
\end{tabular}
\end{center}
\caption{Ablation study of hyperparameters $r$ and temperature $t$ in the proposed MTA loss function. The object detection performance is shown for an audio student with knowledge distilled from a RGB teacher network.}
\label{table:losshyper}
\end{table}

We observe that in this setting, a value of $r=4$ and $t=9$ provides the best result. This suggests that putting more importance to higher valued activations rather than low-valued activations improves the overall performance of the network. 

\subsection{Ranking Loss vs. MTA Loss}
\label{sec:SM_loses}

\begin{table}
\begin{center}
\footnotesize
\begin{tabular}{p{1.2cm}|p{2cm}|p{2cm}}
\toprule
Input & \multicolumn{2}{c}{mAP@ Avg} \\
\cmidrule{2-3}
 & Ranking Loss~\cite{gan2019self} & MTA Loss (Ours) \\
\noalign{\smallskip}\hline\hline\noalign{\smallskip}
RGB & 56.37 & $\mathbf{57.25}$ \\
Thermal & 55.82 & $\mathbf{56.70}$ \\
Depth & 45.85 & $\mathbf{55.41}$ \\
\bottomrule
\end{tabular}
\end{center}
\label{table:singleteacherourloss}
\caption{Our MTA loss function was formulated to distill knowledge from multiple teachers to a single student. Nevertheless, it yields superior performance than Ranking loss employed by Gan~\textit{et~al.}~\cite{gan2019self} for knowledge distillation in a single teacher-student scenario.}
\end{table}

The proposed MM-DistillNet framework exploits the co-occurrence of different modalities representing a scene. Our MTA loss facilitates this process by aligning the intermediate representations of multiple teachers to that of a single student. The previous state-of-the-art~\cite{gan2019self} method employs the Ranking loss to distill knowledge from a single teacher to a single student network. To further demonstrate the capabilities of our proposed MTA loss, we compare its performance with Ranking loss for distillation of knowledge from a single modality-specific teacher to an audio student network. We performed this experiment for all the teacher modalities considered in this work. Although our loss function is designed to align different intermediate representations of modality-specific networks, we observe that it outperforms the Ranking loss formulation even in the single teacher setting.\looseness=-1

\section{Evaluation of Audio Student Pretext Task}
\label{sec:SM_pretext}

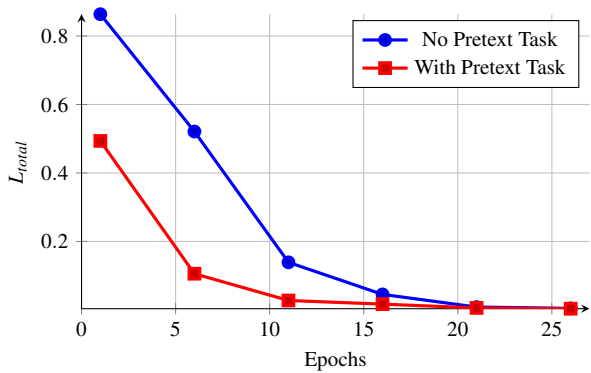
\begin{figure}
    \footnotesize
    \centering
\begin{tikzpicture}
    \pgfplotsset{
        xmin=0, xmax=27,
    }
    \begin{axis}[
        axis lines = left,
        xlabel = Epochs,
        ylabel = $L_{total}$,
        grid=both,
        height = 5.5cm,
        width =\linewidth,
        every axis plot/.append style={very thick}
    ]
        \addplot coordinates {
          (1,  0.8637468)
          (6,  0.5210465)
          (11,  0.138963)
          (16, 0.04530995)
          (21, 0.008364)
          (26, 0.004460582)
        };
        \addlegendentry{No Pretext Task}
        \addplot coordinates {
          (1,  0.49357)
          (6,  0.105744)
          (11,  0.027559)
          (16, 0.016903)
          (21, 0.005821)
          (26, 0.003497)
        };
        \addlegendentry{With Pretext Task}
    \end{axis}
    \end{tikzpicture} 
    \caption{Comparison of training convergence of our MM-DistillNet, with and without the initialization of the student network with weights of the model trained on our proposed self-supervised pretext task.}
    \label{fig:pretext}
\end{figure}

Our MM-DistillNet contains multiple modality-specific teachers and a single student network. Each of the networks are composed of the EfficientNet backbone which has to be initialized with pre-trained weights from large datasets to ease the optimization and achieve better convergence. Since all existing pre-trained models of EfficientNet primarily employ 3-channel RGB images as input, they cannot be used for initializing the audio student network which takes 8-channel spectrograms as input (1-channel spectrogram from each of the 8 microphones in the array). To address this problem, we propose a self-supervised pretext task that provides the audio student network with semantically rich information about the relationship between the audio and visual modalities. The goal of the pretext task is to estimate the number of vehicles present in the RGB image only using sound as input to the network. In the ablation study presented in the main paper, we show that our proposed pretext task improves the performance in terms of detection metrics. In \figref{fig:pretext} of this supplementary material, we present comparisons of the training curves for the MM-DistillNet framework, with and without the initialization of the audio student network with weights of the model trained on the proposed pretext task. We can see that the model with weighted initialized from the pretext task consistently yields a lower loss since the early stages. Moreover, the final loss is $27.55\%$ lower than the model trained from scratch. These results demonstrate that the pretext task not only improves the performance in terms of the metrics, it also accelerates training and leads to faster convergence.

\section{Evaluation in Low Illumination Conditions}
\label{sec:SM_scenarios}

\begin{table*}
\begin{center}
\footnotesize
\begin{tabular}{p{1.2cm}|p{1.5cm}|p{2.5cm}|p{1cm}p{1cm}p{1cm}p{0.8cm}p{0.5cm}}
\toprule
Condition & Vehicle State & Network & mAP@ Avg & mAP@ 0.5 & mAP@ 0.75 & CDx & CDx \\
\noalign{\smallskip}\hline\hline\noalign{\smallskip}
Day & Static & StereoSoundNet~\cite{gan2019self} & $53.48$ & $69.10$ & $52.50$ & $2.77$ & $1.51$ \\
Night & Static & StereoSoundNet~\cite{gan2019self} & $38.13$ & $49.26$ & $34.67$ & $4.34$ & $3.95$ \\
Day & Driving & StereoSoundNet~\cite{gan2019self} & $45.59$ & $69.20$ & $42.84$ & $2.50$ & $1.60$ \\
Night & Driving & StereoSoundNet~\cite{gan2019self} & $28.56$ & $45.18$ & $24.43$ & $3.86$ & $2.77$ \\
\midrule
Day & Static & MM-DistillNet & $63.80$ & $83.90$ & $63.63$ & $1.59$ & $0.78$ \\
Night & Static & MM-DistillNet & $75.10$ & $89.63$ & $73.23$ & $1.43$ & $0.70$ \\
Day & Driving & MM-DistillNet & $55.73$ & $81.51$ & $52.75$ & $1.24$ & $0.76$ \\
Night & Driving & MM-DistillNet & $48.13$ & $67.16$ & $46.07$ & $1.93$ & $1.50$ \\
\bottomrule
\end{tabular}
\end{center}
\vspace{-0.2cm}
\caption{Performance comparison of MM-DistillNet in different illumination conditions as well as driving and static (data collection vehicle) states. Our framework outperform the previous state-of-the-art even in static-day conditions, where the RGB teacher performs the best.}
\label{table:resultsdiffconditions}
\end{table*}

\begin{figure*}
\footnotesize
\centering
\setlength{\tabcolsep}{0.2cm}
{\renewcommand{\arraystretch}{0.4}
\begin{tabular}{p{0.4cm}p{0.45\textwidth}p{0.45\textwidth}}
& \multicolumn{1}{c}{\textbf{StereoSoundNet~\cite{wang2020score}}} & \multicolumn{1}{c}{\textbf{MM-DistillNet (Ours)}} \\
\\

\textbf{t} & \raisebox{-0.4\height}{\includegraphics[width=\linewidth]{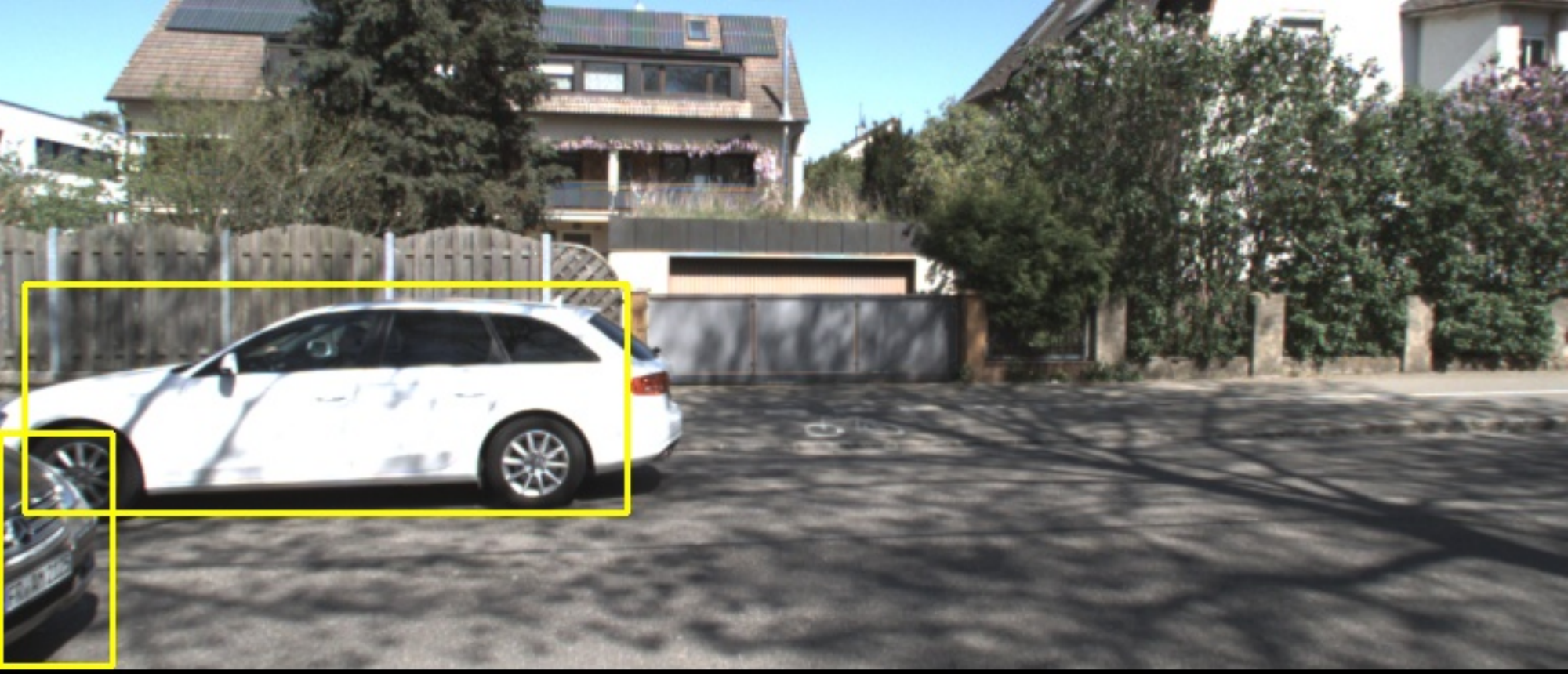}} &  \raisebox{-0.4\height}{\includegraphics[width=\linewidth]{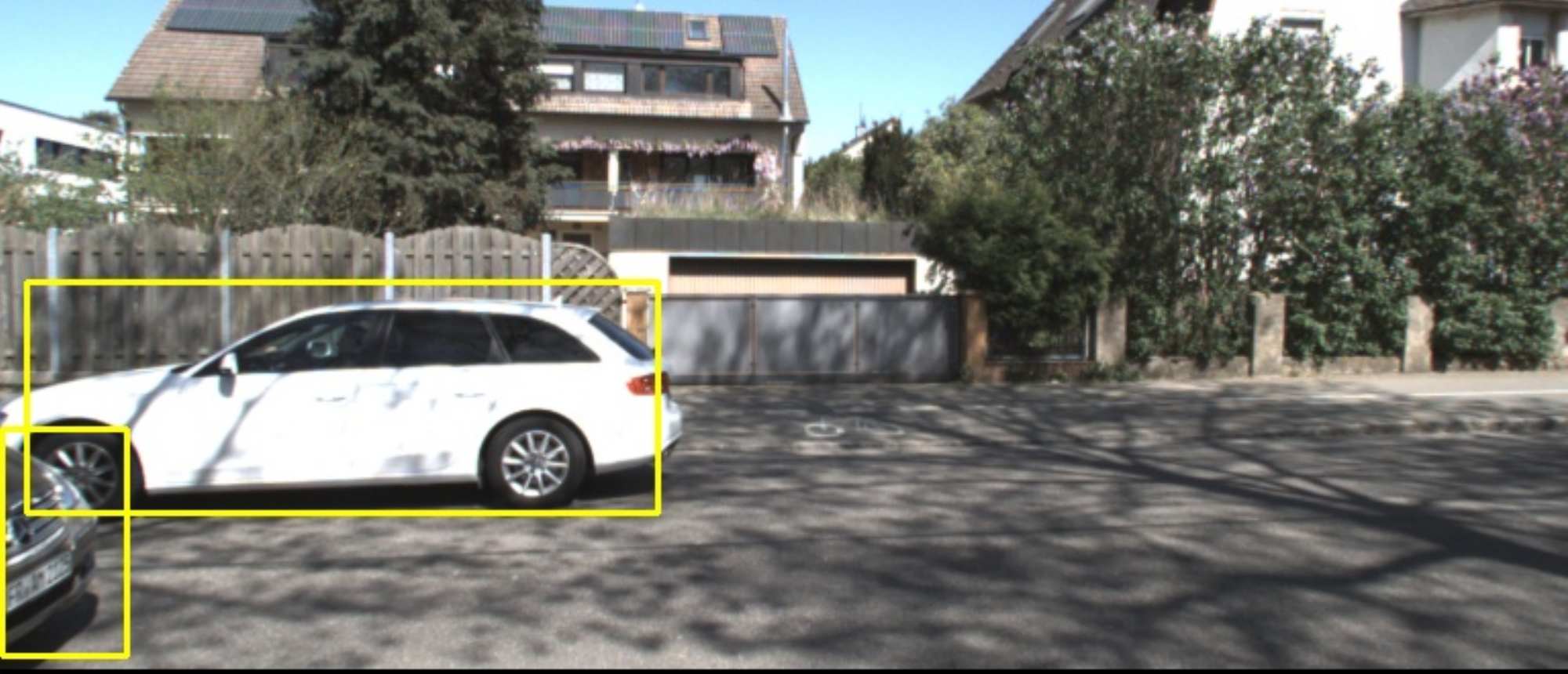}} \\
\\
\textbf{t+1} & \raisebox{-0.4\height}{\includegraphics[width=\linewidth]{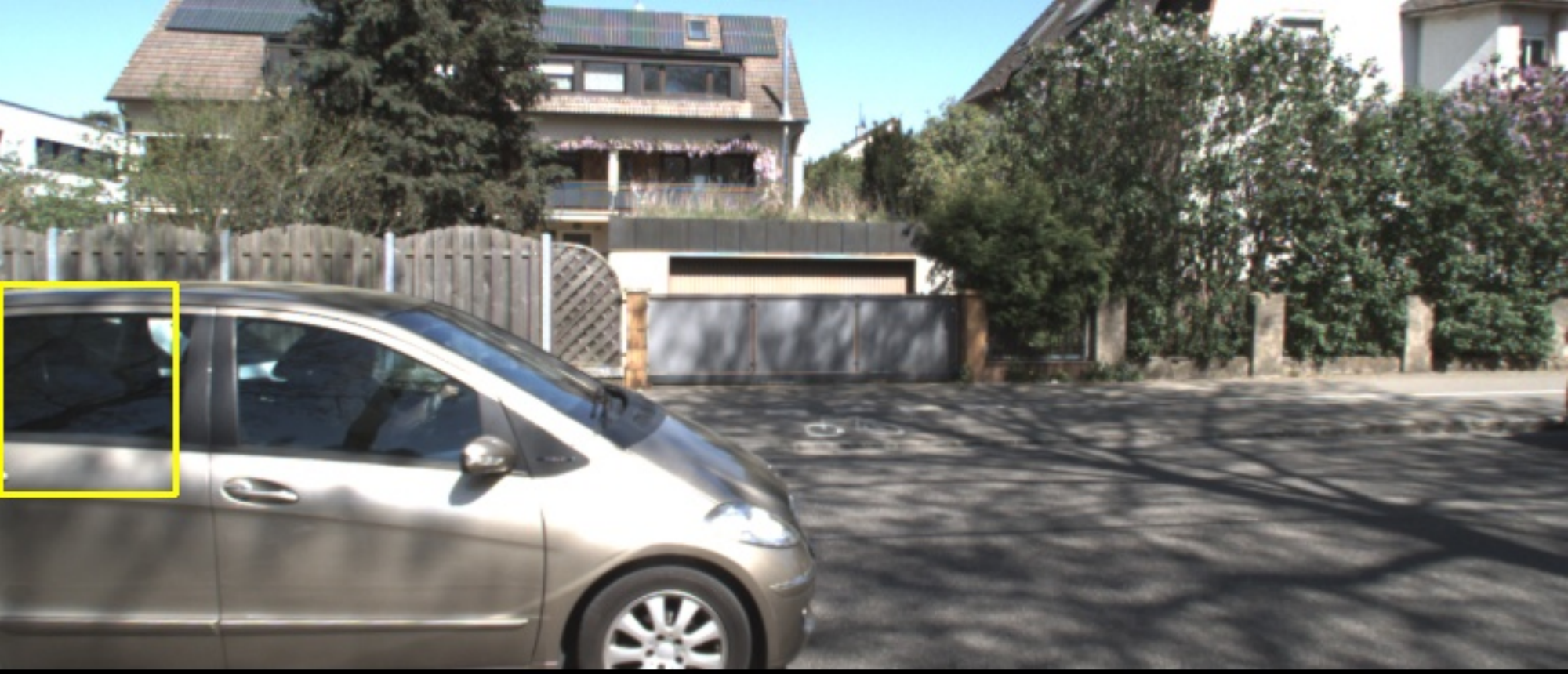}} &  \raisebox{-0.4\height}{\includegraphics[width=\linewidth]{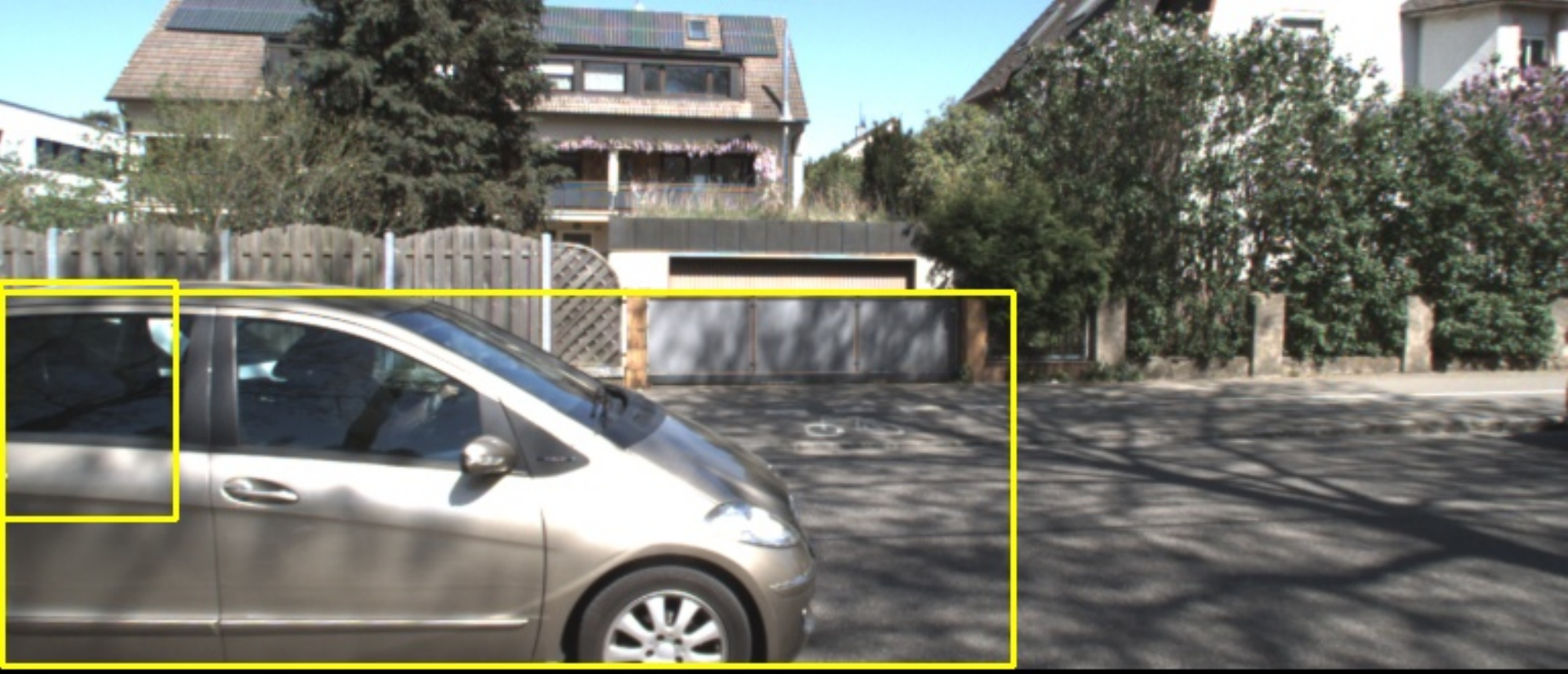}} \\
\\
\end{tabular}}
\caption{Qualitative comparison of detection performance with the previous state-of-the-art StereoSoundNet~\cite{wang2020score} and our MM-DistillNet. We present a scenario with an occluded car to demonstrates the novelty of using sound which overcomes limitations of visual modalities.}
\label{fig:SM_qualitativeoccluded}
\end{figure*}

In this section, we compare the performance our proposed MM-DistillNet and the previous state-of-the-art StereoSoundNet~\cite{gan2019self} in different illumination and driving conditions. StereoSoundNet is trained under the supervision of the RGB teacher, whereas our MM-DistillNet uses the RGB, depth, and thermal teachers. \tabref{table:resultsdiffconditions} presents results in terms of the average precision metric for each of these conditions. We can see that in every scenario our proposed MM-DistillNet substantially outperforms StereoSoundNet, thereby achieving state-of-the-art performance. We observe the largest improvement during night time conditions where StereoSoundNet significantly falls behind. It can also be observed that even during the day when the data collection vehicle is not in motion, our MM-DistillNet achieves over $10\%$ improvement in the mAP @ Average. We observe a performance drop in both the methods from static to driving conditions, which can be attributed to the distortion of sound due to the moving data collection vehicle, wind on the microphone (note that the microphones in the array were not equipped with a wind muff) and high ambient noise conditions. We believe that high fidelity microphones and hard negative mining will help overcome this problem. This experiment demonstrates that incorporating knowledge from multiple modality-specific teachers improves the performance of the audio student, especially while the data collection vehicle is in motion and in low illumination conditions. 

\section{Extended Qualitative Results}
\label{sec:qualitativeS}

\begin{figure*}
\footnotesize
\centering
\setlength{\tabcolsep}{0.2cm}
{\renewcommand{\arraystretch}{0.4}
\begin{tabular}{p{0.45\textwidth}p{0.45\textwidth}}
     \multicolumn{1}{c}{\textbf{StereoSoundNet~\cite{wang2020score}}} & \multicolumn{1}{c}{\textbf{MM-DistillNet (Ours)}} \\
     \\
     \includegraphics[width=\linewidth]{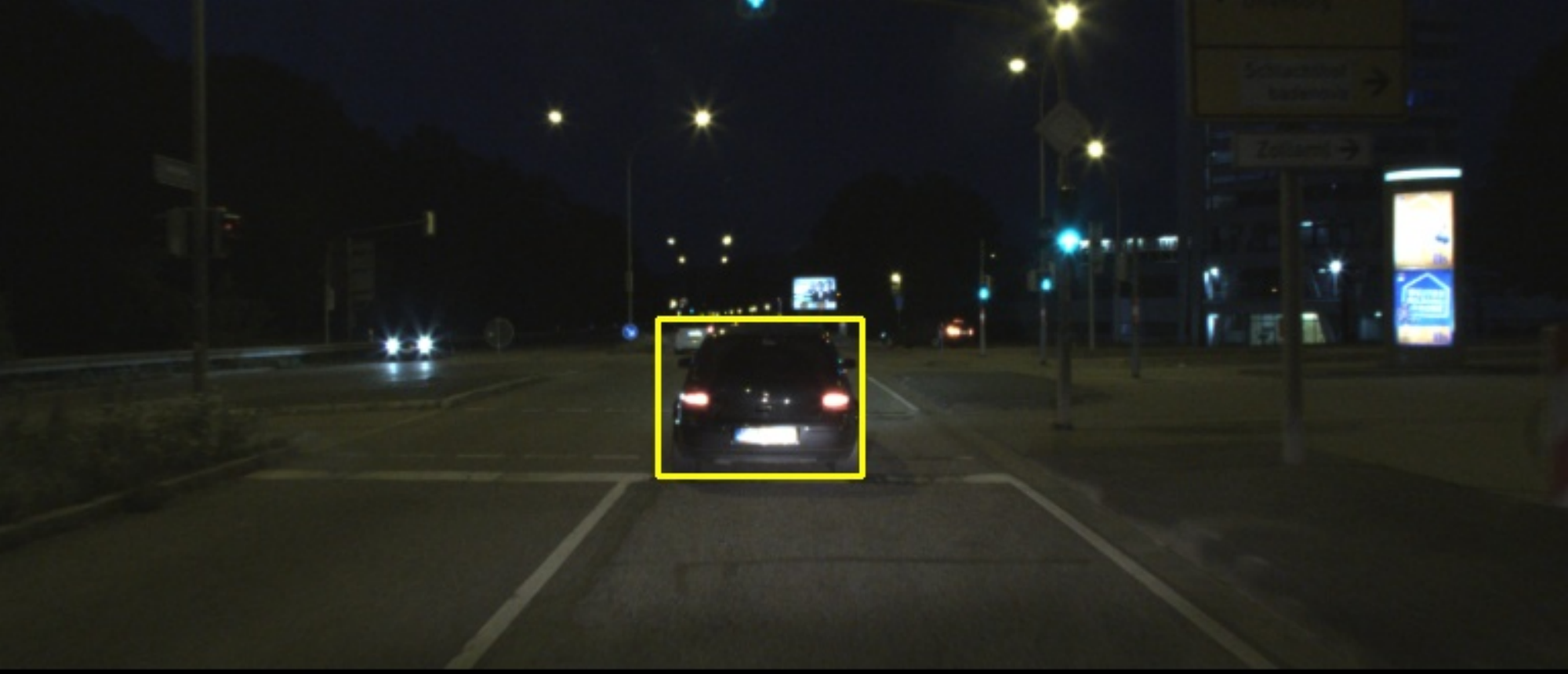} &  \includegraphics[width=\linewidth]{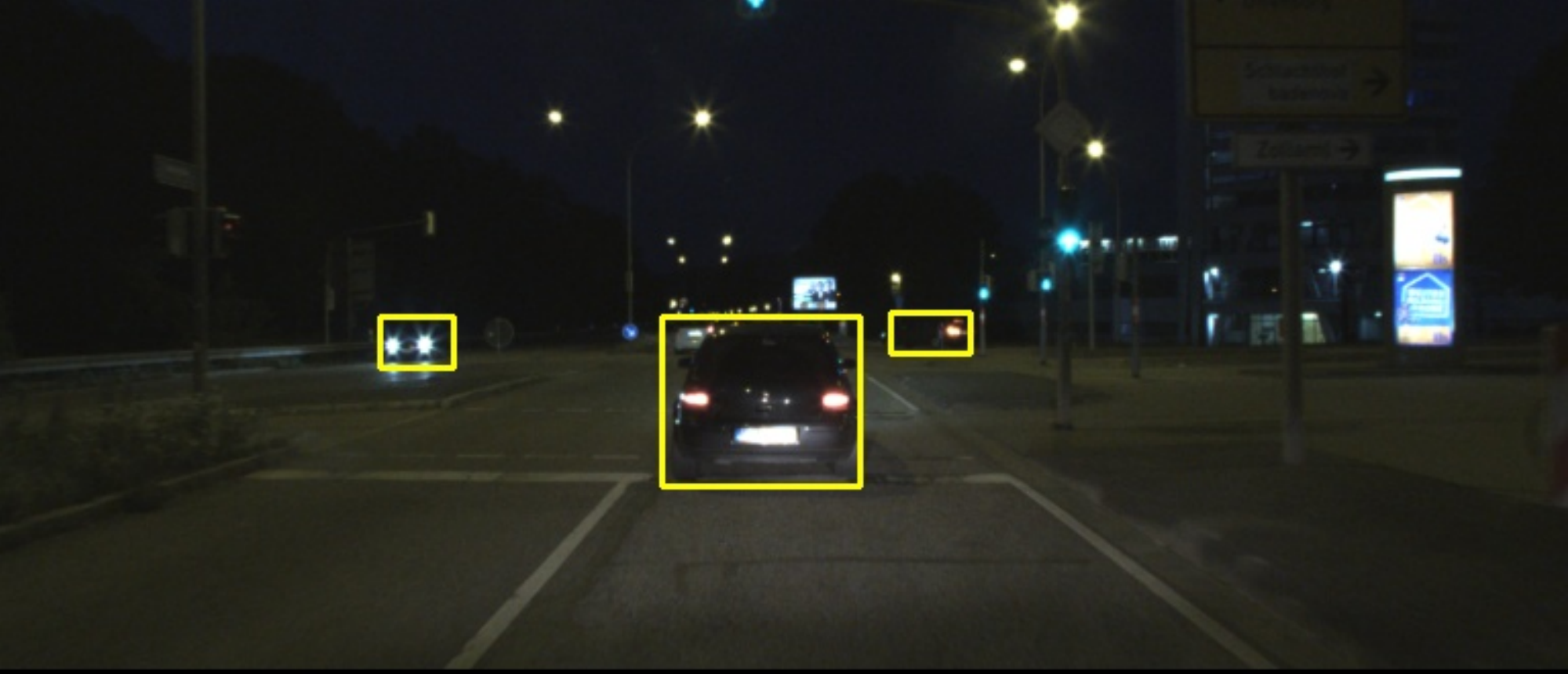} \\
     \includegraphics[width=\linewidth]{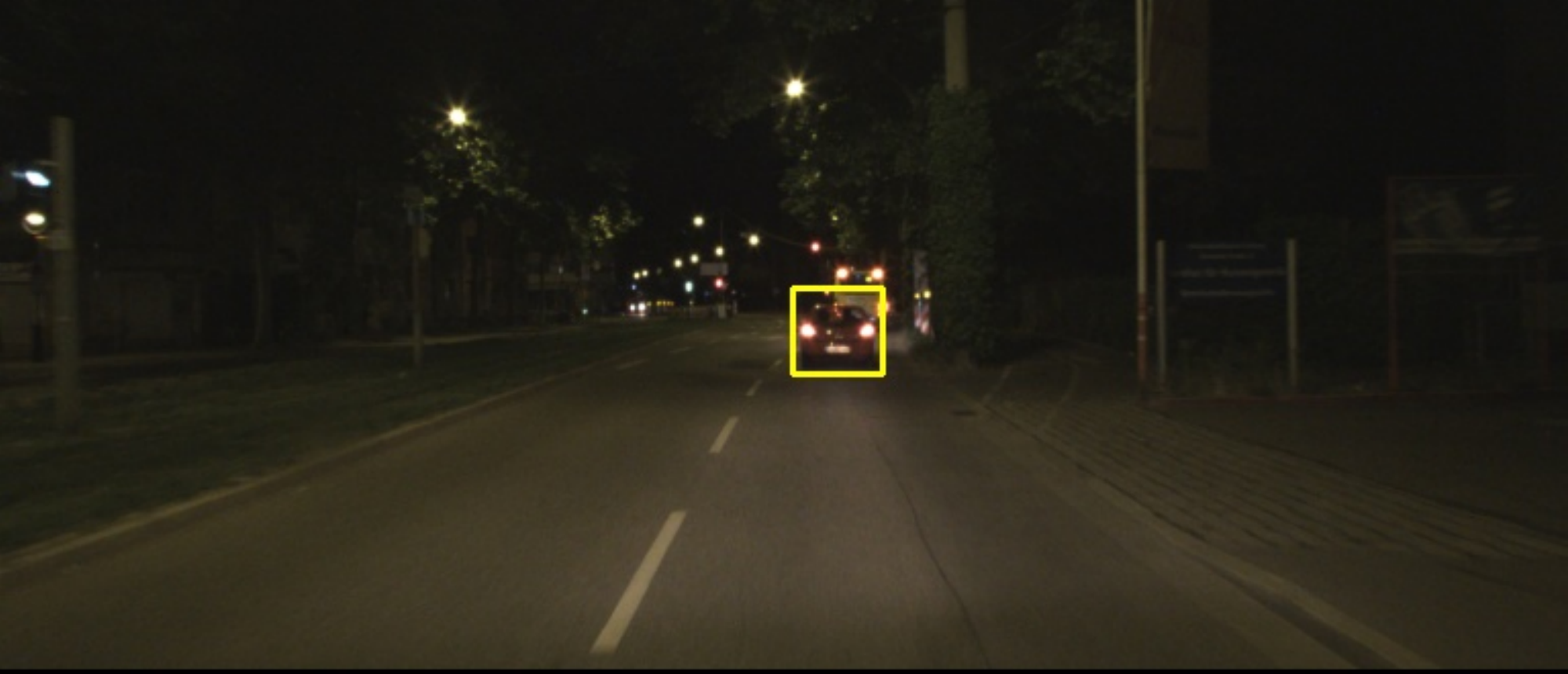} &  \includegraphics[width=\linewidth]{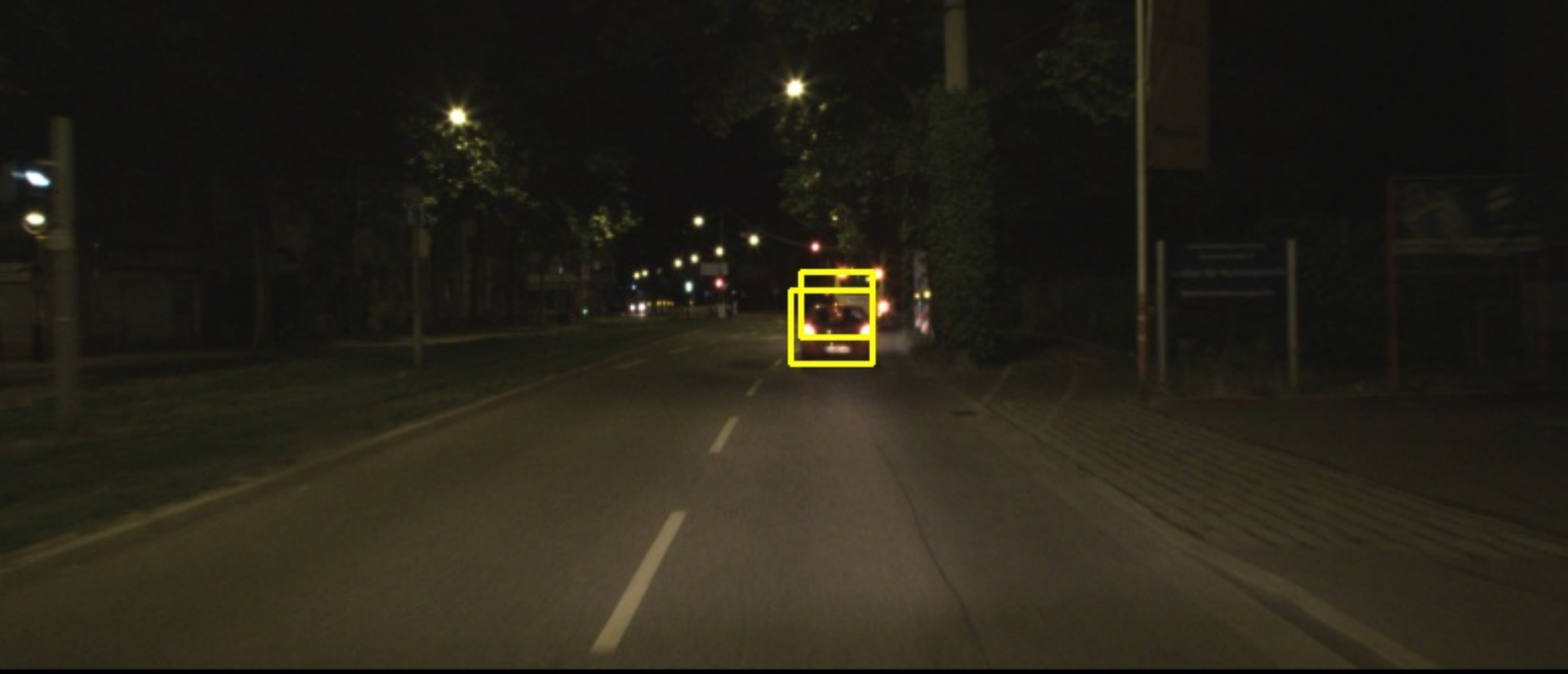} \\
     \includegraphics[width=\linewidth]{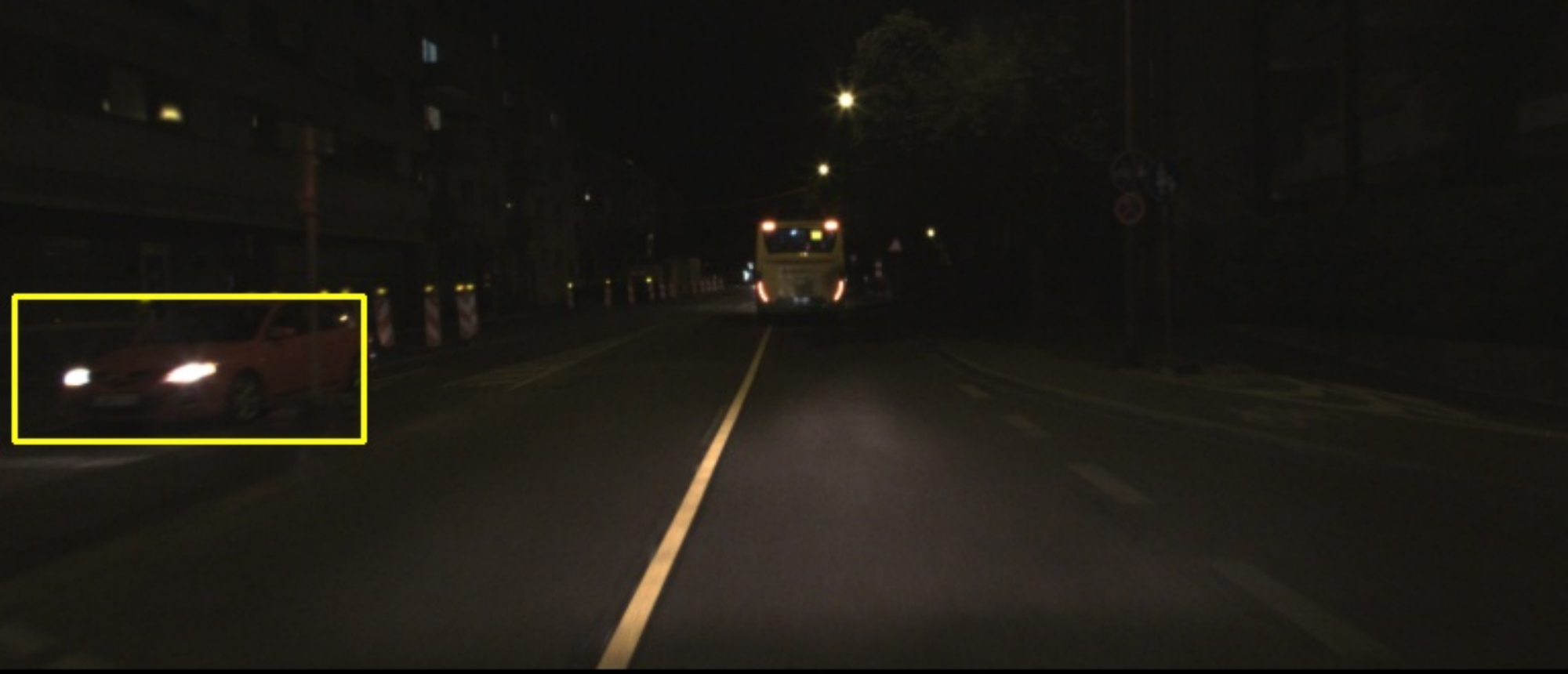} &  \includegraphics[width=\linewidth]{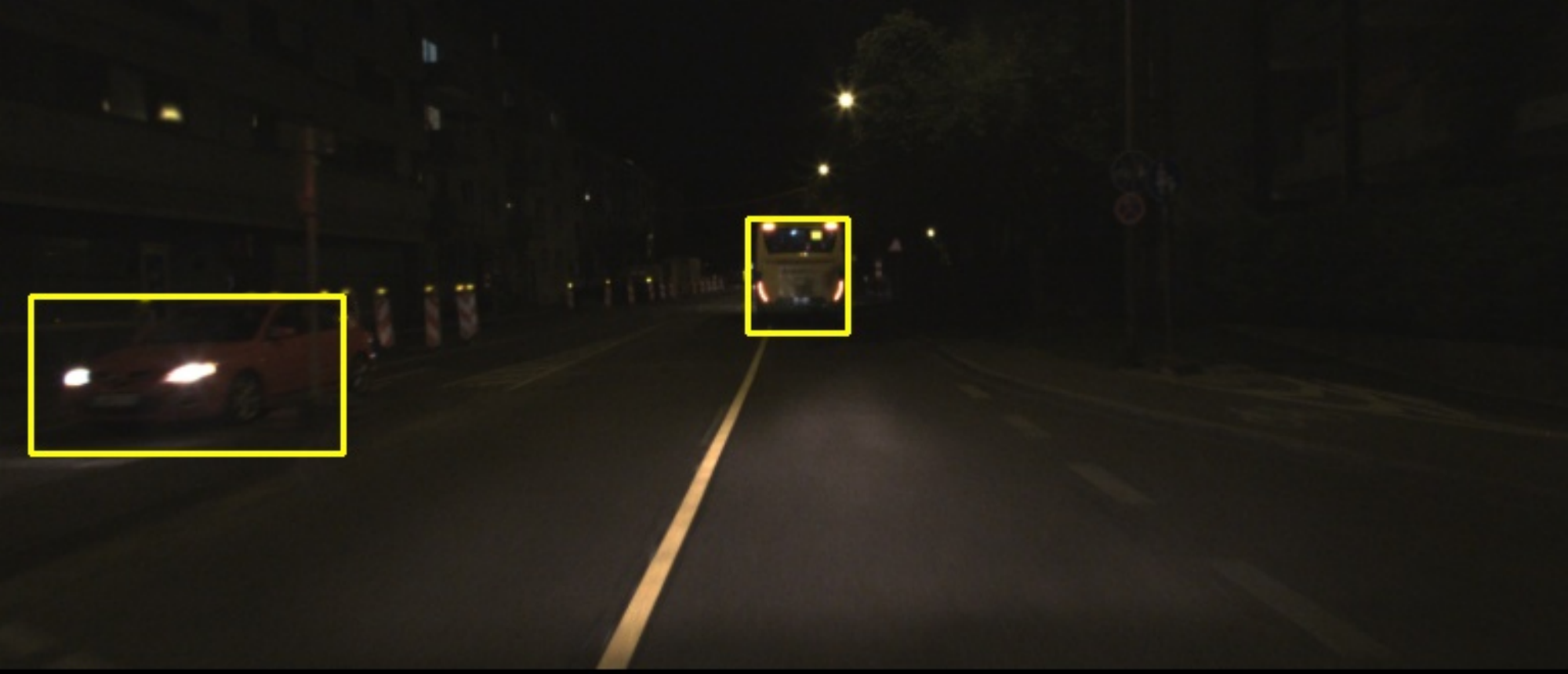} \\
     \includegraphics[width=\linewidth]{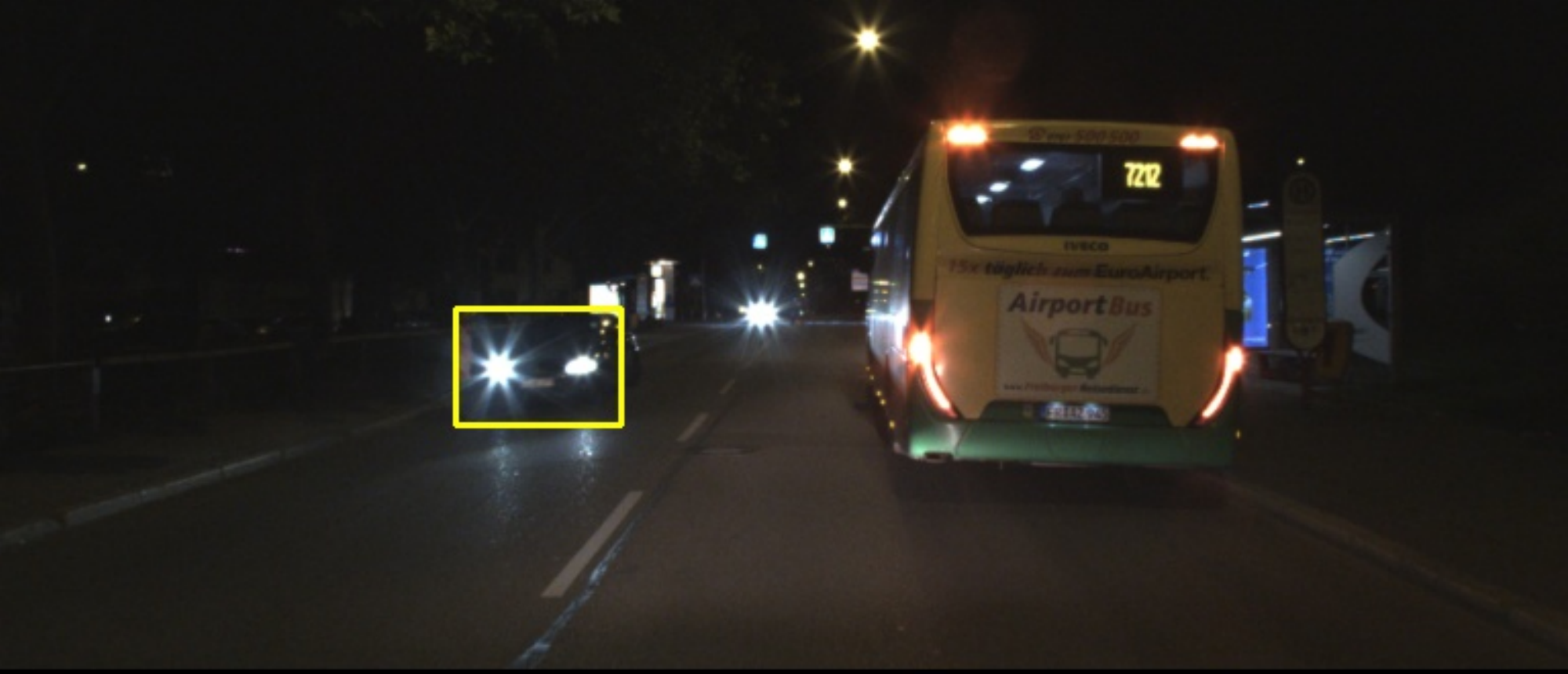} &  \includegraphics[width=\linewidth]{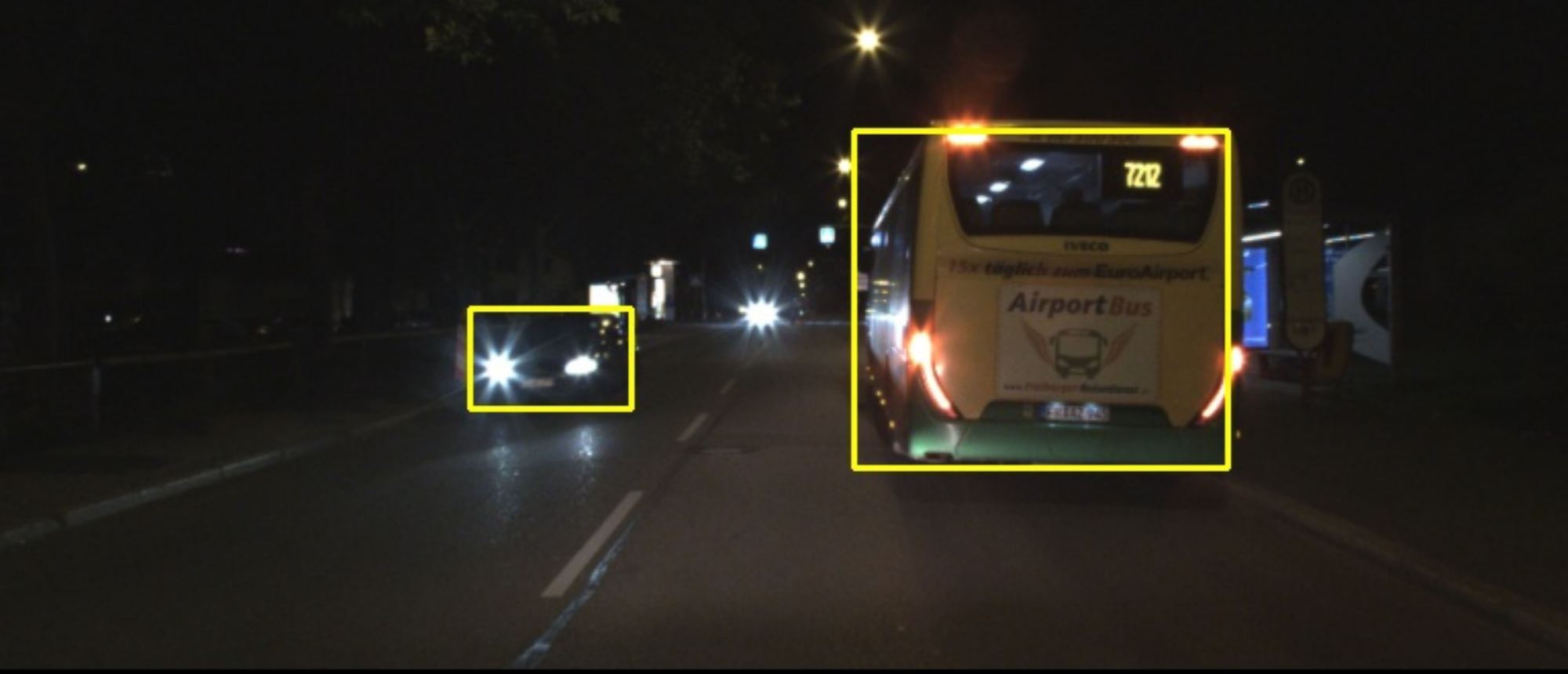} \\
     \includegraphics[width=\linewidth]{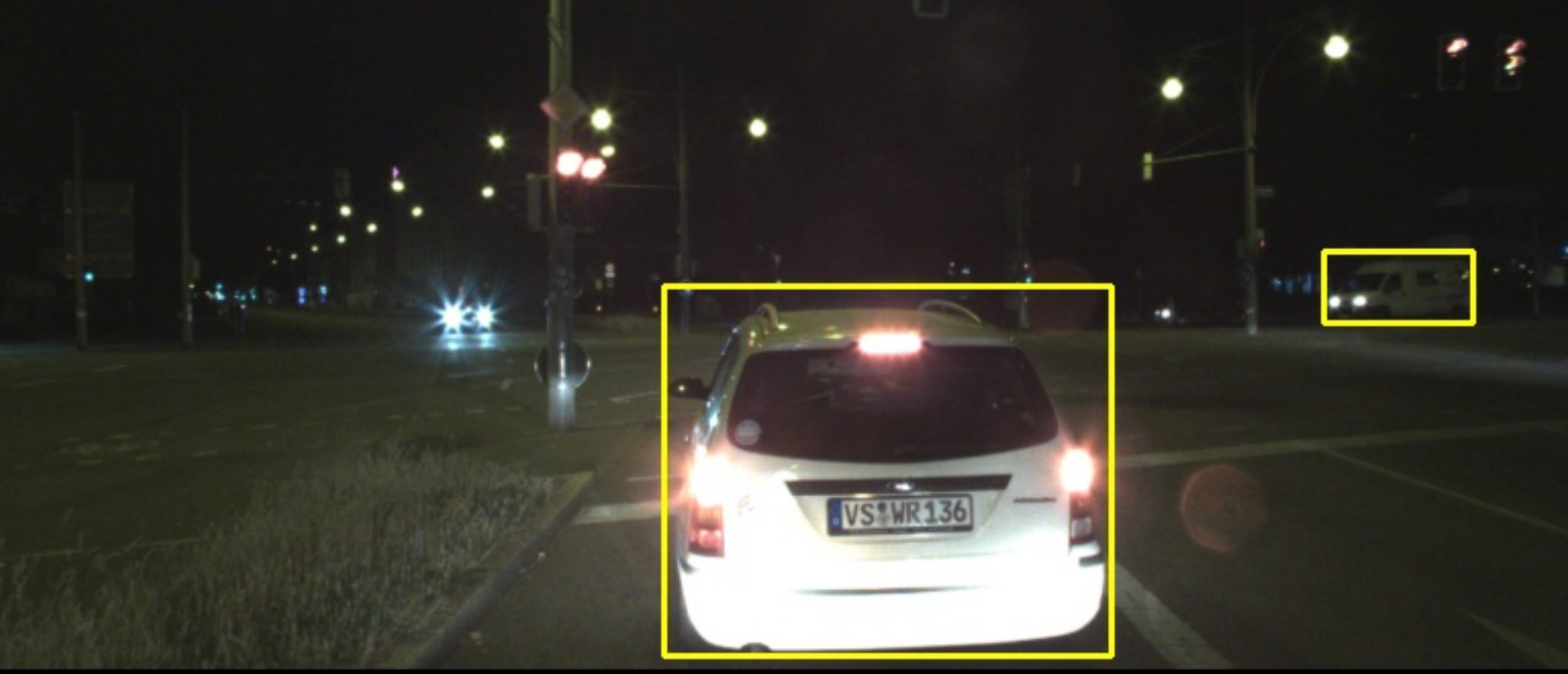} &  \includegraphics[width=\linewidth]{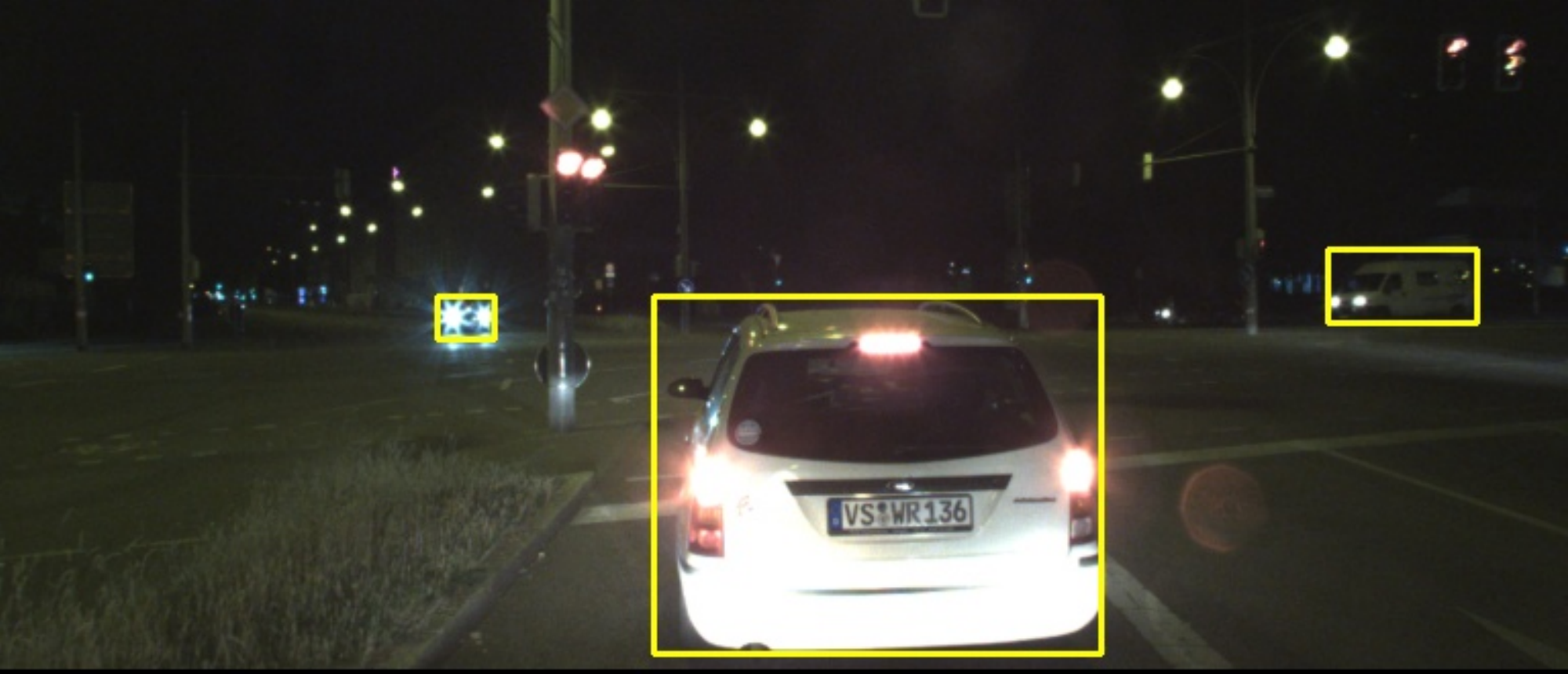} \\
     \includegraphics[width=\linewidth]{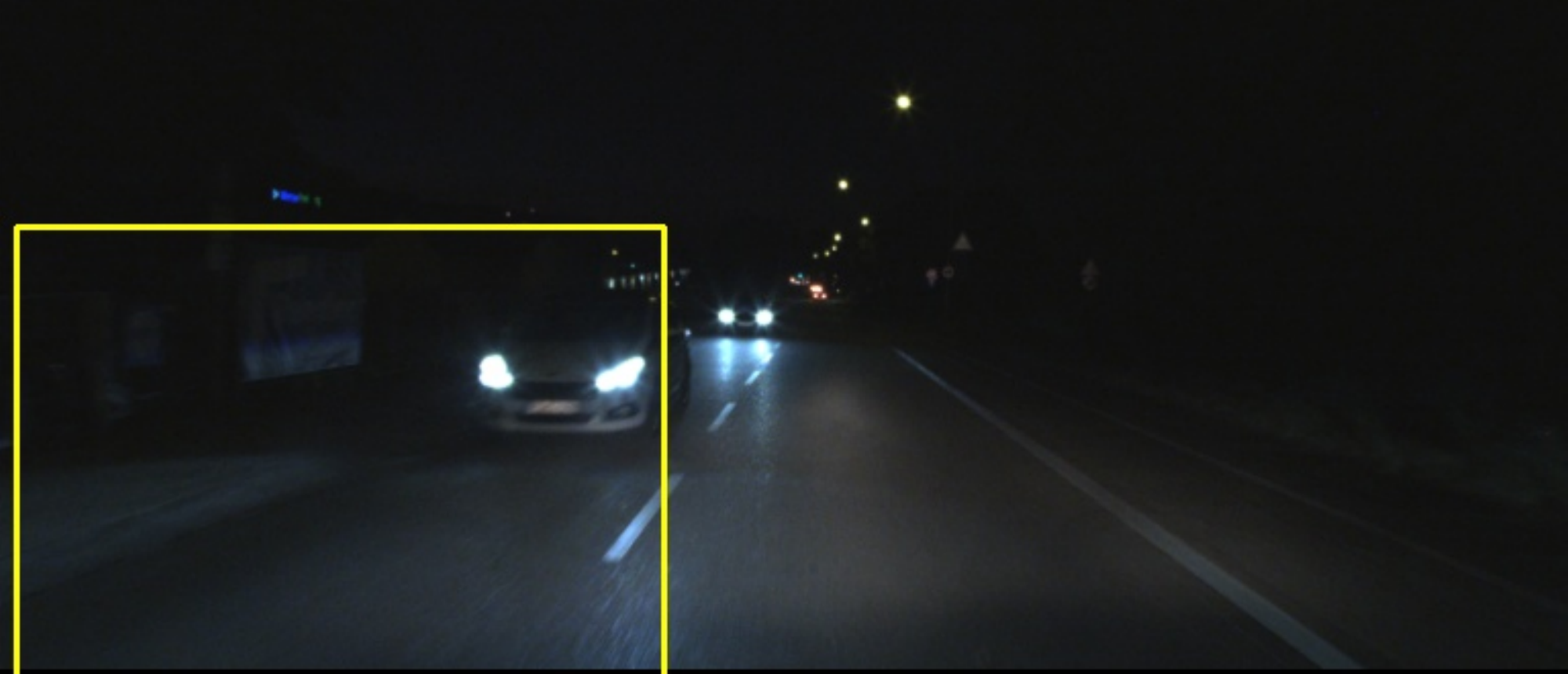} &  \includegraphics[width=\linewidth]{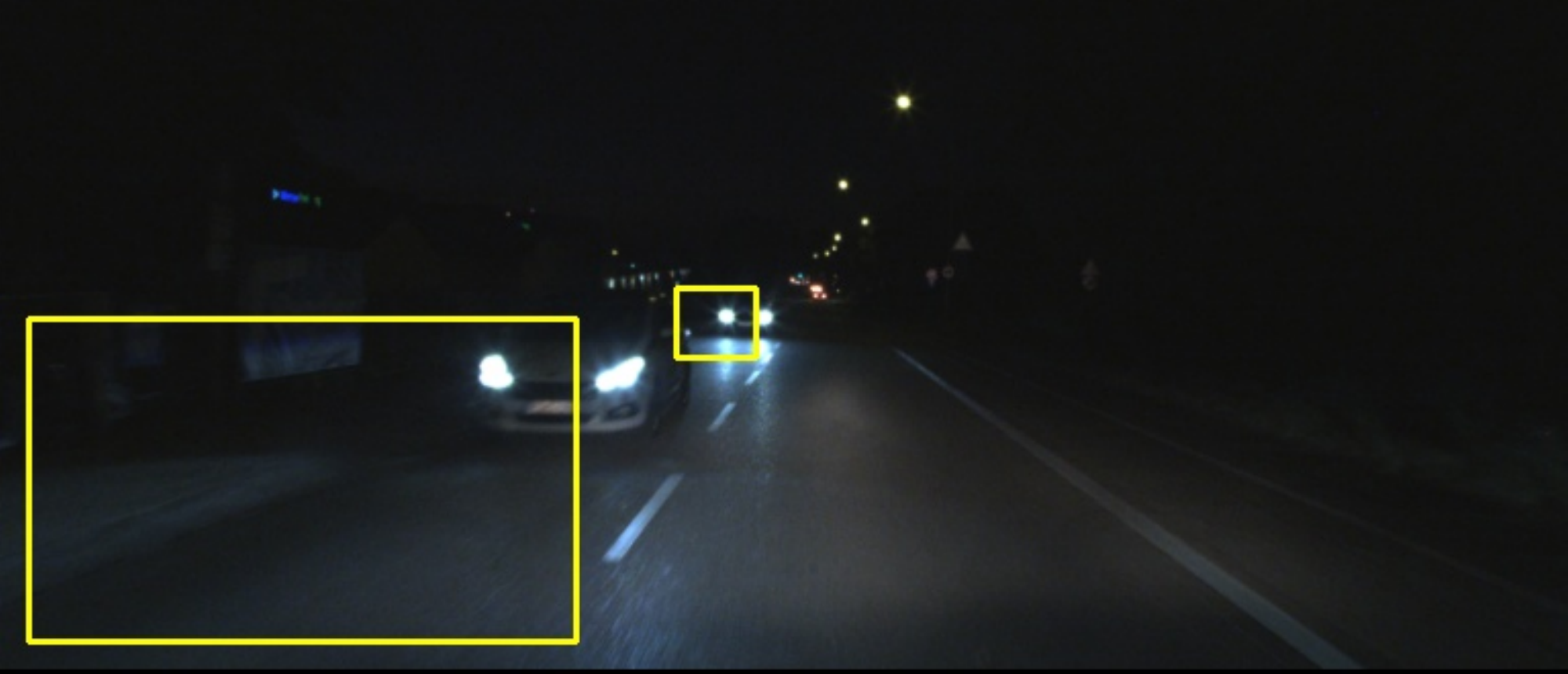} \\
\end{tabular}}
\caption{Qualitative comparisons of predictions in night scenes from our MM-DistillNet and the single student-teacher StereoSoundNet~\cite{wang2020score}. We present hard scenarios in poor lighting condition to demonstrate how our model does not suffer from the day to night domain gap.}
\label{fig:SM_qualitativenight}
\end{figure*}

In this section, we extend the qualitative evaluations of our proposed MM-DistillNet. We provide further results that demonstrate that our MM-DistillNet effectively employs the knowledge of diverse multimodal pre-trained teachers to improve the performance of vehicle detection. We first highlight how the audio modality is able to overcome the visual limitation by detecting occluded cars in \figref{fig:SM_qualitativeoccluded} where the white car at time $t$ is occluded by the vehicle that appears on the left at $t+1$. However, StereoSoundNet fails to detect the foreground car, while our MM-DistillNet precisely detects both the vehicles in the scene, despite the background car not being visible.

\begin{figure*}
\footnotesize
\centering
\setlength{\tabcolsep}{0.05cm}
{\renewcommand{\arraystretch}{0.8}
\begin{tabular}{p{0.3\textwidth}p{0.3\textwidth}}
     \multicolumn{1}{c}{\textbf{StereoSoundNet~\cite{wang2020score}}} & \multicolumn{1}{c}{\textbf{MM-DistillNet (Ours)}} \\
     \includegraphics[width=\linewidth]{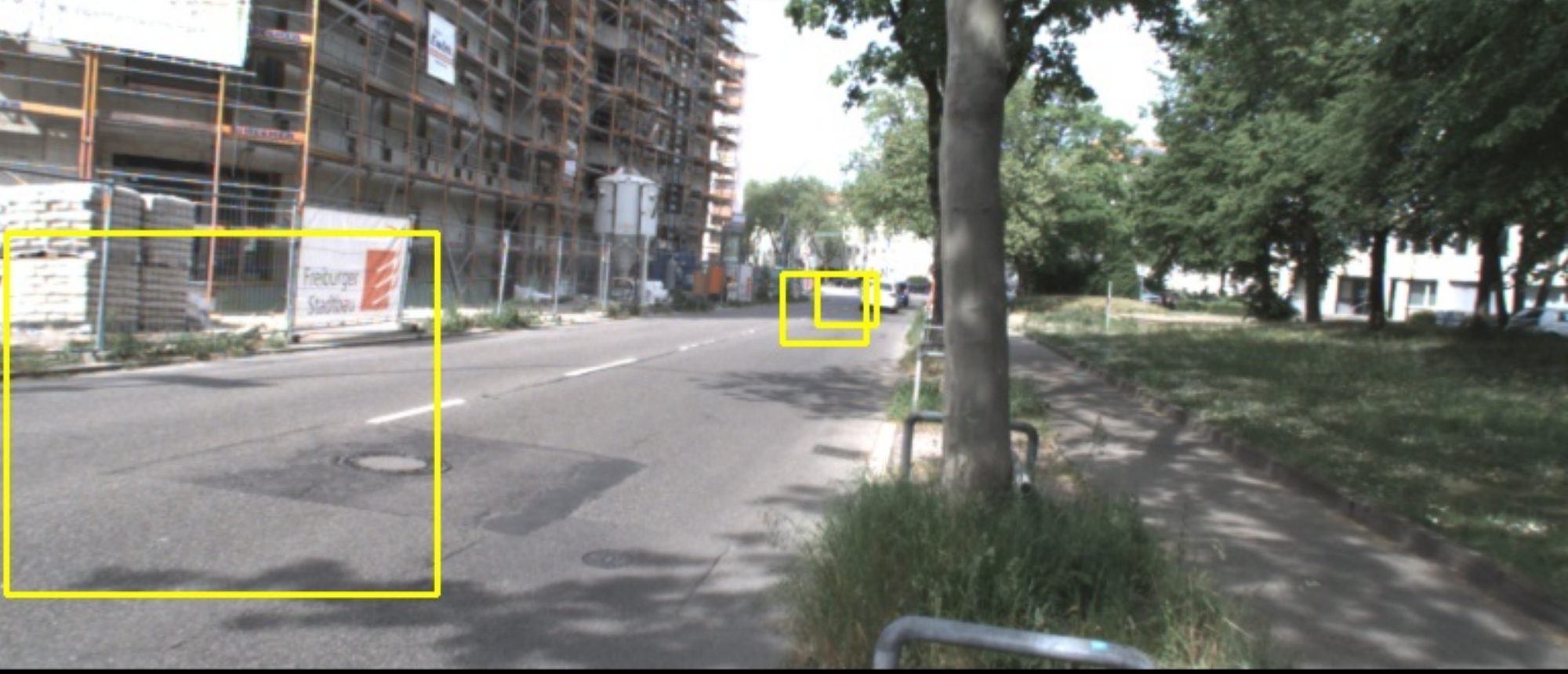} &  \includegraphics[width=\linewidth]{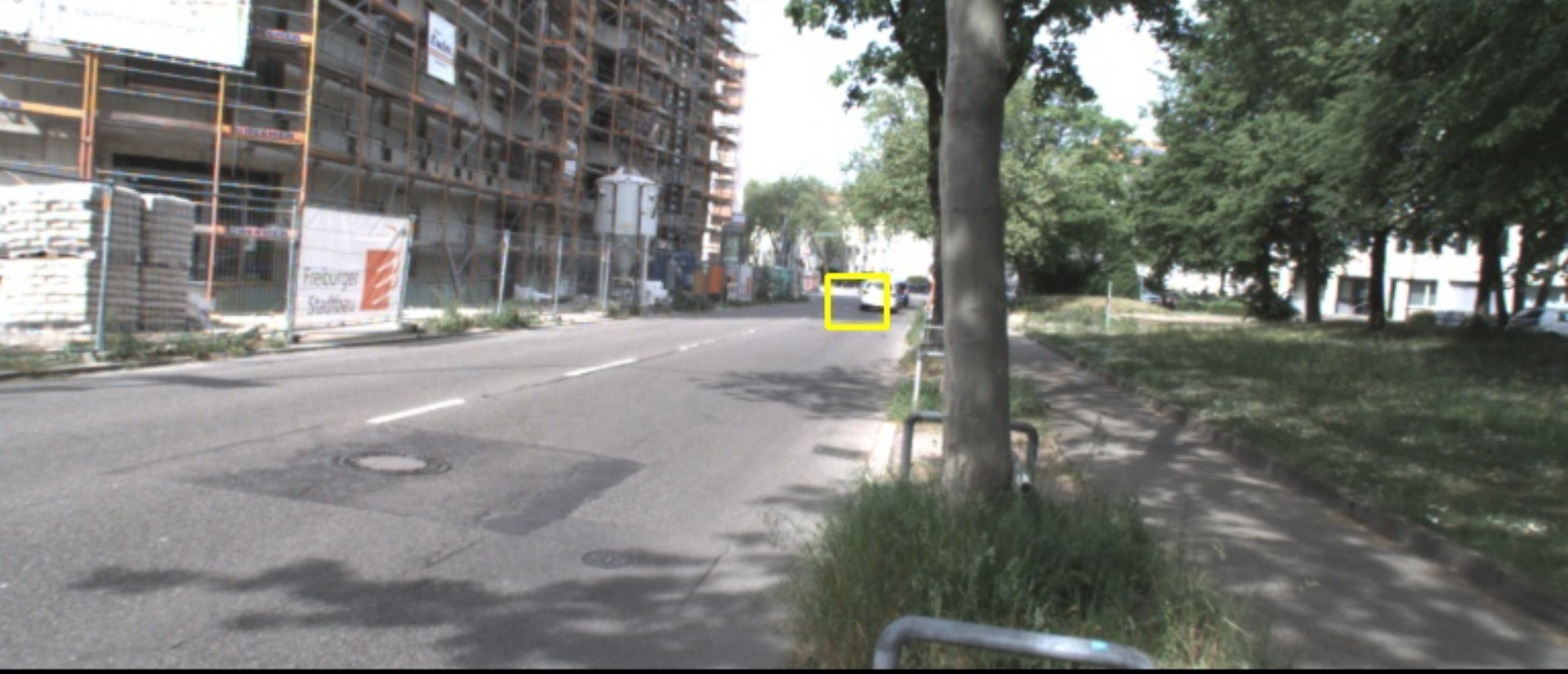} \\
     \includegraphics[width=\linewidth]{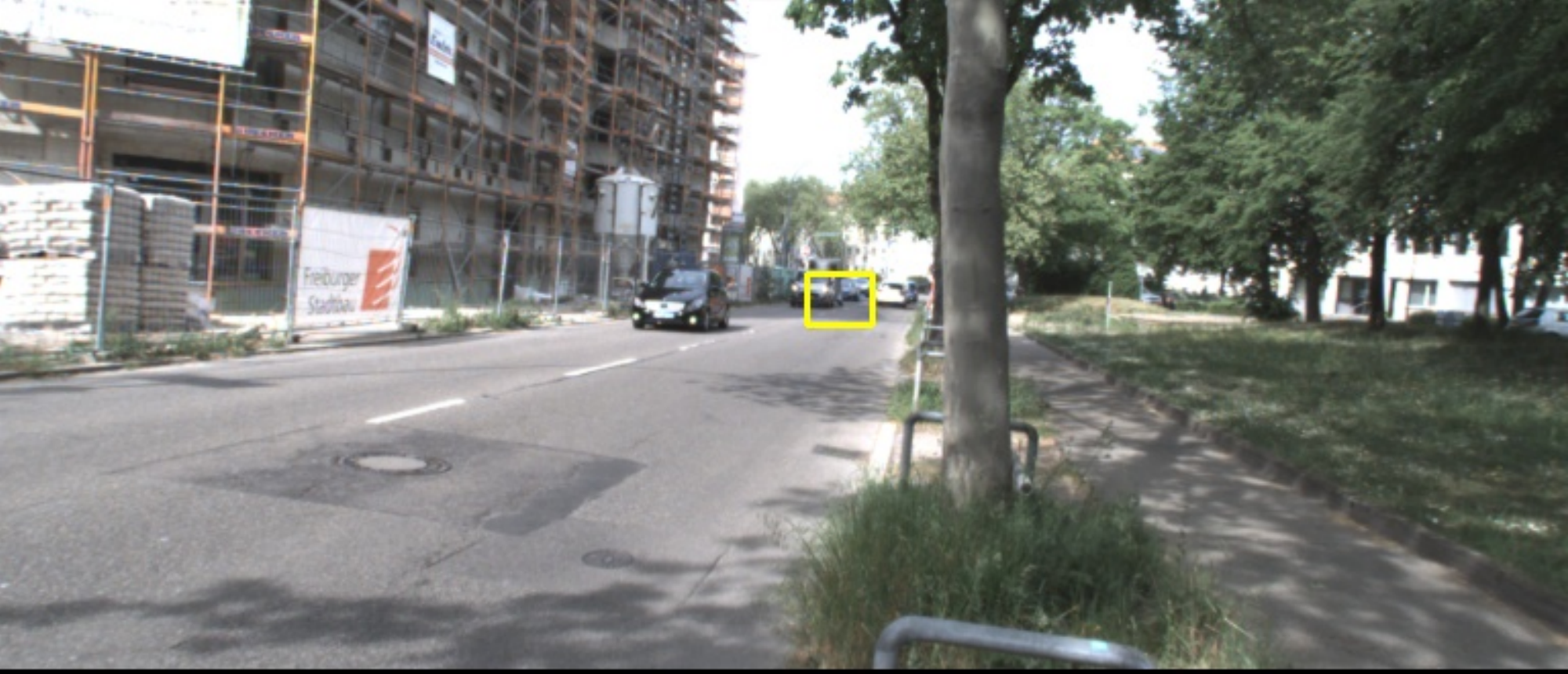} &  \includegraphics[width=\linewidth]{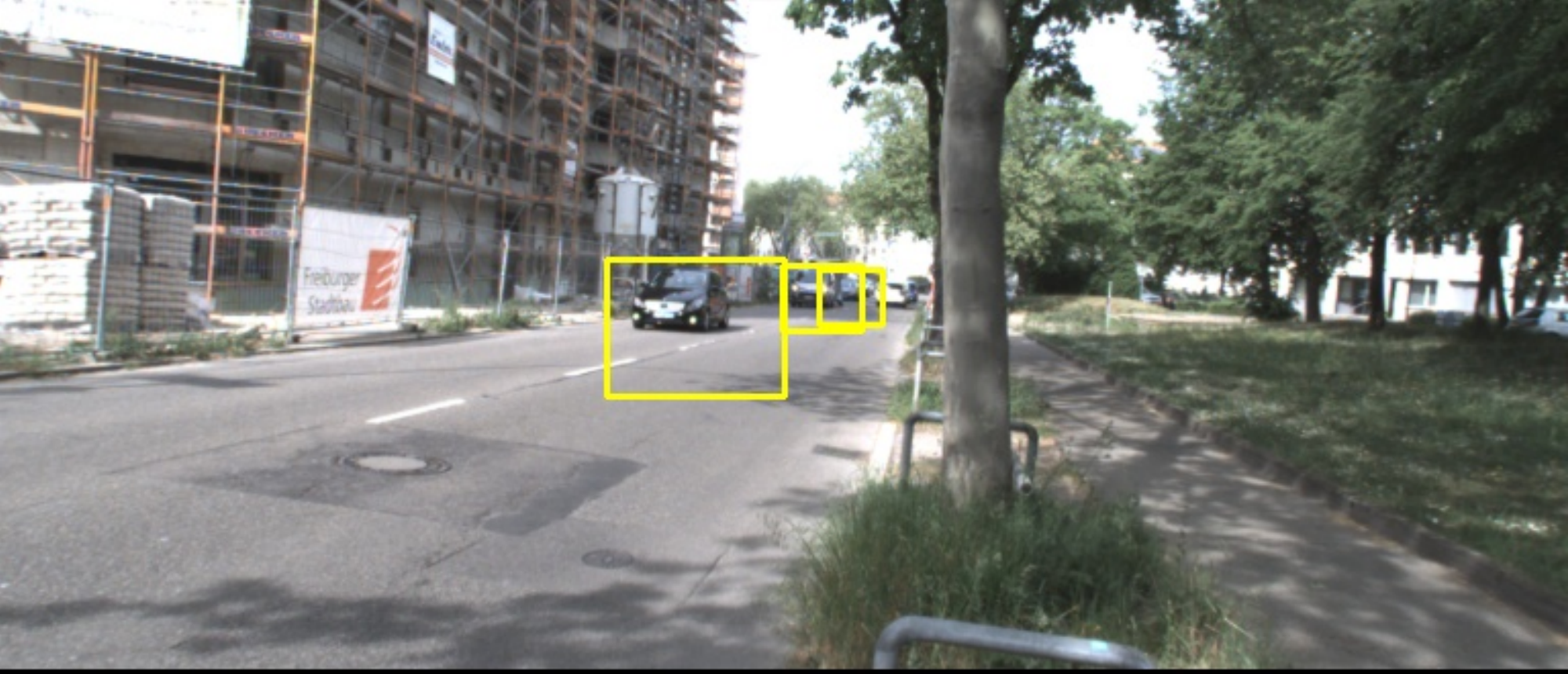} \\
     \includegraphics[width=\linewidth]{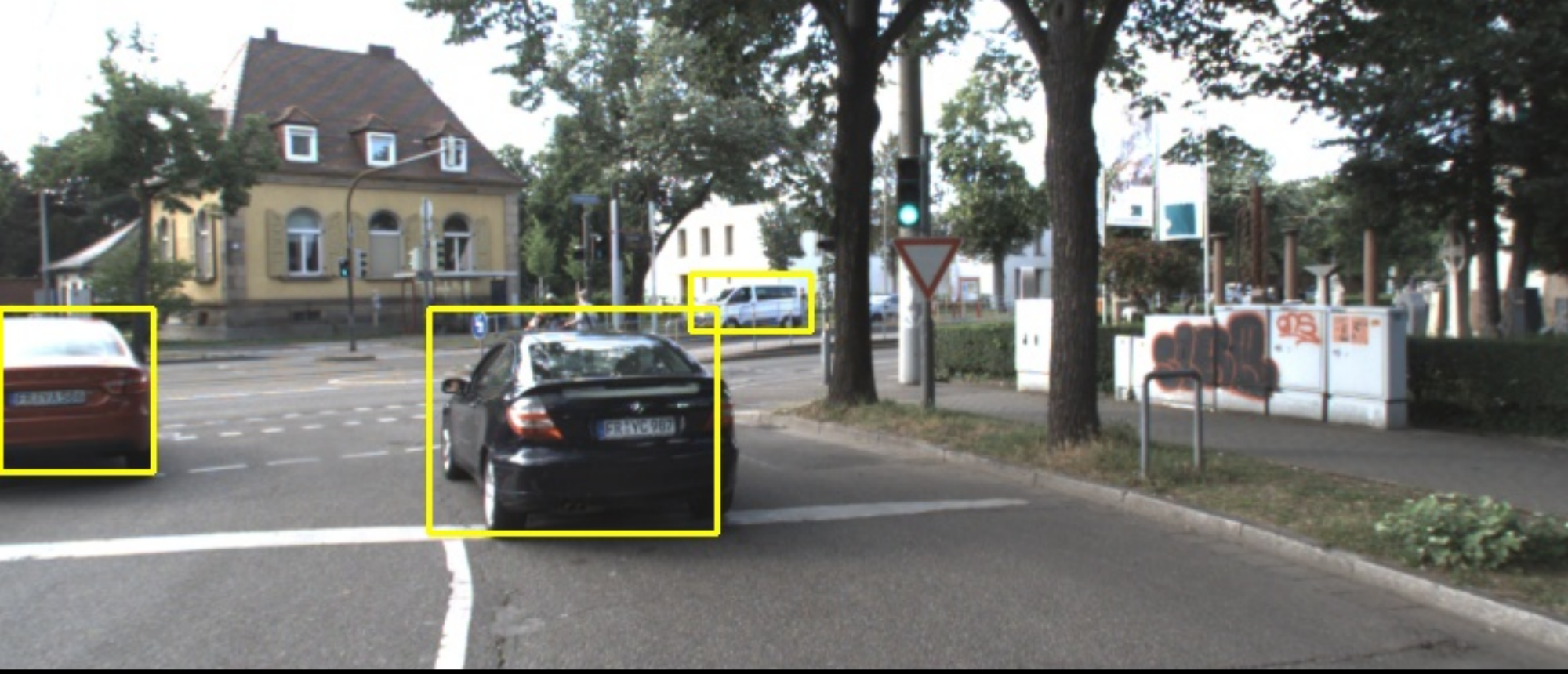} &  \includegraphics[width=\linewidth]{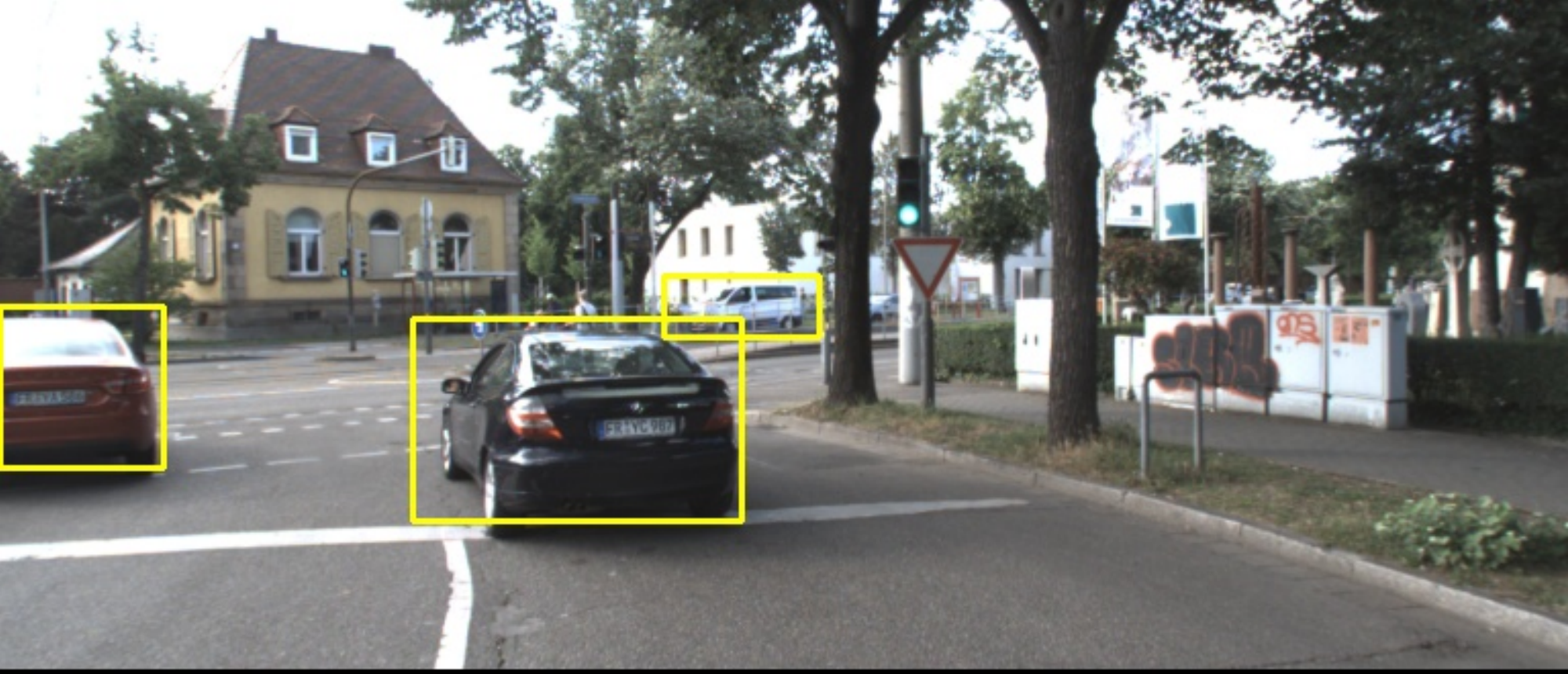} \\
     \includegraphics[width=\linewidth]{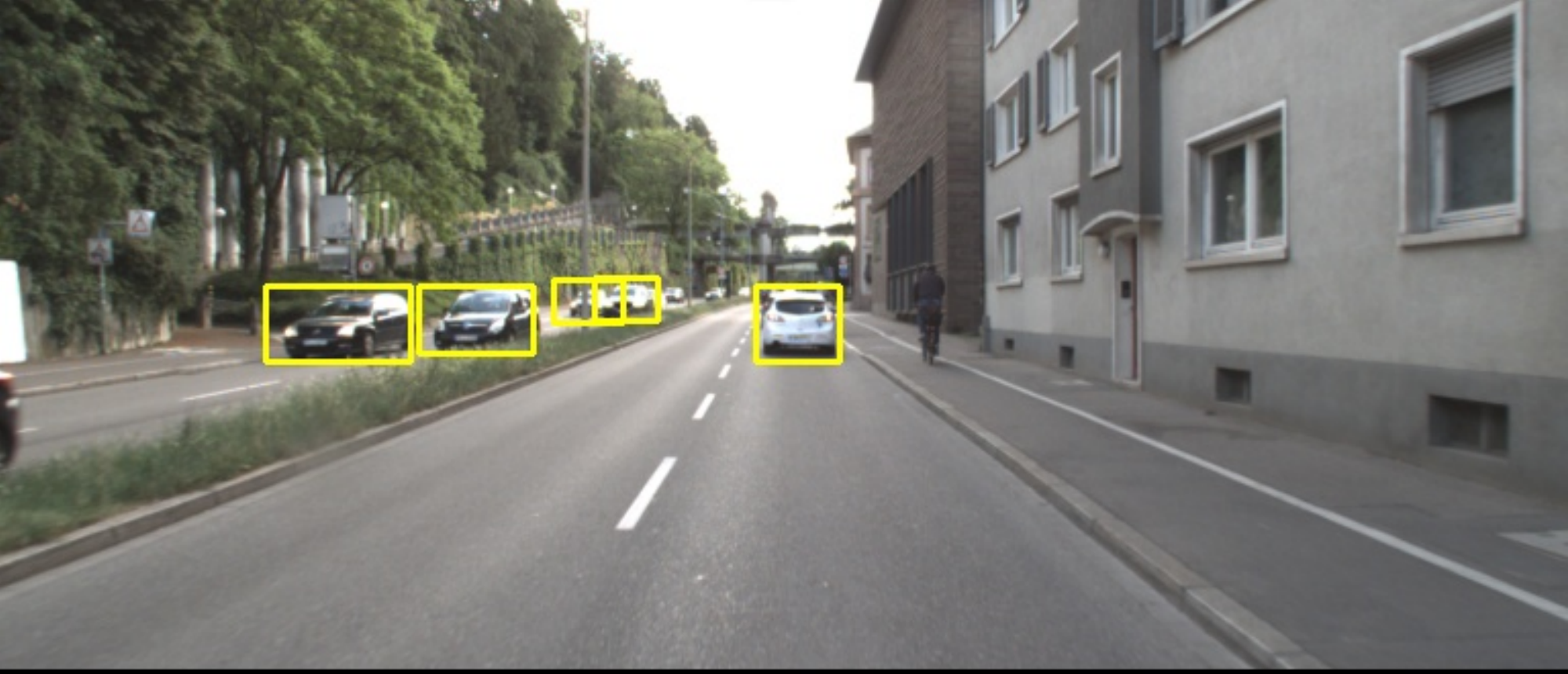} &  \includegraphics[width=\linewidth]{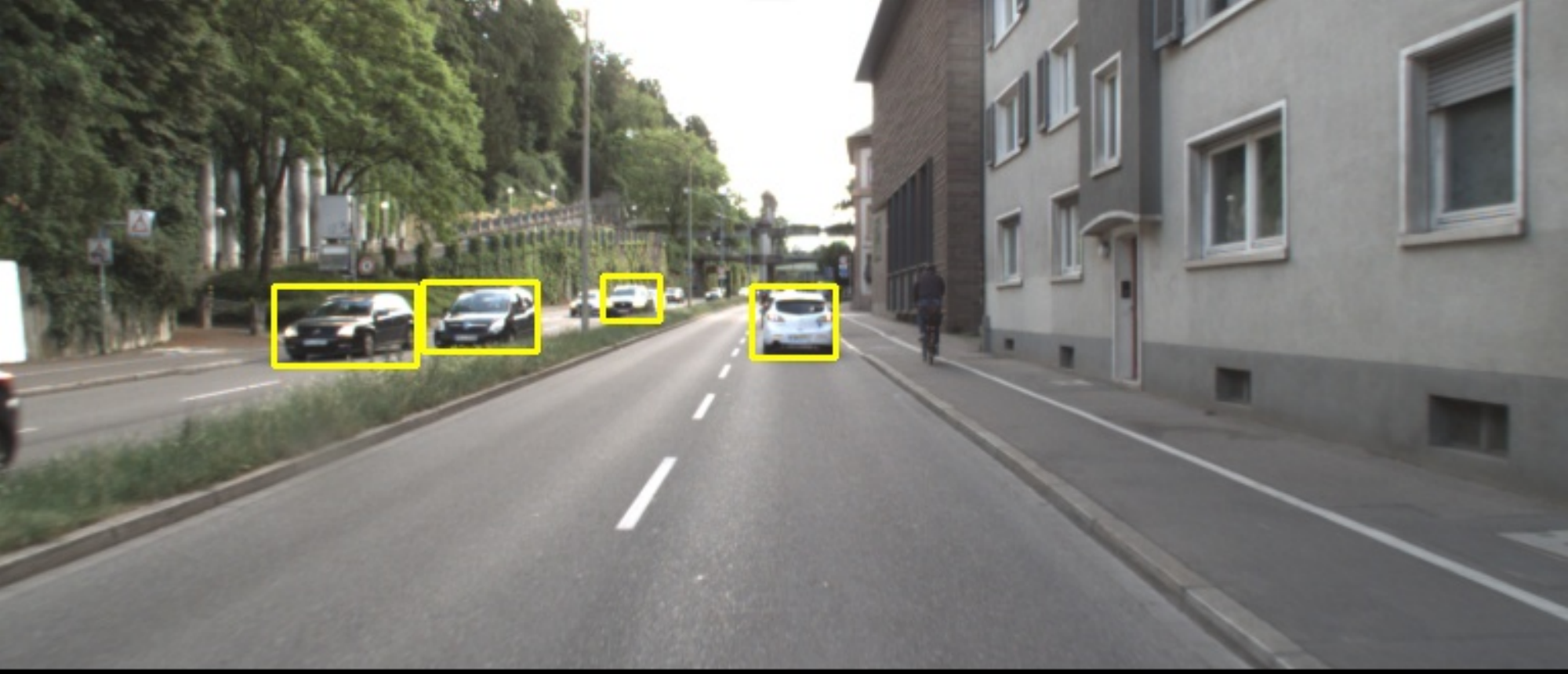} \\
\end{tabular}}
\caption{Qualitative comparisons of predictions of our MM-DistillNet and the previous state-of-the-art StereoSoundNet~\cite{wang2020score} in scenarios with multiple cars. We show that our network is able to simultaneously detect multiple cars, even while the data collection car is moving. Observe that our network is also able to detect very distant cars.}
\label{fig:SM_qualitativegoodmultiplecars}
\end{figure*}

\begin{figure*}
\footnotesize
\centering
\setlength{\tabcolsep}{0.05cm}
{\renewcommand{\arraystretch}{0.8}
\begin{tabular}{p{0.45\textwidth}p{0.45\textwidth}}
     \multicolumn{1}{c}{\textbf{StereoSoundNet~\cite{wang2020score}}} & \multicolumn{1}{c}{\textbf{MM-DistillNet (Ours)}} \\
     \includegraphics[width=\linewidth]{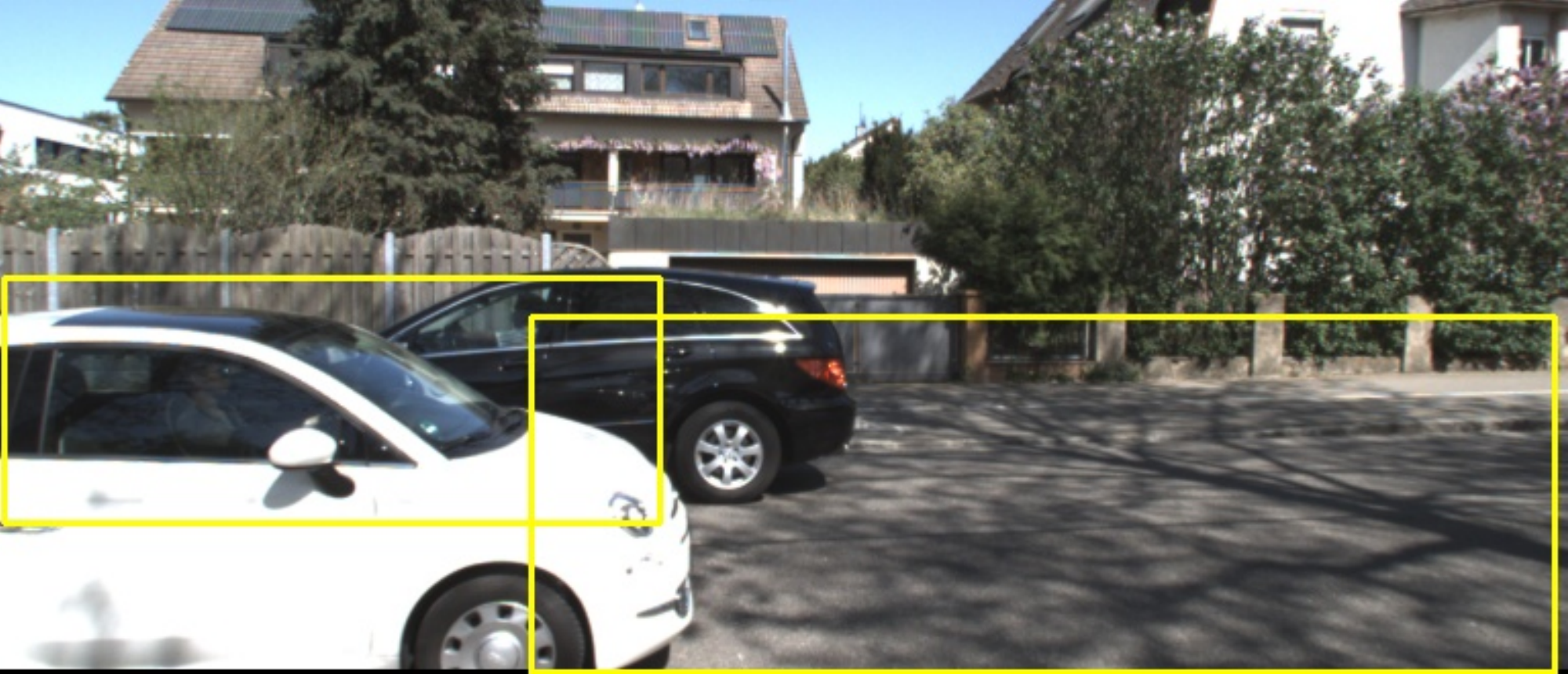} &  \includegraphics[width=\linewidth]{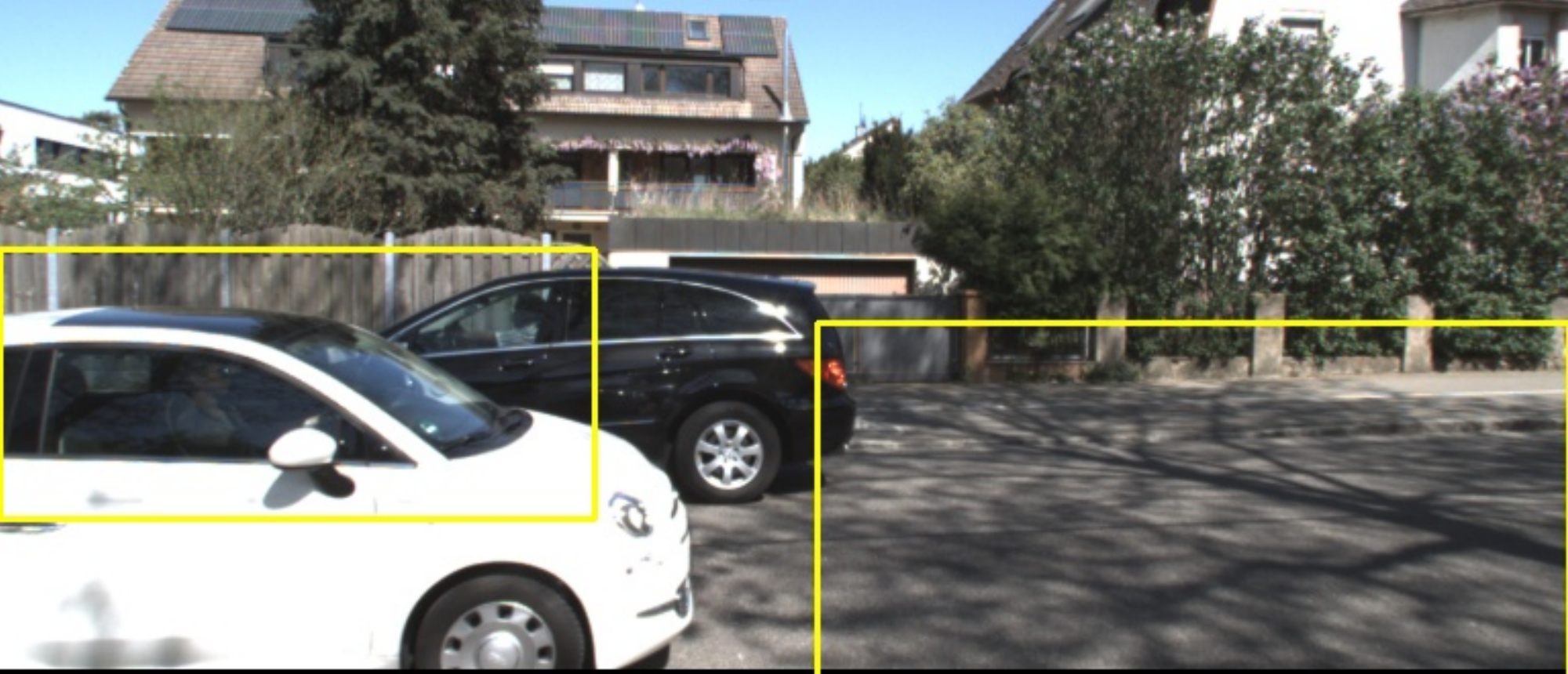} \\
     \includegraphics[width=\linewidth]{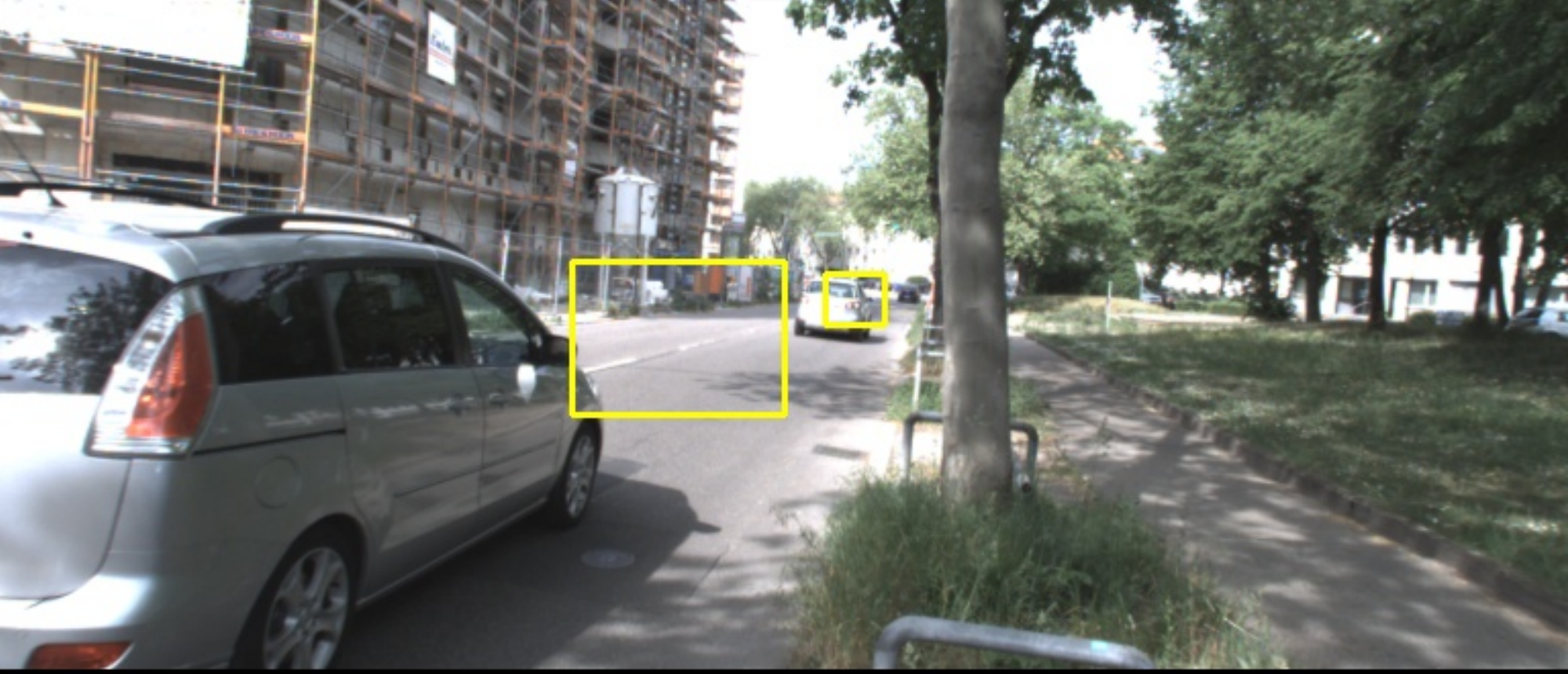} &  \includegraphics[width=\linewidth]{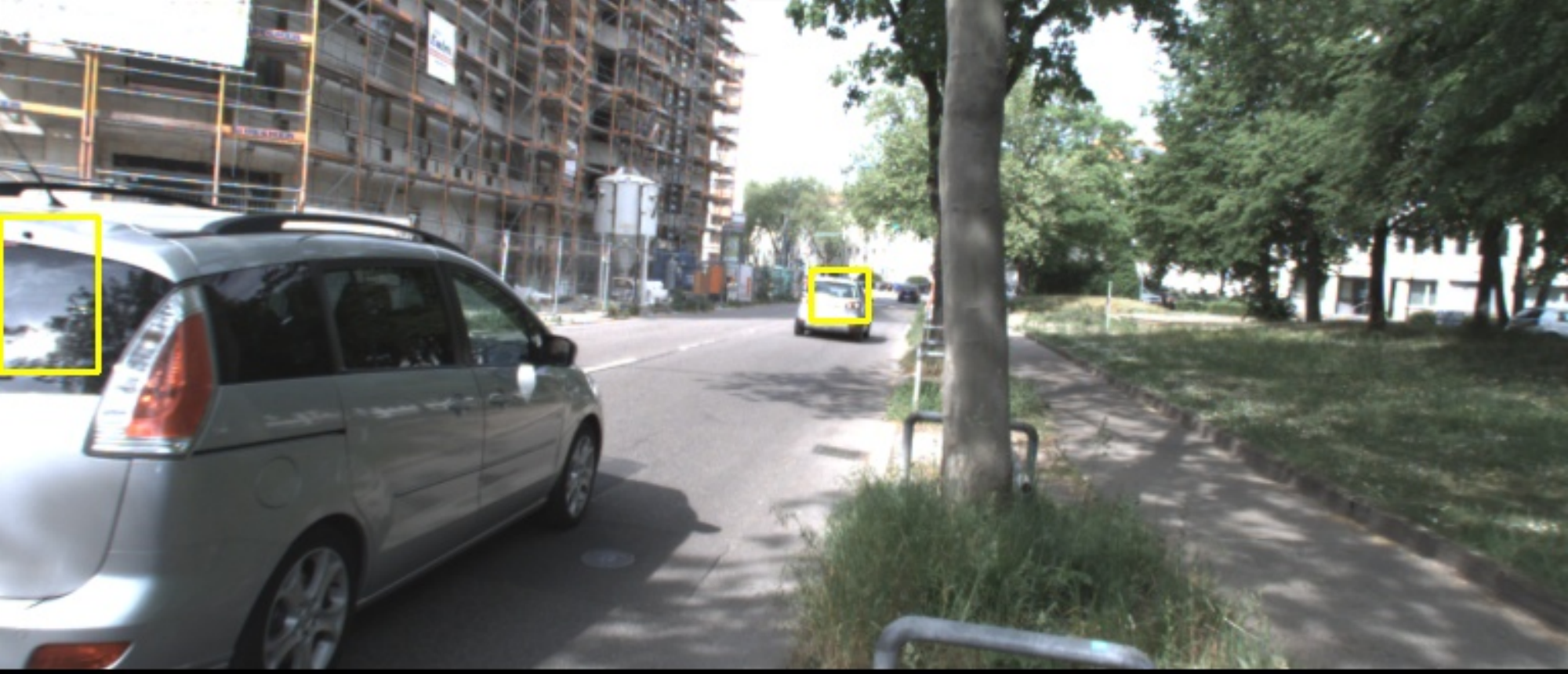} \\
     \includegraphics[width=\linewidth]{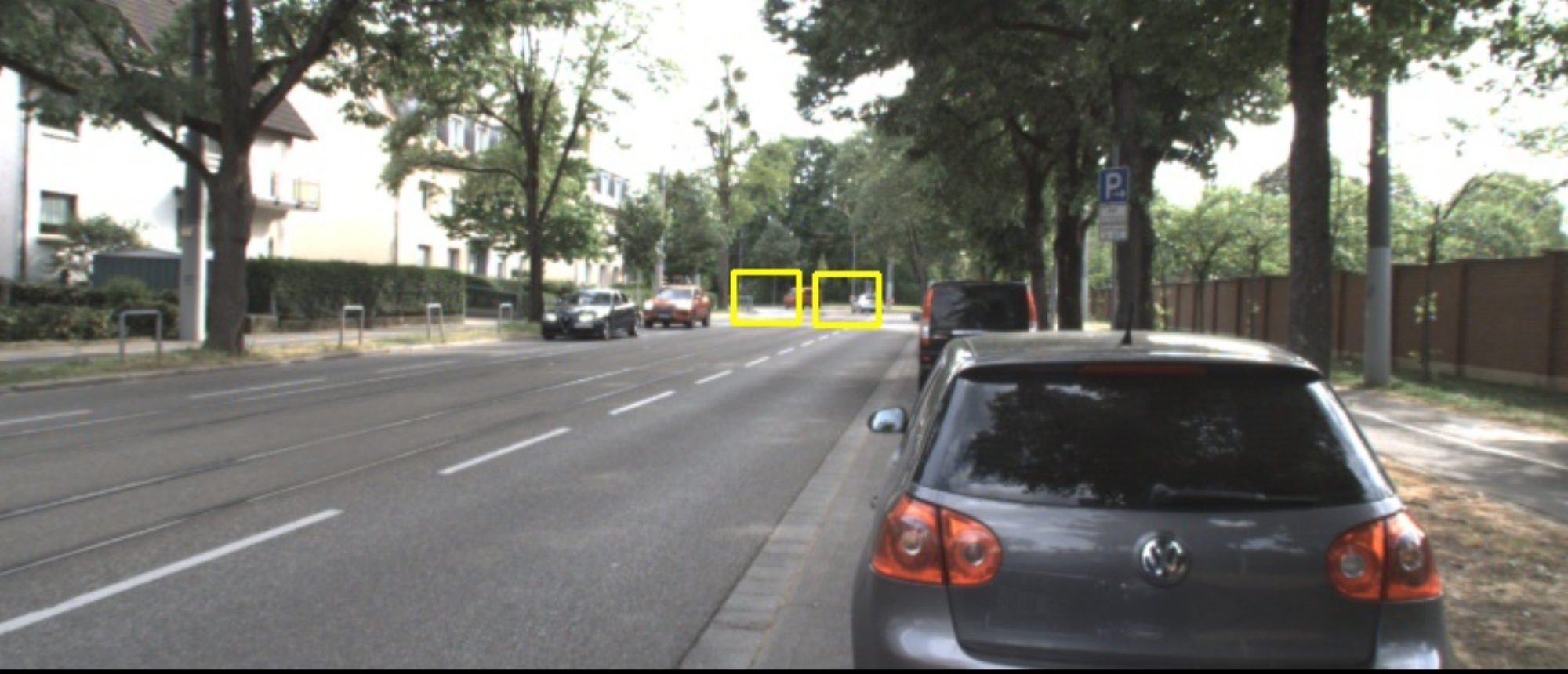} &  \includegraphics[width=\linewidth]{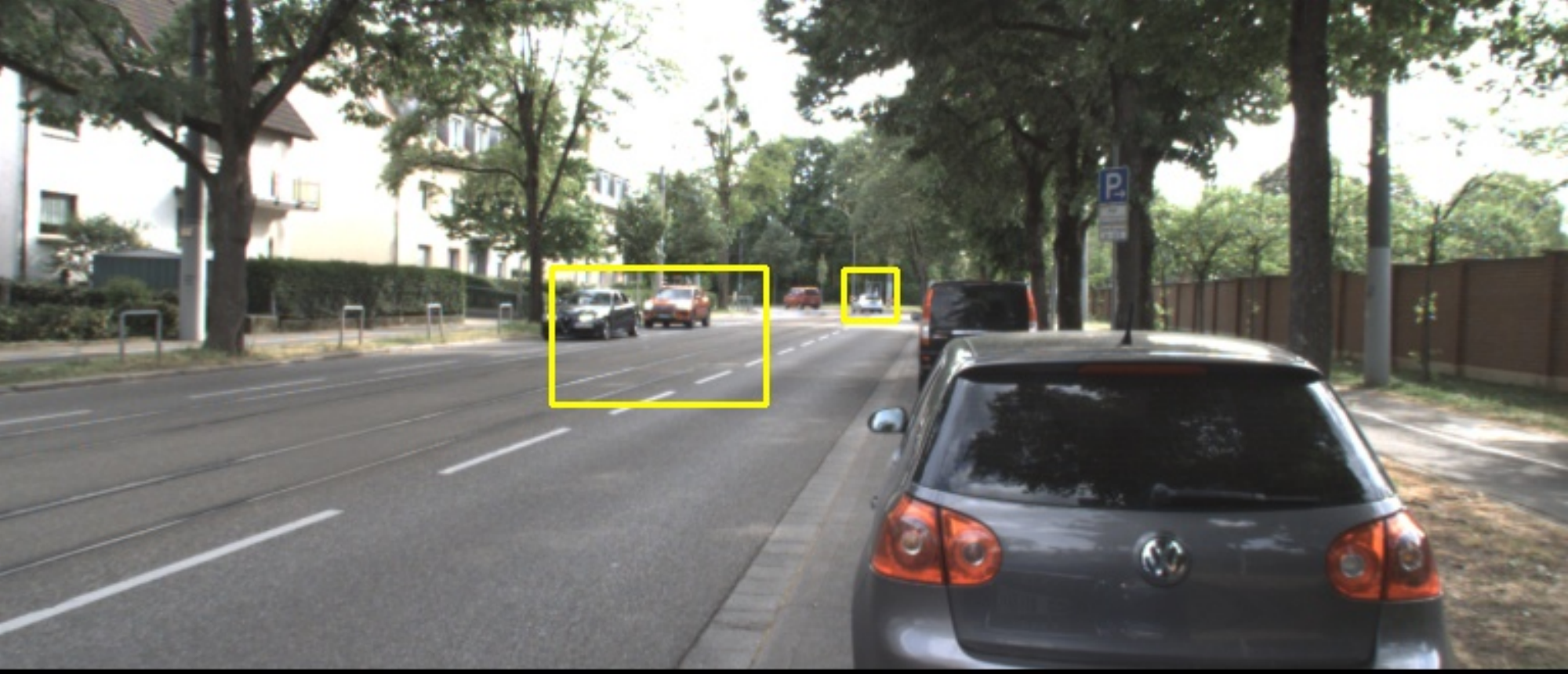} \\
     \includegraphics[width=\linewidth]{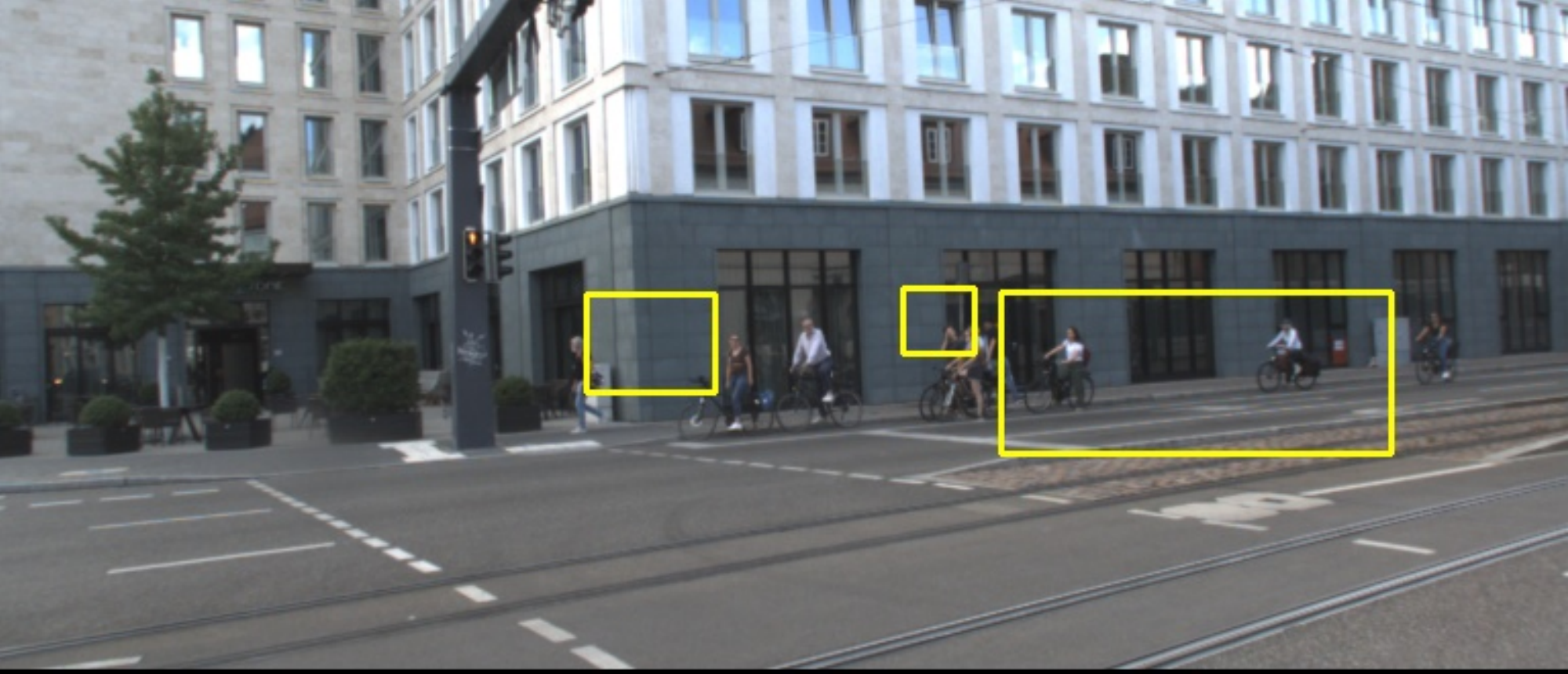} &  \includegraphics[width=\linewidth]{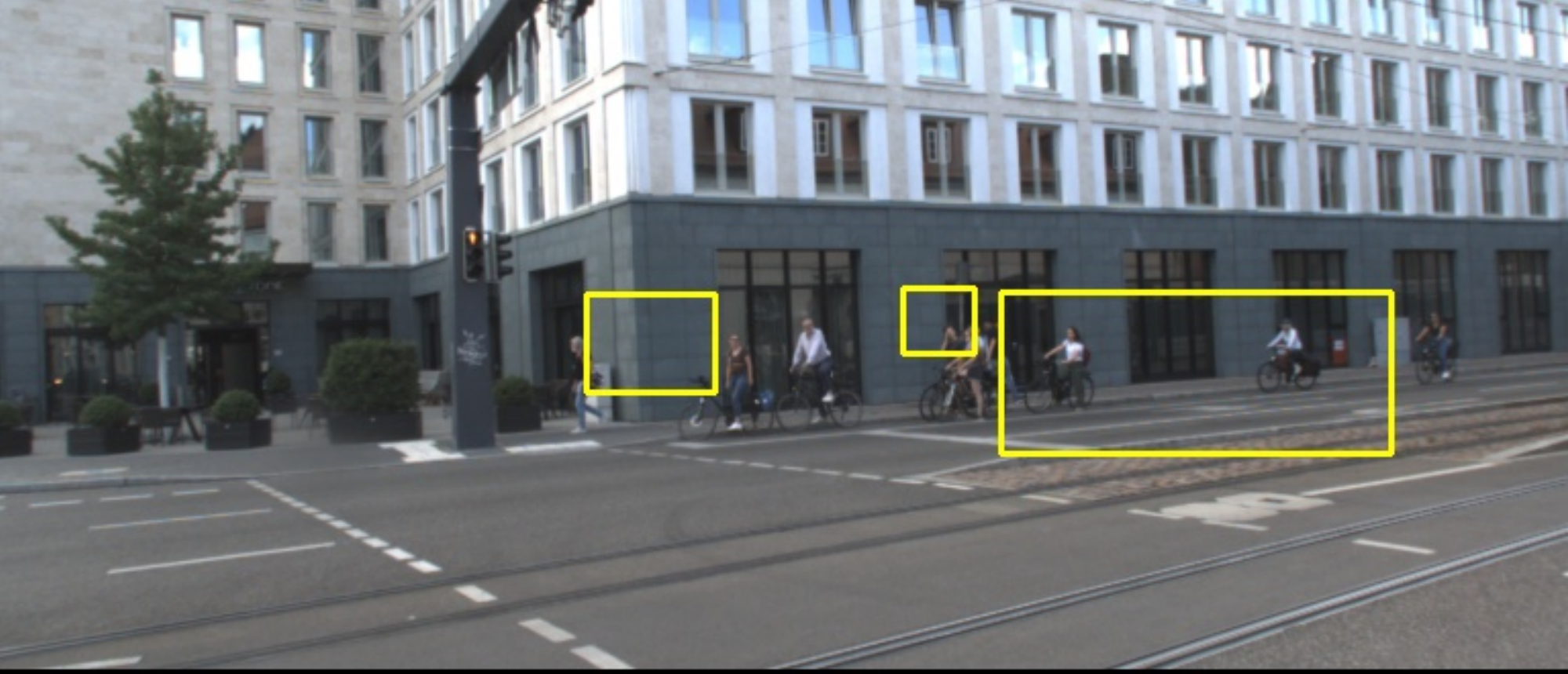} \\
     \includegraphics[width=\linewidth]{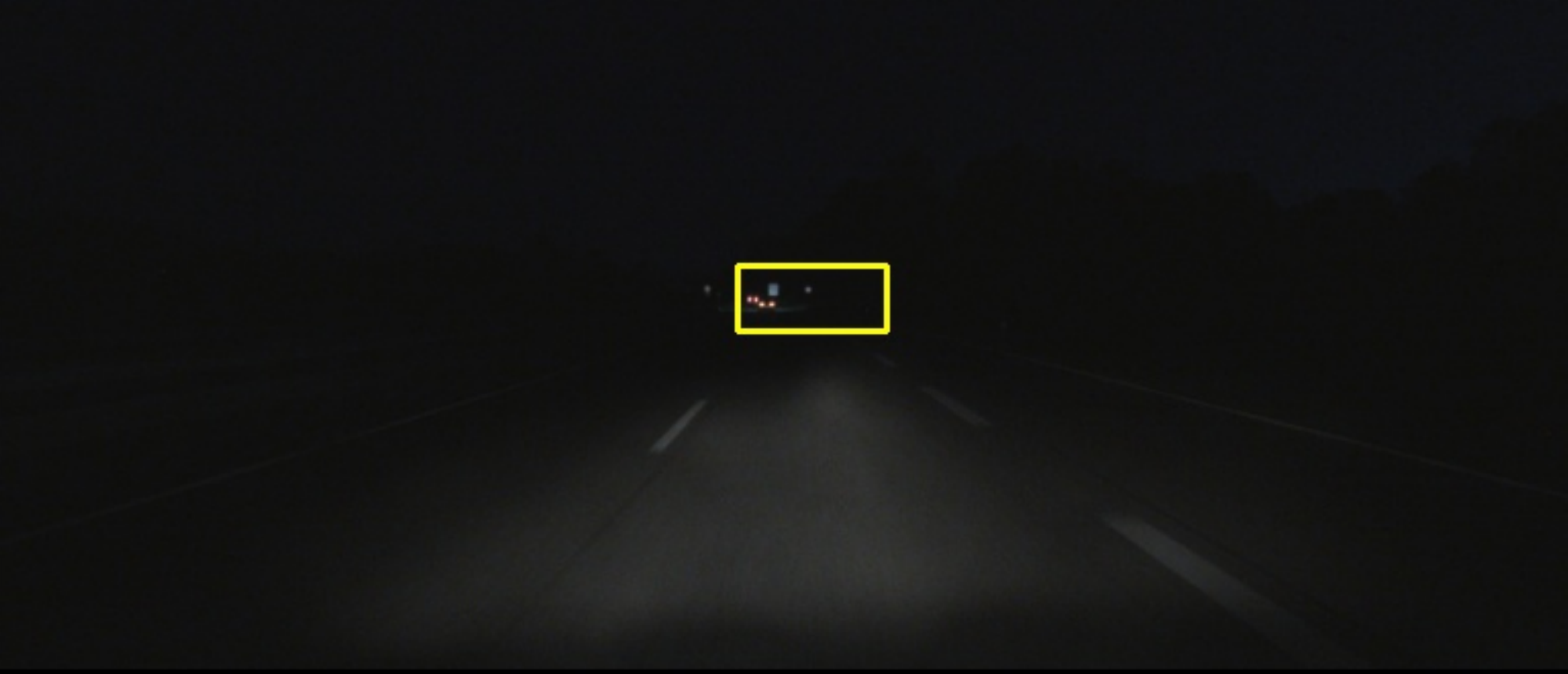} &  \includegraphics[width=\linewidth]{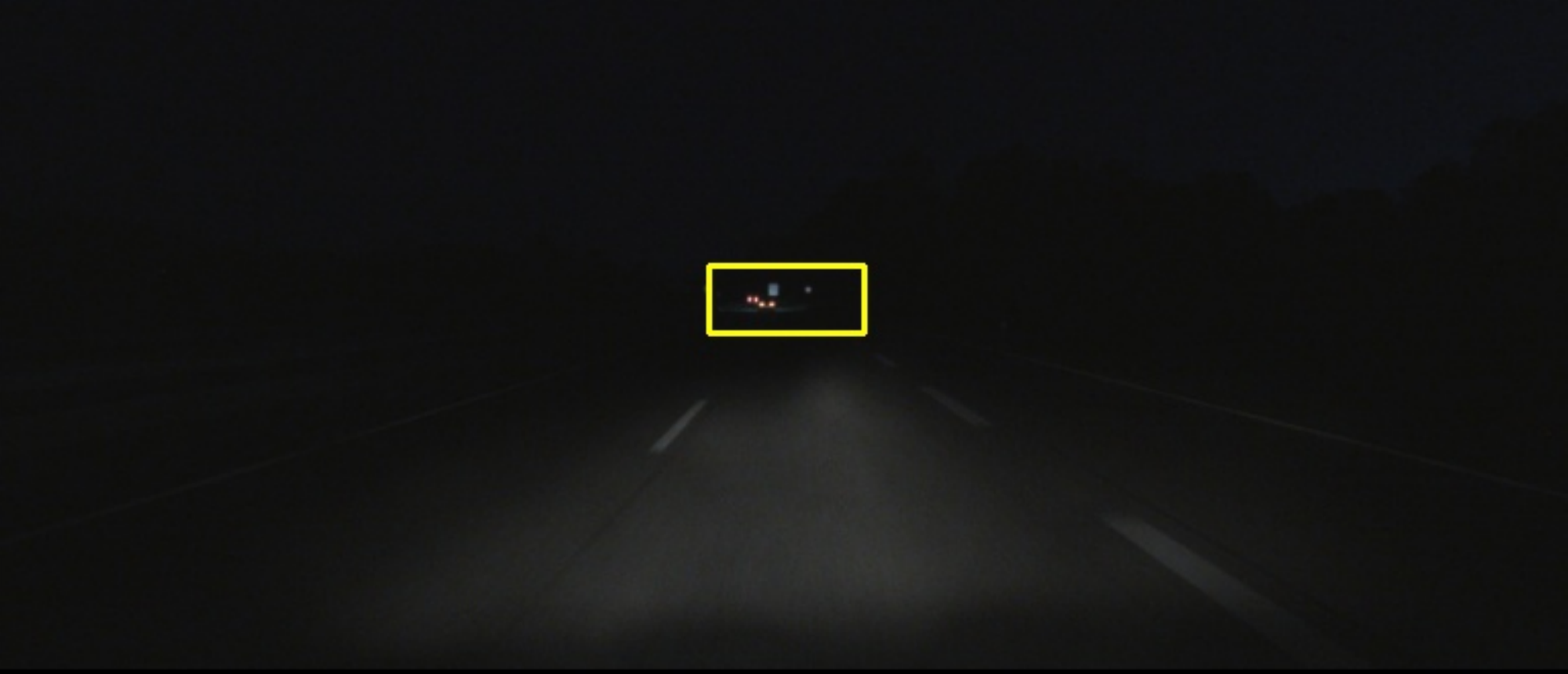} \\
     \includegraphics[width=\linewidth]{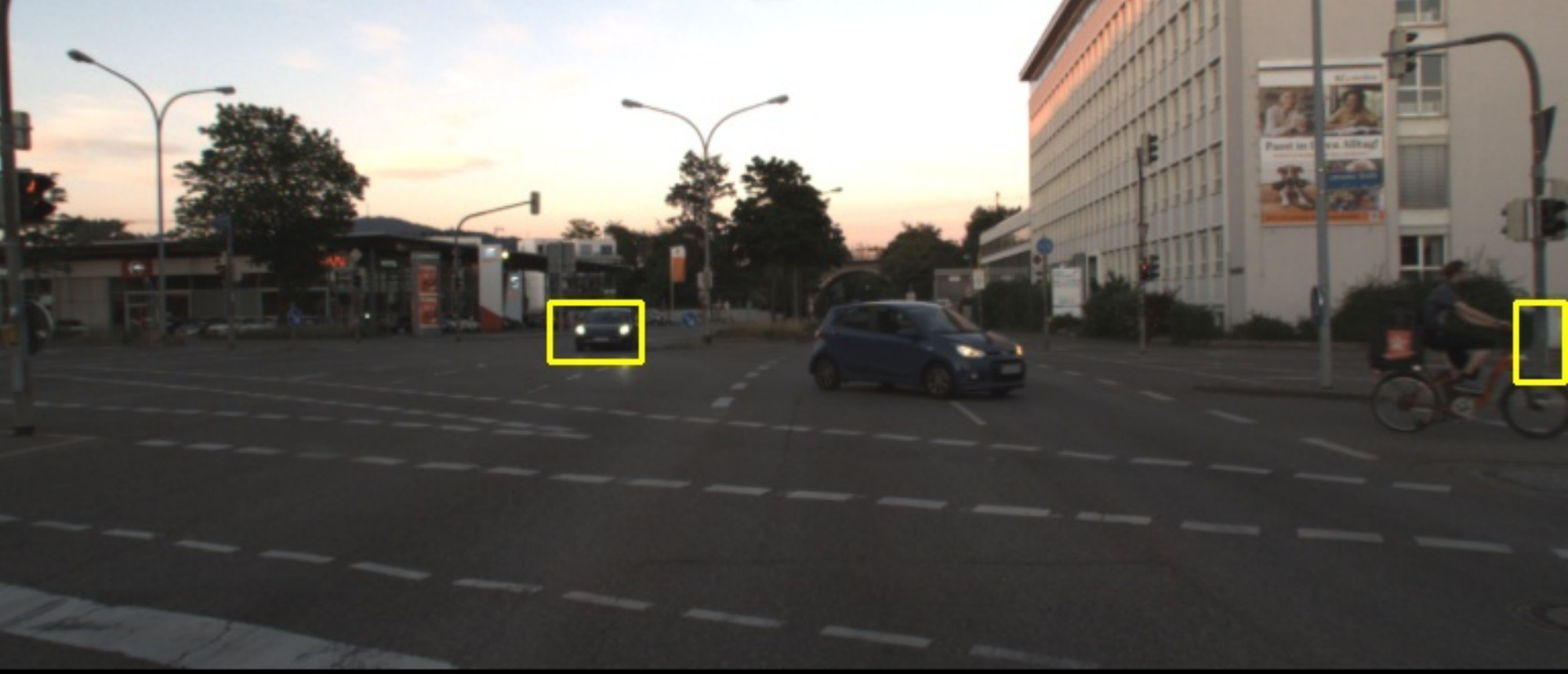} &  \includegraphics[width=\linewidth]{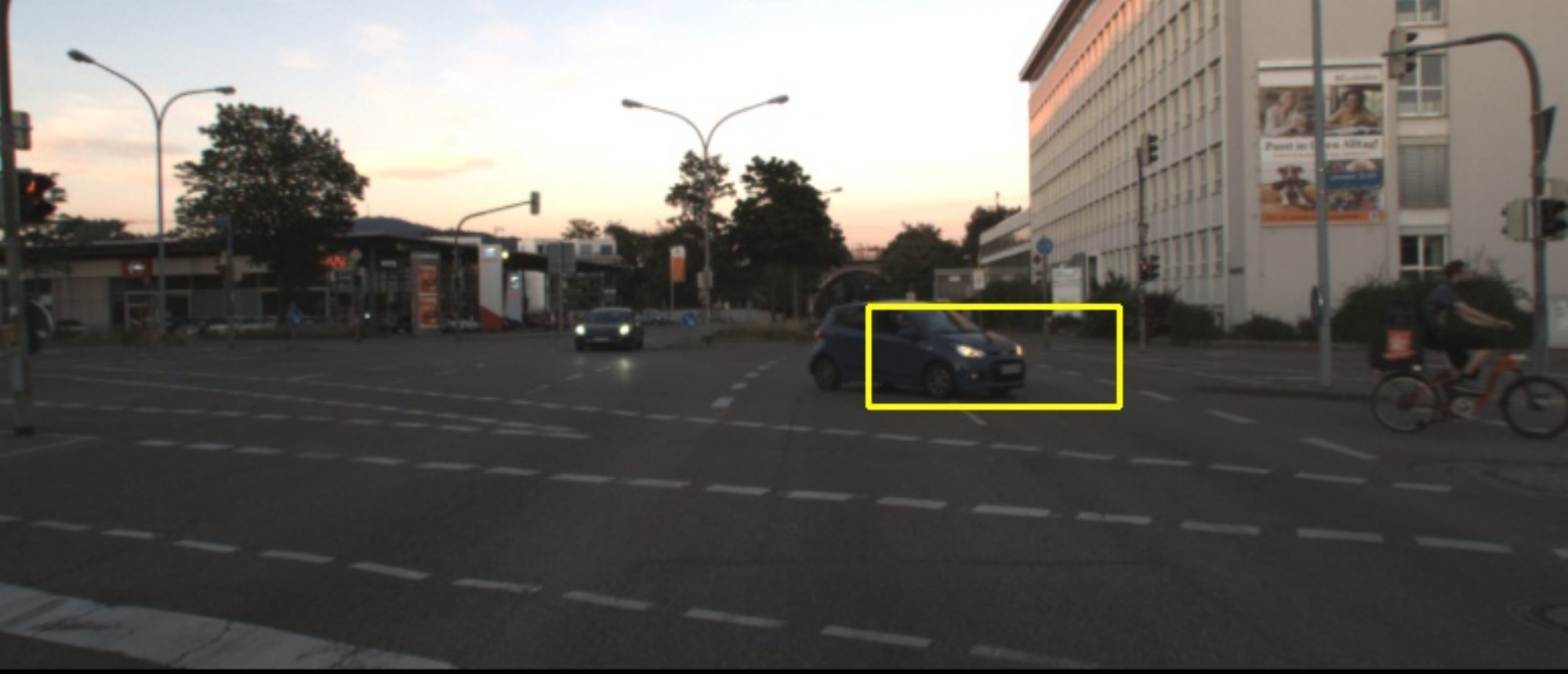} \\
\end{tabular}}
\caption{Qualitative comparisons of predictions from our MM-DistillNet and the previous state-of-the-art StereoSoundNet~\cite{wang2020score}. We present failure cases in these examples that include far away cars whose sounds are not sufficiently captured by the microphones.}
\label{fig:SM_qualitativefailure}
\end{figure*}

\begin{figure*}
\footnotesize
\centering
\setlength{\tabcolsep}{0.05cm}
{\renewcommand{\arraystretch}{0.8}
\begin{tabular}{p{0.195\textwidth}p{0.195\textwidth}p{0.195\textwidth}p{0.195\textwidth}p{0.195\textwidth}}
     \multicolumn{1}{c}{\textbf{RGB Teacher}} & \multicolumn{1}{c}{\textbf{Depth Teacher}} & \multicolumn{1}{c}{\textbf{Thermal Teacher}} & \multicolumn{1}{c}{\textbf{StereoSoundNet~\cite{wang2020score}}} & \multicolumn{1}{c}{\textbf{MM-DistillNet (Ours)}} \\
     \includegraphics[width=\linewidth]{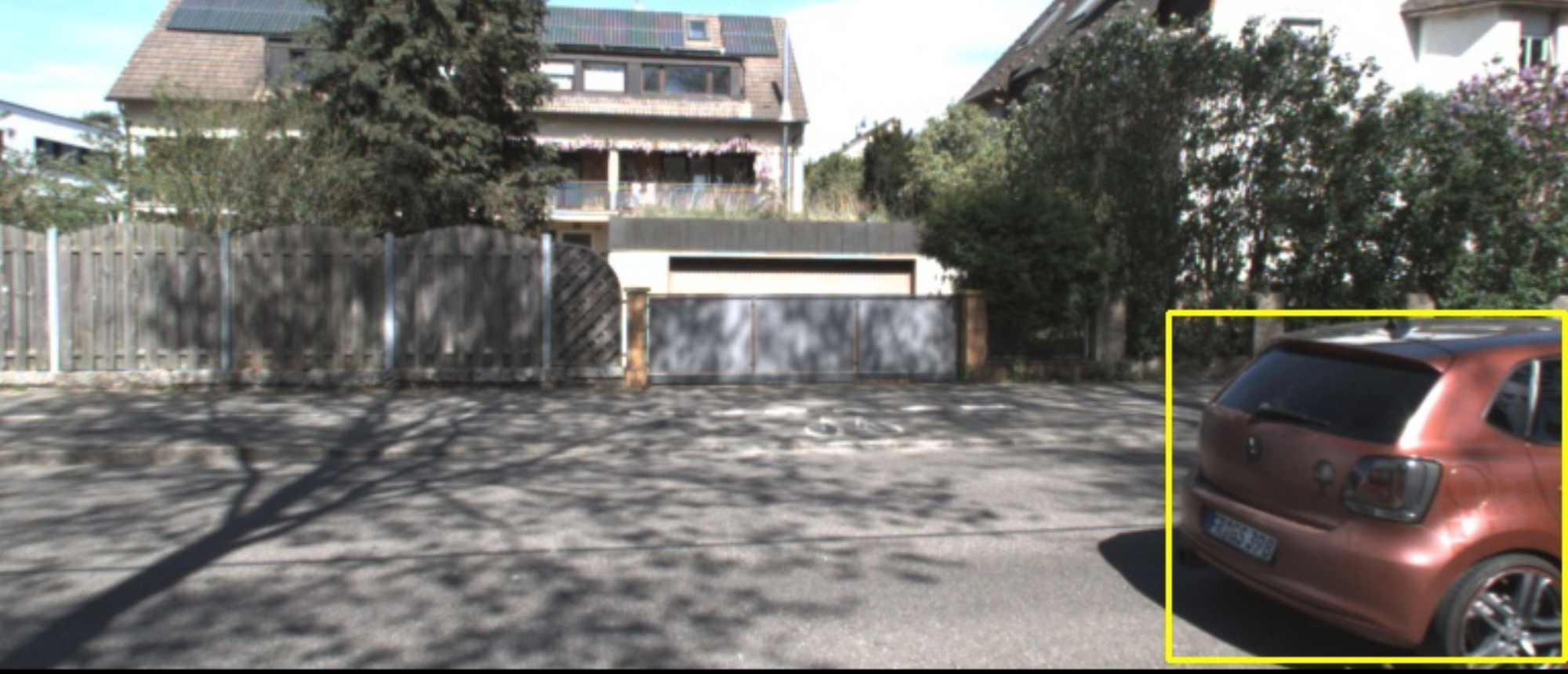} &  \includegraphics[width=\linewidth]{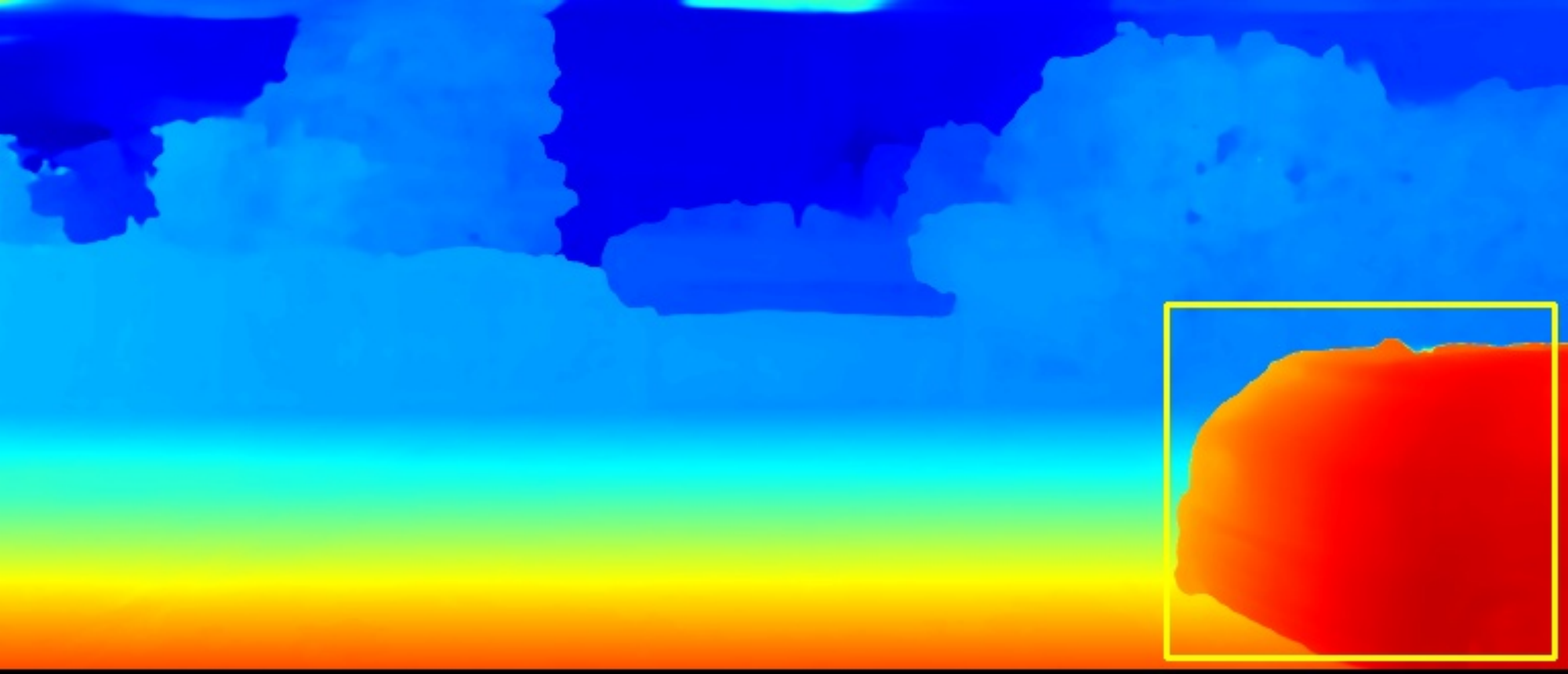} & \includegraphics[width=\linewidth]{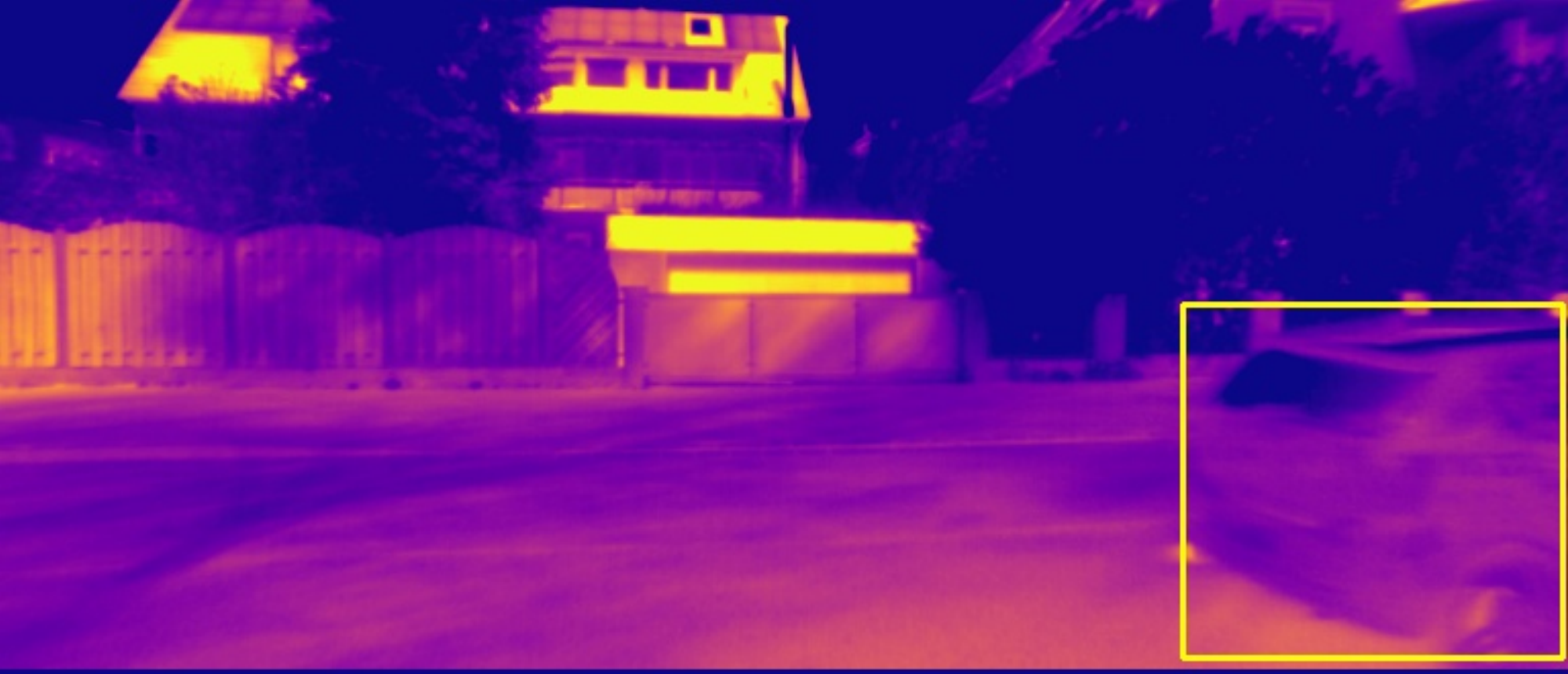} & \includegraphics[width=\linewidth]{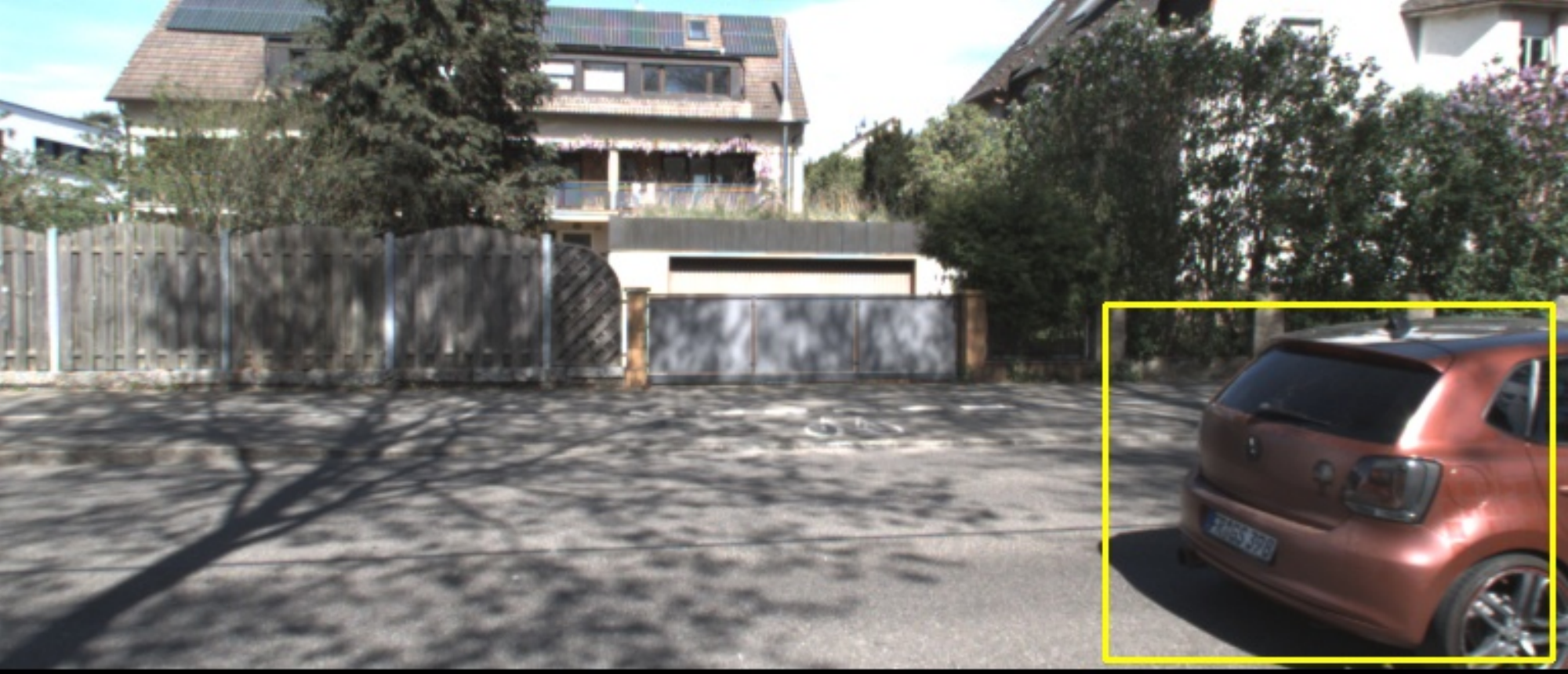} & \includegraphics[width=\linewidth]{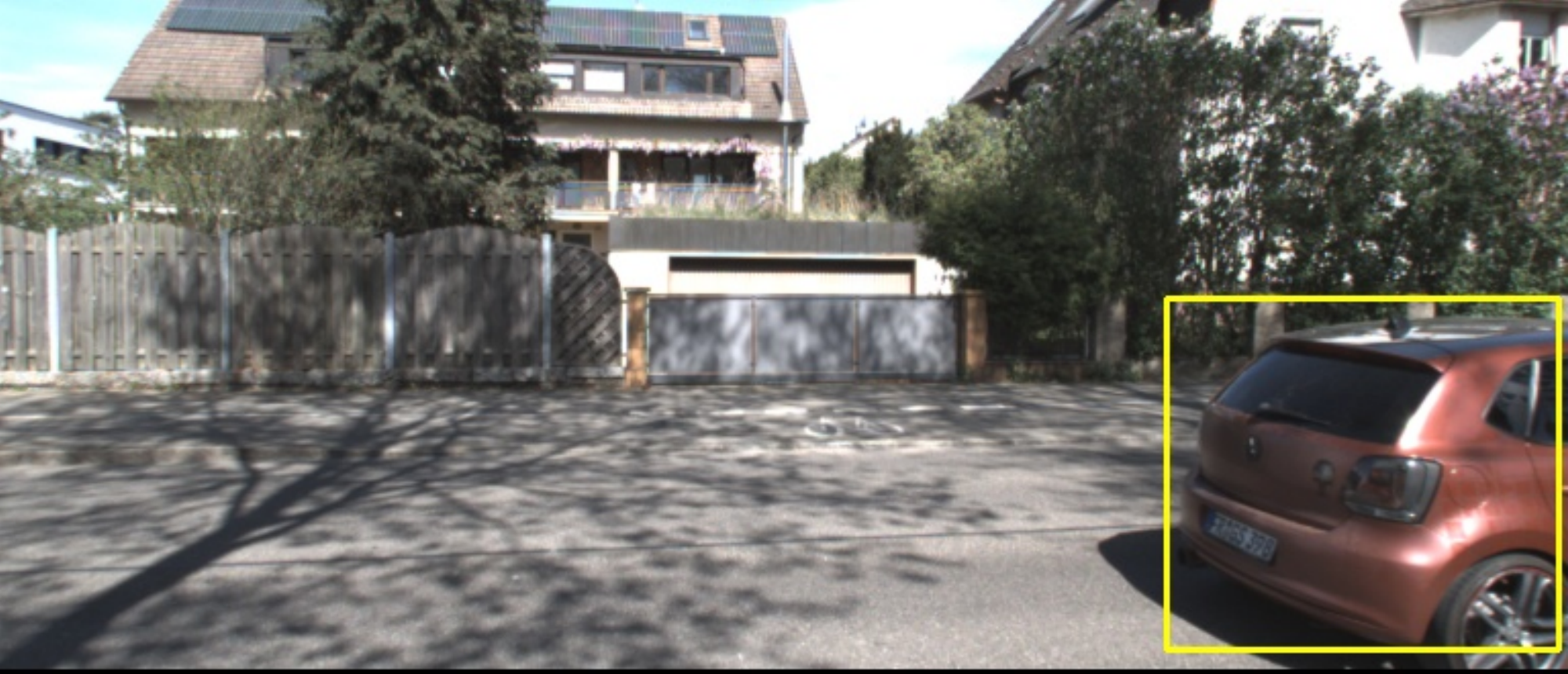} \\
     \includegraphics[width=\linewidth]{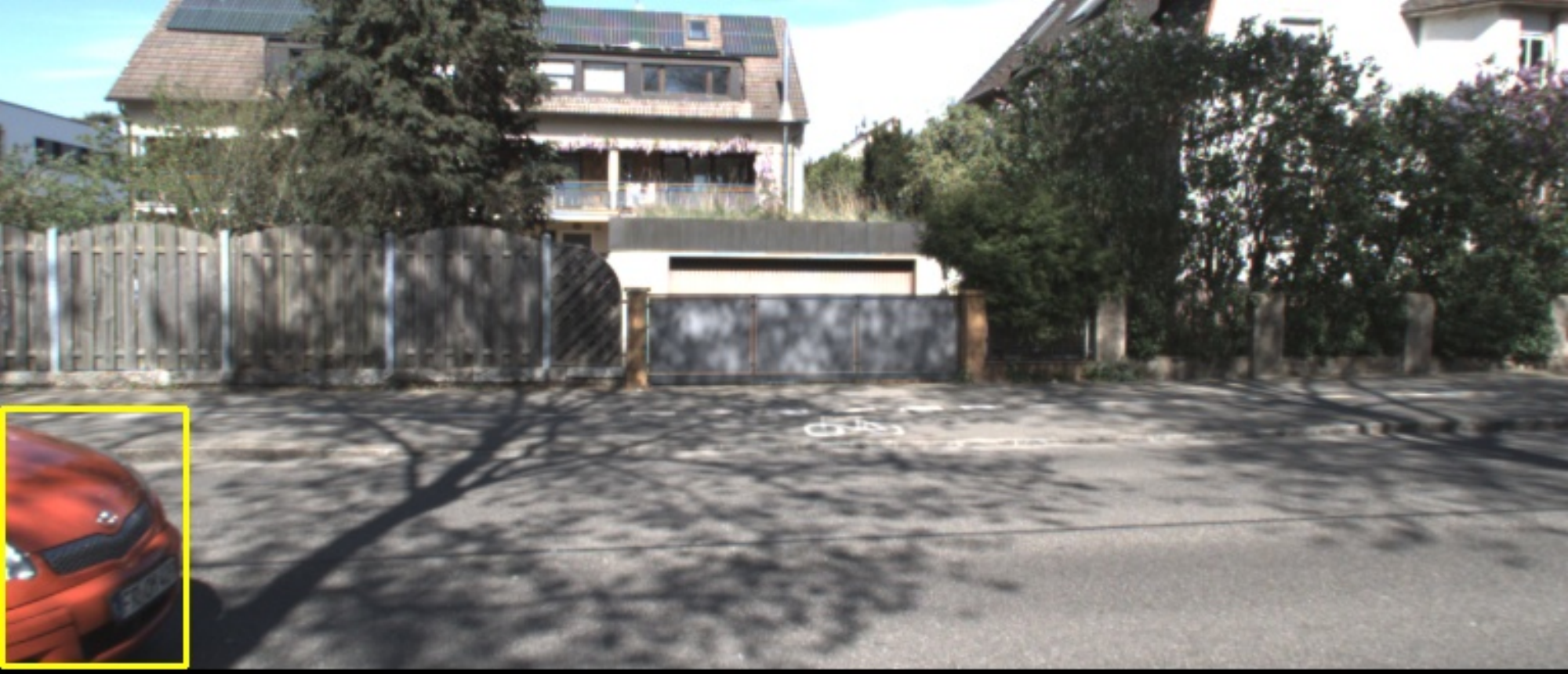} &  \includegraphics[width=\linewidth]{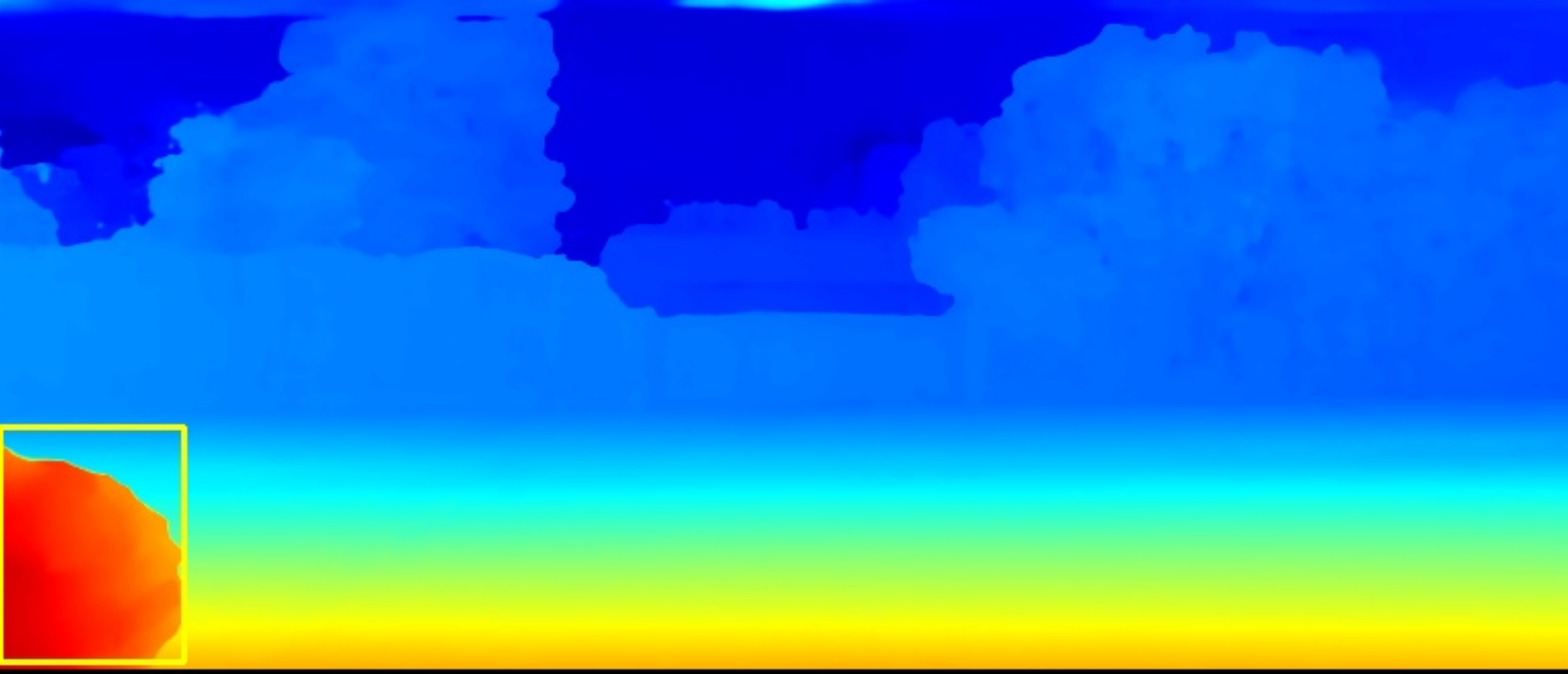} & \includegraphics[width=\linewidth]{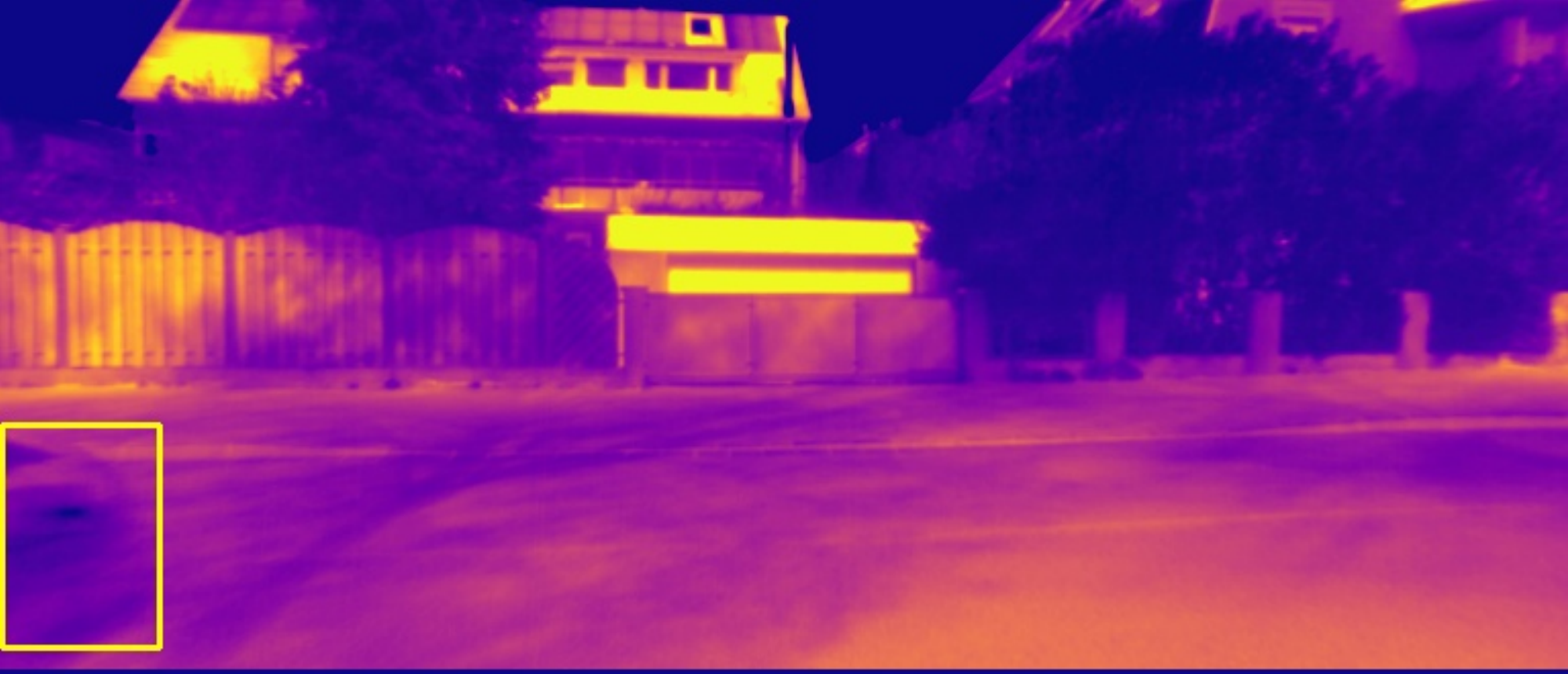} & \includegraphics[width=\linewidth]{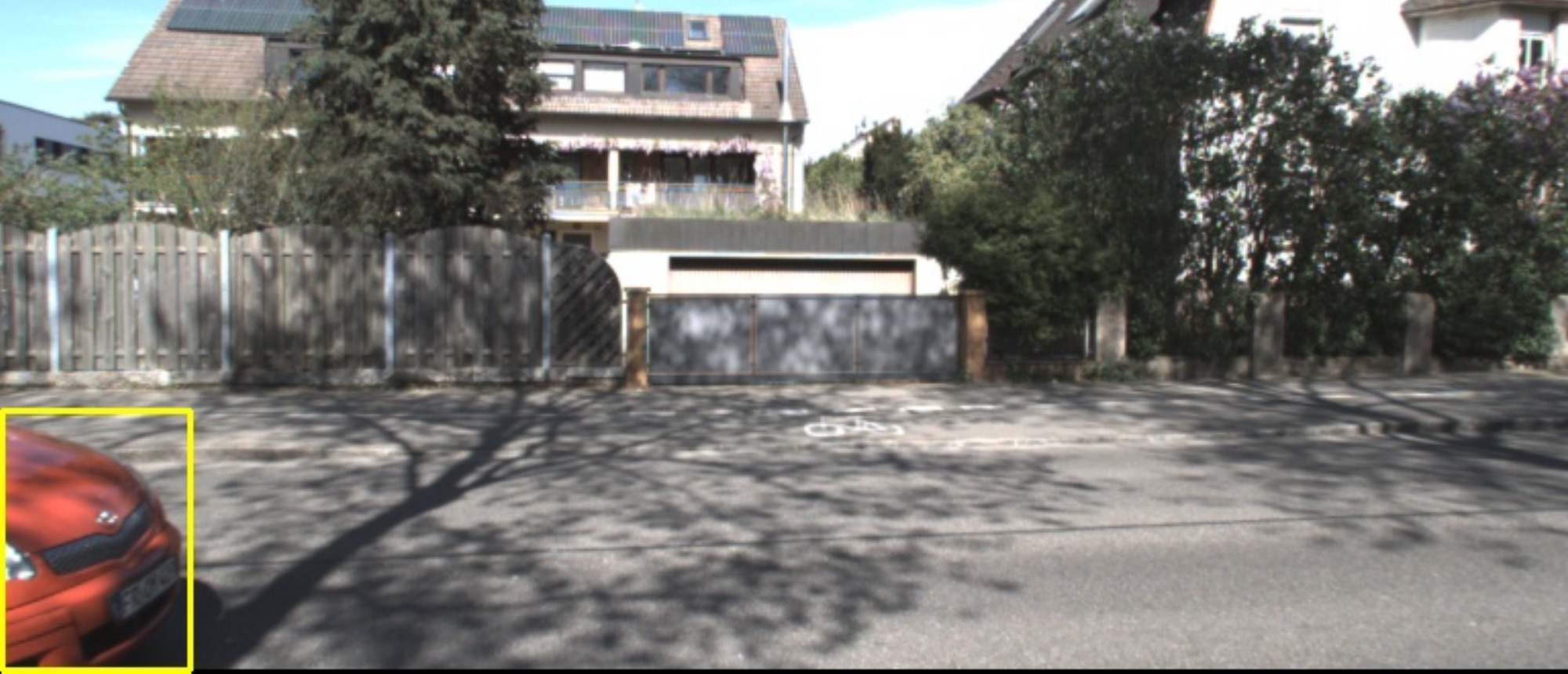} & \includegraphics[width=\linewidth]{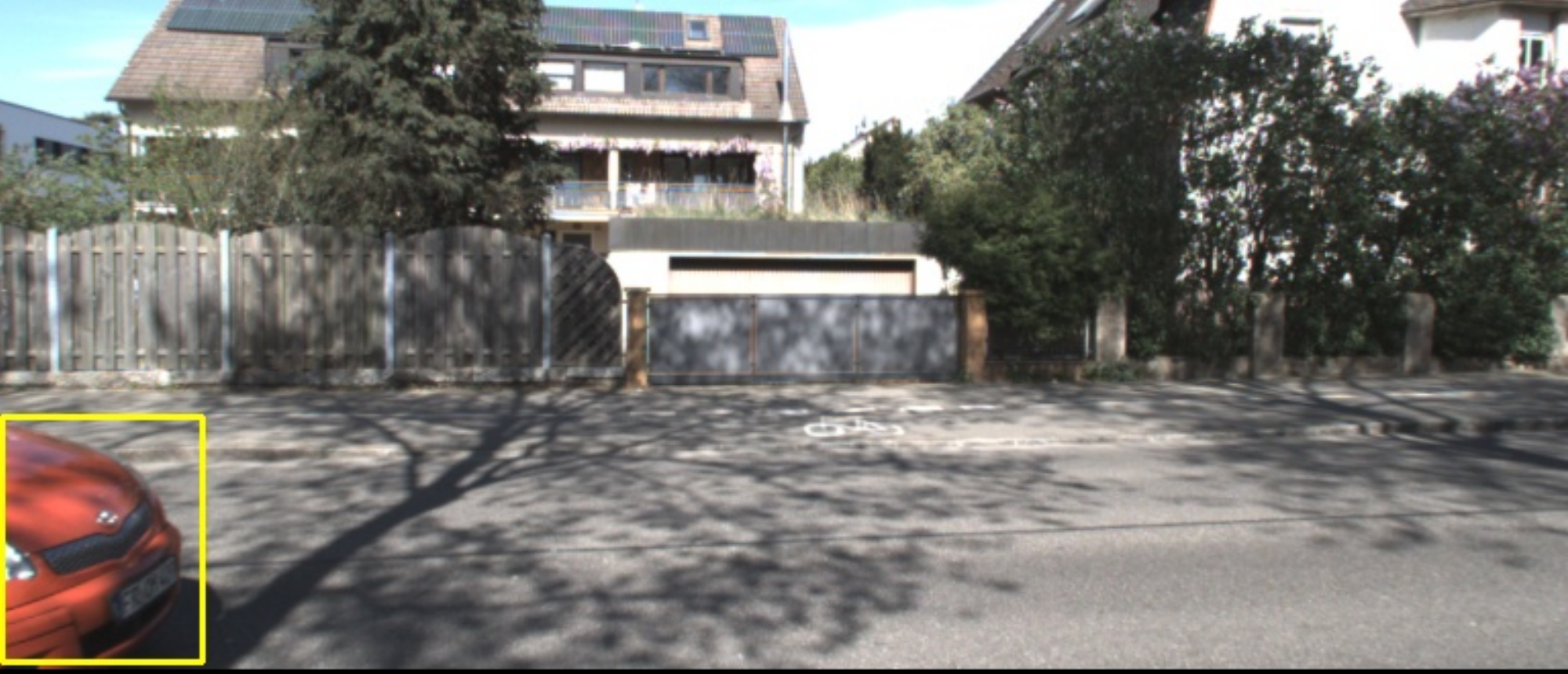} \\
     \includegraphics[width=\linewidth]{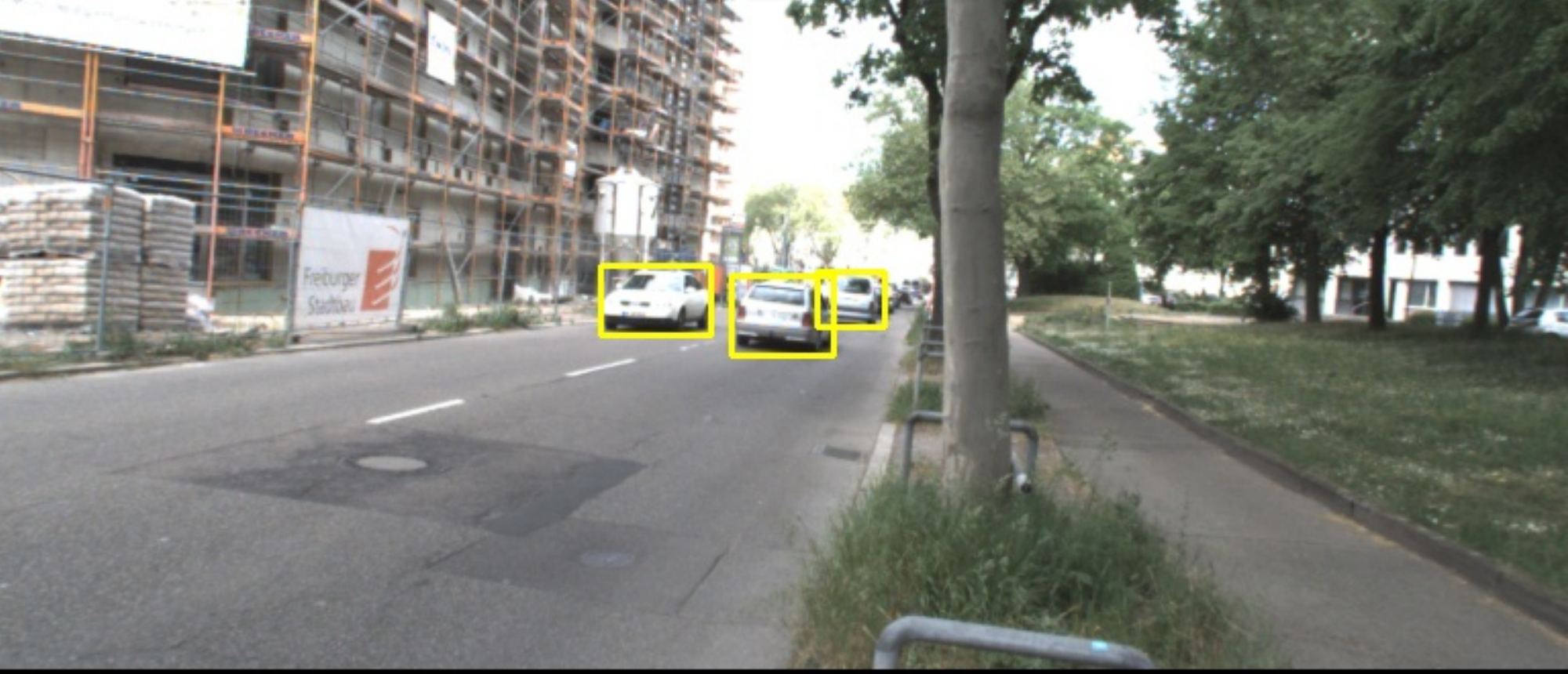} &  \includegraphics[width=\linewidth]{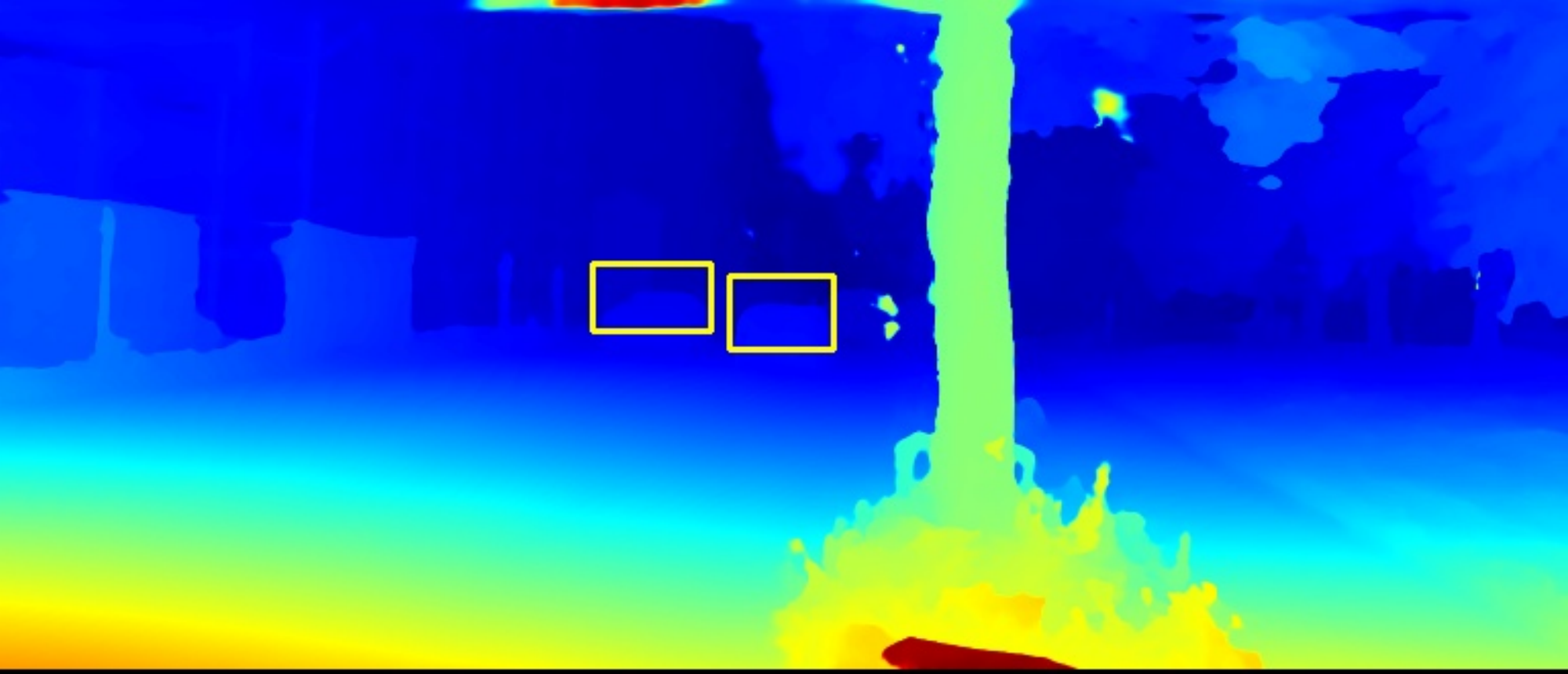} & \includegraphics[width=\linewidth]{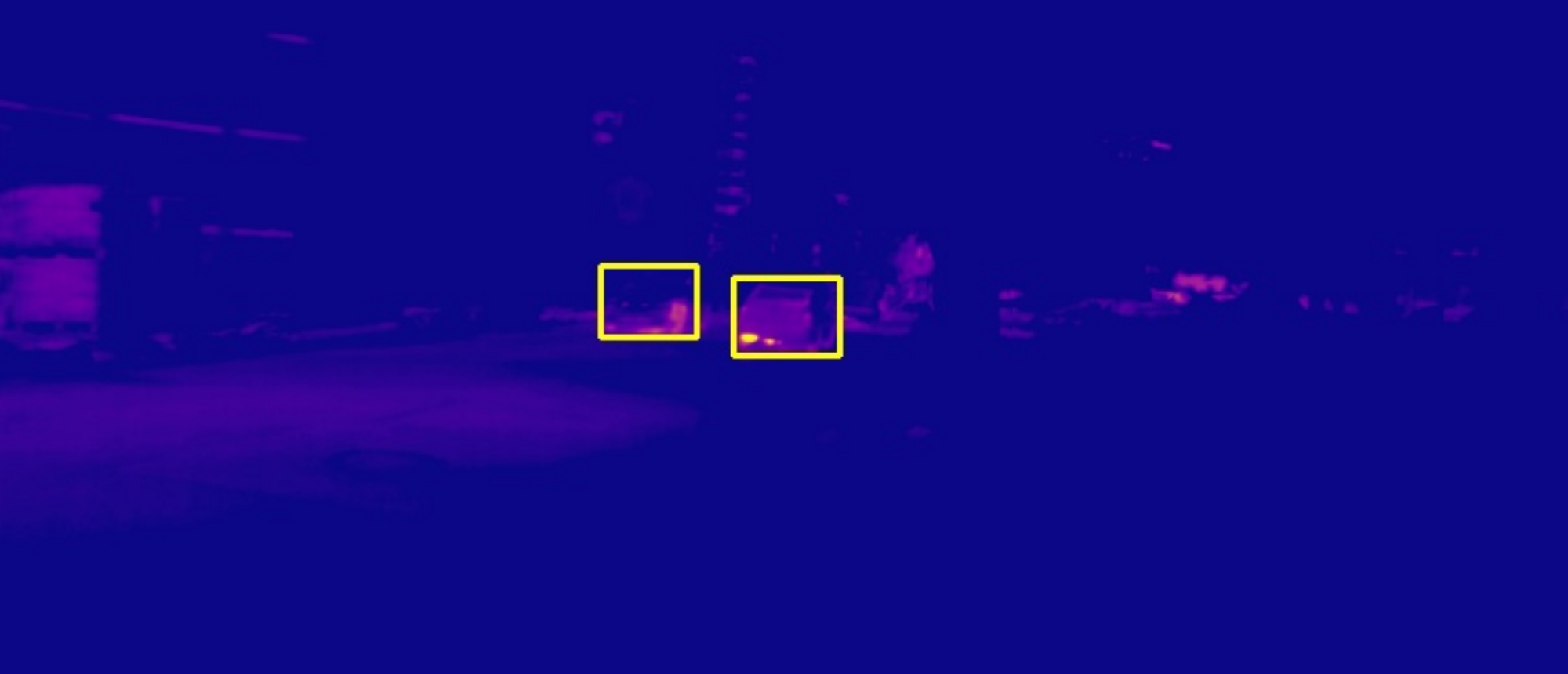} & \includegraphics[width=\linewidth]{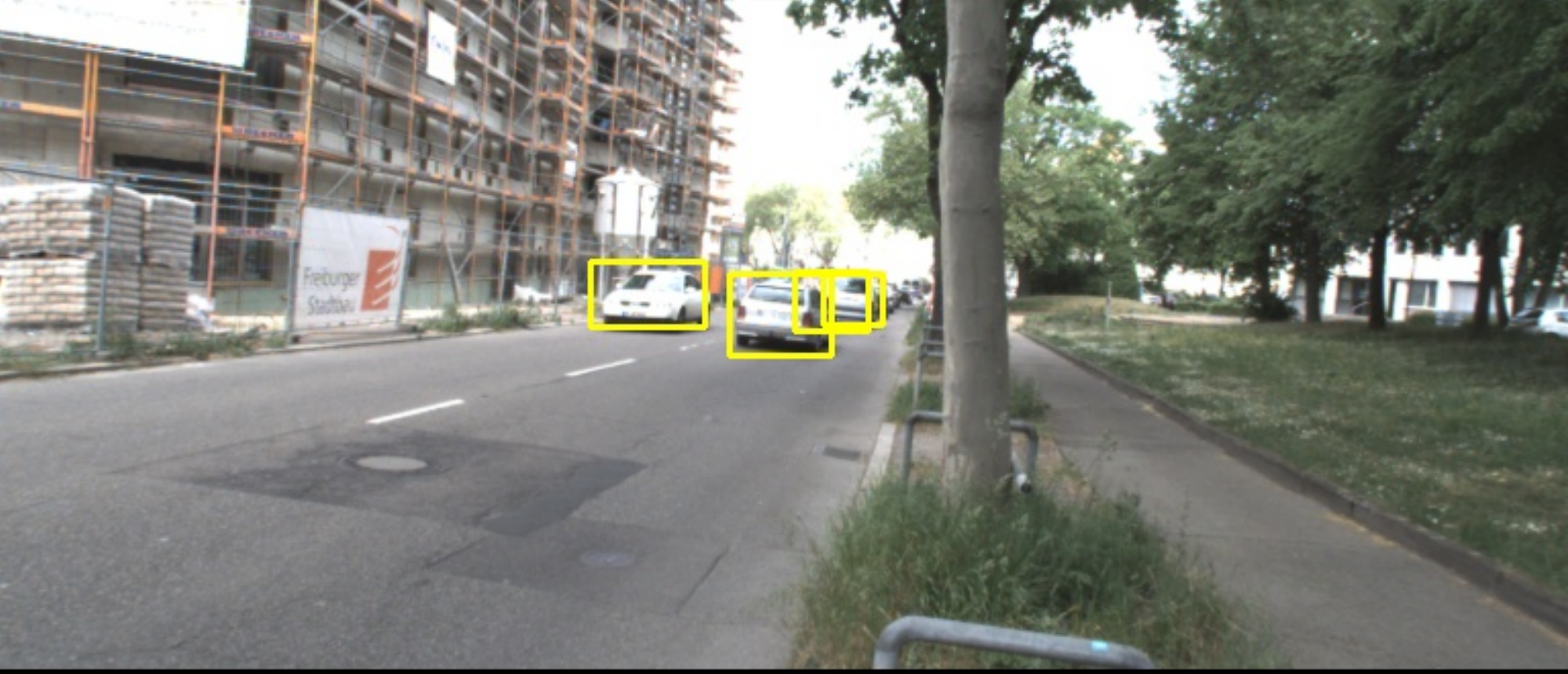} & \includegraphics[width=\linewidth]{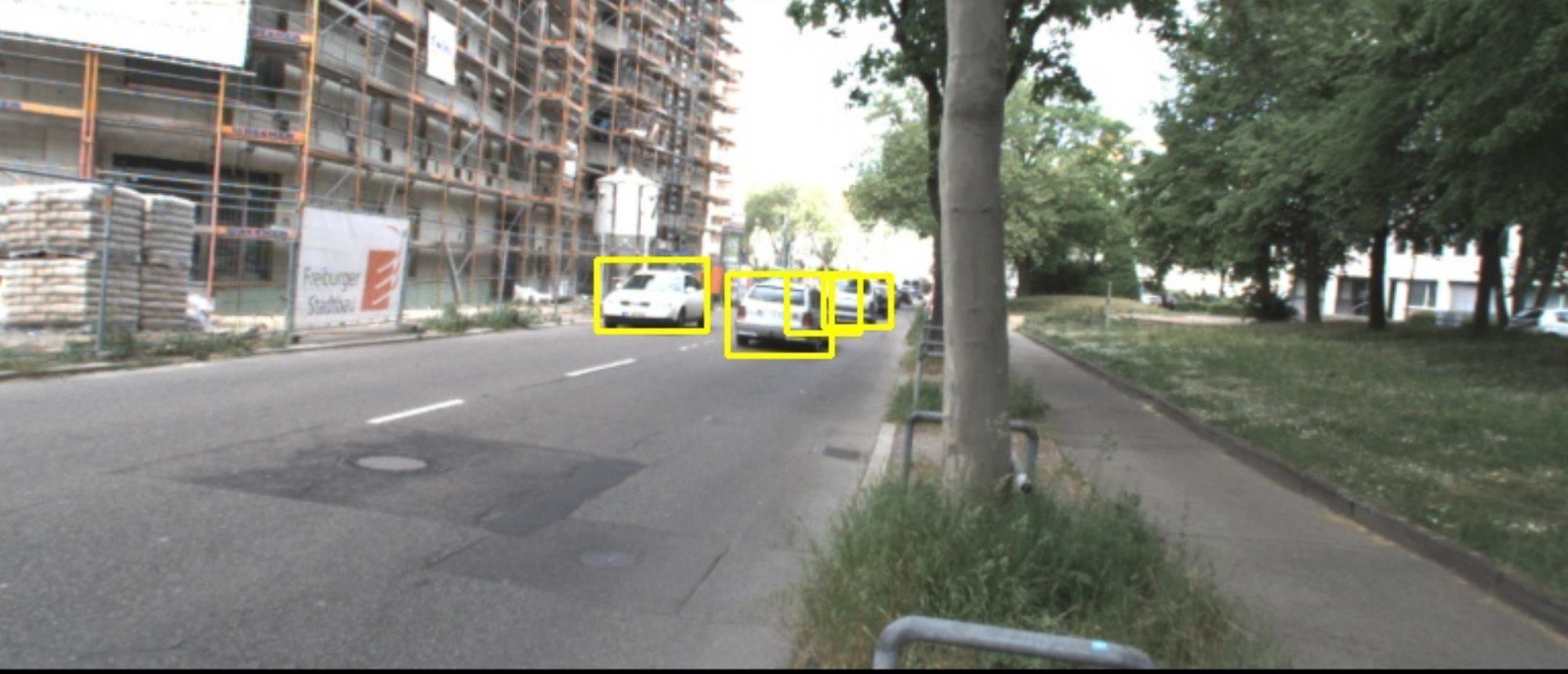} \\
    \includegraphics[width=\linewidth]{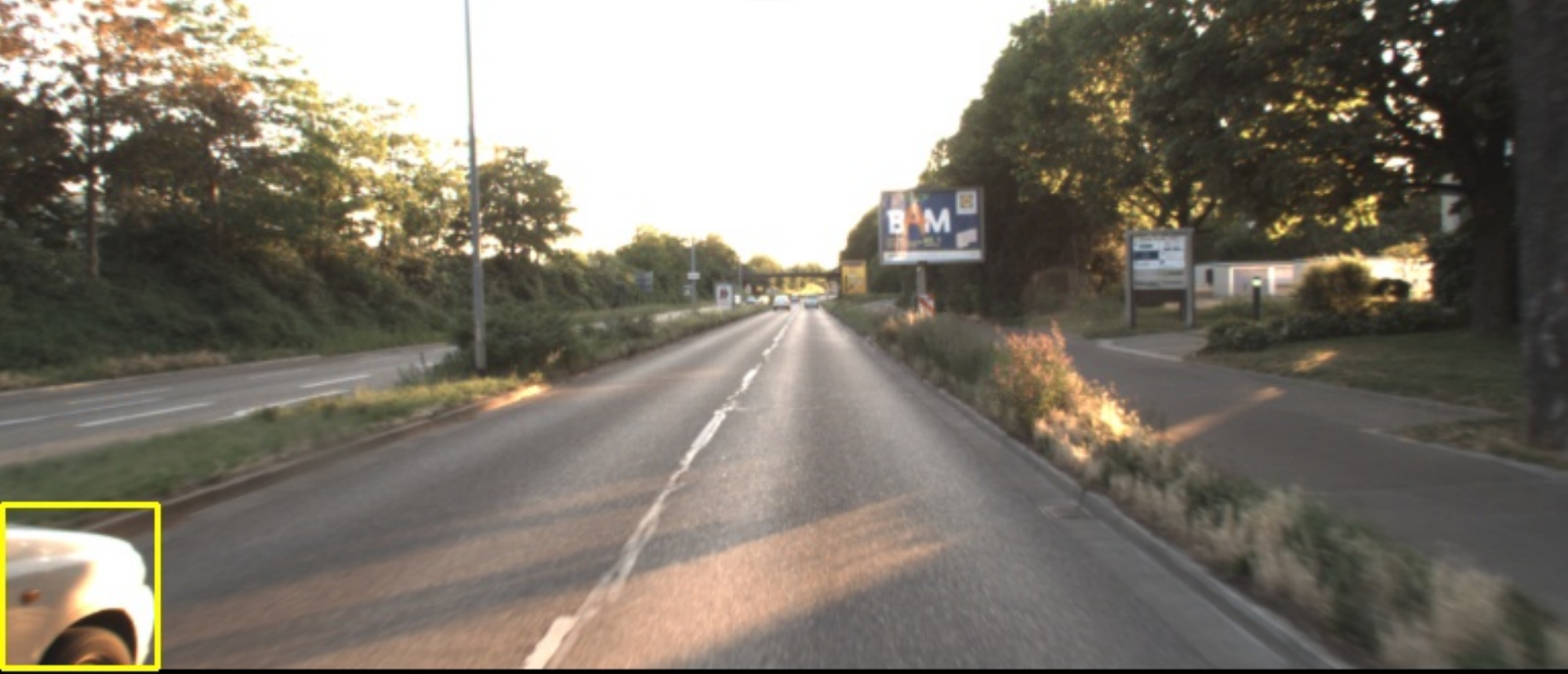} &  \includegraphics[width=\linewidth]{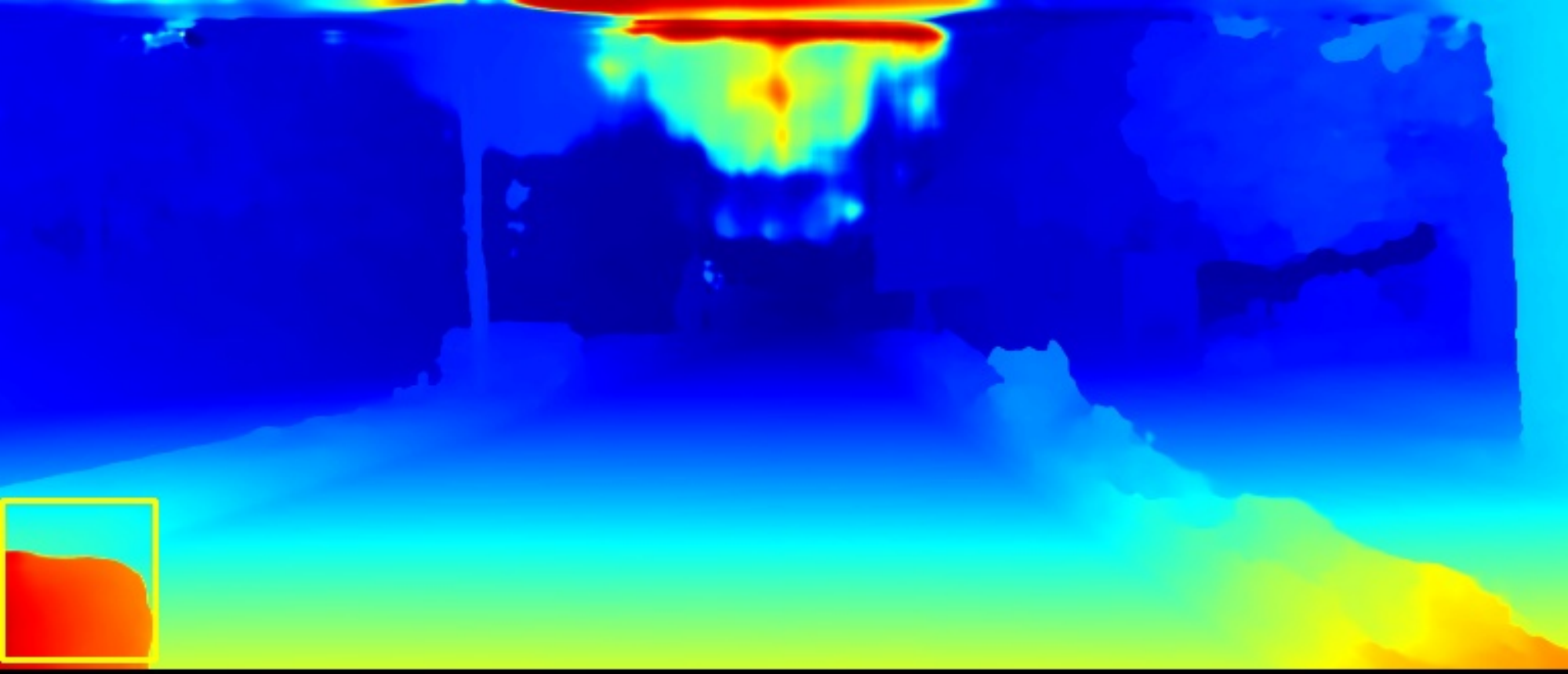} & \includegraphics[width=\linewidth]{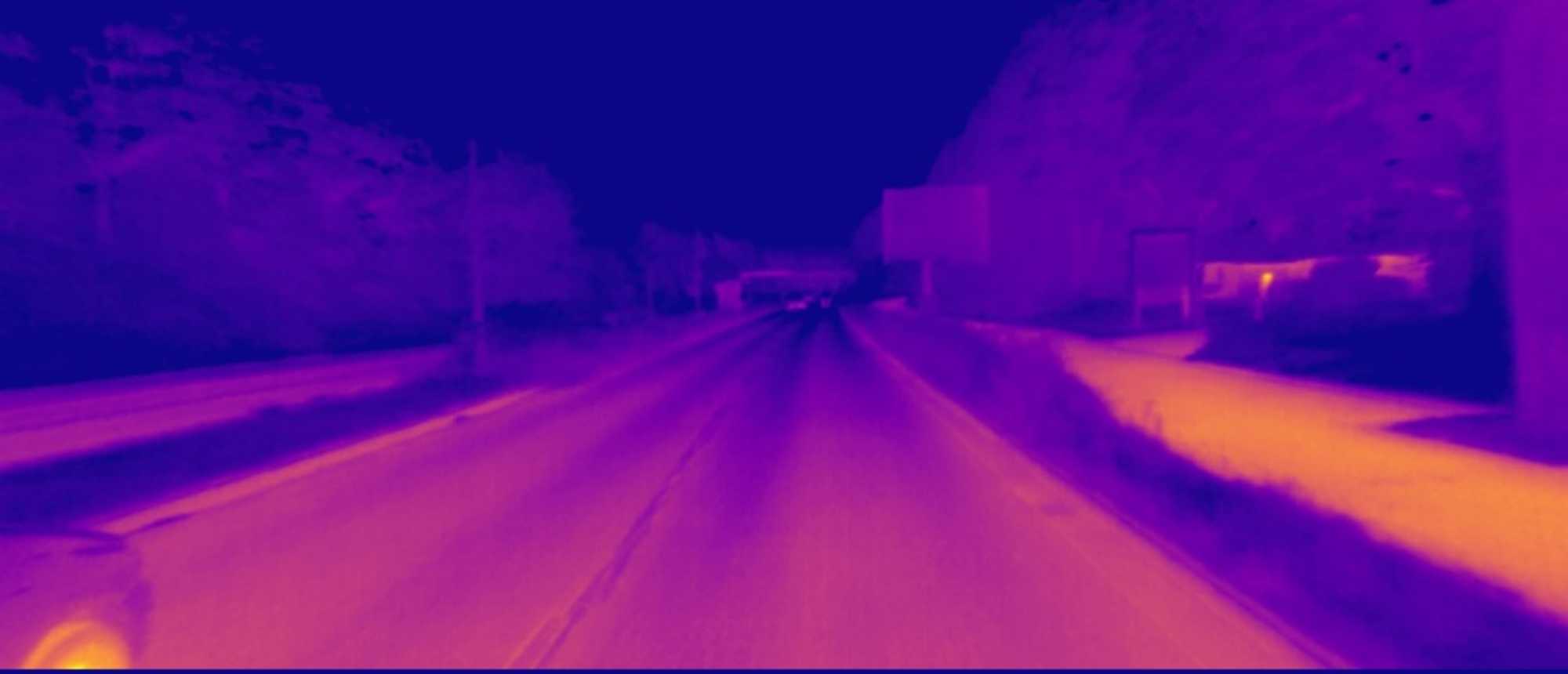} & \includegraphics[width=\linewidth]{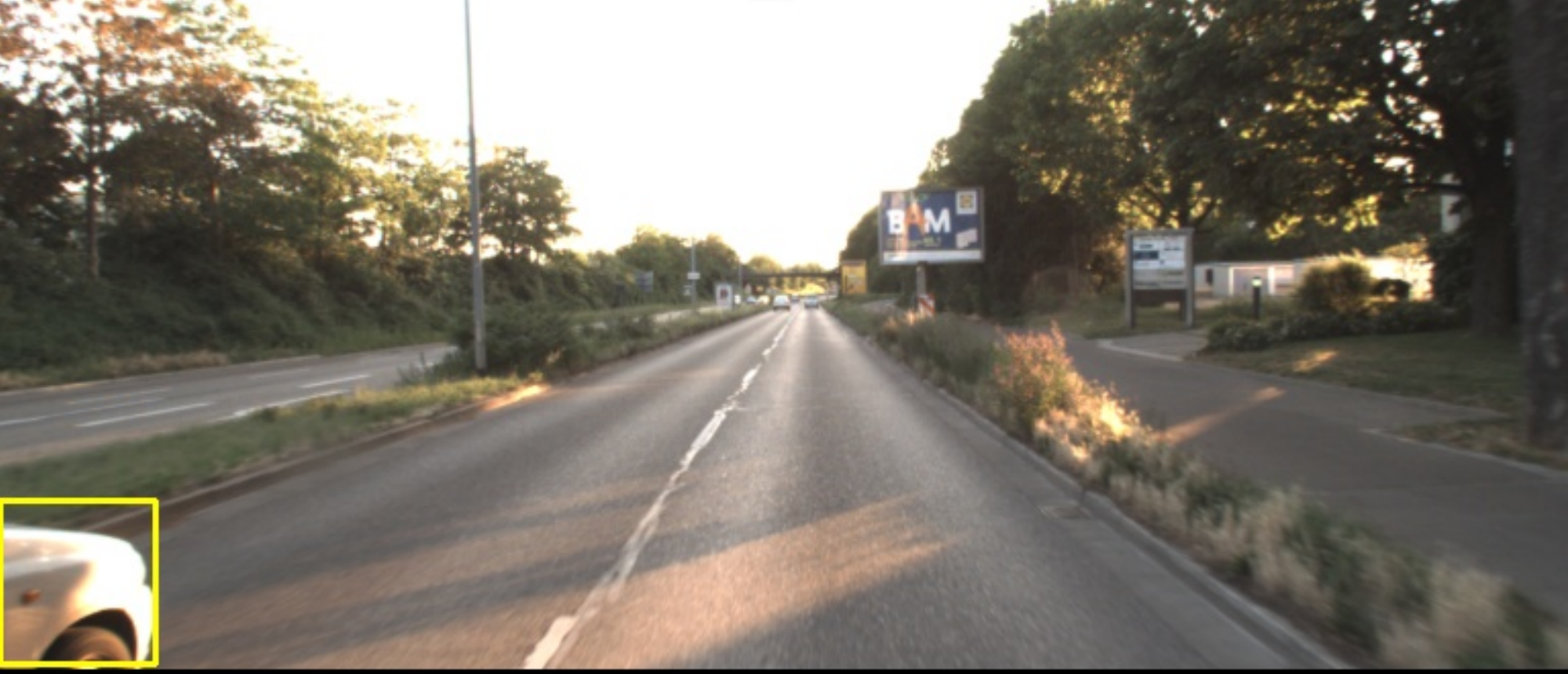} & \includegraphics[width=\linewidth]{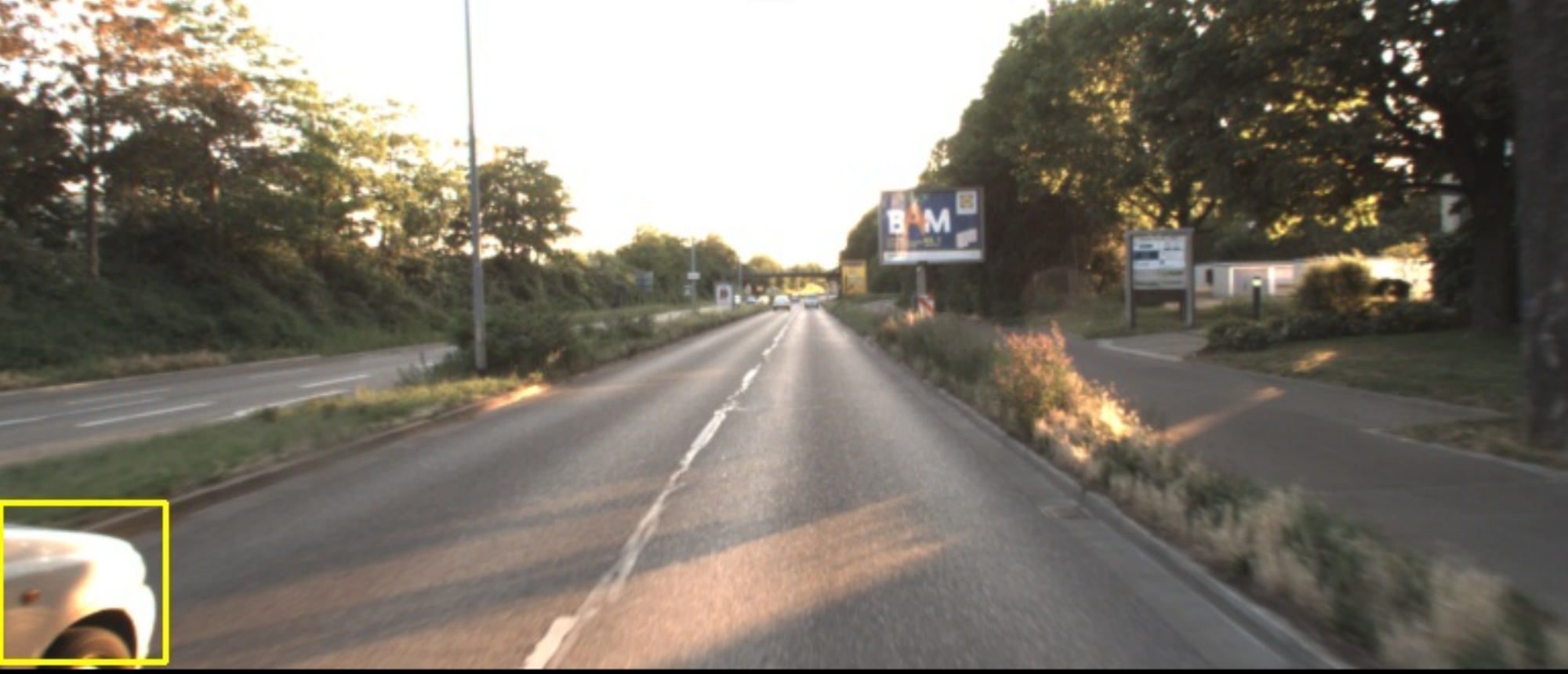} \\
    \includegraphics[width=\linewidth]{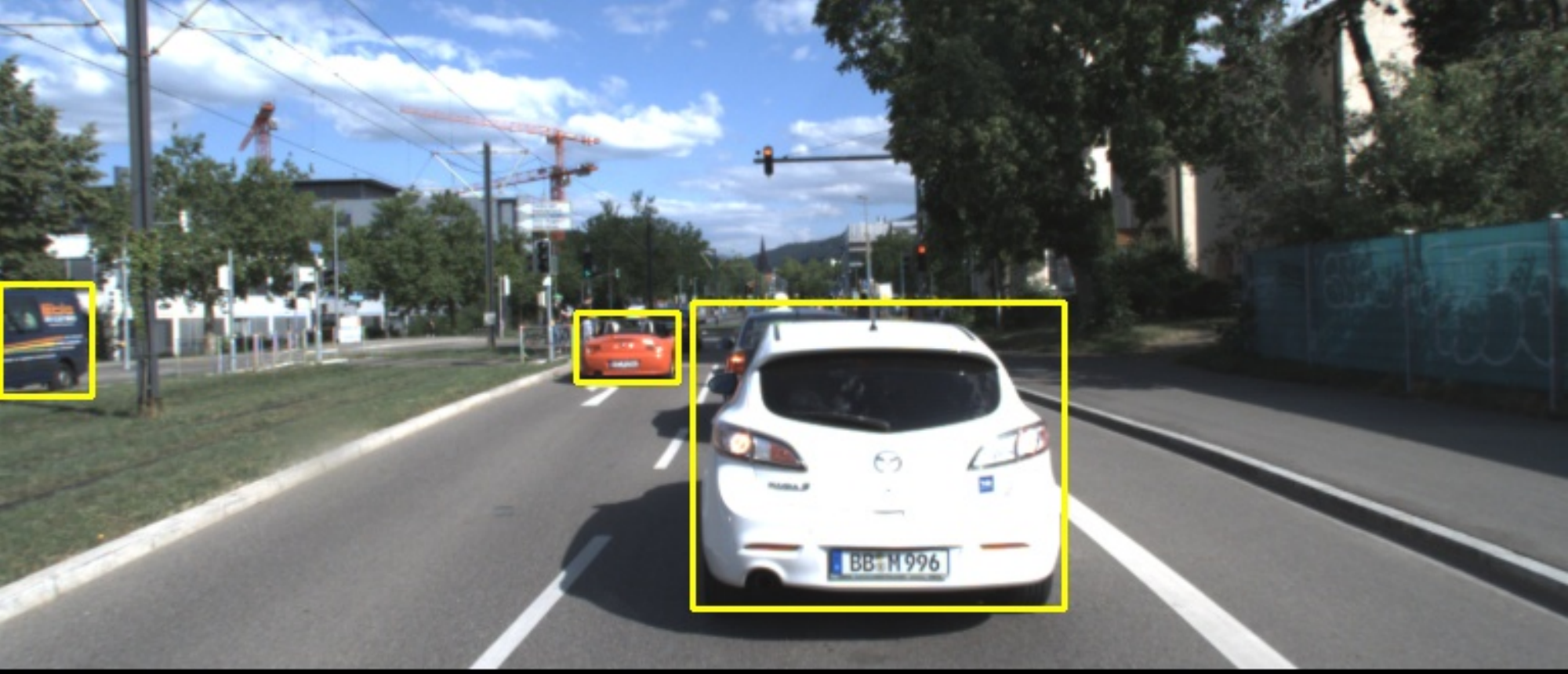} &  \includegraphics[width=\linewidth]{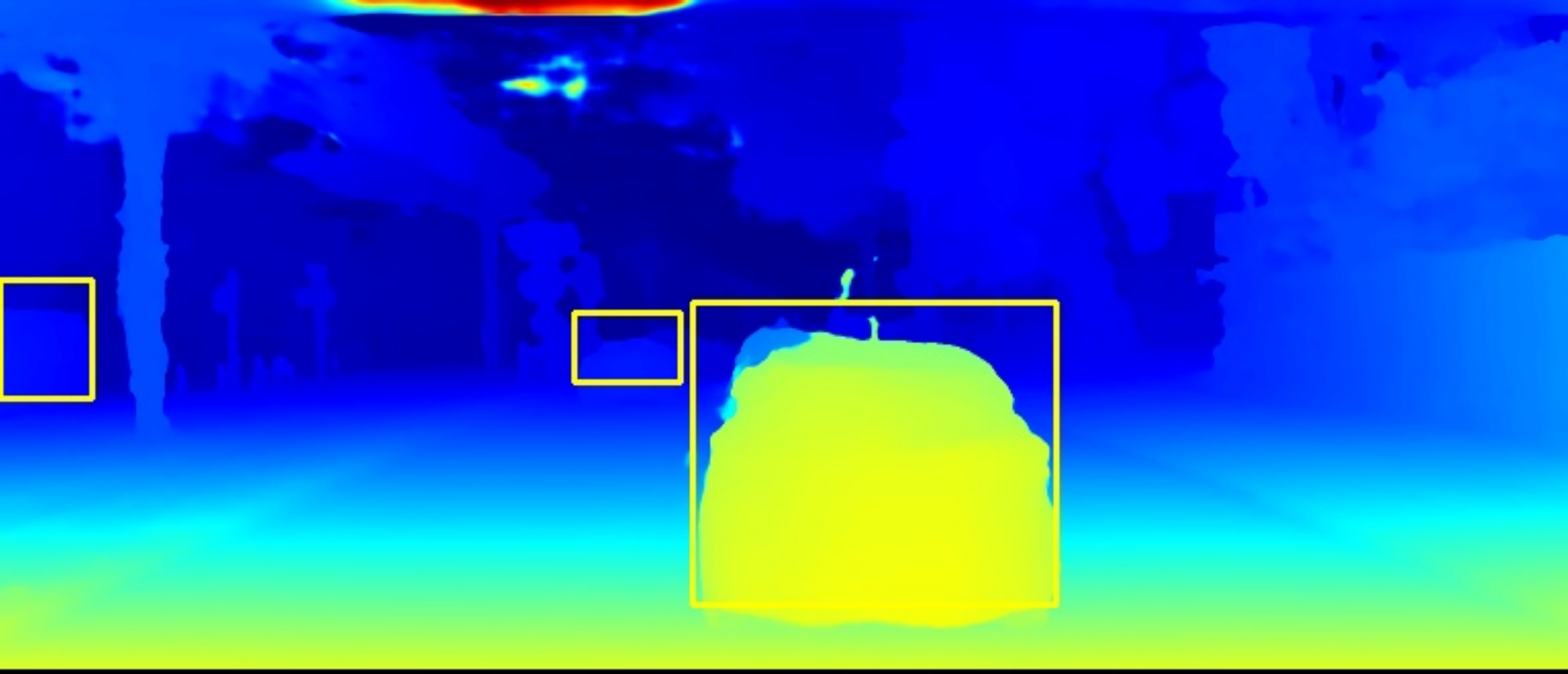} & \includegraphics[width=\linewidth]{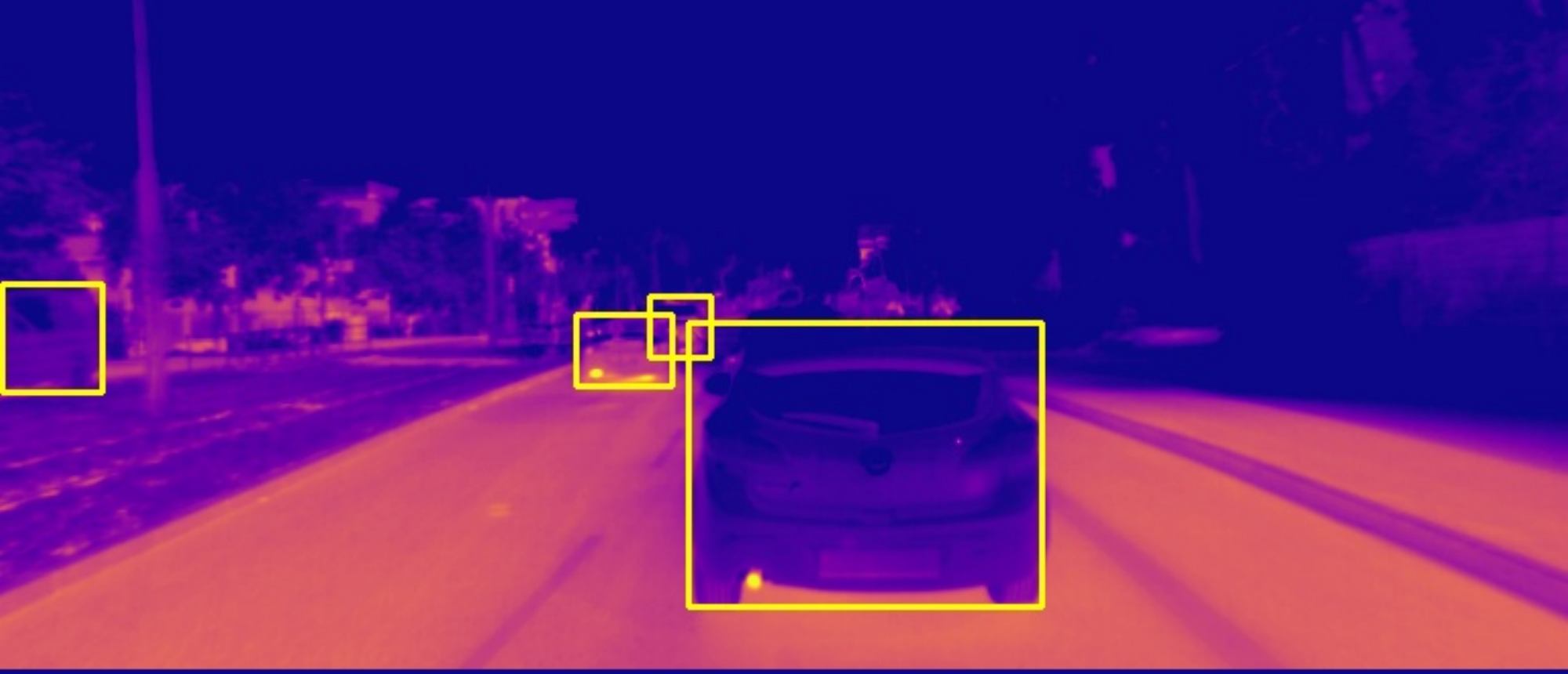} & \includegraphics[width=\linewidth]{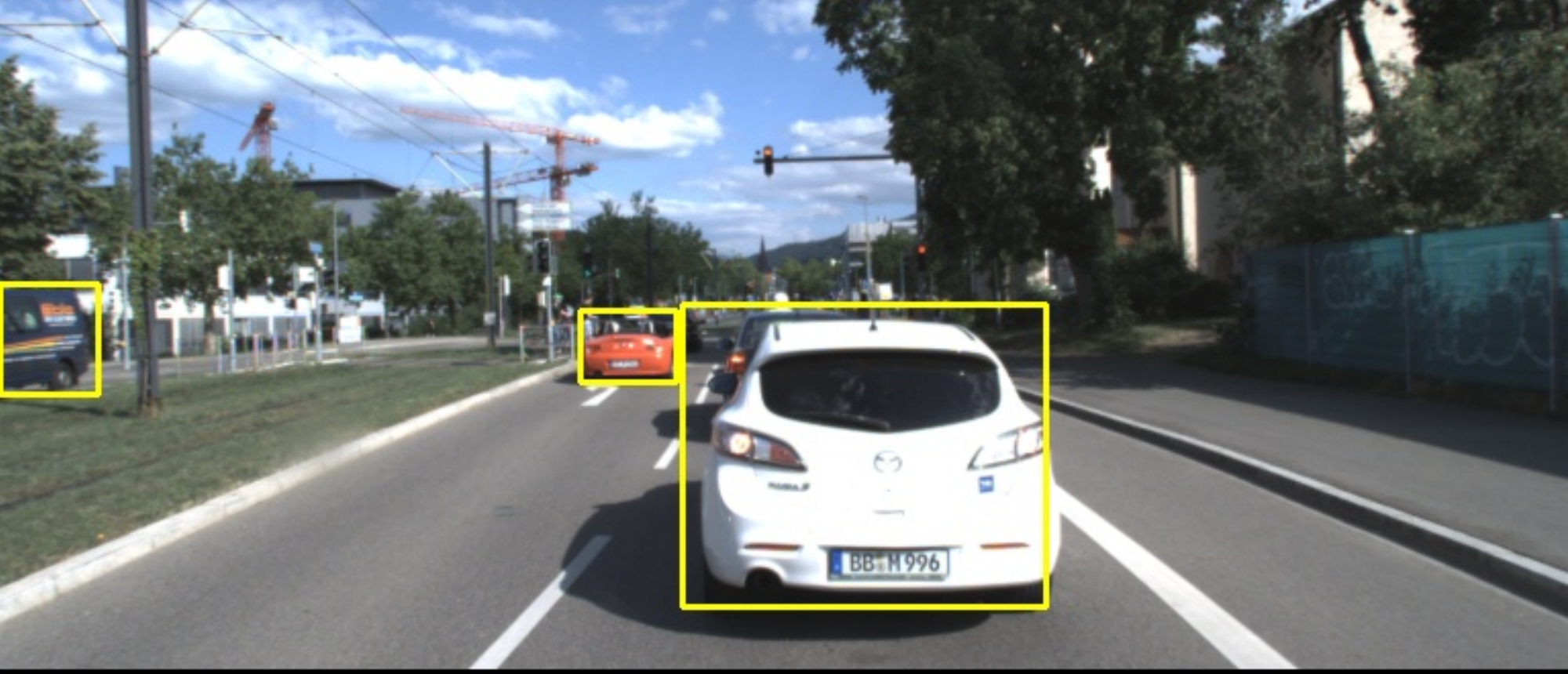} & \includegraphics[width=\linewidth]{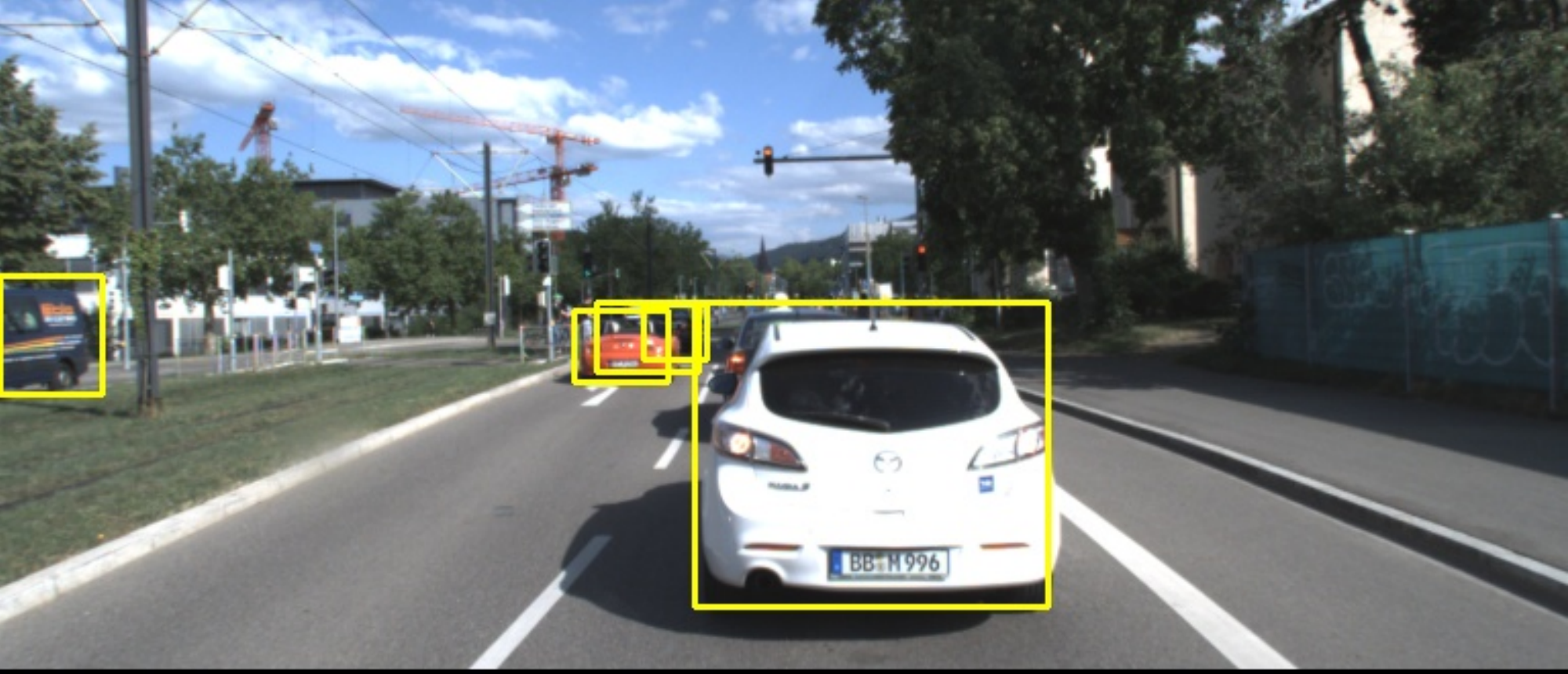} \\
    \includegraphics[width=\linewidth]{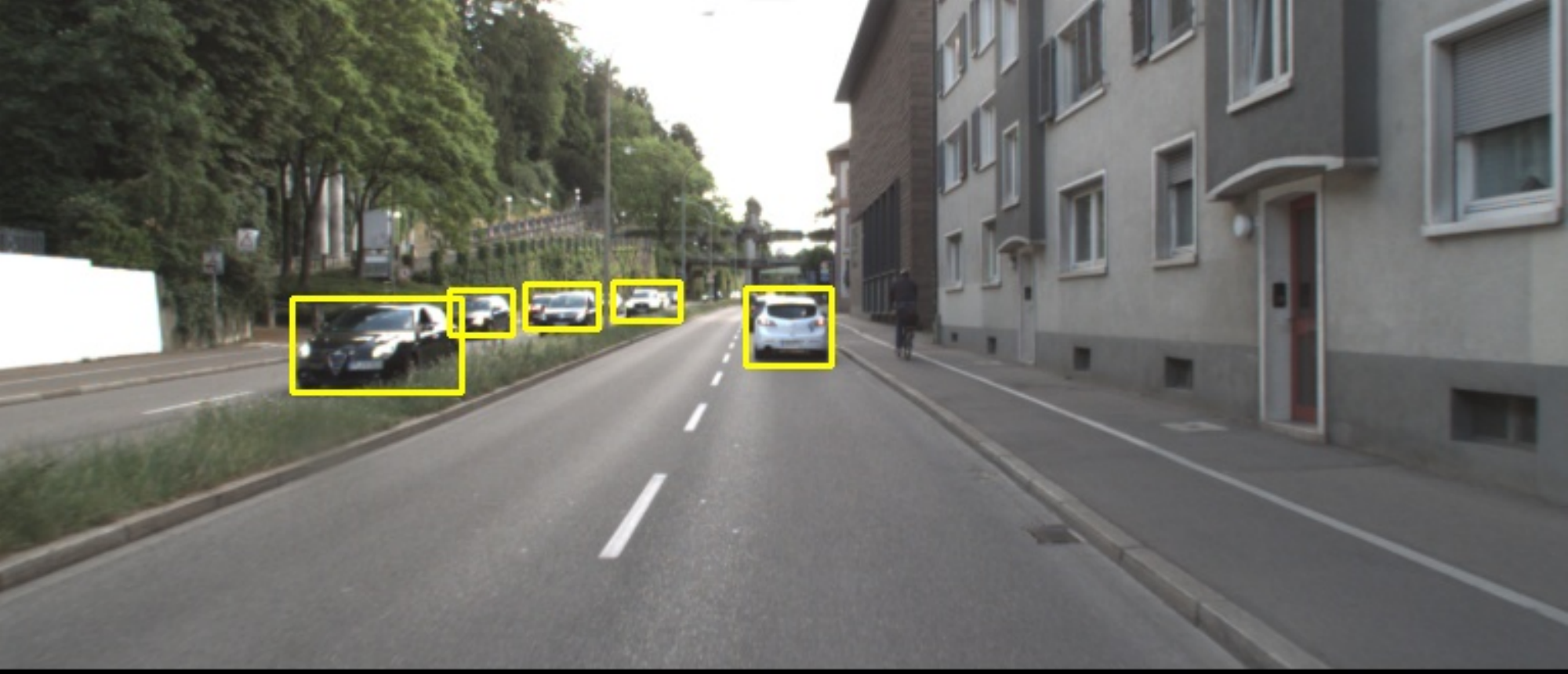} &  \includegraphics[width=\linewidth]{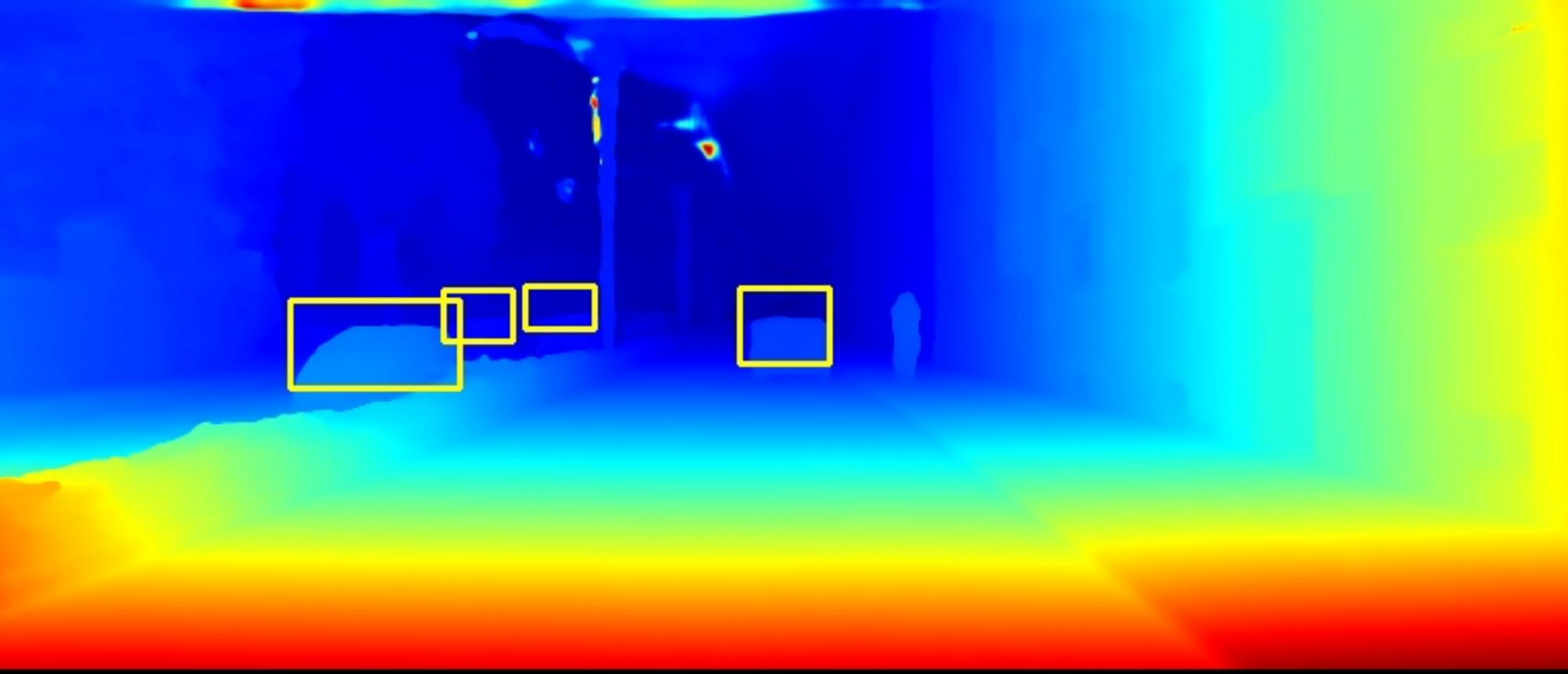} & \includegraphics[width=\linewidth]{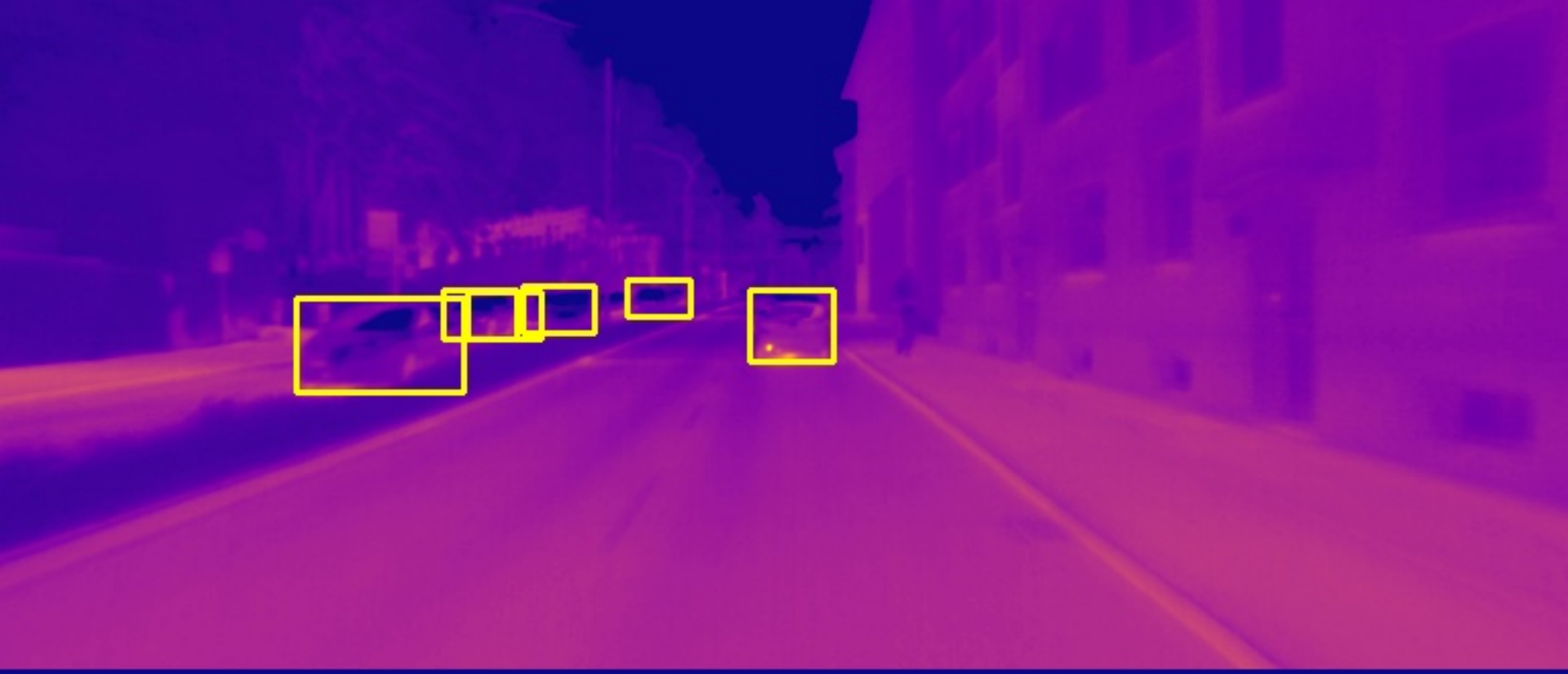} & \includegraphics[width=\linewidth]{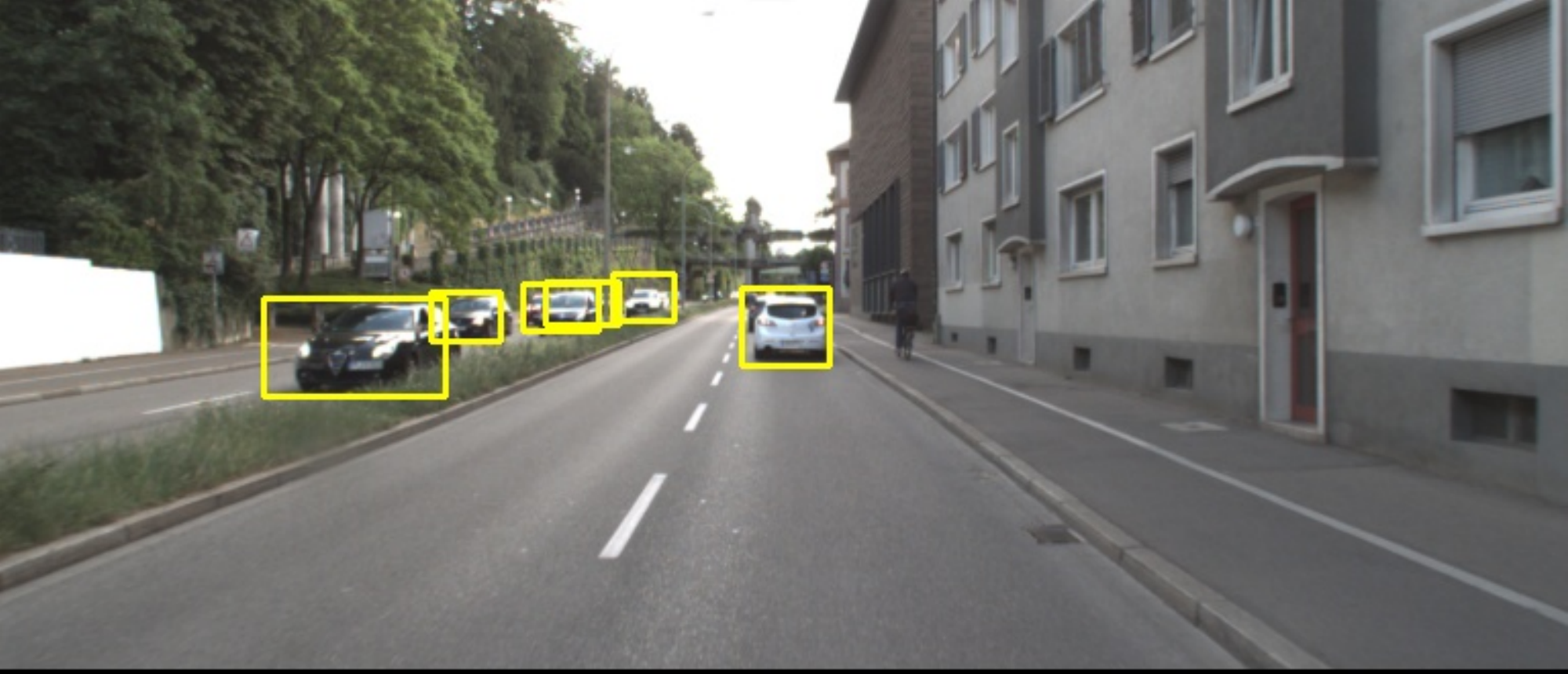} & \includegraphics[width=\linewidth]{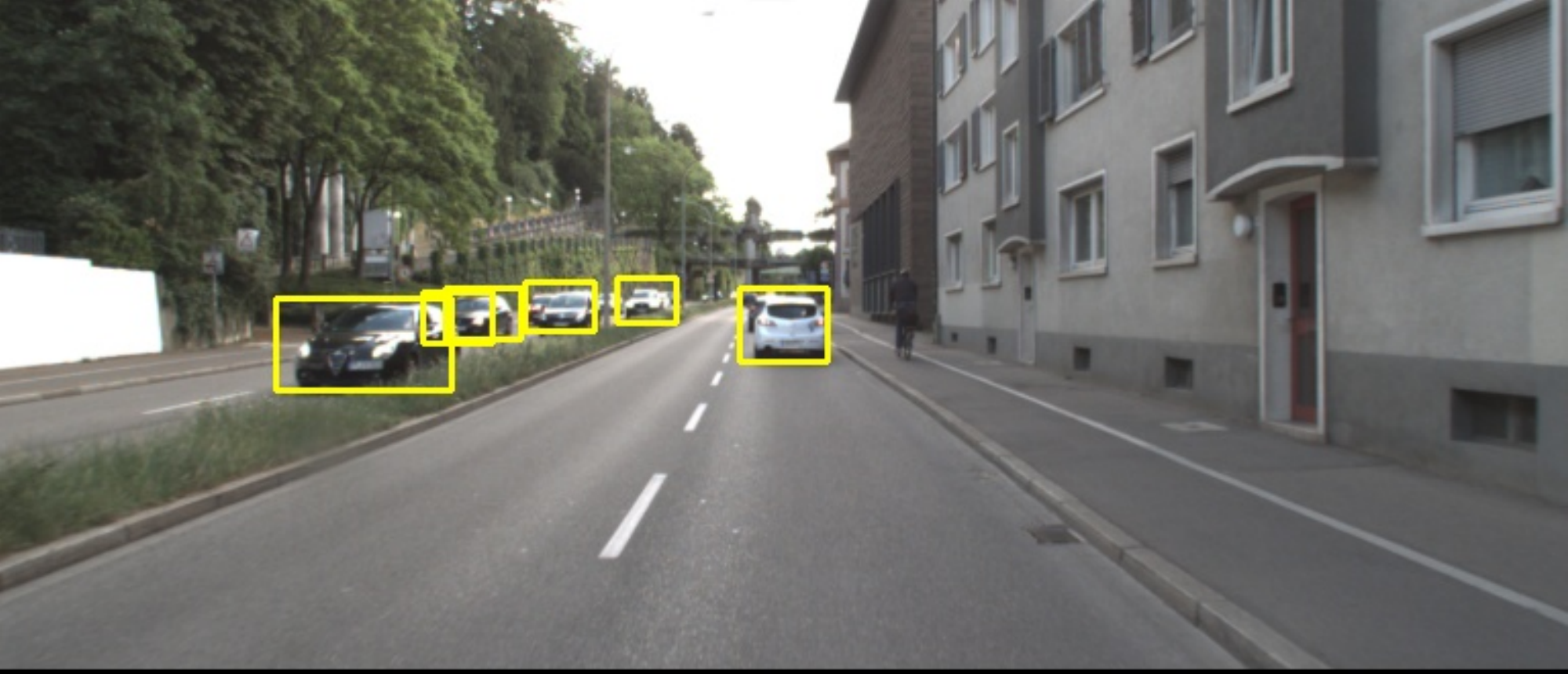} \\
    \includegraphics[width=\linewidth]{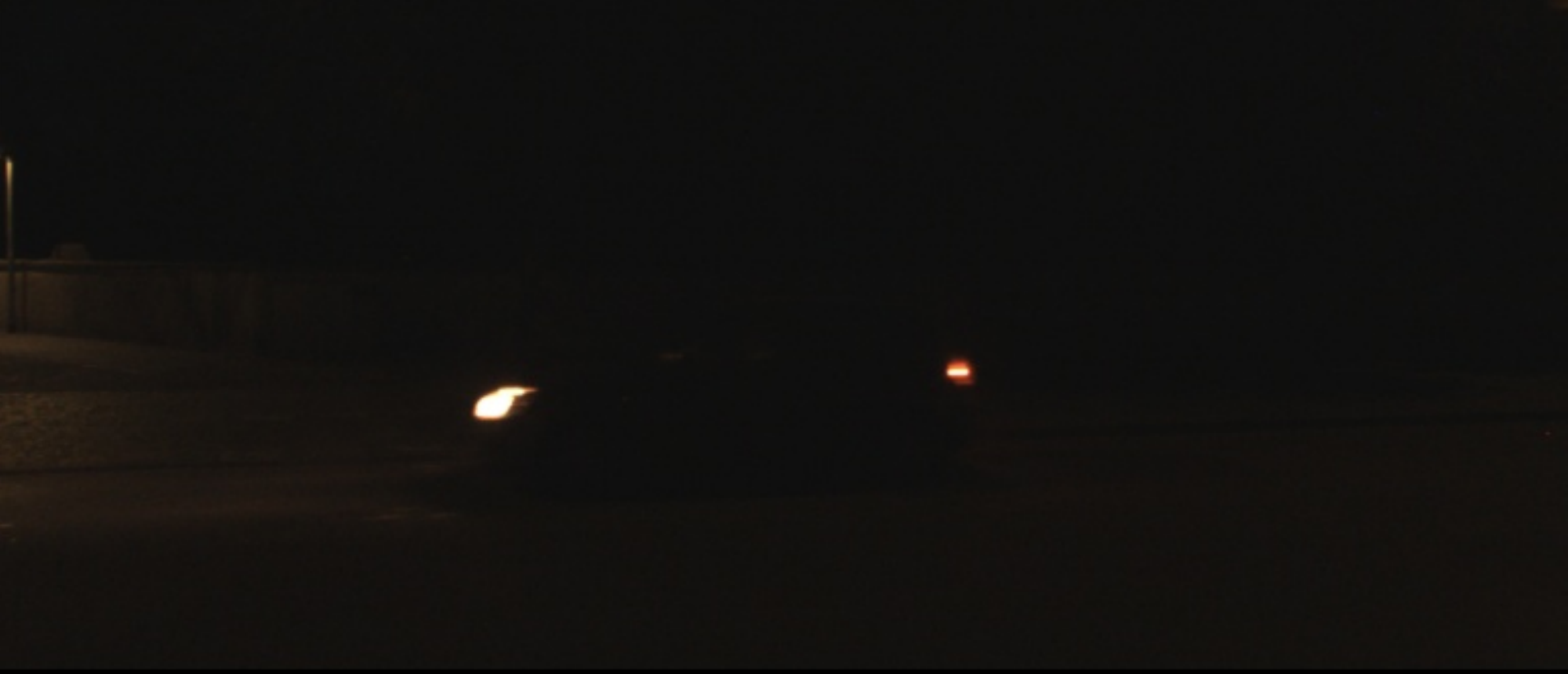} &  \includegraphics[width=\linewidth]{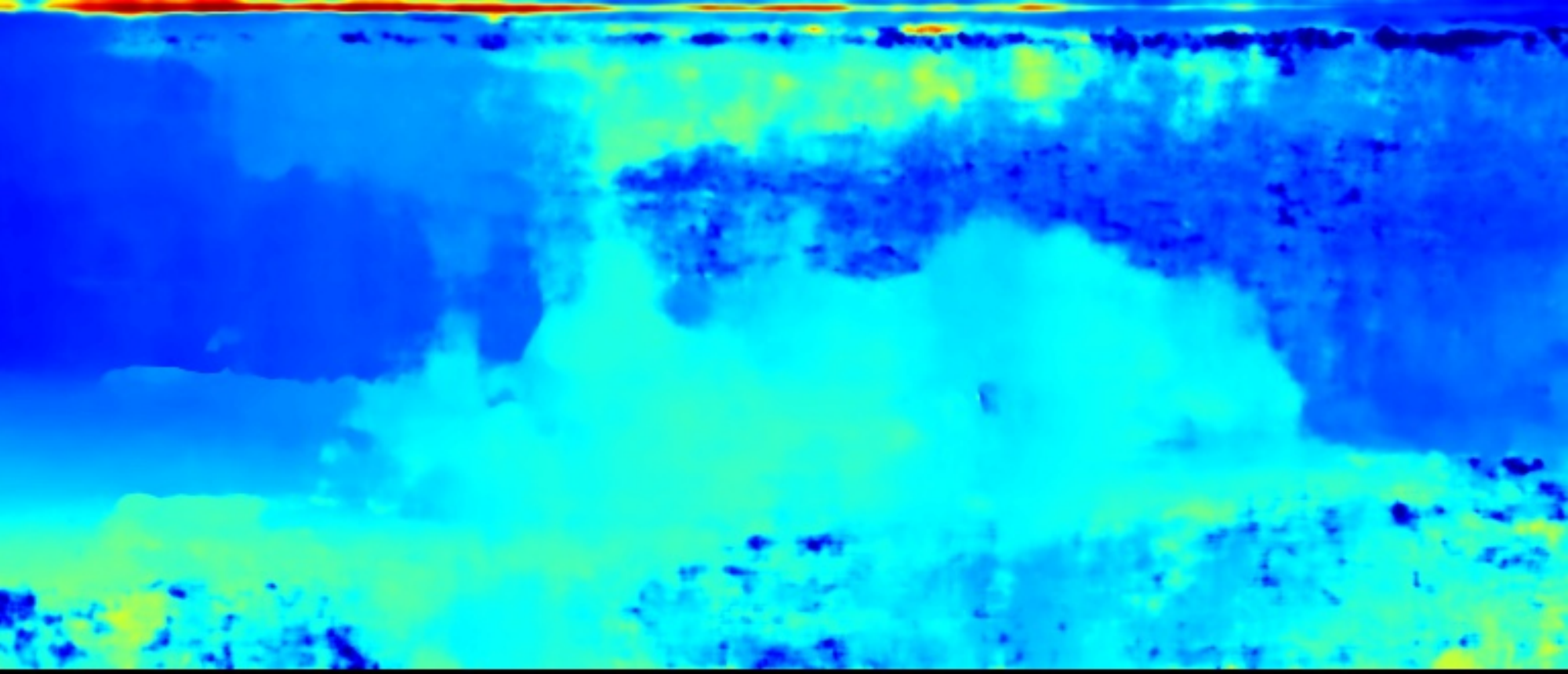} & \includegraphics[width=\linewidth]{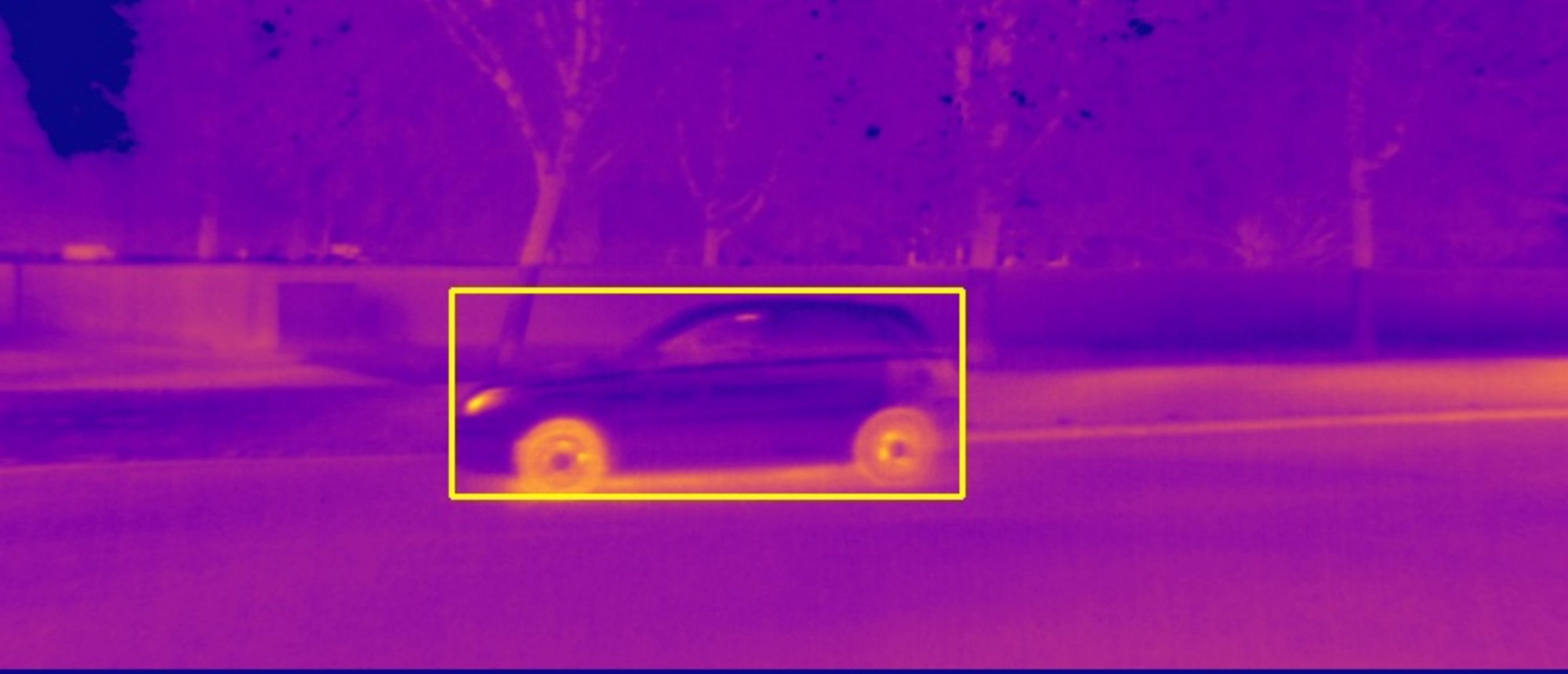} & \includegraphics[width=\linewidth]{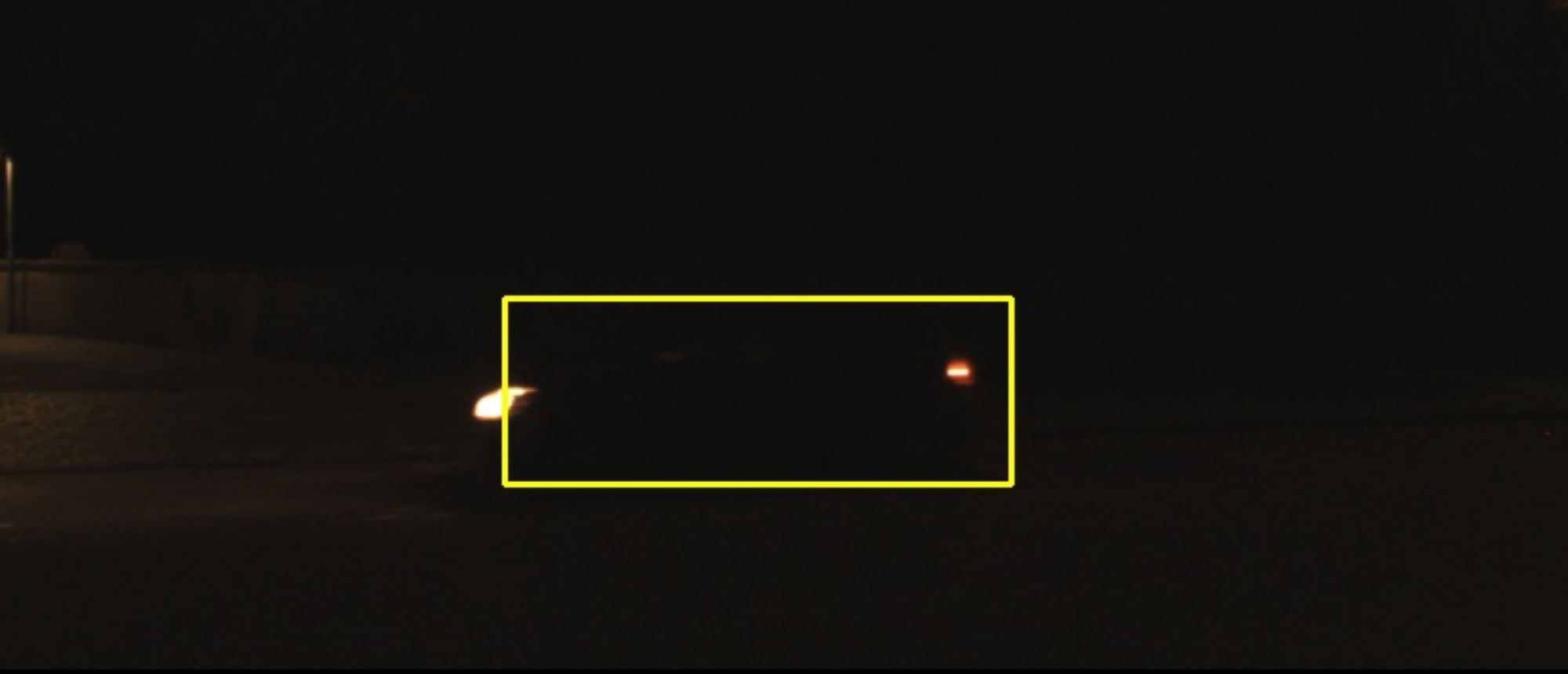} & \includegraphics[width=\linewidth]{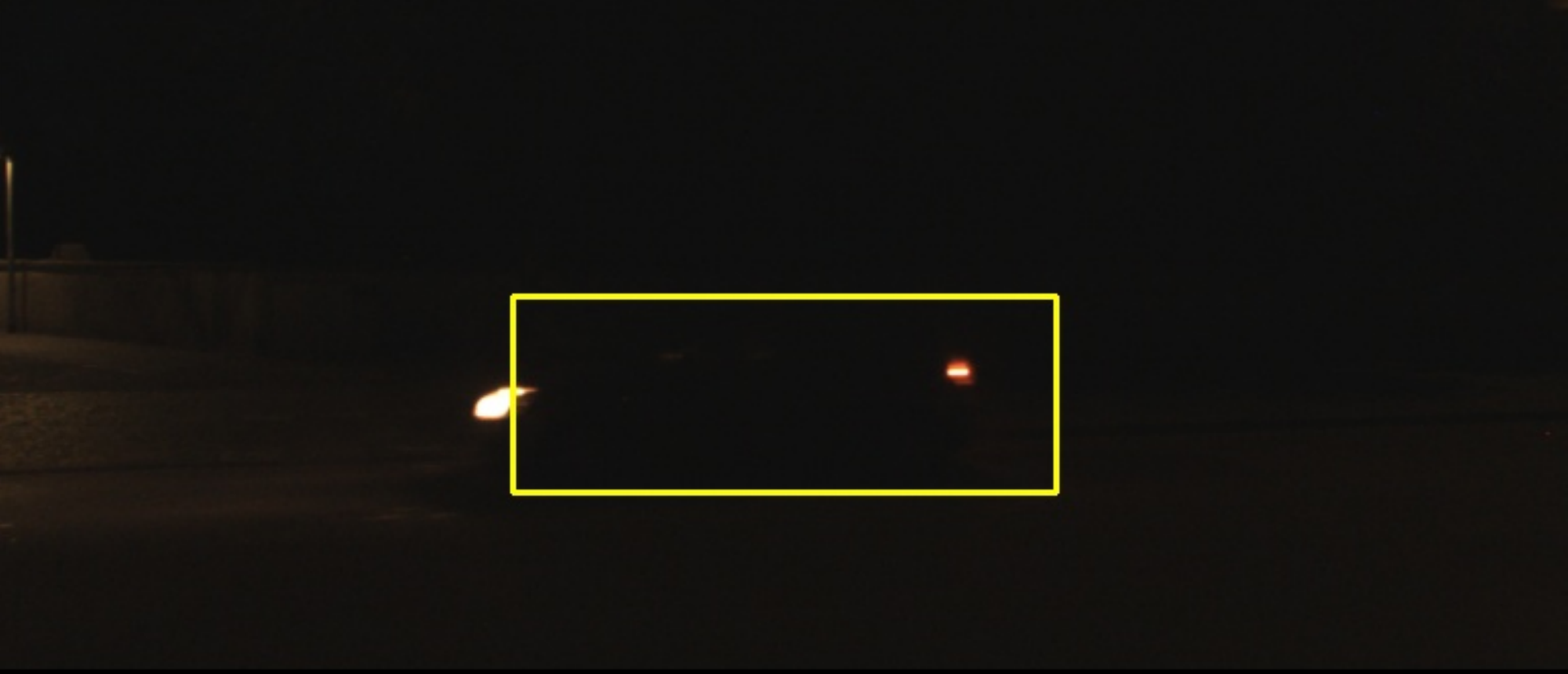} \\
     \includegraphics[width=\linewidth]{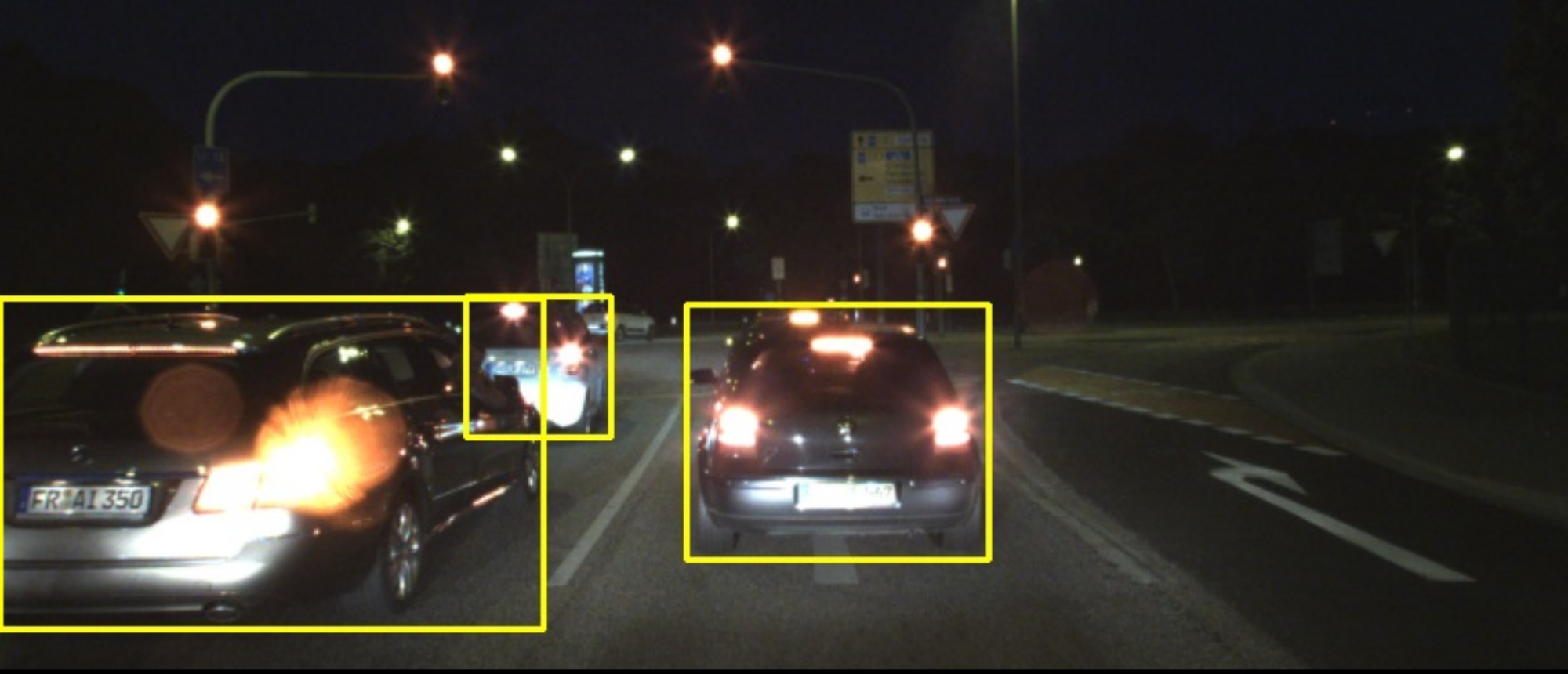} &  \includegraphics[width=\linewidth]{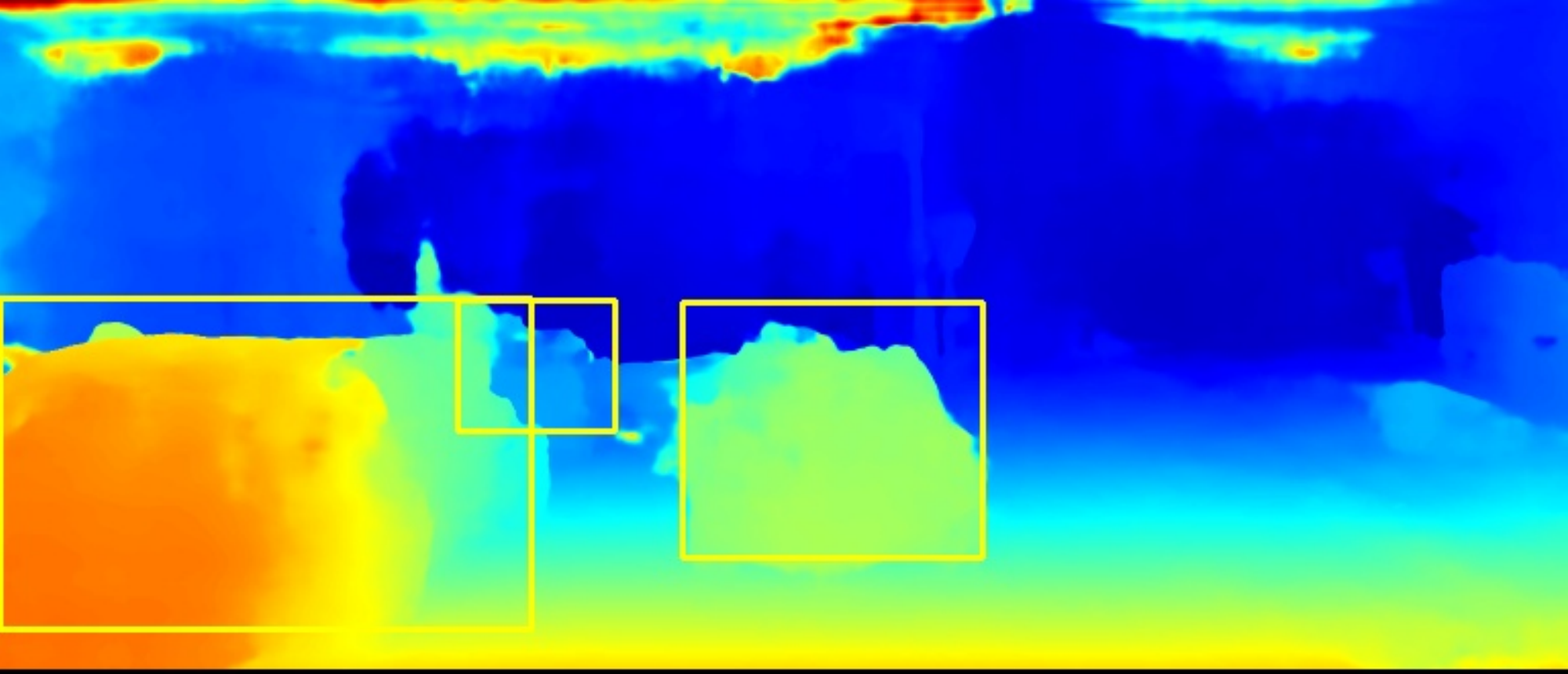} & \includegraphics[width=\linewidth]{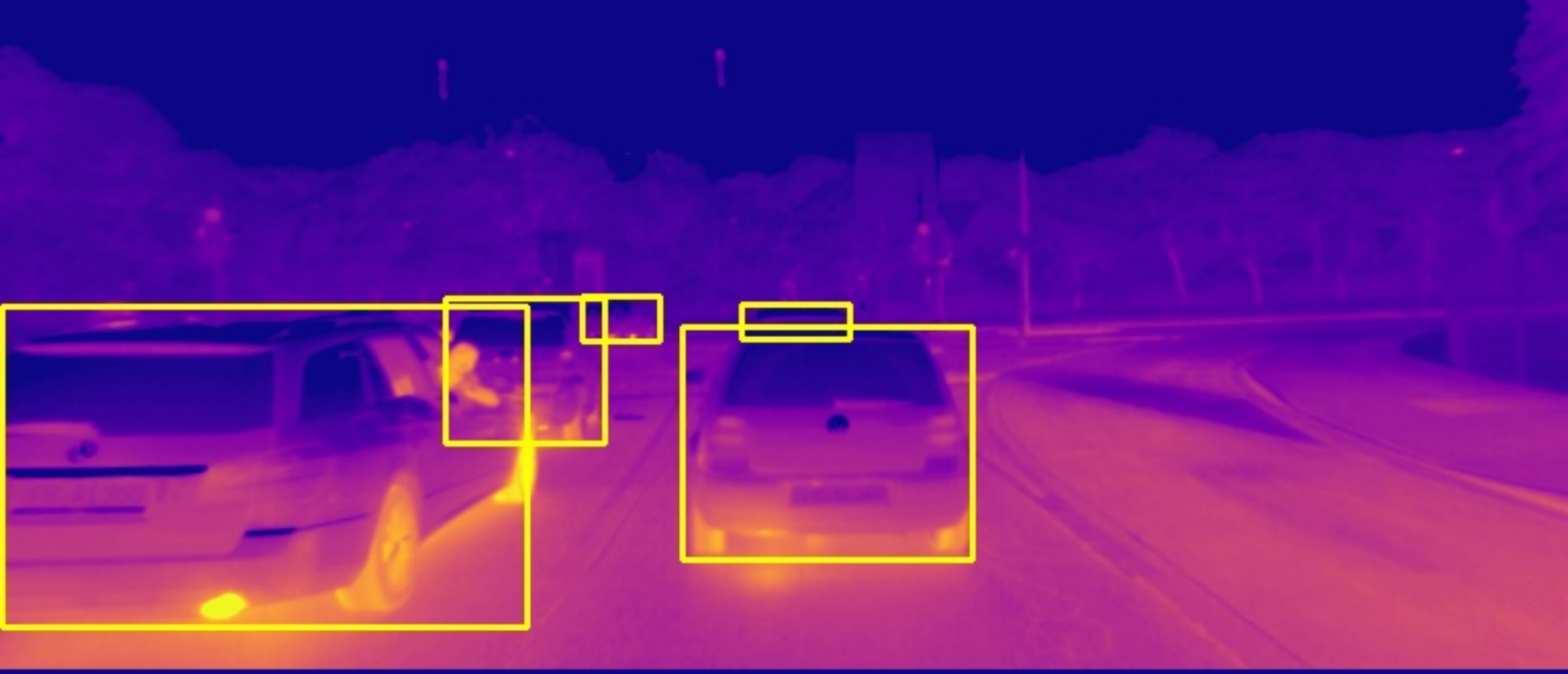} & \includegraphics[width=\linewidth]{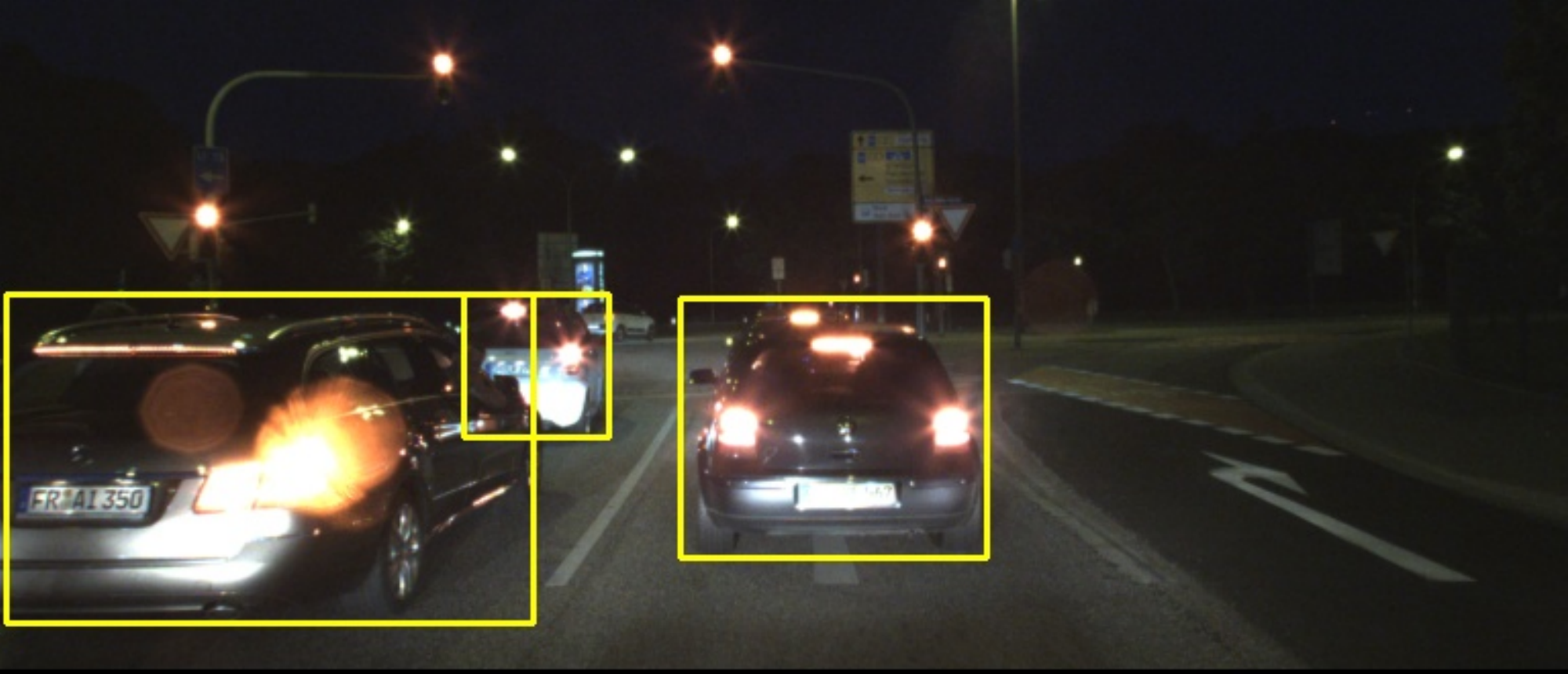} & \includegraphics[width=\linewidth]{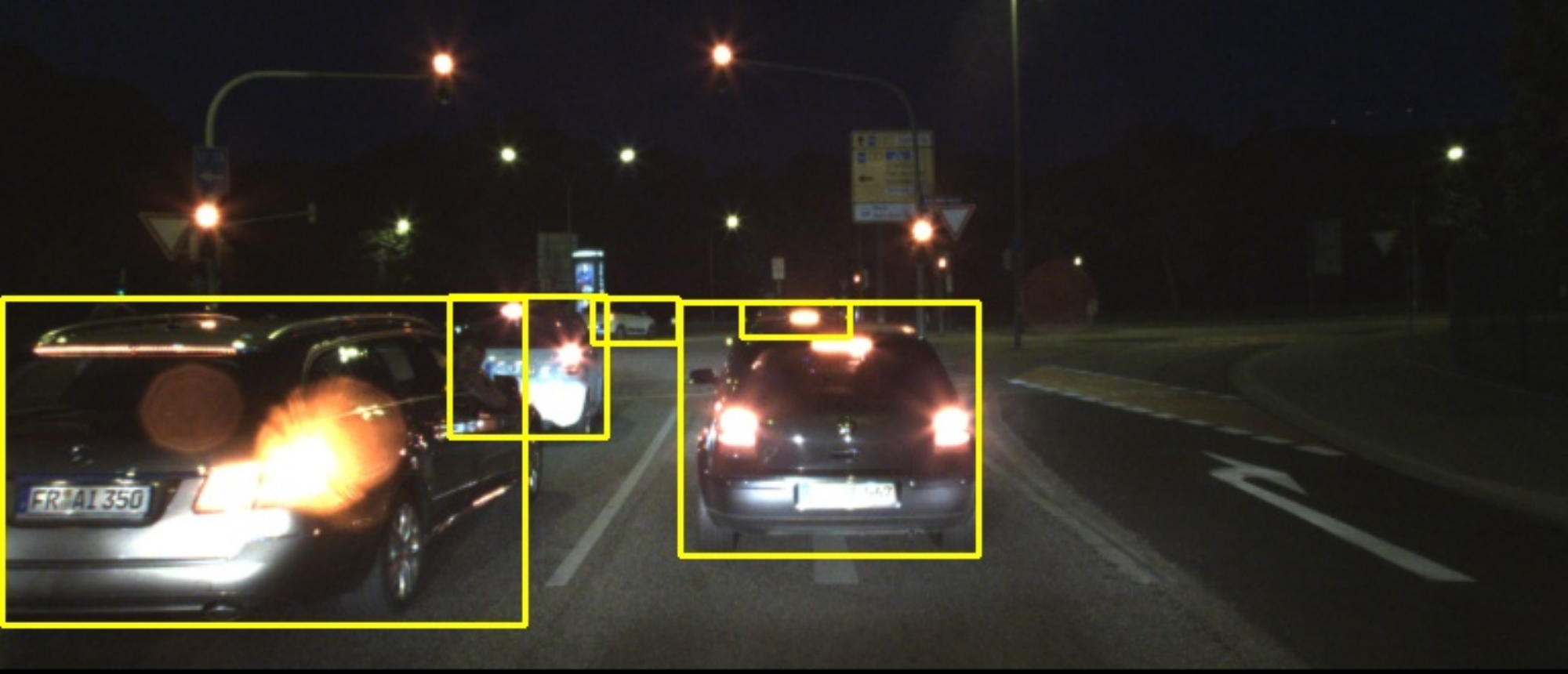} \\
     \includegraphics[width=\linewidth]{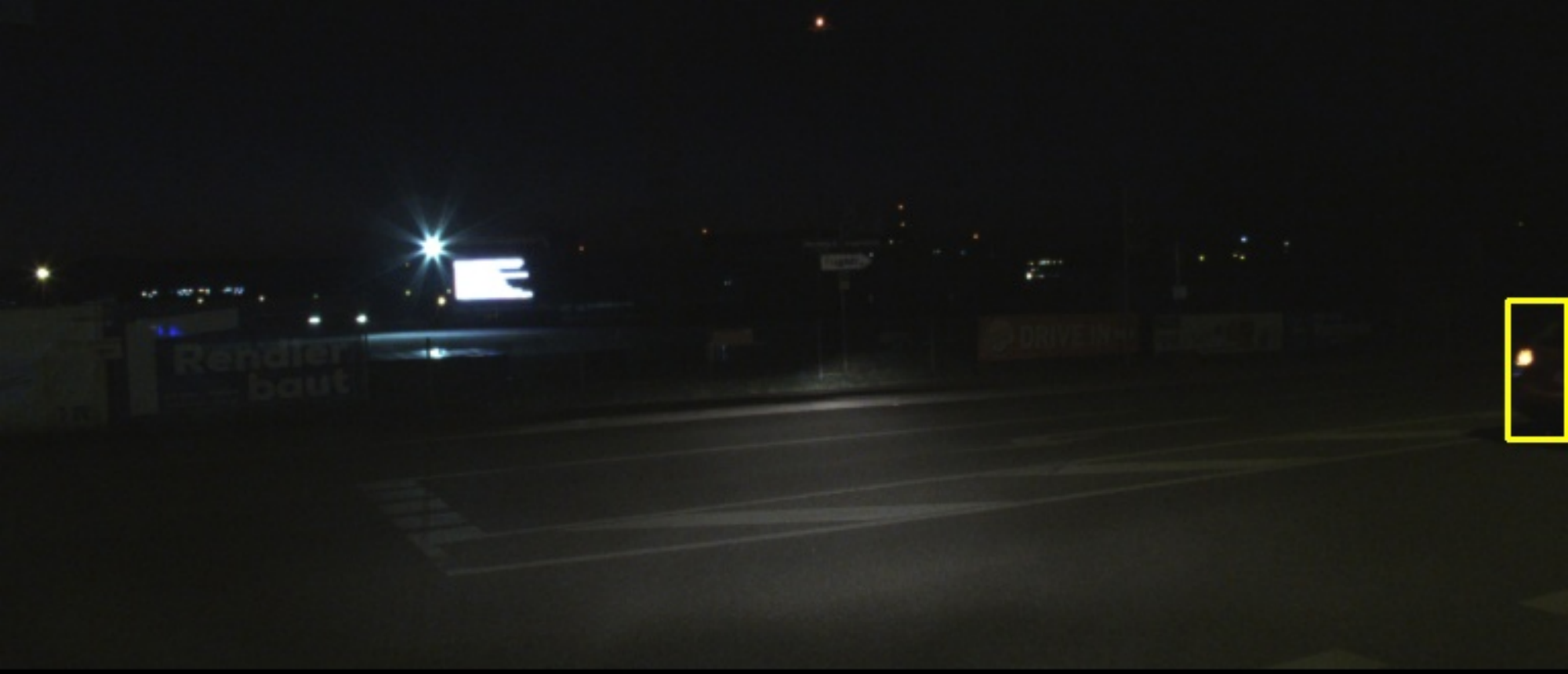} &  \includegraphics[width=\linewidth]{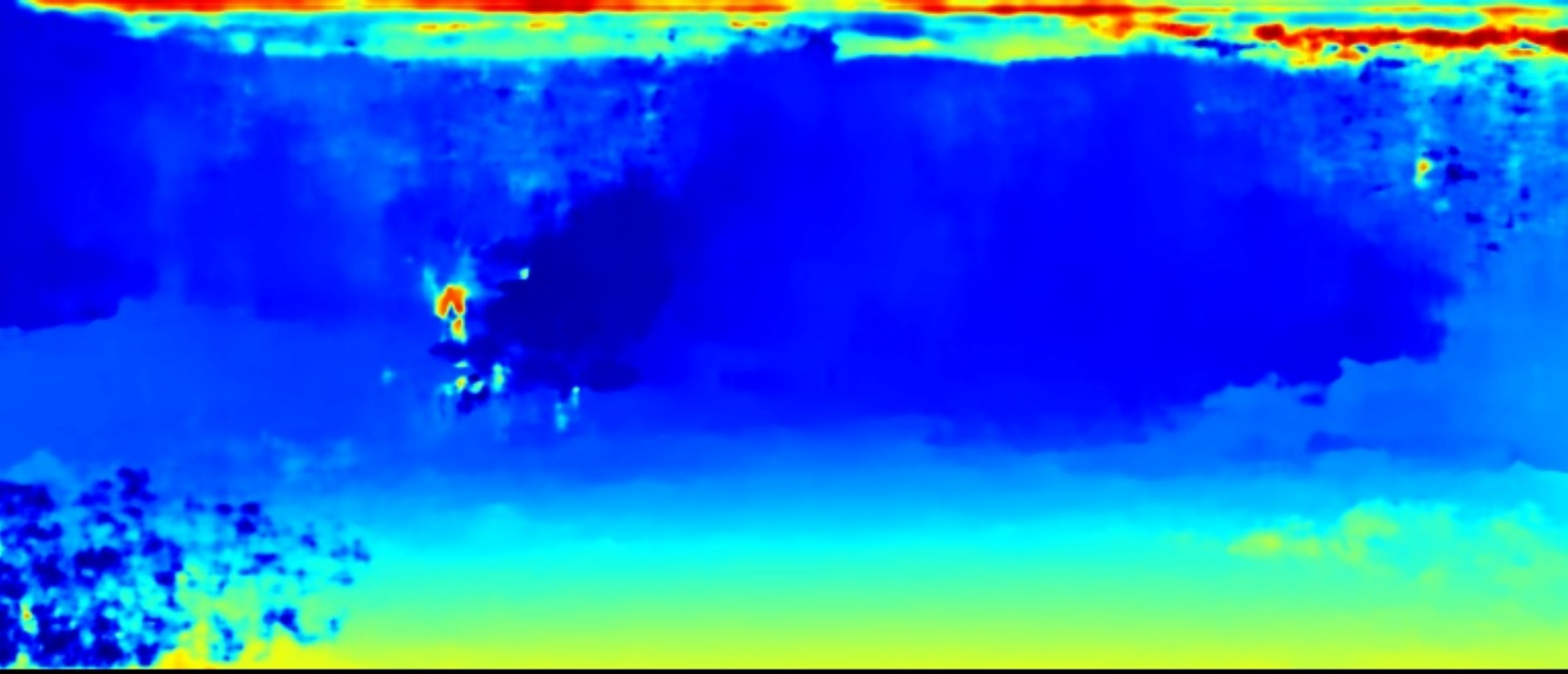} & \includegraphics[width=\linewidth]{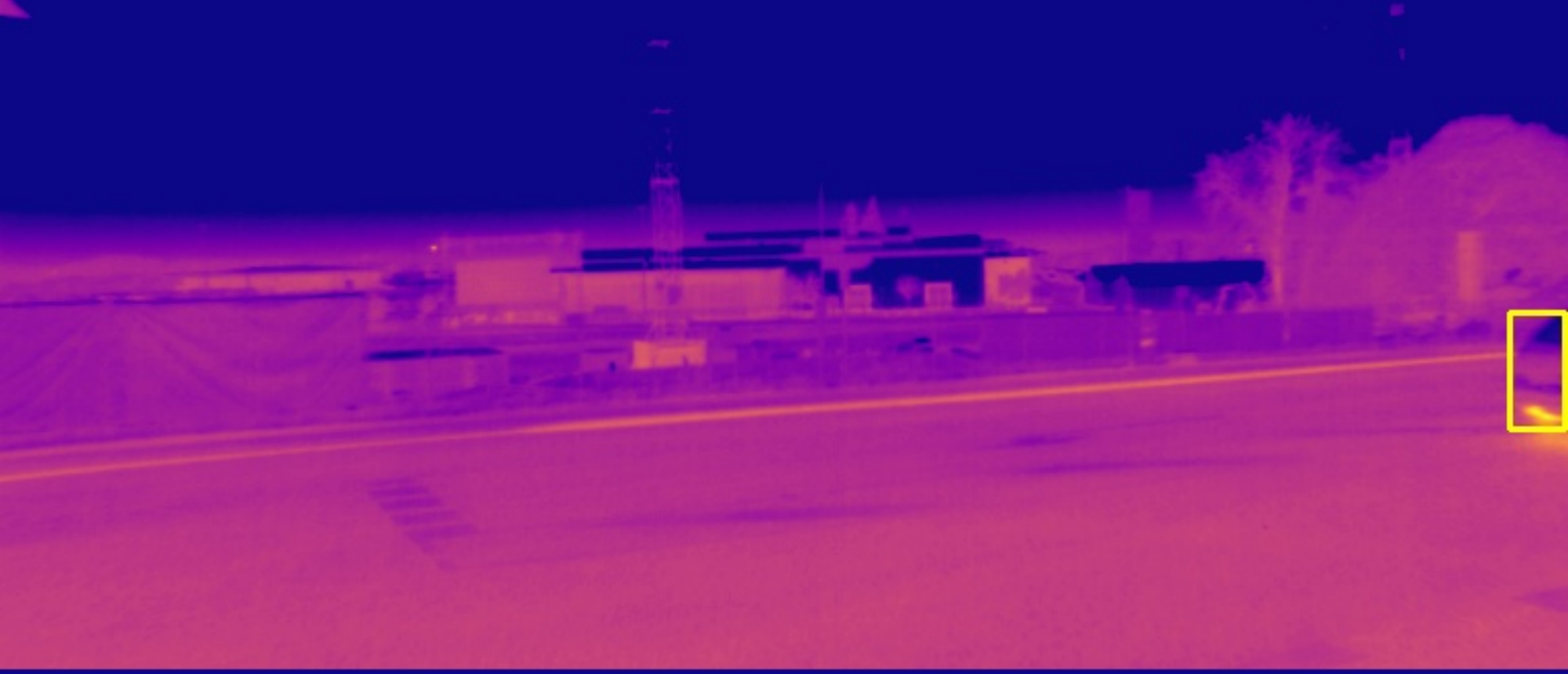} & \includegraphics[width=\linewidth]{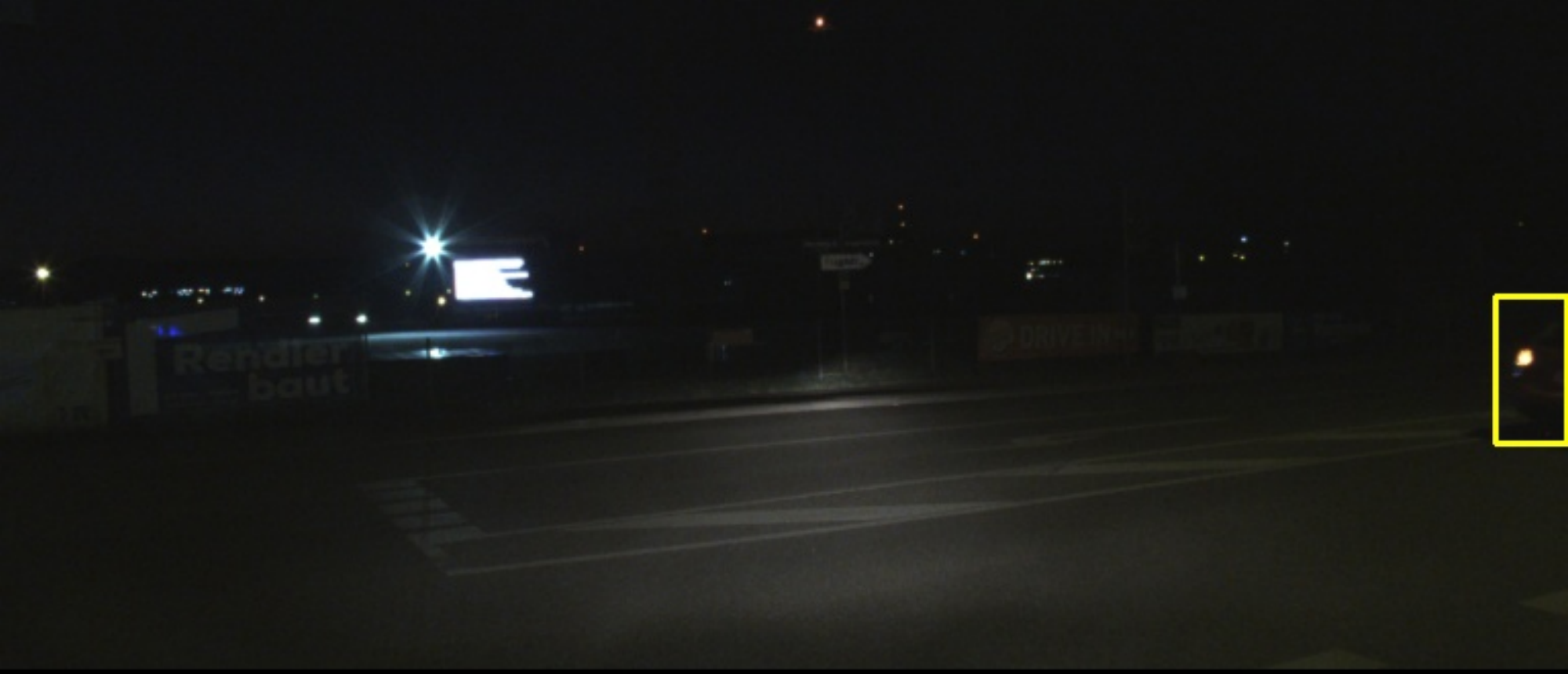} & \includegraphics[width=\linewidth]{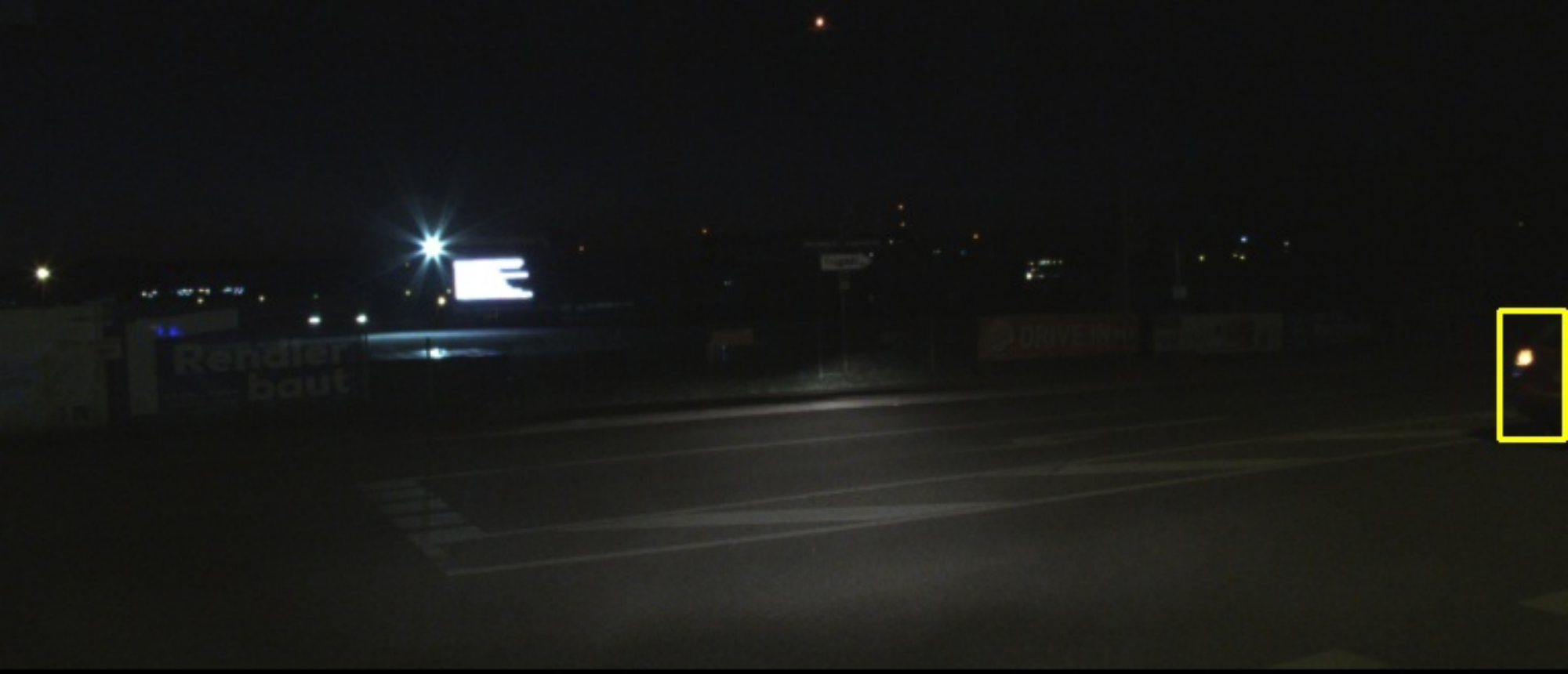} \\
\end{tabular}}
\caption{Qualitative comparisons of predictions from individual modality-specific teachers with the previous state-of-the-art StereoSoundNet~\cite{wang2020score}, and our MM-DistillNet. Our network consistently detects moving vehicles even in the scenes where the baselines fail.}
\label{fig:SM_qualitative}
\end{figure*}

In \figref{fig:SM_qualitativenight}, we present comparison of both methods during nighttime where poor light conditions can be observed. Our MM-DistillNet simultaneously detects multiple cars, even the distant ones, while StereoSoundNet often fails to detect beyond one vehicle. These results demonstrate the novelty of distilling multimodal knowledge in our MM-DistillNet as it shows substantial robustness in poor light conditions, thereby successfully overcoming the limitations of distilling knowledge only from RGB images. Additionally, we qualitatively evaluate the performance of detecting multiple vehicles simultaneously in \figref{fig:SM_qualitativegoodmultiplecars}. Simultaneously detecting multiple vehicles with only sound is an extremely challenging task due to its low spatial resolution. By distilling knowledge from multiple modality-specific teachers, we show that it is not only feasible to detect vehicles simultaneously without relying on the arduous data labeling process, the performance of the detections also substantially improves. Further enhancing this ability will enable a broad spectrum of applications of these audio approaches in real world scenarios. Even though our MM-DistillNet is able to provide very promising results for object detection and tracking, the audio modality suffers from limitations of its own that are highlighted in examples shown in \figref{fig:SM_qualitativefailure}. We observe that occasionally, distant objects are not detected and multiple distant objects are detected as a single object. We believe that a microphone with better sensitivity as well as more examples of this phenomena in the training set will enable our approach to improve the performance in these conditions. Finally, we compare the performance of our MM-DistillNet and the modality-specific RGB, depth, and thermal teachers in \figref{fig:SM_qualitative}. The results show the weaknesses and strengths of each of the selected modalities. Especially in night scenarios, we can observe how the thermal teacher contributes to the distillation of knowledge to the audio student as it reliably detects cars that are not visible in the RGB images, due to low illumination conditions.

\end{document}